%% file: iclr2027_arxiv.tex
\def\ICLRArxivVersion{}

\documentclass{article} 
\usepackage{iclr2027_conference,times}
\newif\ificlrarxivversion
\ifdefined\ICLRArxivVersion
  \iclrarxivversiontrue
  \ClearShipoutPicture
\else
  \iclrarxivversionfalse
\fi

\input{math_commands.tex}

\usepackage{url}
\usepackage{booktabs}
\usepackage{array}
\usepackage{multirow}
\newcommand{\ml}[1]{\begingroup\def\\{}\mbox{#1}\endgroup}
\usepackage{enumitem}
\usepackage{graphicx}
\usepackage{float}
\usepackage{wrapfig}
\usepackage{needspace}
\usepackage{capt-of}
\usepackage{placeins}
\usepackage{hyperref}
\definecolor{linkblue}{RGB}{0,51,102}
\hypersetup{colorlinks=true,allcolors=linkblue,pdfborder={0 0 0}}
\makeatletter
\renewcommand{\hyper@natlinkbreak}[2]{#1}
\makeatother

\newcommand{\promptboxed}{\textbackslash boxed\{\}}
\newcommand{\promptfill}[1]{%
  \par\noindent#1\par
}
\newenvironment{promptbox}{%
  \par\begingroup
  \footnotesize\ttfamily
  \raggedright
  \setlength{\parindent}{0pt}%
  \setlength{\parskip}{0.25em}%
  \hyphenpenalty=10000
  \exhyphenpenalty=10000
  \tolerance=5000
  \emergencystretch=2em
  \noindent\ignorespaces
}{%
  \par\endgroup
}

\makeatletter
\newcommand{\wrapfloatcaptionsetup}{%
  \setlength{\abovecaptionskip}{6pt}%
  \setlength{\belowcaptionskip}{0pt}}
\makeatother

\title{Diagnosing On-Policy Self-Distillation for Reasoning Language Models}

\ificlrarxivversion
\author{Yang Li\textsuperscript{1}, Gongle Xue\textsuperscript{1},
Yuheng Yuan\textsuperscript{1}, Yijia Guo\textsuperscript{1},
Shizhe Zhang\textsuperscript{1}, Liwen Hu\textsuperscript{1},
Lei Ma\textsuperscript{1} \\
\textsuperscript{1}\,Peking University \\
\texttt{ly0376@stu.pku.edu.cn}
}

\hypersetup{pdfauthor={Yang Li, Gongle Xue, Yuheng Yuan, Yijia Guo,
Shizhe Zhang, Liwen Hu, Lei Ma}}
\else

\author{Antiquus S.~Hippocampus, Natalia Cerebro \& Amelie P. Amygdale \thanks{ Use footnote for providing further information
about author (webpage, alternative address)---\emph{not} for acknowledging
funding agencies.  Funding acknowledgements go at the end of the paper.} \\
Department of Computer Science\\
Cranberry-Lemon University\\
Pittsburgh, PA 15213, USA \\
\texttt{\{hippo,brain,jen\}@cs.cranberry-lemon.edu} \\
\And
Ji Q. Ren \& Yevgeny LeNet \\
Department of Computational Neuroscience \\
University of the Witwatersrand \\
Joburg, South Africa \\
\texttt{\{robot,net\}@wits.ac.za} \\
\AND
Coauthor \\
Affiliation \\
Address \\
\texttt{email}
}
\fi

\newcommand{\nth}{\text{\textup{NTh}}}
\newcommand{\thmode}{\text{\textup{Th}}}

\ificlrarxivversion
  \iclrfinalcopy
\else
\fi
\begin{document}

\maketitle
\ificlrarxivversion
  \fancyhead{} 
  \renewcommand{\headrulewidth}{0pt}
\fi

\begin{abstract}
On-policy self-distillation (OPSD) has attracted growing interest as a promising approach to improve the reasoning ability of language models.
Without external rewards nor a separate stronger teacher, the self-teacher with privileged information could provide dense signals on student's trajectories.
However, its behavior in language reasoning remains unclear, with reported outcomes ranging from modest gains to behavioral collapse. 
In this work, we diagnose OPSD for mathematical and coding reasoning tasks across models spanning 0.6B--8B parameters. 
We conduct controlled experiments and token-level analyses to fully delve into OPSD.
We point out that teacher's signal is shaped by reasoning mode alignment and the complete teacher prefix, rather than by privileged semantics alone. 
OPSD improves reasoning only in narrow compatibility regimes. Otherwise, it produces ineffective length growth, stable degradation, or behavioral collapse. 
Token-level analysis shows that teacher's signal is not stable and does not predict downstream performance. 
Based on these results, we argue that OPSD is a sensitive algorithm rather than a generally reliable reasoning improvement post-training method.
\end{abstract}

\section{Introduction}
\label{sec:intro}

Post-training is significant for improving the reasoning capabilities of large language models (LLMs). Reinforcement learning with verifiable rewards (RLVR), such as Group Relative Policy Optimization (GRPO; \citealp{deepseek_math}), optimizes policy models using automatically checked results. However, RLVR typically provides only a sparse outcome reward after a complete trajectory has been generated. For long reasoning tasks, waiting for full responses at each step can make training slow and expensive.

On-policy distillation (OPD; \citealp{gkd,opd_survey}) offers a complementary algorithm. Instead of assigning a single reward to an entire response, OPD uses a stronger teacher to provide a next-token distribution at every prefix visited by the student. This produces dense supervision along the student's own trajectories, but requires inference from a separate and usually more capable teacher. Moreover, when the teacher and student use different tokenizers, their token-level distributions may not be directly aligned.

On-policy self-distillation (OPSD; \citealp{opsd}) removes the external teacher by using a frozen copy of the student itself. The self-teacher receives privileged information, such as a reference answer or solution, while the student conditions only on the original problem. OPSD is appealing since it provides dense on-policy supervision without external rewards or a stronger model. However, the design also creates a fundamental ambiguity. Because the teacher and student share the same parameters, their difference comes entirely from their conditioning contexts. The resulting teacher--student divergence may encode useful information about the solution, but it may also arise from mismatched reasoning modes or prompts. Moreover, the teacher must score tokens generated by the student that it might not generate itself. Minimizing this divergence therefore need not correspond to improving the student's unprivileged reasoning ability.

This ambiguity is especially important for long reasoning tasks. OPSD training can lead to different performances. In some settings, it improves answer accuracy. In others, it produces ineffective response length growth, stable performance degradation, or behavioral collapse \citep{opsd,opsd_rethink,opsd_privileged_rethink,many_faces_opd}. These results leave unclear what determines OPSD's behavior and under what conditions it is reliable. We therefore ask three questions: \textbf{First, how does privileged information shape the teacher signal?} \textbf{Second, under what joint conditions does OPSD improve reasoning ability?} \textbf{Third, what can training dynamics reveal about training quality and downstream performance?}

To answer these questions, we systematically evaluate OPSD on mathematical and coding reasoning across models and checkpoints spanning 0.6B--8B parameters. We examine its sensitivity to model, teacher, and training settings through controlled experiments and token-level analyses. We also compare OPSD with GRPO under matched data and training steps.
Our main findings are:
\begin{itemize}[leftmargin=*,itemsep=\parskip,parsep=0pt,topsep=\parskip]
    \item \textbf{The OPSD signal is shaped by prefix distribution mismatch rather than privileged semantics alone.} Reasoning mode mismatch can induce behavioral shifts, while removing privileged information under this mismatch often causes only limited changes. An appropriate teacher--student distribution gap can improve performance, but the gap alone does not indicate useful supervision.
    \item \textbf{OPSD gains are modest and confined to model- and configuration-specific regimes.} Its effect depends jointly on model ability, rollout length, training duration, teacher context, divergence objective, and initialization stage. Different KL objectives also affect models differently.
    \item \textbf{Initial signal statistics and training dynamics identify collapse but do not predict downstream performance.} Initial forward KL does not predict the direction of performance change, and numerically stable optimization can still degrade performance. Failure modes include collapse, ineffective length growth, and broken formatting. Loss- and entropy-based token-selection methods do not provide a consistent remedy across models.
\end{itemize}

\section{Related Work}
\label{sec:related_work}
  
\noindent\textbf{RLVR.}
Reinforcement learning with verifiable rewards (RLVR) optimizes reasoning policies using automatically verified outcome rewards and has been widely adopted for reasoning post-training~\citep{deepseek_r1,dapo}. Among these methods, GRPO~\citep{deepseek_math} samples $G$ responses for each question $Q$ and replaces PPO's learned critic~\citep{ppo} with group-normalized advantages:
\begin{equation}
{\small
\begin{gathered}
\mathcal{J}_{\mathrm{GRPO}}(\theta)
= \mathbb{E}\!\Biggl[\frac{1}{G}\sum_{g=1}^{G}\frac{1}{|y^{S,g}|}\sum_i
\Bigl(
\min\!\left(\rho_{g,i}\hat A_g,
\operatorname{clip}(\rho_{g,i},1-\varepsilon,1+\varepsilon)\hat A_g\right)
-\beta D_{\mathrm{KL}}(\pi_\theta\|\pi_{\mathrm{ref}})\Bigr)\Biggr],\\
\hat A_g=\frac{r_g-\bar r}{s_r+\epsilon},\qquad
\rho_{g,i}=\frac{\pi_\theta(y_i^{S,g}\mid Q,y_{<i}^{S,g})}
{\pi_{\mathrm{old}}(y_i^{S,g}\mid Q,y_{<i}^{S,g})}.
\end{gathered}
}
\label{eq:grpo_related}
\end{equation}
\noindent Here, the verifiable outcome reward $r_g$ determines a sequence-level advantage shared across the response. DAPO introduces Clip-Higher, Dynamic Sampling, Token-Level Policy Gradient Loss, and Overlong Reward Shaping \citep{dapo}. GSPO defines its importance ratio using sequence likelihood and applies clipping at the sequence level \citep{gspo}. Recent analyses identify response-length bias in GRPO and suggest that RLVR often reinforces successful trajectories already supported by the base model rather than inducing entirely new reasoning capabilities \citep{understand_r1_zero,does_rl_really_incentivize_reasoning}. Although RLVR has been widely used in reasoning post-training \citep{deepseek_math,deepseek_r1,dapo}, it typically requires on-policy rollouts and reward based optimization for each update. This motivates dense alternatives that provide token-level supervision on the student's own trajectories.

\noindent\textbf{Following work of OPSD.}
On-policy self-distillation (OPSD) provides a dense distributional target at every visited student prefix, in contrast to the sequential level outcome signal used by RLVR. Let $Q$ denote a question, $C$ privileged context, and $y^S\sim\pi_\theta(\cdot\mid Q)$ a student-generated rollout. At response position $i$, the student and frozen teacher distributions are
\[
\pi_i^S(v)
=
\pi_\theta(v\mid Q,y_{<i}^S),
\qquad
\pi_i^T(v)
=
\pi_{\theta_0}(v\mid Q,C,y_{<i}^S),
\]
respectively. Vanilla OPSD minimizes their divergence on the on-policy trajectory:
\begin{equation}
\mathcal{L}_{\mathrm{OPSD}}(\theta)
=
\mathbb{E}_{(Q,C)\sim\mathcal{D},\,
y^S\sim\pi_\theta(\cdot\mid Q)}
\left[
\frac{1}{N}\sum_{i=1}^{N}
D\!\left(\pi_i^T,\pi_i^S\right)
\right].
\label{eq:opsd_related}
\end{equation}
The teacher is initialized from the initial student model and receives privileged information unavailable to the student \citep{opsd}. Unlike RLVR's supervision which originates from sequential level reward, OPSD directly transfers the teacher's next-token distribution at each token generated by the student.
Recent work extends OPSD by changing which part of the signal should be trusted. Position-Weighted OPSD assigns greater weight to positions with more reliable teacher supervision \citep{pw_opsd}. Purified OPSD applies a pointwise mutual information correction to reduce bias in the privileged target \citep{purified_opsd}. $\beta$-OPSD interpolates the privileged teacher and reference distributions and incorporates return-to-go credit assignment \citep{beta_opsd}.

Other methods combine dense self-distillation with correctness or reward signals. RLSD uses rewards to determine the direction of token updates and the privileged teacher--student gap to scale their magnitude \citep{rlsd}. SRPO routes correct responses to GRPO and failed responses to self-distillation \citep{srpo}. These approaches provide dense signals from self-distillation as a complement to RLVR.

\noindent\textbf{Rethinking OPD and OPSD.}
As OPD and OPSD have been applied more broadly, some work examines when dense on-policy supervision improves reasoning and when it instead changes behavior or degrades the underlying policy. Privileged supervision can harm reasoning models when applied to long trajectories \citep{opsd_rethink}. Correct reference solutions are not always beneficial, and apparent improvements may reflect recovery of reasoning behavior already present in the base model rather than acquisition of new capabilities \citep{opsd_privileged_rethink}. \citet{rethink_opd} find that effective OPD requires compatible teacher--student thinking patterns and a teacher with genuinely stronger capabilities. \citet{many_faces_opd} further show that OPSD is sensitive to teacher and loss design, and question whether the student actually internalizes privileged information.

Beyond documenting these results, mechanistic analyses of OPD attribute them to signal quality, teacher--student mismatch, and trajectory dynamics. \citet{demystifying_opd_regulations} interpret OPD as an exploration catalyst whose effectiveness is limited by signal quality, teacher--student mismatch, and length exploitation. Rich privileged context may suppress verbalized uncertainty and thereby harm generalization \citep{self_distill_degrade}. Teacher supervision can concentrate near the beginning of a response \citep{fast_prefix_opd}, while increasing prefix drift makes later teacher feedback less relevant to the student's trajectory \citep{prune_opd}. Length inflation, repetition, and truncation can consequently produce training collapse \citep{demystifying_opd_length}.

Our study builds on prior work in two ways. First, we extend several diagnostic perspectives developed for OPD to the OPSD setting. Second, we provide a joint and comprehensive diagnosis of OPSD across reasoning mode, context content, rollout length, divergence loss, model ability, and post-training stage, spanning downstream outcomes, frozen token-level signals, and training dynamics. Together, these diagnostics provide a comprehensive empirical basis for designing and evaluating future OPSD training.

\section{Preliminaries}
\label{sec:preliminaries}
Let $Q\sim\mathcal{D}$ denote a question and $y^S=(y_1^S,\ldots,y_N^S)$ a student-generated rollout, with $y_{<i}^S$ denoting the student prefix before position $i$. $S$ and $T$ identify student and teacher quantities, respectively.

\subsection{On-Policy Distillation (OPD)}
\label{subsec:opd}

For a student rollout $y^S\sim\pi_\theta(\cdot\mid Q)$, let $\pi_i^S(v)=\pi_\theta(v\mid Q,y_{<i}^S)$ and $\pi_i^T(v)=\pi_\phi(v\mid Q,y_{<i}^S)$ denote the student and fixed teacher distributions on the visited student prefix $y_{<i}^S$. OPD minimizes
\begin{equation}
\label{eq:opd_obj}
\mathcal{L}_{\text{OPD}}(\theta)
=
\mathbb{E}_{Q\sim\mathcal{D},\,y^S\sim\pi_\theta(\cdot\mid Q)}
\left[
\frac{1}{N}\sum_{i=1}^{N}
D\!\left(\pi_i^T,\,\pi_i^S\right)
\right].
\end{equation}
Here $D$ is a divergence between next-token distributions. Each rollout is sampled from the current student. At every response position, the teacher conditions on the student-generated prefix and computes its next-token distribution. This provides dense supervision for OPD along the rollout trajectory. Gradients are not propagated through the rollout sampling process.

\subsection{On-Policy Self-Distillation (OPSD)}
\label{subsec:opsd}

OPSD \citep{opsd} specializes OPD by using a frozen copy $\pi_{\theta_0}$ of the initial student as the teacher and providing it with privileged context $C$. For each pair $(Q,C)$, the student conditions only on $Q$, whereas the teacher conditions on $Q\oplus C:=\mathcal{T}_{\mathrm{teacher}}(Q,C)$. The privileged information $C$ may contain a reference solution or verified answer unavailable to the student. On a student-generated prefix $y_{<i}^S$, the teacher distribution is $\pi_i^T(v)=\pi_{\theta_0}(v\mid Q,C,y_{<i}^S)$, while the student distribution remains as in OPD. 
OPSD instantiates $D$ in Equation~\ref{eq:opd_obj} as a clipped forward KL. It caps each vocabulary contribution $d_{i,v}=\pi_i^T(v)\bigl[\log\pi_i^T(v)-\log\pi_i^S(v)\bigr]$ before summation:
\begin{equation}
\label{eq:opsd_clipped}
\mathcal{L}_{\text{OPSD}}^{\text{clip}}(\theta)
=
\mathbb{E}_{(Q,C)\sim\mathcal{D},\,y^S\sim\pi_\theta(\cdot\mid Q)}
\left[
\frac{1}{N}\sum_{i=1}^{N}
\sum_{v\in\mathcal{V}}
\min\!\left(d_{i,v},\tau_{\text{clip}}\right)
\right].
\end{equation}
The clipping limits large contributions from individual vocabulary items.

\begin{figure}[t]
    \centering
    \includegraphics[width=\textwidth]{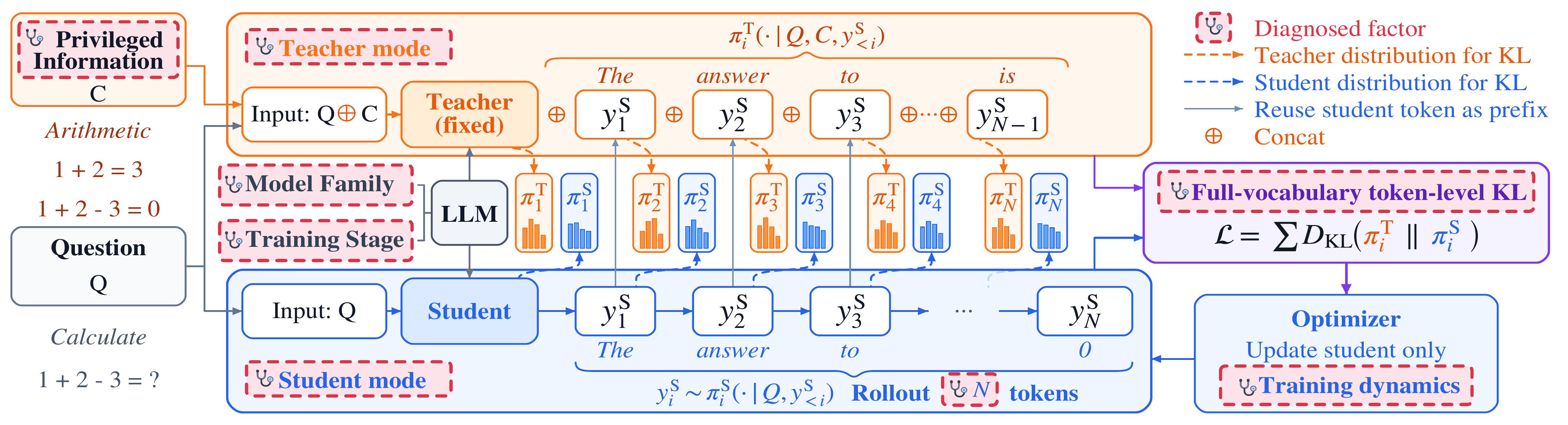}\par
    \vspace{-\parskip}
    \caption{Overview of our diagnosis. We design a comprehensive diagnostic pipeline to study how OPSD behaves in different settings and what can token-level dynamics reveal during training.}
    \label{fig:overview}
\end{figure}

\section{Empirical Methodology and Setup}
\label{sec:setup}

We write $m_S,m_T\in\{\nth,\thmode\}$ for the student and teacher training or inference modes. 
The student uses $m_S$ for on-policy generation. 
The teacher uses $m_T$ when scoring the student's rollout. 
In the paper, we abbreviate \textsc{Think} as \thmode{} and \textsc{No-think} as \nth{}.
We mainly train Qwen3 and OLMo-3 models spanning 0.6B--8B parameters. 
We additionally use DeepSeek-R1-Distill-Qwen-1.5B \citep{deepseek_r1}, Falcon-H1R-7B \citep{falcon-h1r}, MiMo-7B-RL \citep{mimo_7b_rl}, Qwen3.5-4B \citep{qwen3.5}, and Qwen3-4B-Thinking-2507(Qwen3-4B-Thinking,\cite{qwen3}) for additional tests.

The main OPSD experiments mainly use OpenThoughts Math 30k \citep{opsd}.
Default OPSD setup uses a frozen same-model teacher, full solution teacher context. We set 1024-token prompt and rollout caps, 100 training steps, learning rate $10^{-6}$, forward KL, $\tau_{\mathrm{clip}}=0.05$, seed~42, and AdamW\citep{adamw} optimizer.
We evaluate on AIME~2024, AIME~2025, AIME~2026, and HMMT~February~2025, which are sufficiently challenging to evaluate model reasoning ability. 
We sample eight responses per question. Evaluation metrics are defined in Appendix~\ref{app:evaluation_details}. Complete model-specific training and evaluation defaults are given in Appendix~\ref{app:training_details}. Detailed Results are given in Appendix~\ref{app:per_dataset_results}.

For several experiments, we additionally test whether the observed conclusions hold on coding tasks.
For the default coding OPSD experiments, we train on OpenCodeReasoning \citep{opencodereasoning} with a 2048-token prompt cap and use the solution code as privileged information.
The remaining OPSD defaults follow the mathematical reasoning setup above unless otherwise specified.
We evaluate on three coding benchmarks: HumanEvalPlus \citep{humanevalplus_and_mbppplus}, MBPPPlus \citep{humanevalplus_and_mbppplus}, and LiveCodeBench v6 \citep{livecodebench}.
We sample four responses per problem.
Experiment configurations are described in the corresponding sections.

We use several phrases to describe model's some performances. \textbf{Ineffective deliberation} denotes little decline of accuracy accompanied by substantial evaluation length growth. \textbf{Stable degradation} denotes that as training step or student rollout length increases, accuracy declines gradually while evaluation rollout length grows. \textbf{Behavioral collapse} denotes a large performance degradation accompanied by failures such as repetition, hallucination, missing of \texttt{<eos>} token, corrupted text, or loss of boxed answer formatting.

\section{Experimental Results}
\label{sec:diagnosis}

We first establish a valid comparison setting. We then examine training steps, rollout length, teacher context, loss design, model ability, pipeline stage, and OPSD versus GRPO. \autoref{fig:overview} summarizes the experimental pipeline. Per-dataset results and robustness checks are in Appendix~\ref{app:per_dataset_results} and Appendix~\ref{app:seed_sensitivity}.

\subsection{Valid Comparison Setting}
\label{subsec:mode_results}

\noindent Reasoning mode mismatch confounds attribution because Qwen3's \thmode{} and \nth{} modes differ in both reasoning ability and response style. It's necessary to confirm the training mode.

\begin{table}[t]
  \caption{Controls for establishing a valid comparison setting. \textbf{(a)} Reasoning mode results averaged over Qwen3-1.7B/4B/8B. For Avg@8 and Pass@8, $\Delta$ is the absolute OPSD--Base change in percentage points. For Length, it is the relative change. Eval is the inference mode of both Base and OPSD, and S/T denote student/teacher modes. \textbf{(b)} Qwen3-4B feature rates. Self-correction includes Hmm/Actually/No/But. \textbf{(c)} No-privilege same-prefix control for mismatched \nth{}/\thmode{}, evaluated in \thmode{} mode. Parentheses show change from the corresponding full solution checkpoint. Per-model results are in \autoref{tab:qwen_mode_extended}. Per-dataset results are in \autoref{tab:mode_per_dataset}.}
  \label{tab:qwen_mode}
  \label{tab:mode_discourse}
  \label{tab:qwen_mode_no_privilege}
  \vspace{4pt}
  \centering
  \setlength{\tabcolsep}{3.5pt}
  \scriptsize
  \textbf{(a) Reasoning mode comparison}\par\vspace{2pt}%
  \providecommand{\opsda}[1]{\makebox[2.85em][c]{#1}}
  \providecommand{\opsdb}[1]{\makebox[3.55em][c]{#1}}
  \providecommand{\opsdc}[1]{\makebox[4.2em][c]{#1}}
  \begin{tabular*}{\linewidth}{@{\extracolsep{\fill}}lccc ccc c ccc c ccc @{}}
  \toprule
  \multirow{2}{*}[-2pt]{Model} & \multirow{2}{*}[-2pt]{S} & \multirow{2}{*}[-2pt]{T} & \multirow{2}{*}[-2pt]{Eval}
    & \multicolumn{3}{c}{Avg@8} & & \multicolumn{3}{c}{Pass@8} & & \multicolumn{3}{c}{Length} \\
  \cmidrule(lr){5-7}\cmidrule(lr){9-11}\cmidrule(lr){13-15}
  & & & & \opsda{Base} & \opsda{OPSD} & \opsda{$\Delta$} & & \opsda{Base} & \opsda{OPSD} & \opsda{$\Delta$} & & \opsdb{Base} & \opsdb{OPSD} & \opsdc{$\Delta$} \\
  \midrule
  \multirow{6}{*}{Qwen3 mean} & \thmode & \thmode & \thmode & \opsda{53.5} & \opsda{50.7} & \opsda{$-2.8$}  & & \opsda{72.2} & \opsda{71.1} & \opsda{$-1.1$}  & & \opsdb{18013} & \opsdb{21250} & \opsdc{$+17.9\%$} \\
  & \thmode & \nth & \thmode & \opsda{53.5} & \opsda{36.6} & \opsda{$-16.9$} & & \opsda{72.2} & \opsda{59.5} & \opsda{$-12.7$} & & \opsdb{18013} & \opsdb{13291} & \opsdc{$-26.0\%$} \\
  & \nth & \thmode & \nth & \opsda{15.8} & \opsda{19.6} & \opsda{$+3.8$}  & & \opsda{32.2} & \opsda{39.7} & \opsda{$+7.5$}  & & \opsdb{5510}  & \opsdb{14093} & \opsdc{$+154.8\%$} \\
  & \nth & \nth & \nth & \opsda{15.8} & \opsda{15.2} & \opsda{$-0.6$}  & & \opsda{32.2} & \opsda{28.6} & \opsda{$-3.6$}  & & \opsdb{5510}  & \opsdb{13634} & \opsdc{$+144.3\%$} \\
  & \nth & \thmode & \thmode & \opsda{53.5} & \opsda{56.2} & \opsda{$+2.7$}  & & \opsda{72.2} & \opsda{73.1} & \opsda{$+0.9$}  & & \opsdb{18013} & \opsdb{19527} & \opsdc{$+8.4\%$} \\
  & \thmode & \nth & \nth & \opsda{15.8} & \opsda{14.0} & \opsda{$-1.8$}  & & \opsda{32.2} & \opsda{27.8} & \opsda{$-4.4$}  & & \opsdb{5510}  & \opsdb{5104}  & \opsdc{$-8.7\%$} \\
  \bottomrule
  \end{tabular*}
  \vspace{5pt}

  \begin{tabular}{@{}>{\centering\arraybackslash}p{0.49\linewidth}@{\hspace{0.02\linewidth}}>{\centering\arraybackslash}p{0.49\linewidth}@{}}
  \textbf{(b) Qwen3-4B feature rates}\par\vspace{2pt}%
  {\setlength{\tabcolsep}{1.4pt}%
  \renewcommand{\arraystretch}{1.03}%
  \begin{tabular*}{\linewidth}{@{\extracolsep{\fill}}lcccc@{}}
  \toprule
  Stylistic feature & \nth{} Base & \nth{}/\thmode{} & \nth{}/\nth{} & \thmode{} base \\
  \midrule
  Contains ``wait'' & 31.6\% & 47.0\% & 18.2\% & 100.0\% \\
  Self-correction & 35.7\% & 49.9\% & 21.0\% & 99.7\% \\
  Contains \texttt{<think>} & 0.0\% & 0.1\% & 0.0\% & 100.0\% \\
  \bottomrule
  \end{tabular*}}
  &
  \textbf{(c) No-privilege control}\par\vspace{2pt}%
  {\setlength{\tabcolsep}{2.0pt}%
  \renewcommand{\arraystretch}{1.03}%
  \begin{tabular*}{\linewidth}{@{\extracolsep{\fill}}lccc@{}}
  \toprule
  Model & Avg@8 & Pass@8 & Length \\
  \midrule
  Qwen3-1.7B & 41.0\,(+0.2) & 62.5\,(-0.8) & 20486\,(+2.4\%) \\
  Qwen3-4B   & 60.9\,(-2.1) & 75.0\,(-4.2) & 19715\,(+2.2\%) \\
  Qwen3-8B   & 64.3\,(-0.5) & 80.8\,(+4.1) & 19699\,(+2.1\%) \\
  \bottomrule
  \end{tabular*}}
  \end{tabular}
\end{table}

\autoref{tab:qwen_mode}, panel (a), shows a directional effect: a \nth{} teacher suppresses a \thmode{} student's accuracy and length, whereas a \thmode{} teacher increases both for a \nth{} student. Although there is no explicit \texttt{<think>} leakage, the latter still shows more \thmode{}-associated discourse features (panel (b)).

\noindent The no-privilege same-prefix control in panel (c) supports this interpretation. Across model sizes, the no-privilege versus full solution means differ only slightly (55.4/72.8/20.0k vs 56.2/73.1/19.5k), so the apparent gain primarily reflects mode transfer rather than privileged semantics.

\noindent Because mismatch separates the rollout and evaluation policies and relies on a Qwen-specific mode switch, making comparisons across model families unavailable, all subsequent experiments are trained and tested on matched native modes, with matched \thmode{}/\thmode{} for hybrid reasoning models.

\subsection{Training Steps and Maximum Student Rollout Length}
\label{subsec:training_steps}
\label{subsec:length_results}

\begin{figure}[t]
\centering
\includegraphics[width=\linewidth]{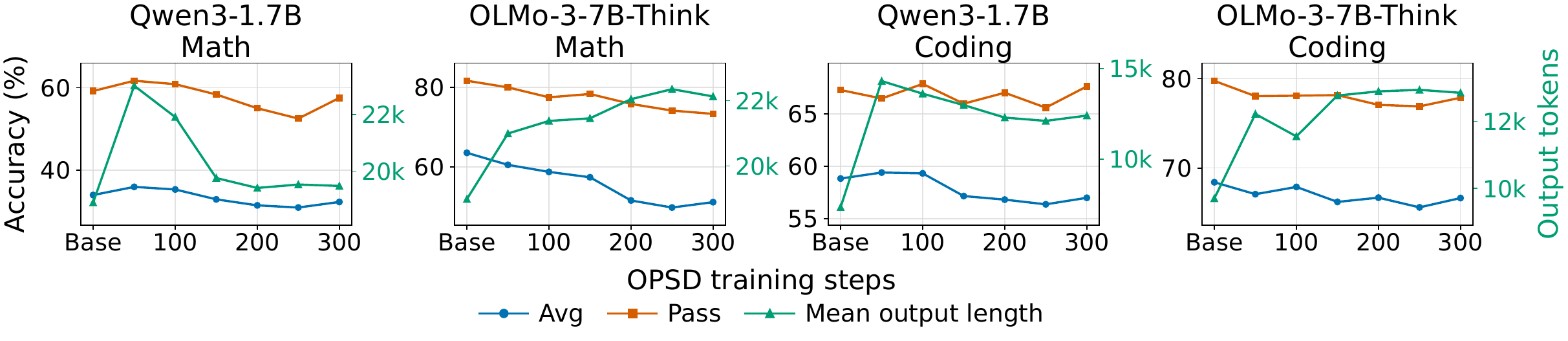}
\vspace{-14pt}
\setlength{\abovecaptionskip}{0pt}
\caption{Stability over training steps on math and coding tasks. Math and coding per-dataset results are in \autoref{tab:opsd_extend300_per_dataset} and \autoref{tab:opsd_extend300_coding_per_dataset}, respectively.}
\label{fig:opsd_extend300}
\par\vspace{8pt}
\includegraphics[width=\linewidth,trim=7pt 7pt 7pt 7pt,clip]{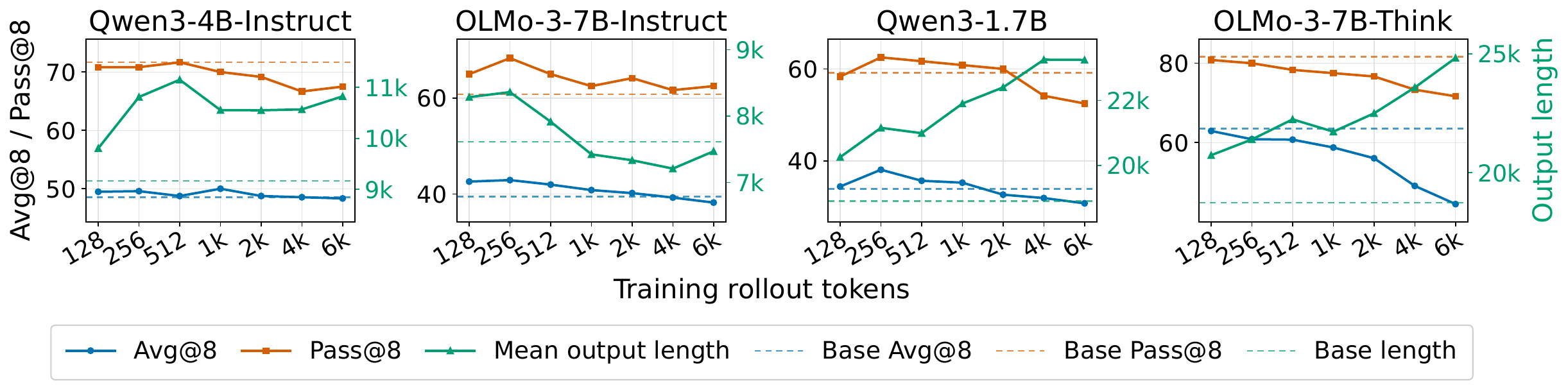}
\vspace{-6pt}
\caption{OPSD vs.\ rollout cap for four models. Solid: OPSD. Dashed: Base. Left axes: Avg@8/Pass@8. Right: Length. Per-dataset results are in \autoref{tab:length_per_dataset_nothink} and \autoref{tab:length_per_dataset}.}
\label{fig:length_sweeps}
\end{figure}

We first ask whether more training steps can improve final performance. We train Qwen3-1.7B and OLMo-3-7B-Think for up to 300 steps on math and coding and evaluate every 50 steps (\autoref{fig:opsd_extend300}).

Across both math and coding tasks, the trends are similar. Avg and Pass generally decline as the number of training steps increases, while output length increases in nearly all settings. These results indicate that OPSD performance gradually degrades with training steps growing.

\newsavebox{\arxivpositionselectiontab}
\newsavebox{\arxivpositionselectionwrap}
\sbox{\arxivpositionselectiontab}{%
\scriptsize
\renewcommand{\arraystretch}{1.04}%
\setlength{\tabcolsep}{1.8pt}%
\begin{tabular}{@{}lccc@{}}
\toprule
Positions & Avg@8 & Pass@8 & Length \\
\midrule
First-256 & \textbf{49.5}\,(+0.7) & 71.3\,(+0.8) & \textbf{21.4k}\,(+14.7\%) \\
Uni-256 & 47.1\,(-1.7) & \textbf{72.1}\,(+1.6) & 22.0k\,(+17.9\%) \\
Final-256 & 47.8\,(-1.0) & 69.2\,(-1.3) & 22.4k\,(+20.1\%) \\
\bottomrule
\end{tabular}
}
\sbox{\arxivpositionselectionwrap}{%
  \begin{minipage}[t]{\wd\arxivpositionselectiontab}
  \vspace{0pt}%
  \wrapfloatcaptionsetup
  \setlength{\abovecaptionskip}{0pt}%
  \setlength{\belowcaptionskip}{6pt}%
  \raggedright
  \captionof{table}{Results of position loss. Parentheses: change from Base. Per-dataset results are in \autoref{tab:position_selection_per_dataset}.}
  \label{tab:position_selection_summary}
  \usebox{\arxivpositionselectiontab}%
  \end{minipage}}

\begingroup
\dimen0=\dimexpr\linewidth-\wd\arxivpositionselectionwrap-10pt\relax
\parshape=1 0pt \dimen0
\noindent\raisebox{\ht\strutbox}[0pt][0pt]{%
  \makebox[0pt][l]{\hspace{\dimexpr\linewidth-\wd\arxivpositionselectionwrap\relax}\usebox{\arxivpositionselectionwrap}}%
}We next compare loss token positions with a 256-position supervision budget. First-256 uses the equivalent OPSD-256 condition, while Uni-256 and Final-256 use a 1024-token rollout and retain uniformly sampled or final valid response positions (Appendix~\ref{app:seed_position}). First-256 gives the highest Avg@8 and shortest outputs while Final-256 performs the worst (\autoref{tab:position_selection_summary}).
\par
\parshape=3 0pt \dimen0 0pt \dimen0 0pt \linewidth
\noindent With the comparison restricted to matched modes, we then vary the maximum student training rollout length over $\{128$, $256$, $512$, $1024$, $2048$, $4096$, $6144\}$ tokens(\autoref{fig:length_sweeps}). For instruct models, performance decreases as rollout length increases: short or intermediate rollout lengths are competitive with or better than the base, whereas the longest settings lose accuracy.
\par
\endgroup
\noindent Under matched \thmode{}/\thmode{}, both models exhibit stable degradation. Prior work reports that privileged OPSD does not improve reasoning models. We find that small gains remain possible, but only for some models and only at short rollout lengths.
\par

\subsection{Privileged Information}
\label{sec:signal_interventions}
\label{subsec:teacher_context}
We next vary privileged information to test whether target-specific information alone explains training utility. 
The primary conditions use target problems from OpenThoughts Math 30k. 
Besides the full solution, \textsc{Answer} provides only the correct final answer, while \textsc{Unrelated} provides a deduplicated problem--solution pair from DAPO-Math-17k \citep{dapo}. 
The unrelated pair changes the privileged information without providing evidence relevant to the target problem. 
The student receives only the target problem in all conditions.

For context type $z\in\{\textsc{Solution},\textsc{Answer},\textsc{Unrelated}\}$, the teacher input is $Q  ,  C^{(z)}:=\mathcal{T}_{\mathrm{teacher}}(Q,C^{(z)})$. Because the intervention changes a complete prefix, it estimates the overall context effect rather than separately identifying length or target problem position. We vary these three contexts across five model settings at maximum rollout lengths of 256 and 1024 tokens (\autoref{tab:teacher_context}).

\begin{table}[ht]
\caption{Effect of privileged information averaged across five model settings. The five models are Qwen3-1.7B, Qwen3-4B, Qwen3-4B-Instruct, OLMo-3-7B-Think, and OLMo-3-7B-Instruct. Parentheses report change from Base in percentage points for Avg and Pass and relative percent for Length. Per-model and per-dataset results are in \autoref{tab:context_per_dataset}.}
\label{tab:teacher_context}
\centering
\setlength{\tabcolsep}{3.0pt}
\fontsize{8pt}{9.6pt}\selectfont
\providecommand{\tcacc}[2]{#1\,(#2)}
\providecommand{\tclen}[2]{#1\,(#2)}
\begin{tabular}{@{}l ccc ccc@{}}
\toprule
\multirow{2}{*}[-2pt]{Teacher context}
& \multicolumn{3}{c}{Rollout length: 256}
& \multicolumn{3}{c}{Rollout length: 1024} \\
\cmidrule(lr){2-4}\cmidrule(lr){5-7}
& Avg@8 & Pass@8 & Length & Avg@8 & Pass@8 & Length \\
\midrule
Full solution      & \tcacc{\textbf{50.5}}{$+0.9$} & \tcacc{\textbf{71.5}}{$+1.0$} & \tclen{16.2k}{$+12.9\%$} & \tcacc{\textbf{48.6}}{$-1.0$} & \tcacc{\textbf{69.3}}{$-1.2$} & \tclen{16.4k}{$+12.2\%$} \\
Unrelated solution & \tcacc{48.9}{$-0.7$} & \tcacc{69.7}{$-0.8$} & \tclen{15.4k}{$+6.0\%$} & \tcacc{48.5}{$-1.0$} & \tcacc{67.7}{$-2.8$} & \tclen{18.0k}{$+24.5\%$} \\
Answer only        & \tcacc{45.9}{$-3.7$} & \tcacc{65.7}{$-4.8$} & \tclen{16.3k}{$+6.4\%$} & \tcacc{44.6}{$-5.0$} & \tcacc{65.7}{$-4.8$} & \tclen{17.7k}{$+15.1\%$} \\
\bottomrule
\end{tabular}
\end{table}
\vspace{\dimexpr-\parskip+5pt\relax}

\newsavebox{\longcontexttab}
\newlength{\contextcontrolwidth}
\setlength{\contextcontrolwidth}{0.44\linewidth}
\sbox{\longcontexttab}{%
\fontsize{7.2pt}{7.8pt}\selectfont
\setlength{\tabcolsep}{1.3pt}%
\renewcommand{\arraystretch}{0.96}%
\begin{tabular*}{\contextcontrolwidth}{@{\extracolsep{\fill}}l*{6}{c}@{}}
\toprule
& \multicolumn{3}{c}{Qwen3-1.7B} & \multicolumn{3}{c}{OLMo-3-7B-Think} \\
\cmidrule(lr){2-4}\cmidrule(lr){5-7}
Teacher context & Avg@8 & Pass@8 & Length & Avg@8 & Pass@8 & Length \\
\midrule
None (Base)        & 34.0 & 59.2 & 18.6k & 63.5 & 81.7 & 18.7k \\
Answer only        & 35.1 & 58.3 & 26.5k & 57.6 & \textbf{80.8} & 24.2k \\
Unrelated solution & \textbf{36.8} & 55.8 & 22.5k & \textbf{59.5} & 77.5 & 23.2k \\
Full solution      & 35.3 & \textbf{60.8} & 22.4k & 58.8 & 77.5 & 21.7k \\
CoT + solution     & 30.8 & 55.8 & 18.6k & 58.6 & 77.5 & 20.8k \\
\bottomrule
\end{tabular*}%
}
\newsavebox{\answeronlytab}
\newsavebox{\contextcontrolswrap}
\newcount\contextremaininglines
\newcount\contextlineheight
\sbox{\answeronlytab}{%
\fontsize{7.2pt}{7.8pt}\selectfont
\setlength{\tabcolsep}{1.3pt}%
\renewcommand{\arraystretch}{0.96}%
\begin{tabular*}{\contextcontrolwidth}{@{\extracolsep{\fill}}l*{6}{c}@{}}
\toprule
& \multicolumn{3}{c}{Qwen3-1.7B} & \multicolumn{3}{c}{OLMo-3-7B-Think} \\
\cmidrule(lr){2-4}\cmidrule(lr){5-7}
Answer condition & Avg@8 & Pass@8 & Length & Avg@8 & Pass@8 & Length \\
\midrule
None (Base)       & 34.0 & 59.2 & 18.6k & 63.5 & 81.7 & 18.7k \\
Random problems   & 35.7 & 53.3 & 27.9k & 58.4 & 79.2 & 24.8k \\
Correct integer   & 39.9 & 63.3 & 23.6k & 57.4 & 76.7 & 24.1k \\
\midrule
Equivalent form   & 41.5 & 64.2 & 24.0k & 58.4 & 80.8 & 23.8k \\
Wrong (fixed)     & 42.2 & 68.3 & 23.6k & 58.9 & 77.5 & 23.6k \\
Wrong (random)    & 40.8 & 61.7 & 23.6k & 56.9 & 79.2 & 23.8k \\
\bottomrule
\end{tabular*}}
\sbox{\contextcontrolswrap}{%
  \begin{minipage}[t]{\contextcontrolwidth}
  \vspace{0pt}%
  \setlength{\parskip}{0pt}%
  \wrapfloatcaptionsetup
  \setlength{\abovecaptionskip}{2pt}%
  \setlength{\belowcaptionskip}{2pt}%
  \scriptsize
  \raggedright
  \captionof{table}{Results at a 1024-token rollout cap. Per-dataset results are in \autoref{tab:context_per_dataset}.}
  \label{tab:teacher_context_1p7b_cot}%
  \noindent\usebox{\longcontexttab}\par
  \vspace{4pt}%
  \captionof{table}{Answer-only controls at step 50. Settings and per-dataset results are in Appendix~\ref{app:answer_only_controls}.}
  \label{tab:answer_only_controls}%
  \noindent\usebox{\answeronlytab}\par
  \end{minipage}}
\Needspace{\dimexpr\ht\contextcontrolswrap+\dp\contextcontrolswrap\relax}
\begingroup
\dimen0=\dimexpr\linewidth-\wd\contextcontrolswrap-6pt\relax
\contextremaininglines=\dimexpr\ht\contextcontrolswrap+\dp\contextcontrolswrap-\parskip+\baselineskip-1sp\relax
\contextlineheight=\baselineskip
\divide\contextremaininglines by \contextlineheight
\parshape=1 0pt \dimen0
\noindent\raisebox{\ht\strutbox}[0pt][0pt]{%
  \makebox[0pt][l]{\hspace{\dimexpr\linewidth-\wd\contextcontrolswrap\relax}\usebox{\contextcontrolswrap}}%
}Context variants exhibit a non-monotonic ordering.
At both rollout lengths in \autoref{tab:teacher_context}, Unrelated solution outperforms Answer only.
Furthermore, we apply \textsc{CoT+Solution} to Qwen3-1.7B and OLMo-3-7B-Think by concatenating the OpenThoughts CoT trace with the gold solution on the teacher side, using exactly the same target problems and row order as Full solution while increasing the teacher prompt budget from 1024 to 8192 tokens.
For Qwen3-1.7B, Avg@8 and Pass@8 both decrease. For OLMo-3-7B-Think, neither metric improves (\autoref{tab:teacher_context_1p7b_cot}).
Thus, providing more detailed teacher-side information does not necessarily yield better training performance.
\par
\advance\contextremaininglines by -\prevgraf
\ifnum\contextremaininglines<0 \contextremaininglines=0 \fi
\edef\contextparshape{\the\numexpr\contextremaininglines+1\relax}
\loop\ifnum\contextremaininglines>0
  \edef\contextparshape{\contextparshape\space 0pt \the\dimen0}%
  \advance\contextremaininglines by -1
\repeat
\edef\contextparshape{\contextparshape\space 0pt \the\linewidth}%
\expandafter\parshape\contextparshape\relax
\noindent For Answer Only, we examine three transformations of the supplied answers on an integer-answer subset. (i) Simple reformulations while keeping the value unchanged, such as equivalent fractions or decimals. (ii) Wrong answers obtained through a fixed mapping. (iii) Wrong answers obtained through randomized mappings. We compare these variants with three baselines: the base model, Answer Only on randomly sampled problems, and Answer Only on the same integer answer subset(\autoref{tab:answer_only_controls}). The results show that providing the original correct integers does not consistently yield better Avg@8 or Pass@8 than the transformed answers. Moreover, OLMo-3-7B-Think remains degradation across all tested conditions, indicating that changing the privileged answers does not resolve the lack of improvement for this model. Detailed settings and results are reported in Appendix~\ref{app:answer_only_controls}.
\par
\endgroup

\subsection{Loss Design}
\label{subsec:divergence_objectives}
Following the OPSD objective \citep{opsd}, at position $i$ the interior branch $0<\beta<1$ uses the generalized-JSD loss
\begin{equation}
\label{eq:jsd_beta}
\ell_{\beta}(\pi_i^S,\pi_i^T)
=
\beta D_{\mathrm{KL}}(\pi_i^T\|m_{\beta,i})
+(1-\beta)D_{\mathrm{KL}}(\pi_i^S\|m_{\beta,i}),
\quad
m_{\beta,i}=(1-\beta)\pi_i^S+\beta\pi_i^T.
\end{equation}
The endpoints are defined separately:
\[
\ell_0 = D_{\mathrm{KL}}(\pi_i^T\|\pi_i^S)\ \text{(forward KL)},
\quad
\ell_1 = D_{\mathrm{KL}}(\pi_i^S\|\pi_i^T)\ \text{(reverse KL)}.
\]
\noindent
\begin{minipage}[t]{0.43\linewidth}
\vspace{0pt}%
We use $\beta=0$ by default and ablate $\beta\in\{0.5,1\}$, which means we use the forward KL as default and ablate the JSD KL and reverse KL respectively.

The result shows that only a few trained checkpoints improve over their corresponding base models, and the leading objective varies by model (\autoref{tab:divergence_main}). Changing divergence direction redistributes outcomes but cannot provide a universal remedy.

Vocabulary truncation, teacher-favored position selection, and student entropy masking do not consistently improve performance across models (Appendix~\ref{subsec:entropy_training} and Appendix~\ref{app:selective_supervision}).
\end{minipage}\hfill
\begin{minipage}[t]{0.55\linewidth}
  \vspace{0pt}%
  \wrapfloatcaptionsetup
  \setlength{\abovecaptionskip}{0pt}%
  \setlength{\belowcaptionskip}{4pt}%
  \raggedright
  \captionof{table}{Divergence-objective sensitivity after 100 steps of OPSD-1024 under different KL losses. Per-model and per-dataset results are in \autoref{tab:divergence_macro} and \autoref{tab:divergence_per_dataset}.}
  \label{tab:divergence_main}%
  \centering
  {\scriptsize
  \setlength{\tabcolsep}{0.5pt}%
  \renewcommand{\arraystretch}{0.92}%
  \begin{tabular*}{\linewidth}{@{\extracolsep{\fill}}l*{6}{>{\centering\arraybackslash}m{21pt}}@{}}
  \toprule
  & \multicolumn{2}{c}{Fwd. KL} & \multicolumn{2}{c}{JSD} & \multicolumn{2}{c}{Rev. KL} \\
  \cmidrule(lr){2-3}\cmidrule(lr){4-5}\cmidrule(lr){6-7}
  Model & Avg & Pass & Avg & Pass & Avg & Pass \\
  \midrule
  Qwen3-1.7B          & \textbf{$+1.3$} & \textbf{$+1.6$} & $-2.7$ & $-4.2$ & $-7.5$ & $-11.7$ \\
  Qwen3-4B            & \textbf{$-4.4$} & \textbf{$-3.4$} & $-5.7$ & \textbf{$-3.4$} & $-30.3$ & $-22.5$ \\
  Qwen3-4B-Thinking   & $-9.6$ & $-3.4$ & $-7.4$ & $-4.2$ & \textbf{$-1.2$} & \textbf{$-1.7$} \\
  OLMo-3-7B-Think     & $-4.7$ & $-4.2$ & $-1.4$ & $0.0$ & \textbf{$-1.1$} & \textbf{$+0.8$} \\
  OLMo-3-7B-Instruct  & $+1.3$ & \textbf{$+1.7$} & $-2.3$ & $-2.5$ & \textbf{$+3.0$} & $+0.9$ \\
  Falcon-H1R-7B       & \textbf{$-0.4$} & \textbf{$0.0$} & $-1.1$ & $-0.9$ & $-1.4$ & \textbf{$0.0$} \\
  MiMo-7B-RL          & $-11.3$ & $-10.0$ & \textbf{$-3.0$} & \textbf{$-1.7$} & $-5.4$ & \textbf{$-1.7$} \\
  \bottomrule
  \end{tabular*}}
\end{minipage}

\subsection{Initial Model Ability}
\label{subsec:initial_capability}

We ask whether OPSD's performance depends on the model itself. We compare Base, OPSD-256, and OPSD-1024 across eleven models on math and coding under their matched student--teacher mode and standard full solution or solution code privileged information.

\begin{figure}[h]
\centering
\includegraphics[width=0.98\textwidth,trim=4pt 12pt 4pt 8pt,clip]{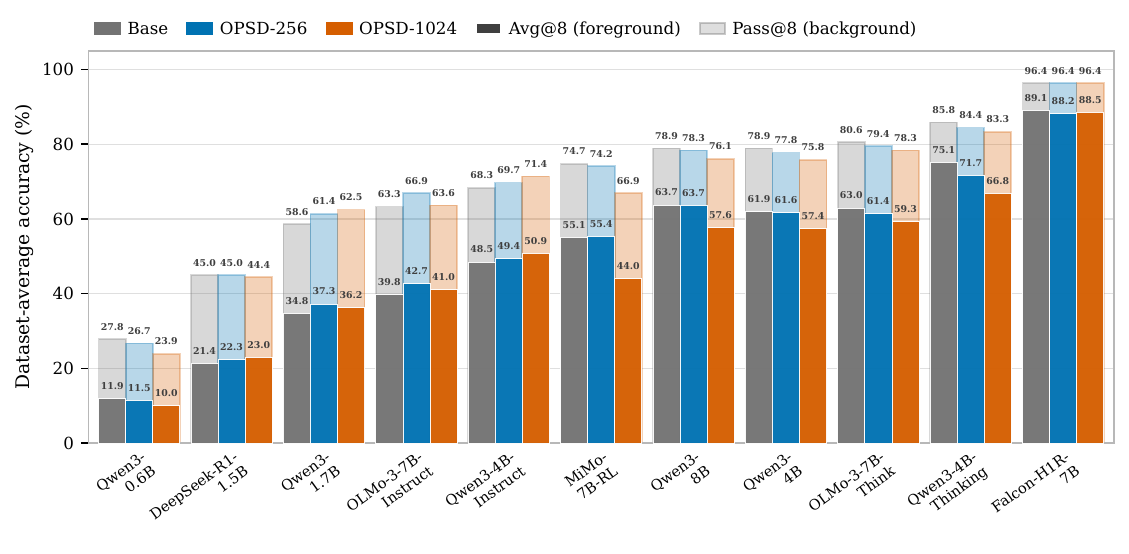}
\par\vspace{-4pt}
{\small (a) Mathematical reasoning}\par\vspace{4pt}
\includegraphics[width=0.98\textwidth,trim=4pt 12pt 4pt 8pt,clip]{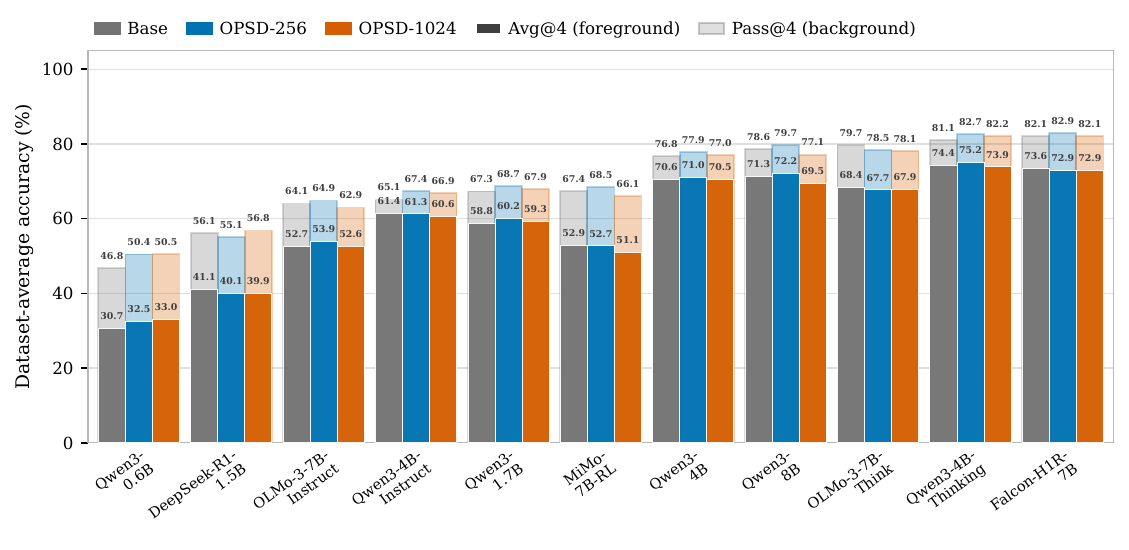}
\par\vspace{-4pt}
{\small (b) Coding}
\caption{OPSD results across baseline performance levels on (a) math and (b) coding. Each model cluster reports Base, OPSD-256, and OPSD-1024. Dark foreground bars show Avg@$k$ and light background bars show Pass@$k$, with $k=8$ for math and $k=4$ for coding. Models are ordered by Base Pass@$k$ within each panel. Per-dataset math and coding results are in \autoref{tab:capability_per_dataset} and \autoref{tab:capability_coding_per_dataset}, respectively.}
\label{fig:initial_capability}
\end{figure}

The effect of initial model ability differs across tasks. On math, stronger models generally show larger performance drops after OPSD (\autoref{fig:initial_capability}a). On coding, stronger models can still benefit, although the improvements are typically modest (\autoref{fig:initial_capability}b). For example, consider Qwen3-8B and Qwen3-4B-Thinking. On math, both show lower Avg@8 under OPSD-256 and decline further under OPSD-1024. On coding, both obtain modest gains under OPSD-256, but their Avg@4 falls below Base under OPSD-1024. These results suggest that OPSD's performance depends on the task and is also associated with the model's initial ability and rollout cap.

\subsection{Sensitivity to Post-Training Stage}
\label{sec:pipeline}

\noindent We next test a separate source of sensitivity: the post-training stage at which OPSD is applied. For math, the SFT stage uses 15k steps on the \texttt{cot} split of OpenMathReasoning, and the GRPO stage uses DAPO-Math-17k. For coding, SFT uses 15k steps on OpenCodeReasoning, and GRPO uses APPS \citep{apps}. OPSD uses OpenCodeReasoning for the Base, SFT, and SFT+GRPO pipeline comparisons, while the released Qwen3 checkpoints use APPS for a matched-data OPSD comparison. We evaluate coding performance on HumanEval+, MBPP+, and LiveCodeBench v6. Full configurations appear in Appendix~\ref{app:sft_configuration} and Appendix~\ref{app:grpo_configuration}.

\begin{table*}[h]
\caption{Post-training pipeline results on mathematical reasoning and coding. More math results are in \autoref{tab:sft_initialization}. Per-dataset math results are in \autoref{tab:sft_grpo_per_dataset} and \autoref{tab:sft_grpo_opsd_per_dataset}. Per-dataset coding results are in \autoref{tab:sft_grpo_coding_per_dataset}.}
\label{tab:sft_grpo_control}
\centering
{\fontsize{6.4pt}{7.0pt}\selectfont
\setlength{\tabcolsep}{1.4pt}%
\renewcommand{\arraystretch}{0.95}%
\begin{minipage}[t]{0.34\textwidth}
\centering
\textbf{Math (Qwen3-4B)}\par\vspace{2pt}
\begin{tabular*}{\linewidth}{@{\extracolsep{\fill}}lcccc@{}}
\toprule
Model & Avg@8 & Pass@8 & Format & Length \\
\midrule
Qwen3-4B-Base & 7.29 & 20.00 & 87.19 & 3.0k \\
$+$ OPSD & 4.79 & 17.50 & 74.59 & 10.3k \\
$+$ SFT & 34.69 & 60.00 & 89.27 & 14.3k \\
$+$ SFT $+$ OPSD & 1.98 & 8.33 & 3.44 & 32.9k \\
$+$ SFT $+$ GRPO & 33.65 & 60.83 & 94.79 & 14.2k \\
$+$ SFT $+$ GRPO $+$ OPSD & 3.54 & 14.17 & 5.62 & 37.6k \\
\midrule
Qwen3-4B & 62.3 & 79.2 & 97.1 & 17.8k \\
$+$ OPSD & 57.9 & 75.8 & 86.4 & 19.9k \\
\bottomrule
\end{tabular*}
\end{minipage}\hfill
\begin{minipage}[t]{0.32\textwidth}
\centering
\textbf{Coding (Qwen3-1.7B)}\par\vspace{2pt}
\begin{tabular*}{\linewidth}{@{\extracolsep{\fill}}lccc@{}}
\toprule
Model & Avg@4 & Pass@4 & Length \\
\midrule
Qwen3-1.7B-Base & 29.62 & 53.06 & 4.2k \\
$+$ OPSD & 25.51 & 48.68 & 5.6k \\
$+$ SFT & 41.79 & 57.09 & 9.7k \\
$+$ SFT $+$ OPSD & 24.29 & 43.49 & 26.0k \\
$+$ SFT $+$ GRPO & 38.21 & 54.14 & 7.5k \\
$+$ SFT $+$ GRPO $+$ OPSD & 22.11 & 43.10 & 23.1k \\
\midrule
Qwen3-1.7B & 58.83 & 67.29 & 7.3k \\
$+$ OPSD & 58.57 & 67.36 & 13.4k \\
\bottomrule
\end{tabular*}
\end{minipage}\hfill
\begin{minipage}[t]{0.32\textwidth}
\centering
\textbf{Coding (Qwen3-4B)}\par\vspace{2pt}
\begin{tabular*}{\linewidth}{@{\extracolsep{\fill}}lccc@{}}
\toprule
Model & Avg@4 & Pass@4 & Length \\
\midrule
Qwen3-4B-Base & 43.47 & 60.55 & 3.4k \\
$+$ OPSD & 33.92 & 58.87 & 8.1k \\
$+$ SFT & 64.18 & 73.50 & 6.3k \\
$+$ SFT $+$ OPSD & 61.25 & 70.30 & 10.5k \\
$+$ SFT $+$ GRPO & 62.18 & 72.32 & 6.2k \\
$+$ SFT $+$ GRPO $+$ OPSD & 58.54 & 70.71 & 10.5k \\
\midrule
Qwen3-4B & 70.62 & 76.83 & 7.0k \\
$+$ OPSD & 69.18 & 77.60 & 10.1k \\
\bottomrule
\end{tabular*}
\end{minipage}}
\end{table*}

\noindent Across math and coding, starting from a pretrained checkpoint and applying OPSD after task-specific SFT or SFT+GRPO does not further improve model capability (\autoref{tab:sft_grpo_control}). 
Instead, performance decreases and can even collapse, with the collapse especially evident on math. 
In contrast to the collapse induced by OPSD, GRPO largely preserves model capability after SFT. 
When OPSD is applied to fully trained released Qwen3 models, however, it produces only small changes in accuracy, accompanied by increased output length. 
Together, these results show that OPSD is more sensitive than SFT or GRPO to the stage and quality of its initialization.

\subsection{Training Compute Tradeoff: OPSD versus GRPO}
\label{sec:opsd_grpo_compute}
Finally, we compare 50-step OPSD and GRPO on Qwen3-1.7B / 4B / 8B in both math and coding. Within each domain, the two methods use exactly the same training datasets in the same order: integer answer problems from OpenMathReasoning for math and problems from APPS for coding. OPSD conditions its teacher on the reference solution, whereas GRPO uses answer or execution based verification. \autoref{tab:opsd_grpo_compute} reports accuracy and training GPU-hours. Full configurations are reported in Appendix~\ref{app:released_grpo}.

\begin{table}[H]
\centering
\caption{Accuracy and training compute for 50-step OPSD and GRPO. GPU h denotes training GPU-hours. Per-dataset math results are in \autoref{tab:released_grpo_per_dataset} and \autoref{tab:omr_opsd_per_dataset}. Per-dataset coding results are in \autoref{tab:apps_opsd_grpo_per_dataset}.}
\label{tab:opsd_grpo_compute}
{\fontsize{6.7pt}{7.5pt}\selectfont
\setlength{\tabcolsep}{0.8pt}%
\renewcommand{\arraystretch}{1.08}%
\begin{tabular*}{\textwidth}{@{\extracolsep{\fill}}l*{8}{c}@{\hspace{6pt}}l*{8}{c}@{}}
\toprule
\multicolumn{9}{c}{\textbf{Math (Avg@8 / Pass@8)}}
& \multicolumn{9}{c}{\textbf{Coding (Avg@4 / Pass@4)}} \\
\cmidrule(lr){1-9}\cmidrule(lr){10-18}
& \multicolumn{2}{c}{Base} & \multicolumn{3}{c}{OPSD$_{50}$} & \multicolumn{3}{c}{GRPO$_{50}$}
& & \multicolumn{2}{c}{Base} & \multicolumn{3}{c}{OPSD$_{50}$} & \multicolumn{3}{c}{GRPO$_{50}$} \\
\cmidrule(lr){2-3}\cmidrule(lr){4-6}\cmidrule(lr){7-9}
\cmidrule(lr){11-12}\cmidrule(lr){13-15}\cmidrule(lr){16-18}
Model & Avg & Pass & Avg & Pass & GPU h & Avg & Pass & GPU h
& Model & Avg & Pass & Avg & Pass & GPU h & Avg & Pass & GPU h \\
\midrule
Qwen3-1.7B
& 34.0 & 59.2
& \textbf{38.0} & \textbf{63.3} & 0.66
& 37.1 & \textbf{63.3} & 114.27
& Qwen3-1.7B
& \textbf{58.8} & 67.3
& 57.5 & 65.3 & 1.08
& 58.8 & \textbf{67.8} & 101.93 \\
Qwen3-4B
& 62.3 & 79.2
& 60.1 & \textbf{80.0} & 1.13
& \textbf{62.8} & 78.3 & 171.81
& Qwen3-4B
& 70.6 & 76.8
& 69.8 & 76.7 & 1.19
& \textbf{71.4} & \textbf{77.4} & 69.81 \\
Qwen3-8B
& \textbf{64.1} & 78.3
& 60.8 & 79.2 & 2.18
& 63.4 & \textbf{82.5} & 279.64
& Qwen3-8B
& 71.3 & 78.6
& \textbf{71.9} & \textbf{78.8} & 1.77
& 70.5 & 76.6 & 62.14 \\
\bottomrule
\end{tabular*}}
\end{table}

On math, although both algorithms underperform the base model on several metrics, their results exhibit a scale-dependent crossover: OPSD is more competitive at smaller parameter scales, whereas GRPO attains a clearer relative advantage as model capacity grows. On coding, GRPO preserves or slightly improves the base checkpoint at 1.7B and 4B, while at 8B OPSD gives a small improvement and GRPO declines. In some settings, OPSD improves model performance at substantially lower computational cost, but overall it is less stable than GRPO.


\FloatBarrier
\section{Analysis: From Token-Level Signals to Optimization Dynamics}
\label{sec:token_analysis}

We analyze the initial token-level signal across reasoning mode, checkpoint type, teacher context, and response position, then ask what training trajectories can diagnose. Each model uses 2048 prompts with two on-policy rollouts ($n=4096$). Completions are capped at 1024 tokens except in the 6144-token response position analysis. Settings and per-model results are in Appendix~\ref{app:token_analysis}.

\subsection{Mismatched Mode Effect}
\label{subsec:token_mode_analysis}

\newsavebox{\modesignaltab}
\begingroup
\setlength{\emergencystretch}{2em}%
\sbox{\modesignaltab}{%
  \scriptsize
  \setlength{\tabcolsep}{2.8pt}%
  \renewcommand{\arraystretch}{0.90}%
  \begin{tabular}{@{}ccccccc@{}}
  \toprule
  S & T & Fwd. KL & $D>.05(\%)$ & Agree.(\%) & Enc.(\%) & Disc.(\%) \\
  \midrule
  \nth & \nth & 0.100 & 17.8 & 94.3 & 32.3 & 37.9 \\
  \nth & \thmode & 0.193 & 24.4 & 93.3 & 14.0 & 70.9 \\
  \thmode & \nth & 0.468 & 42.9 & 87.5 & 15.3 & 63.0 \\
  \thmode & \thmode & 0.144 & 20.6 & 92.9 & 34.5 & 36.0 \\
  \bottomrule
  \end{tabular}}%
\dimen0=\dimexpr\linewidth-\wd\modesignaltab-10pt\relax
\parshape=12
  0pt \dimen0 0pt \dimen0 0pt \dimen0 0pt \dimen0 0pt \dimen0
  0pt \dimen0 0pt \dimen0 0pt \dimen0 0pt \dimen0 0pt \dimen0
  0pt \dimen0
  0pt \linewidth
\noindent\raisebox{\ht\strutbox}[0pt][0pt]{%
  \makebox[0pt][l]{\hspace{\dimexpr\linewidth-\wd\modesignaltab\relax}%
  \begin{minipage}[t]{\wd\modesignaltab}
  \wrapfloatcaptionsetup
  \setlength{\abovecaptionskip}{0pt}%
  \captionof{table}{Qwen3-1.7B grid. S/T denote student/teacher modes. $D>.05(\%)$ is the percentage of valid response positions with unclipped forward KL above 0.05. Agree. is top-1 agreement; Enc./Disc. are the percentages with positive/negative sampled token log advantage.}
  \label{tab:token_signal_summary}%
  \usebox{\modesignaltab}%
  \end{minipage}}}%
\autoref{tab:token_signal_summary} shows that mismatched pairs have substantially higher initial forward KL and high divergence position rates than matched pairs, despite high top-1 agreement. Mode mismatch amplifies the initial signal: both forward KL and the share of high divergence positions increase relative to matched pairs. The effect is asymmetric, with a larger shift for \thmode{}/\nth{} than for \nth{}/\thmode{}.
The unmatched teacher prefers to discourage more tokens than the matched teacher and generated bigger KL loss. These results support that unmatched mode transfer over privileged semantic transfer. Subsequent work should consider the mode mismatch effect or keep the mode matched.
\par
\endgroup

\begin{figure}[t]
\centering
\begin{minipage}[t]{0.485\linewidth}
  \centering
  \vspace{0pt}%
  \includegraphics[width=\linewidth,height=.695\linewidth,trim=7pt 8pt 6pt 6pt,clip]{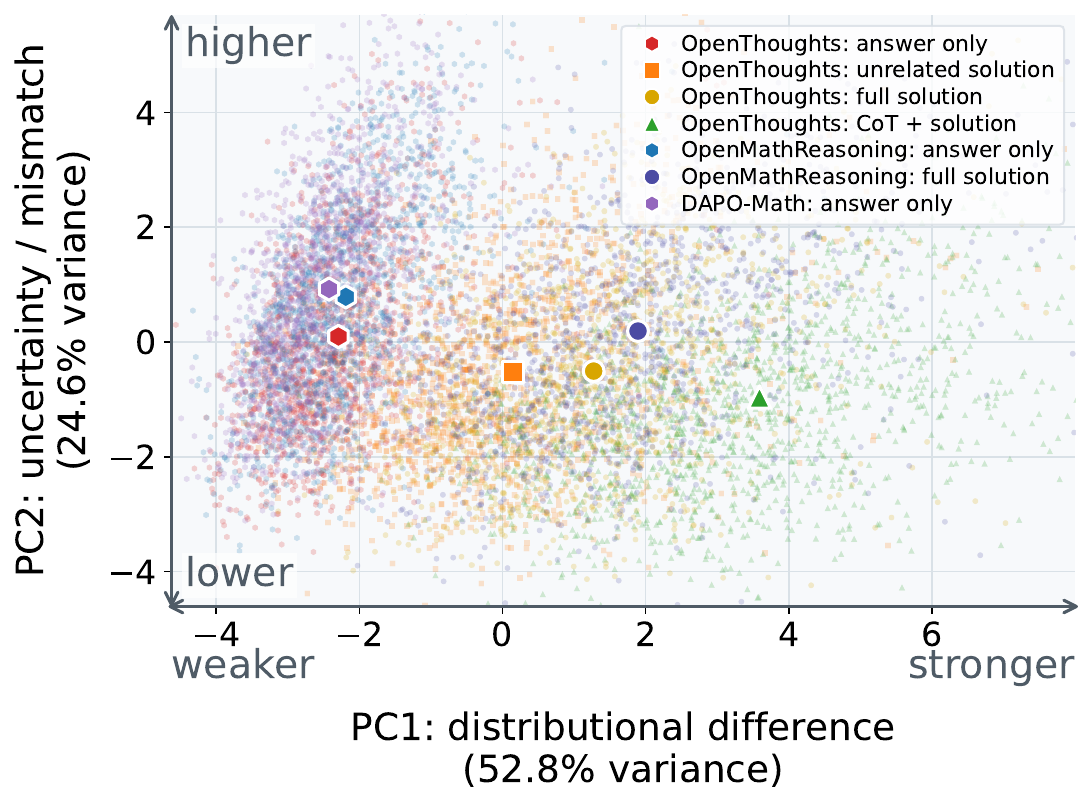}
  \par\vspace{1pt}%
  \textbf{(a)} Context condition PCA
\end{minipage}\hfill
\begin{minipage}[t]{0.485\linewidth}
  \centering
  \vspace{0pt}%
  \includegraphics[width=\linewidth,height=.695\linewidth,trim=7pt 7pt 7pt 7pt,clip]{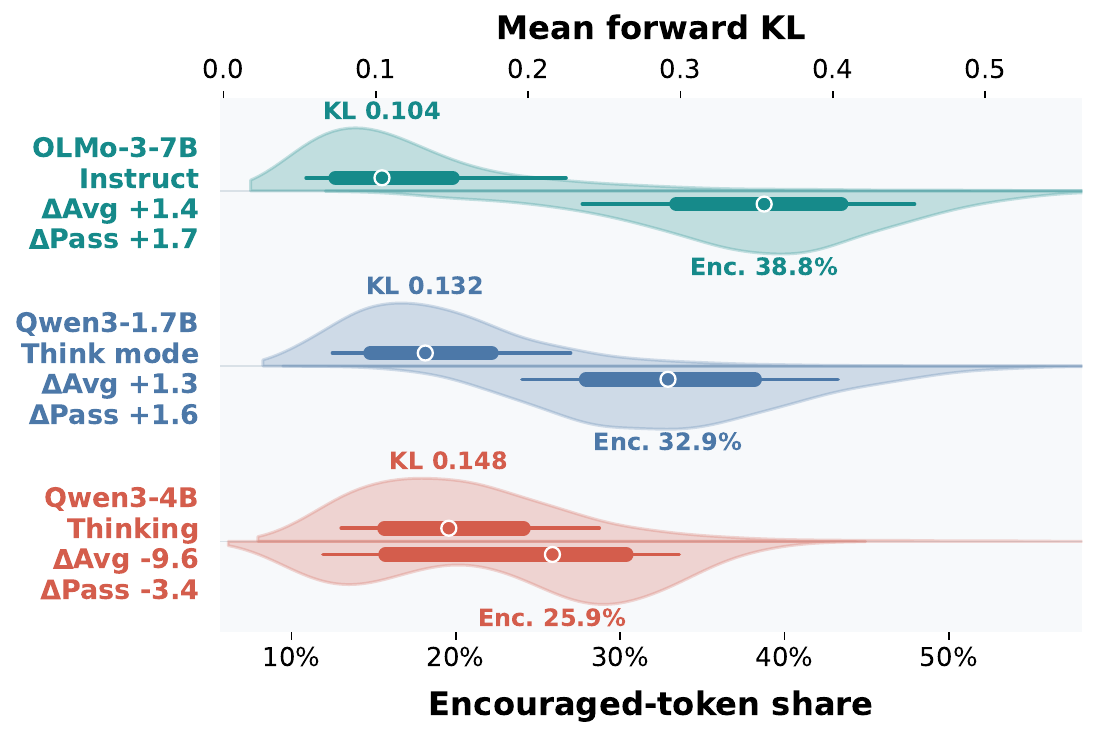}
  \par\vspace{1pt}%
  \textbf{(b)} Full solution signal contrast
\end{minipage}
\caption{Privileged context effects on the initial teacher--student signal. \textbf{(a)} PCA of teacher's signal on student rollout across teacher context conditions. \textbf{(b)} Full solution OpenThoughts signal distributions across models with different downstream OPSD performances.}
\label{fig:teacher_context_signals}
\end{figure}

\subsection{Privileged information Effect}
\label{subsec:signal_performance}

To isolate privileged information, we rescore fixed student rollouts under four conditions as training, Answer only, Full solution, CoT+Solution, and Unrelated solution. To explore whether dataset influences the effect, we add OpenMathReasoning for answer only, full solution and DAPO-Math-17k for answer only.
We change privileged information and teacher prompt length while keeping student's input same in all conditions . Panel (a) of \autoref{fig:teacher_context_signals} applies PCA after within-model standardization. Panel (a) of \autoref{tab:token_context_signal} reports raw macro averages over ten models.

\newsavebox{\checkpointtypetab}
\newsavebox{\checkpointtypewrap}
\begingroup
\setlength{\emergencystretch}{2em}%
\sbox{\checkpointtypetab}{%
  \scriptsize
  \setlength{\tabcolsep}{3.2pt}%
  \renewcommand{\arraystretch}{0.90}%
  \begin{tabular}{@{}lcccc@{}}
  \toprule
  Model & Mean Fwd. KL & Disc. (\%) & HE20 KL (\%) & HE20 $|$adv.$|$ (\%) \\
  \midrule
  Instruction & 0.151 & 44.1 & \textbf{57.2} & \textbf{70.5} \\
  Reasoning & 0.139 & \textbf{61.0} & 47.9 & 64.6 \\
  \bottomrule
  \end{tabular}}%
\sbox{\checkpointtypewrap}{%
  \begin{minipage}[t]{\wd\checkpointtypetab}
  \vspace{-\ht\strutbox}
  \wrapfloatcaptionsetup
  \setlength{\abovecaptionskip}{0pt}%
  \setlength{\belowcaptionskip}{2pt}%
  \raggedright
  \captionof{table}{Different types of model signal analysis.}
  \label{tab:checkpoint_type_signal}%
  \usebox{\checkpointtypetab}%
  \end{minipage}}%
\noindent The signal shifts steadily from Answer only to CoT+Solution. Even for different datasets, the signals are still clustered in the nearby area.
This indicates that the type of privileged information has a greater impact on the signal.
Unrelated solution also produces a clear shift despite containing no task-relevant content.
These show that teacher's signal therefore reflects the full teacher prefix, different type of privileged information would yield different signal and impact on final performance.
\par
To compare whether different model types concentrate their initial signals at high-entropy positions, we use two concentration metrics. Let $\mathcal{I}$ denote all valid response positions in a trajectory, and let $\mathcal{H}_{20}\subset\mathcal{I}$ denote the highest-entropy 20\% of positions within each completion. Define $D_i^{\mathrm{fwd}}=D_{\mathrm{KL}}(\pi_i^T\|\pi_i^S)$ and $a_i=\log\pi_i^T(y_i^S)-\log\pi_i^S(y_i^S)$. 
\par\begingroup
\setlength{\parskip}{2pt}%
\setlength{\abovedisplayskip}{2pt}%
\setlength{\abovedisplayshortskip}{2pt}%
\setlength{\belowdisplayskip}{2pt}%
\setlength{\belowdisplayshortskip}{2pt}%
\vspace{-6pt}
\noindent
\begin{minipage}[t]{0.47\linewidth}
\begin{equation}
\mathrm{HE20\ KL}\ (\%)
=100\,\frac{\sum_{i\in\mathcal{H}_{20}}D_i^{\mathrm{fwd}}}
{\sum_{i\in\mathcal{I}}D_i^{\mathrm{fwd}}},
\label{eq:he20_kl_share}
\end{equation}
\end{minipage}\hfill%
\begin{minipage}[t]{0.47\linewidth}
\begin{equation}
\mathrm{HE20}\ \lvert\mathrm{adv.}\rvert\ (\%)
=100\,\frac{\sum_{i\in\mathcal{H}_{20}}\lvert a_i\rvert\vphantom{D_i^{\mathrm{fwd}}}}
{\sum_{i\in\mathcal{I}}\lvert a_i\rvert\vphantom{D_i^{\mathrm{fwd}}}}.
\label{eq:he20_abs_adv_share}
\end{equation}
\end{minipage}\par
\endgroup
The first quantity measures how much of the full distribution mismatch is concentrated at the highest-entropy positions, since $D_i^{\mathrm{fwd}}$ compares the teacher and student probabilities over the entire vocabulary. The second instead measures the concentration of disagreement on the tokens actually sampled by the student, using the absolute teacher--student log probability difference $\lvert a_i\rvert$.

\dimen0=\dimexpr\linewidth-\wd\checkpointtypewrap-10pt\relax
\parshape=6
  0pt \dimen0 0pt \dimen0 0pt \dimen0 0pt \dimen0
  0pt \dimen0
  0pt \linewidth
\noindent\raisebox{\dimexpr\ht\strutbox-8pt\relax}[0pt][0pt]{%
  \makebox[0pt][l]{\hspace{\dimexpr\linewidth-\wd\checkpointtypewrap\relax}\usebox{\checkpointtypewrap}}%
}The known answer can act as an instruction to reconsider the sampled reasoning, making the Instruct and Reasoning models contrast another prefix context effect. For Qwen3-4B and OLMo-3-7B, mean forward KL is similar, but Reasoning model has more discouraged tokens while Instruct model concentrates more signal at high-entropy positions (\autoref{tab:checkpoint_type_signal}). The results are consistent with instruction following turning privileged information into targeted corrections that can make OPSD more effective. The results agree with \citet{opsd_rethink}'s fork suppression analysis. Our controls further show that prefix mismatch can shift the signal even with task irrelevant context.
\par
\endgroup

A stronger signal is not necessarily more useful. Panel (b) of \autoref{fig:teacher_context_signals} contrasts two settings that improve after full solution OPSD with one that degrades. 
Their rollout-level signal distributions overlap, so neither raw mean forward KL nor the encouraged token share predicts the final performance. 
On matched rows, CoT+Solution increases initial divergence but reduces Avg@8 relative to the full solution (\autoref{tab:same_corpus_cot_signal}, panel (b)). Training-time forward KL has the same limitation: its change varies across models and rollout caps without tracking downstream performance (\autoref{tab:token_downstream}).

Privileged information should therefore be treated as a context-induced distributional intervention. Adding context can strengthen the teacher--student difference without making the supervision more useful. Context construction should control the nature and magnitude of this difference, not maximize context length, information, or initial KL.

\FloatBarrier

\begin{table}[!ht]
\caption{Privileged context effects. \textbf{(a)} OpenThoughts signal averages over ten models. $D>0.05$ is the percentage of valid response positions with unclipped forward KL above 0.05. Abs. adv. and Agree. are the mean absolute sampled token log advantage and top-1 agreement rate. \textbf{(b)} First batch forward KL and Avg@8 under a 1024-token rollout cap. Only teacher context and its prefix cap change. Full per-model signal results are in \autoref{tab:token_context_signal_app}; downstream per-dataset results are in \autoref{tab:context_per_dataset}.}
\label{tab:token_context_signal}
\label{tab:same_corpus_cot_signal}
\centering
\begin{minipage}[t]{0.485\linewidth}
  \vspace{0pt}%
  \centering
  \scriptsize
  \textbf{(a) Context signal}\par\vspace{2pt}%
  \setlength{\tabcolsep}{2.4pt}%
  \renewcommand{\arraystretch}{0.90}%
  \begin{tabular*}{\linewidth}{@{\extracolsep{\fill}}lcccc@{}}
  \toprule
  Context & Fwd. KL & $D>0.05$ & Abs. adv. & Agree. \\
  \midrule
  Answer only        & 0.016 & 2.9  & 0.039 & 98.0 \\
  Unrelated solution & 0.054 & 14.9 & 0.100 & 95.0 \\
  Full solution      & 0.108 & 20.3 & 0.151 & 93.3 \\
  CoT + solution     & 0.179 & 30.8 & 0.224 & 90.6 \\
  \bottomrule
  \end{tabular*}%
\end{minipage}\hfill
\begin{minipage}[t]{0.485\linewidth}
  \vspace{0pt}%
  \centering
  \scriptsize
  \textbf{(b) Signal and utility}\par\vspace{2pt}%
  \setlength{\tabcolsep}{2.0pt}%
  \renewcommand{\arraystretch}{0.90}%
  \begin{tabular*}{\linewidth}{@{\extracolsep{\fill}}llccc@{}}
  \toprule
  Model & Context & Prefix cap & Fwd. KL & Avg@8 \\
  \midrule
  Qwen3-1.7B & Full solution & 1024 & 0.141 & 35.3 \\
              & CoT + solution & 8192 & 0.218 & 30.8 \\
  OLMo-3-7B-Think & Full solution & 1024 & 0.115 & 58.8 \\
                   & CoT + solution & 8192 & 0.205 & 58.6 \\
  \bottomrule
\end{tabular*}%
\end{minipage}
\wrapfloatcaptionsetup
\setlength{\abovecaptionskip}{2pt}%
\end{table}

\subsection{Teacher--Student Differences across Response Positions}
\label{subsec:response_position_signal}

Because OPSD supervises tokens throughout the sampled response, we use four metrics to characterize how teacher--student differences and the supervision signal vary with token position along student rollouts. The top-1 disagreement rate is $\Pr[\arg\max_v \pi_i^T(v)\ne\arg\max_v \pi_i^S(v)]$. The mean absolute sampled-token log ratio is $\mathbb{E}[|a_i|]$, where $a_i=\log\pi_i^T(y_i^S)-\log\pi_i^S(y_i^S)$. We also measure mean forward KL, $\mathbb{E}[D_{\mathrm{KL}}(\pi_i^T\|\pi_i^S)]$, and SNR, $\mathrm{SNR}=\mu_{\mathrm{KL}}/\sigma_{\mathrm{KL}}$, where $\mu_{\mathrm{KL}}$ and $\sigma_{\mathrm{KL}}$ are the mean and standard deviation of token-level forward KL within each position window. The first three metrics capture differences in the preferred token, the log-probability of the sampled token, and the full predictive distribution, respectively. We measure these metrics before training, using the same responses and weights across position windows (Appendix~\ref{app:token_length_windows}).

\begin{figure}[!ht]
\centering
\includegraphics[width=\linewidth]{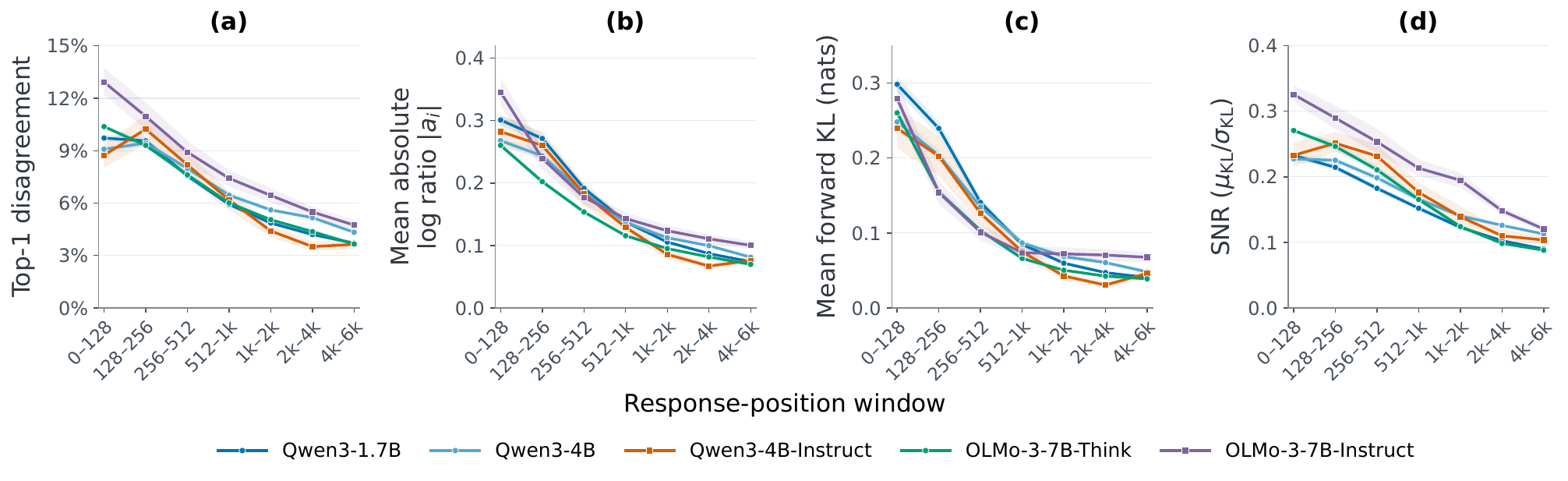}
\caption{(a) Top-1 disagreement rate, (b) mean absolute sampled-token log ratio, (c) mean forward KL, and (d) SNR across response positions. Here $\mathrm{k}=1024$, and shading denotes pointwise 95\% confidence intervals.}
\label{fig:position_intervention}
\end{figure}

All four metrics show an overall decrease toward later positions across all five models (\autoref{fig:position_intervention}).
At later positions, the teacher and student agree more often on the selected token (\autoref{fig:position_intervention}(a)) and assign more similar probabilities to tokens sampled by student(\autoref{fig:position_intervention}(b)).
Their full predictive distributions are also closer (\autoref{fig:position_intervention}(c); \autoref{tab:position_intervention_control}).
The SNR values indicate a lower mean KL relative to its token-level variability at these positions (\mbox{\autoref{fig:position_intervention}(d)}).
These results indicate that, at initialization, OPSD's privileged teacher and student tend to make more similar predictions at later positions in long rollouts, with the supervision signal concentrated early in the response.

Consistently, First-256 outperforms Final-256 under the same loss token budget (\autoref{tab:position_selection_summary}).
Together with \citet{early_stop_opd}, this result suggests a common limitation of OPD and OPSD: once a rollout takes a wrong branch, the weaker signal at later tokens on the student's trajectory may be insufficient or even harmful.
OPSD does not need to compute the loss over an entire trajectory. Applying the loss only to the early tokens of each trajectory can achieve better performance, and it takes less training time and computation.

\begin{table}[t]
\caption{Forward KL and downstream change under full solution OPSD. KL First10 and Final10 average KL loss over the first and final ten steps. $\Delta$ Avg and $\Delta$ Pass are Avg@8 and Pass@8 changes from Base. DS-Qwen-1.5B refers to DeepSeek-R1-Distill-Qwen-1.5B.}
\label{tab:token_downstream}
\centering
\scriptsize
\setlength{\tabcolsep}{2.4pt}%
\renewcommand{\arraystretch}{0.88}%
\begin{tabular*}{\linewidth}{@{\extracolsep{\fill}}lcccccccc@{}}
\toprule
& \multicolumn{4}{c}{256-token rollout cap} & \multicolumn{4}{c}{1024-token rollout cap} \\
\cmidrule(lr){2-5}\cmidrule(lr){6-9}
Model
& \multicolumn{1}{c}{KL First10}
& \multicolumn{1}{c}{KL Final10}
& \multicolumn{1}{c}{$\Delta$ Avg}
& \multicolumn{1}{c}{$\Delta$ Pass}
& \multicolumn{1}{c}{KL First10}
& \multicolumn{1}{c}{KL Final10}
& \multicolumn{1}{c}{$\Delta$ Avg}
& \multicolumn{1}{c}{$\Delta$ Pass} \\
\midrule
DS-Qwen-1.5B & 0.062 & $0.073\uparrow$ & $+0.7$ & $+0.9$ & 0.032 & $0.040\uparrow$ & $+0.6$ & $-0.8$ \\
Falcon-H1R-7B                  & 0.112 & $0.087\downarrow$ & $-0.6$ & $0.0$  & 0.077 & $0.054\downarrow$ & $-0.4$ & $0.0$ \\
Qwen3-0.6B                     & 0.125 & $0.334\uparrow$ & $+0.2$ & $+0.8$ & 0.073 & $0.212\uparrow$ & $-2.6$ & $-6.7$ \\
MiMo-7B-RL                     & 0.175 & $0.340\uparrow$ & $-0.3$ & $+0.8$ & 0.098 & $0.243\uparrow$ & $-11.3$ & $-10.0$ \\
OLMo-3-7B-Think                & 0.193 & $0.377\uparrow$ & $-2.7$ & $-1.7$ & 0.108 & $0.251\uparrow$ & $-4.7$ & $-4.2$ \\
OLMo-3-7B-Instruct             & 0.206 & $0.204\downarrow$ & $+3.4$ & $+7.5$ & 0.120 & $0.151\uparrow$ & $+1.3$ & $+1.7$ \\
Qwen3-1.7B                     & 0.261 & $0.458\uparrow$ & $+4.1$ & $+3.3$ & 0.142 & $0.336\uparrow$ & $+1.3$ & $+1.6$ \\
Qwen3-4B                       & 0.226 & $0.412\uparrow$ & $-1.4$ & $-3.4$ & 0.137 & $0.329\uparrow$ & $-4.4$ & $-3.4$ \\
Qwen3-4B-Thinking              & 0.246 & $0.341\uparrow$ & $-5.1$ & $-3.4$ & 0.156 & $0.325\uparrow$ & $-9.6$ & $-3.4$ \\
Qwen3-4B-Instruct              & 0.260 & $0.232\downarrow$ & $+1.1$ & $-0.9$ & 0.164 & $0.161\downarrow$ & $+1.5$ & $-1.7$ \\
\bottomrule
\end{tabular*}%
\wrapfloatcaptionsetup
\end{table}

\subsection{OPSD Training Dynamics}
\label{subsec:training_dynamics}

\begin{figure}[H]
  \centering
  \includegraphics[width=\linewidth]{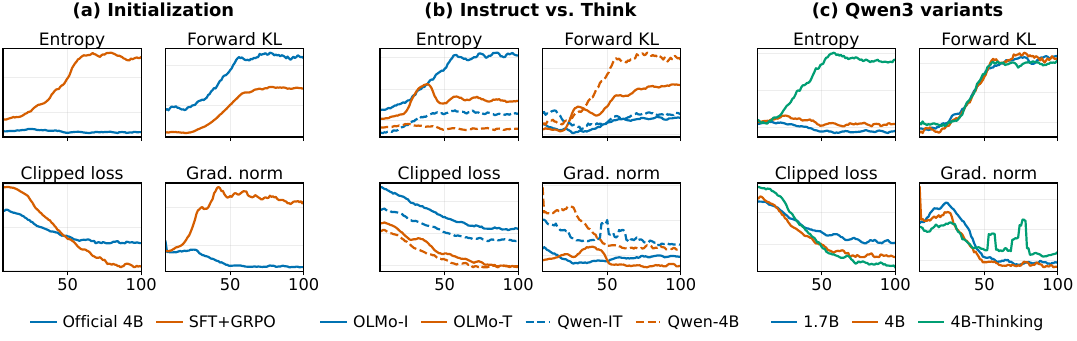}
  \vspace{-14pt}
  \caption{OPSD training dynamics across (a) initialization, (b) Instruct/Think checkpoints, and (c) Qwen3 variants. Results for each window in the initialization comparison are in \autoref{tab:sft_grpo_opsd_train}.}
  \label{fig:grpo_opsd_dynamics}
\end{figure}

We examine whether training-time statistics can diagnose OPSD behavior before evaluation. Specifically, we track student entropy, forward KL to privileged teacher, the clipped distillation loss, and gradient norm, which characterize output uncertainty, student--teacher discrepancy, and optimization behavior.  
\autoref{fig:grpo_opsd_dynamics} compares these trajectories across different initializations, Instruct and Think checkpoints, and Qwen3 variants to explore further.

The initialization contrast exhibits the clearest collapse signature: student entropy and gradient norm rise together for the SFT+GRPO actor, whereas both remain bounded or decrease for the Qwen3-4B. 
The Instruct/Think comparison shows consistent mode-dependent differences: within both OLMo and Qwen, Instruct checkpoints exhibit lower forward KL and less negative clipped loss than Think checkpoints, without the joint entropy--gradient rise associated with collapse. 
The Qwen3 variants further show that similar forward KL values can correspond to substantially different downstream performances. Thus, the joint entropy--gradient rise is a useful online warning of collapse, but no single trace reliably predicts final performance.

\FloatBarrier
\section{Conclusion}
\label{sec:conclusion}

In this paper, we design a comprehensive diagnostic pipeline to study the behavior of OPSD.
Controlled experiments and token-level analyses show that OPSD is a sensitive algorithm.
The teacher's signal of OPSD is driven by contextual distribution differences rather than privileged semantics alone. 
Both of the training configuration such as rollout cap and model itself including model type affect the performance of OPSD.
Training dynamics can be used to warn of OPSD collapse, but not predict downstream performance.

Compared with RLVR, OPSD requires no external reward or stronger privileged teacher, provides dense supervision, and uses less computation. 
However, OPSD offers no free lunch. Its efficiency comes at the cost of relying on unverified, training-state-dependent supervision. 
OPSD should therefore be viewed as a complementary post-training method rather than a complete replacement for RLVR. 
Its effect and generalization remain to be tested on larger models and more diverse tasks.

%

%

\newpage
\bibliographystyle{iclr2027_conference}
\bibliography{iclr2027_conference}

\newpage
\appendix
\input{appendix_arxiv.tex}

\end{document}

%% file: math_commands.tex
\usepackage{amsmath,amsfonts,bm}

\def\eqref#1{equation~\ref{#1}}

\def\1{\bm{1}}

\DeclareMathAlphabet{\mathsfit}{\encodingdefault}{\sfdefault}{m}{sl}
\SetMathAlphabet{\mathsfit}{bold}{\encodingdefault}{\sfdefault}{bx}{n}



%% file: appendix_arxiv.tex
\setcounter{section}{0}
\setcounter{secnumdepth}{4}
\setlength{\textfloatsep}{4pt plus 1pt minus 1pt}
\setlength{\floatsep}{4pt plus 1pt minus 1pt}
\setlength{\intextsep}{5pt plus 1pt minus 1pt}
\setlength{\abovecaptionskip}{2pt}
\setlength{\belowcaptionskip}{2pt}
\setlength{\parskip}{.5pc}
\setlength{\abovedisplayskip}{4pt plus 1pt minus 2pt}
\setlength{\belowdisplayskip}{4pt plus 1pt minus 2pt}
\setlength{\abovedisplayshortskip}{2pt plus 1pt}
\setlength{\belowdisplayshortskip}{2pt plus 1pt}

\phantomsection
\section*{Appendix Contents}
\pdfbookmark[1]{Appendix Contents}{appendix.contents}
\begingroup
\setlength{\parindent}{0pt}
\setlength{\parskip}{3pt}
\newcommand{\appendixtocentry}[1]{%
  \noindent
  \hyperref[#1]{\makebox[2em][l]{\textbf{\ref*{#1}}}\nameref*{#1}}%
  \dotfill
  \hyperref[#1]{\pageref*{#1}}\par
}
\appendixtocentry{app:training_details}
\appendixtocentry{app:main_results}
\appendixtocentry{app:analysis_results}
\appendixtocentry{app:additional_experiments}
\appendixtocentry{app:robustness_checks}
\appendixtocentry{app:per_dataset_results}
\endgroup
\medskip

\clearpage
\section{Experimental Setup}
\label{app:training_details}

We report the training, evaluation, and systems settings used throughout the paper. Each ablation changes only the named factor.

\subsection{Training Framework and Systems Configuration}
\label{app:framework}

We implement OPSD with PyTorch, Hugging Face Transformers, and TRL. Distributed jobs are launched with Accelerate and use the DeepSpeed implementation of ZeRO-3 \citep{zero} to shard the trainable student, optimizer states, and the frozen teacher across data-parallel workers. All main experiments train with full parameter student updates. No parameter-efficient adapters are used. Both student and teacher dropout are disabled. Student forward passes use PyTorch scaled dot-product attention, with non-reentrant gradient checkpointing and the kv cache disabled during training. Parameters and activations use bfloat16, and TF32 matrix multiplication is enabled. Training runs use 2--8 NVIDIA A800 GPUs.

The evaluated models come from the Qwen3, Qwen3.5, DeepSeek-R1, OLMo-3, Falcon-H1R, and MiMo families. A colocated inference engine generates on-policy rollouts: vLLM \citep{vllm} for Qwen3 and DeepSeek-R1-Distill, and SGLang \citep{sglang} for Qwen3.5-4B, OLMo-3, Falcon-H1R-7B, and MiMo-7B-RL. The rollout copy of the student is synchronized after every optimizer update. Each worker uses tensor parallelism of 1 for generation and data parallelism otherwise. The vLLM path uses PyTorch~2.6.0, Transformers~4.57.0, TRL~0.22.1, Accelerate~1.7.0, DeepSpeed~0.18.9, and vLLM~0.8.5. The SGLang path uses PyTorch~2.8 and SGLang~0.5.4. Both backends use the same experimental settings.

\subsection{Training Datasets and Preprocessing}
\label{app:training_data}

\subsubsection{OPSD Data}
For math, OPSD uses the OpenThoughts Math 30k training split unless noted otherwise. The public version contains about 29k rows. Each row provides a mathematics problem, reference solution, final answer, and separate CoT trace. Student and teacher prompts are initialized independently with the model's native chat template and designated reasoning mode. Only the user-side text is reproduced below. Before tokenization and length filtering, we remove examples from OpenThoughts and OpenMathReasoning that overlap with the test sets. We keep examples for which both prompts contain at most 1024 tokens and filter longer prompts without truncation. Each model--mode configuration is filtered separately because its tokenizer and template differ. Under full solution conditioning, the resulting sets contain approximately 22k examples for the standard Qwen3 models, 22k for Qwen3-4B-Thinking(Qwen-3-4B-Thinking-2507), 21k for Qwen3.5-4B with a \thmode{} teacher, 22k for DeepSeek-R1-Distill-Qwen-1.5B, 21k for OLMo-3-7B-Think, 22k for OLMo-3-7B-Instruct, 16k for Falcon-H1R-7B, and 22k for MiMo-7B-RL. Teacher context controls are filtered under their modified teacher templates. The data sampler uses seed~42.

For the CoT+Solution intervention, we use approximately 22k Qwen3-1.7B rows or 21k OLMo-3-7B-Think rows in the same order as in the full solution condition, concatenate the CoT and gold solution fields in the teacher context, and raise the teacher prompt cap to 8192 tokens. The cross-dataset replication uses the OpenMathReasoning \texttt{cot} split \citep{openmathreasoning} with Qwen3-1.7B. We separate each generated solution into its internal CoT and post-thinking solution, then form subsets with post-thinking lengths of at most 1024 tokens (approximately 610k rows) or 2048--4096 tokens (approximately 15k rows). Within each subset, the solution-only and CoT-plus-solution conditions use identical rows and ordering. The solution-only prompt caps are 1024 and 4096 tokens, respectively. Both augmented conditions use a cap of 12288 tokens.

For coding, OPSD uses OpenCodeReasoning with a 2048-token prompt cap and the solution code as privileged teacher context. The matched comparison with GRPO uses APPS.

\subsubsection{Training Prompts}
The default math student user message is:
\begin{promptbox}
Problem: \{problem\}

\promptfill{Please reason step by step, and put your final answer within \promptboxed.}
\end{promptbox}
All teacher context experiments use this student message. The teacher-side message varies as follows.
Under the full solution teacher context, the teacher user message is:
\begin{promptbox}
Problem: \{problem\}

Here is a reference solution to this problem:\\
=== Reference Solution Begin ===\\
\{solution\}\\
=== Reference Solution End ===

After reading the reference solution above, make sure you truly understand the reasoning behind each step---do not copy or paraphrase it. Now, using your own words and independent reasoning, derive the same final answer to the problem above. Think step by step, explore different approaches, and don't be afraid to backtrack or reconsider if something doesn't work out:

\promptfill{Please reason step by step, and put your final answer within \promptboxed.}
\end{promptbox}
Under CoT+Solution, the reference block contains the dataset-provided CoT followed by the gold solution. The student message and sampled rows remain fixed.
Under answer-only conditioning, the teacher receives the verified final answer without its derivation:
\begin{promptbox}
Problem: \{problem\}

Here is the verified answer:\\
\{answer\}

After understanding the privileged information, solve the problem using your own reasoning.

\promptfill{Please reason step by step and put the final answer within \promptboxed.}
\end{promptbox}
Under the unrelated-solution control, an independently sampled problem--solution pair precedes the target problem and never claims to solve it:
\begin{promptbox}
Problem: \{problem\_B\}

Solution: \{solution\_B\}

Problem: \{problem\_A\}

\promptfill{Please reason step by step, and put your final answer within \promptboxed.}
\end{promptbox}
For each prompt, the current student produces one sampled completion. The frozen teacher then scores exactly the same completion tokens under its own prompt, and the loss is applied only to non-padding response positions. Thus neither reference solution tokens nor prompt tokens are direct prediction targets.

For the no-privilege same-prefix control in \autoref{tab:qwen_mode}, panel~(c), both sides receive the default student message above: the teacher receives no solution, answer, or additional problem. The student generates in \nth{} mode, the teacher scores in \thmode{} mode, and evaluation uses \thmode{} mode. All other optimization, rollout, and model-specific batch settings are the default OPSD settings below.

\subsubsection{Supervised Initialization Data}
For math, the Qwen3-1.7B-Base and Qwen3-4B-Base supervised initialization uses the NVIDIA OpenMathReasoning \texttt{cot} split and excludes the tool-integrated-reasoning, GenSelect, and additional-problem splits. The source split contains approximately 3.2M problem--solution rows. We use \texttt{problem} as the user turn and \texttt{generated\_solution} as the assistant target, initialize both with the Qwen3 chat template in \thmode{} mode, and keep sequences of at most 12288 tokens. The resulting pretokenized dataset contains approximately 2.6M examples. For coding, supervised initialization uses OpenCodeReasoning with the same 12288-token sequence limit. Supervised loss applies to assistant tokens, without sequence packing.

OpenMathReasoning provides the supervised initialization, cross-dataset context replication, and the OPSD--GRPO comparison on the matched dataset, whereas OpenThoughts provides the problems and privileged teacher context for the main OPSD experiments. The datasets are not mixed within a training batch.

The teacher is a frozen copy of the initial student and receives no gradient or parameter updates. Teacher behavior therefore changes only through context or reasoning mode. The objective contains no task reward, correctness signal, policy-gradient advantage, or auxiliary supervised loss.

\subsection{Supervised Initialization Configuration}
\label{app:sft_configuration}

We apply SFT to Qwen3-1.7B-Base and Qwen3-4B-Base using the same optimization settings across tasks. All runs use a 12288-token sequence limit, DeepSpeed ZeRO-2 without CPU offload, a global batch of 64 sequences, 15k updates (960k sampled sequences), and checkpoints every 500 updates. We use AdamW with peak learning rate $1\times10^{-5}$, zero weight decay, gradient-norm clipping at 1.0, and a cosine schedule with 10\% linear warmup. Training uses bfloat16, gradient checkpointing, no sequence packing, and seed~42. Each base checkpoint is formatted with the chat template from its corresponding Qwen3 instruction checkpoint, with thinking enabled.

Qwen3-1.7B-Base uses 4 A800 GPUs with per-device batch size 2 and gradient accumulation 8. Qwen3-4B-Base uses per-device batch size 1, with 8 GPUs and gradient accumulation 8 for math, or 4 GPUs and gradient accumulation 16 for coding. All runs have the same global batch and sample budget.

\subsection{GRPO Configuration}
\label{app:grpo_configuration}

The GRPO recipes use full-parameter actor updates, one PPO epoch per update, DAPO clip ratios $0.2/0.28$, actor weight decay $0.01$, gradient-norm clipping at 1.0, and no entropy loss. They use a low-variance KL loss with a frozen reference and coefficient $0.01$, no reward-side KL, and no overlong response penalty. Reference parameters are offloaded between scoring passes. Prompts use the Qwen3 \thmode{} template, and rollouts use vLLM with temperature $0.6$, top-$p=0.95$, and top-$k=20$.

The pipeline controls initialize from the Qwen3-1.7B-Base and Qwen3-4B-Base SFT-15k checkpoints and run for 100 updates. The comparisons with OPSD initialize from Qwen3-1.7B, Qwen3-4B, and Qwen3-8B and use 50 updates with a 24576-token response cap. These comparisons offload actor parameters and optimizer states, with additional activation offloading for 8B.

For math, we sample eight generations per prompt. The reward extracts a boxed answer first and falls back to an \texttt{Answer:} pattern, with correctness checked by Math-Verify. The pipeline controls train on DAPO-Math-17k with a 1024-token prompt cap, train batch size 16, PPO mini-batch size 8, learning rate $5\times10^{-7}$, and a 20480-token response cap. The 1.7B and 4B runs use 4 and 8 A800 GPUs, respectively, and save checkpoints every 25 updates. The comparisons from Qwen3-1.7B, Qwen3-4B, and Qwen3-8B use the OpenMathReasoning subset with integer answers and source solutions capped at 12288 tokens. They use train and PPO mini-batch sizes of 64, learning rate $1\times10^{-6}$, and 4, 8, and 4 A800 GPUs, respectively.

For coding, we combine the APPS train and test splits across all difficulty levels and retain at most 10 execution tests per problem. We use a 2048-token prompt cap, four generations per prompt, execution correctness as the reward, and learning rate $5\times10^{-7}$. Qwen3-1.7B, Qwen3-4B, and Qwen3-8B use train batch sizes 96, 48, and 32 and PPO mini-batch sizes 48, 24, and 32, respectively. Each run uses 4 A800 GPUs. The pipeline controls use the OpenCodeReasoning SFT-15k checkpoints with the corresponding batch sizes, a 24576-token response cap, and actor parameter and optimizer offloading.

In the pipeline controls, subsequent OPSD initializes both student and teacher from the corresponding GRPO checkpoint and retains the task-specific OPSD settings.

\subsection{OPSD Optimization and Rollout Hyperparameters}
\label{app:hyperparameters}

\subsubsection{Default OPSD Configuration}
Unless stated otherwise, OPSD uses a frozen same-model teacher, matched native student--teacher reasoning modes, privileged solution context, a rollout cap of 1024 tokens, and full-parameter student updates. Training runs for 100 steps, with checkpoints saved every 25 updates. We use AdamW \citep{adamw} with peak learning rate $1\times10^{-6}$, zero weight decay, and gradient-norm clipping at $0.1$. The learning rate follows a cosine schedule with linear warmup over the first 10\% of updates. The random and data seeds are 42. The prompt cap is 1024 tokens for math and 2048 for coding.

For the loss in \autoref{eq:opsd_clipped}, we set generalized-JSD parameter $\beta=0$ and a pointwise cap on vocabulary contributions $\tau_{\mathrm{clip}}=0.05$. At temperature 1.1, the implementation computes full vocabulary contributions, caps them coordinatewise, sums over the vocabulary, and averages over selected valid response positions. The top-$k$ ablation in Appendix~\ref{app:topk_loss} restricts vocabulary support. The entropy masking study in Appendix~\ref{subsec:entropy_training} uses the highest- or lowest-entropy response tokens (HE20/HE80 or LE20/LE80). First-256 reuses the equivalent OPSD-256 run, while Uni-256 and Final-256 keep a 1024-token rollout but apply the loss to at most 256 uniformly sampled or final valid positions. The AdvT ablation in Appendix~\ref{app:advt4} applies the loss only at positions where the frozen teacher ranks the sampled token among its top-$4$ or top-$16$ outcomes. Other runs average over all valid response positions. Rollout top-$k$ and loss support $k$ are distinct.

Student rollouts use multinomial sampling with temperature 1.1, top-$p=0.95$, top-$k=20$, min-$p=0$, presence penalty 0, and repetition penalty 1.0. In the rollout length study, we vary only the rollout cap over $\{128,256,512,1024,2048,4096,6144\}$ tokens where reported and keep all other settings fixed. We rebalance the per-device micro-batch and gradient accumulation at longer rollout lengths to preserve the default model-specific global batch in \autoref{tab:training_batch_config}.

We additionally train OPSD on Qwen3-4B-Thinking, DeepSeek-R1-Distill-Qwen-1.5B, Falcon-H1R-7B, and MiMo-7B-RL. \autoref{tab:training_batch_config} gives the distributed configuration for every OPSD checkpoint. A parenthetical teacher denotes a cross-model experiment. Otherwise, student and teacher use the same initial model.

\begin{table}[H]
\caption{Distributed configurations at the default rollout cap of 1024 tokens. Global batch equals micro-batch $\times$ accumulation $\times$ GPUs. The length sweep rebalances micro-batch and accumulation while preserving global batch.}
\label{tab:training_batch_config}
\centering
\small
\setlength{\tabcolsep}{2pt}
\begin{tabular}{@{}lccccc@{}}
\toprule
Student model (teacher if different) & Engine & GPUs & Micro-batch & Accumulation & Global batch \\
\midrule
Qwen3-0.6B                                      & vLLM   & 2 & 16 & 4 & 128 \\
Qwen3-1.7B                                      & vLLM   & 2 &  8 & 4 &  64 \\
Qwen3-4B / Qwen3-4B-Instruct                   & vLLM   & 2 &  4 & 4 &  32 \\
Qwen3-4B-Instruct (Qwen3-4B)                   & vLLM   & 2 &  4 & 4 &  32 \\
\ml{Qwen3-4B-\\Thinking}                   & vLLM   & 2 &  4 & 4 &  32 \\
Qwen3-8B                                        & vLLM   & 4 &  4 & 2 &  32 \\
Qwen3.5-4B                                      & SGLang & 2 &  4 & 4 &  32 \\
\ml{DeepSeek-R1-\\Distill-Qwen-1.5B}            & vLLM   & 2 &  8 & 4 &  64 \\
\mbox{OLMo-3-7B-Think / OLMo-3-7B-Instruct}     & SGLang & 4 &  4 & 4 &  64 \\
Falcon-H1R-7B                                   & SGLang & 4 &  2 & 4 &  32 \\
MiMo-7B-RL                                      & SGLang & 4 &  2 & 4 &  32 \\
\bottomrule
\end{tabular}
\end{table}

Over 100 updates, the configurations in \autoref{tab:training_batch_config} correspond to 12800 sampled trajectories for Qwen3-0.6B, 6400 for Qwen3-1.7B, DeepSeek-R1-Distill-Qwen-1.5B, and the OLMo-3 models, and 3200 for every remaining configuration.

\subsubsection{Main Text Configuration Coverage}
The reasoning mode comparison uses the default OPSD recipe with only the stated student, teacher, and evaluation modes changed. The no-privilege control is defined above. The extended-training experiment changes the maximum step count to 300 and saves/evaluates every 50 updates. The rollout length, loss position, teacher context, and divergence studies change only the named factor from the default recipe. The initial capability study uses the model rows in \autoref{tab:training_batch_config}. The post-training stage study combines the SFT and GRPO recipes above with the default size-matched OPSD recipe, and the compute comparison uses default OPSD versus the released model GRPO recipe. Token analysis sampling and scoring settings are reported separately in Appendix~\ref{app:token_analysis}.

\subsection{Evaluation Protocol and Inference Settings}
\label{app:evaluation_details}

We report Avg@$k$ (the mean correctness of the $k$ sampled responses per problem), Pass@$k$ (the fraction of problems with at least one correct response among those $k$ samples), and Length, with $k=8$ for math and $k=4$ for coding. Format is the fraction of math responses containing a boxed final answer. Aggregate results are unweighted means across four math or three coding benchmarks, computed before rounding. Length is mean output length in tokens unless otherwise specified.
Unless otherwise specified, we evaluate the checkpoint from the final training step.

Evaluation uses zero-shot prompting, seed~42, and the model's native chat template and designated reasoning mode. Base and OPSD use identical decoding settings. \autoref{tab:evaluation_config} lists the model-specific inference defaults.

All runs use 1 A800 GPU with tensor parallelism 1. The vLLM path uses PyTorch~2.6.0, Transformers~4.57.0, and vLLM~0.8.5 \citep{vllm}. The SGLang path uses PyTorch~2.8 with SGLang~0.5.4 for OLMo-3 and MiMo-7B-RL, and SGLang~0.5.10 for Qwen3.5-4B and Falcon-H1R-7B \citep{sglang}.

For math, each test set contains 30 problems, with eight sampled completions per problem. The evaluation user message is:
\begin{promptbox}
\{problem\}

\promptfill{Please reason step by step, and put your final answer within \promptboxed.}
\end{promptbox}
Unlike the training student prompt, evaluation does not prepend a \texttt{Problem:} label. MiMo-7B-RL uses an empty system message. Answers are extracted from the final balanced \verb|\boxed{}| expression and checked with Math-Verify.

For coding, we sample four completions per problem on HumanEvalPlus, MBPPPlus, and LiveCodeBench v6. We use the benchmark-specific coding prompts and evaluate extracted programs against the execution tests. HumanEvalPlus and MBPPPlus use the EvalPlus evaluator, and LiveCodeBench uses its code execution evaluator.

\begin{table}[H]
\caption{Inference settings.}
\label{tab:evaluation_config}
\centering
\setlength{\tabcolsep}{2.2pt}
\renewcommand{\arraystretch}{0.92}
\begin{tabular}{@{}llcccccc@{}}
\toprule
Model and eval mode & Engine & Max new tokens & Context tokens & Temp. & Top-$p$ & Top-$k$ & Pres. pen. \\
\midrule
Qwen3-0.6/1.7/4/8B, \thmode             & vLLM   & 38912 & 40960  & 0.6 & 0.95 & 20  & 0.0 \\
Qwen3-1.7/4/8B, \nth              & vLLM   & 32768 & 40960  & 0.7 & 0.80 & 20  & 0.0 \\
Qwen3-4B-Instruct, \nth           & vLLM   & 32768 & 40960  & 0.7 & 0.80 & 20  & 0.0 \\
Qwen3-4B-Thinking                & vLLM   & 81920 & 90112  & 0.6 & 0.95 & 20  & 0.0 \\
Qwen3.5-4B, \thmode \space or \nth         & SGLang & 81920 & 131072 & 1.0 & 0.95 & 20  & 1.5 \\
DeepSeek-R1-Distill-Qwen-1.5B         & vLLM   & 32768 & 40960  & 0.6 & 0.95 & off & 0.0 \\
OLMo-3-7B-Think                       & SGLang & 32768 & 40960  & 0.6 & 0.95 & off & 0.0 \\
OLMo-3-7B-Instruct                    & SGLang & 32768 & 40960  & 0.6 & 0.95 & off & 0.0 \\
Falcon-H1R-7B                         & SGLang & 65536 & 73728  & 0.6 & 0.95 & off & 0.0 \\
MiMo-7B-RL                            & SGLang & 32768 & 32768 & 0.6 & 0.95 & off & 0.0 \\
\bottomrule
\end{tabular}
\end{table}

\clearpage
\section{Supplementary Experimental Results}
\label{app:main_results}

This section follows Section~5 and reports the corresponding extended comparisons, ablations, and model-level results.

\subsection{Valid Comparison Setting: Extended Reasoning Mode Results}
\label{app:qwen_mode_extended}

\autoref{tab:qwen_mode_extended} gives the model-level results underlying the Qwen3 means in \autoref{tab:qwen_mode}. Qwen3-1.7B, Qwen3-4B, Qwen3-8B, and Qwen3.5-4B use the same grid of six configurations.

\begin{table}[H]
\caption{Model-level reasoning mode results underlying \autoref{tab:qwen_mode}, with the complete Qwen3.5-4B grid. Rows are averaged over four benchmarks. Notation follows \autoref{tab:qwen_mode}. Per-dataset results are in \autoref{tab:mode_per_dataset}.}
\label{tab:qwen_mode_extended}
\centering
\setlength{\tabcolsep}{3.0pt}
\scriptsize
\renewcommand{\arraystretch}{0.92}
\providecommand{\opsda}[1]{\makebox[2.85em][c]{#1}}
\providecommand{\opsdb}[1]{\makebox[3.55em][c]{#1}}
\providecommand{\opsdc}[1]{\makebox[4.2em][c]{#1}}
\begin{tabular*}{\linewidth}{@{\extracolsep{\fill}}lccc ccc c ccc c ccc @{}}
\toprule
\multirow{2}{*}[-2pt]{Model} & \multirow{2}{*}[-2pt]{S} & \multirow{2}{*}[-2pt]{T} & \multirow{2}{*}[-2pt]{Eval}
  & \multicolumn{3}{c}{Avg@8} & & \multicolumn{3}{c}{Pass@8} & & \multicolumn{3}{c}{Length} \\
\cmidrule(lr){5-7}\cmidrule(lr){9-11}\cmidrule(lr){13-15}
& & & & \opsda{Base} & \opsda{OPSD} & \opsda{$\Delta$} & & \opsda{Base} & \opsda{OPSD} & \opsda{$\Delta$} & & \opsdb{Base} & \opsdb{OPSD} & \opsdc{$\Delta$} \\
\midrule
Qwen3-1.7B & \thmode & \thmode & \thmode & \opsda{34.0} & \opsda{35.3} & \opsda{$+1.3$}  & & \opsda{59.2} & \opsda{60.8} & \opsda{$+1.6$}  & & \opsdb{18624} & \opsdb{22380} & \opsdc{$+20.2\%$} \\
Qwen3-1.7B & \thmode & \nth & \thmode & \opsda{34.0} & \opsda{19.2} & \opsda{$-14.8$} & & \opsda{59.2} & \opsda{39.2} & \opsda{$-20.0$} & & \opsdb{18624} & \opsdb{11109} & \opsdc{$-40.4\%$} \\
Qwen3-1.7B & \nth & \thmode & \nth & \opsda{9.8} & \opsda{14.1} & \opsda{$+4.3$}  & & \opsda{20.8} & \opsda{28.3} & \opsda{$+7.5$}  & & \opsdb{3923}  & \opsdb{9731}  & \opsdc{$+148.1\%$} \\
Qwen3-1.7B & \nth & \nth & \nth & \opsda{9.8} & \opsda{9.2} & \opsda{$-0.6$}  & & \opsda{20.8} & \opsda{20.8} & \opsda{$0.0$}   & & \opsdb{3923}  & \opsdb{8463}  & \opsdc{$+115.7\%$} \\
Qwen3-1.7B & \nth & \thmode & \thmode & \opsda{34.0} & \opsda{40.8} & \opsda{$+6.8$}  & & \opsda{59.2} & \opsda{63.3} & \opsda{$+4.1$}  & & \opsdb{18624} & \opsdb{19996} & \opsdc{$+7.4\%$} \\
Qwen3-1.7B & \thmode & \nth & \nth & \opsda{9.8} & \opsda{7.5} & \opsda{$-2.3$}  & & \opsda{20.8} & \opsda{20.0} & \opsda{$-0.8$}  & & \opsdb{3923}  & \opsdb{3003}  & \opsdc{$-23.5\%$} \\
\midrule
Qwen3-4B & \thmode & \thmode & \thmode & \opsda{62.3} & \opsda{57.9} & \opsda{$-4.4$}  & & \opsda{79.2} & \opsda{75.8} & \opsda{$-3.4$}  & & \opsdb{17769} & \opsdb{19949} & \opsdc{$+12.3\%$} \\
Qwen3-4B & \thmode & \nth & \thmode & \opsda{62.3} & \opsda{47.9} & \opsda{$-14.4$} & & \opsda{79.2} & \opsda{69.2} & \opsda{$-10.0$} & & \opsdb{17769} & \opsdb{15108} & \opsdc{$-15.0\%$} \\
Qwen3-4B & \nth & \thmode & \nth & \opsda{19.6} & \opsda{21.5} & \opsda{$+1.9$}  & & \opsda{38.3} & \opsda{45.8} & \opsda{$+7.5$}  & & \opsdb{6694}  & \opsdb{17204} & \opsdc{$+157.0\%$} \\
Qwen3-4B & \nth & \nth & \nth & \opsda{19.6} & \opsda{17.2} & \opsda{$-2.4$} & & \opsda{38.3} & \opsda{32.5} & \opsda{$-5.8$} & & \opsdb{6694}  & \opsdb{15907} & \opsdc{$+137.6\%$} \\
Qwen3-4B & \nth & \thmode & \thmode & \opsda{62.3} & \opsda{63.0} & \opsda{$+0.7$}  & & \opsda{79.2} & \opsda{79.2} & \opsda{$0.0$}   & & \opsdb{17769} & \opsdb{19294} & \opsdc{$+8.6\%$} \\
Qwen3-4B & \thmode & \nth & \nth & \opsda{19.6} & \opsda{16.0} & \opsda{$-3.6$}  & & \opsda{38.3} & \opsda{30.0} & \opsda{$-8.3$}  & & \opsdb{6694}  & \opsdb{5538}  & \opsdc{$-17.3\%$} \\
\midrule
Qwen3-8B & \thmode & \thmode & \thmode & \opsda{64.1} & \opsda{58.9} & \opsda{$-5.2$}  & & \opsda{78.3} & \opsda{76.7} & \opsda{$-1.6$}  & & \opsdb{17647} & \opsdb{21422} & \opsdc{$+21.4\%$} \\
Qwen3-8B & \thmode & \nth & \thmode & \opsda{64.1} & \opsda{42.8} & \opsda{$-21.3$} & & \opsda{78.3} & \opsda{70.0} & \opsda{$-8.3$}  & & \opsdb{17647} & \opsdb{13655} & \opsdc{$-22.6\%$} \\
Qwen3-8B & \nth & \thmode & \nth & \opsda{18.0} & \opsda{23.2} & \opsda{$+5.2$}  & & \opsda{37.5} & \opsda{45.0} & \opsda{$+7.5$}  & & \opsdb{5914}  & \opsdb{15343} & \opsdc{$+159.4\%$} \\
Qwen3-8B & \nth & \nth & \nth & \opsda{18.0} & \opsda{19.1} & \opsda{$+1.1$}  & & \opsda{37.5} & \opsda{32.5} & \opsda{$-5.0$}  & & \opsdb{5914}  & \opsdb{16533} & \opsdc{$+179.5\%$} \\
Qwen3-8B & \nth & \thmode & \thmode & \opsda{64.1} & \opsda{64.8} & \opsda{$+0.7$}  & & \opsda{78.3} & \opsda{76.7} & \opsda{$-1.6$}  & & \opsdb{17647} & \opsdb{19292} & \opsdc{$+9.3\%$} \\
Qwen3-8B & \thmode & \nth & \nth & \opsda{18.0} & \opsda{18.5} & \opsda{$+0.5$}  & & \opsda{37.5} & \opsda{33.3} & \opsda{$-4.2$}  & & \opsdb{5914}  & \opsdb{6771}  & \opsdc{$+14.5\%$} \\
\midrule
Qwen3.5-4B & \thmode & \thmode & \thmode & \opsda{80.6} & \opsda{0.0} & \opsda{$-80.6$} & & \opsda{91.7} & \opsda{0.0} & \opsda{$-91.7$} & & \opsdb{38096} & \opsdb{81920} & \opsdc{$+115.0\%$} \\
Qwen3.5-4B & \thmode & \nth & \thmode & \opsda{80.6} & \opsda{8.6} & \opsda{$-72.0$} & & \opsda{91.7} & \opsda{27.5} & \opsda{$-64.2$} & & \opsdb{38096} & \opsdb{73214} & \opsdc{$+92.2\%$} \\
Qwen3.5-4B & \nth & \thmode & \nth & \opsda{52.7} & \opsda{0.0} & \opsda{$-52.7$} & & \opsda{77.5} & \opsda{0.0} & \opsda{$-77.5$} & & \opsdb{10193} & \opsdb{81920} & \opsdc{$+703.7\%$} \\
Qwen3.5-4B & \nth & \nth & \nth & \opsda{52.7} & \opsda{8.0} & \opsda{$-44.7$} & & \opsda{77.5} & \opsda{35.0} & \opsda{$-42.5$} & & \opsdb{10193} & \opsdb{70710} & \opsdc{$+593.7\%$} \\
Qwen3.5-4B & \nth & \thmode & \thmode & \opsda{80.6} & \opsda{0.0} & \opsda{$-80.6$} & & \opsda{91.7} & \opsda{0.0} & \opsda{$-91.7$} & & \opsdb{38096} & \opsdb{81920} & \opsdc{$+115.0\%$} \\
Qwen3.5-4B & \thmode & \nth & \nth & \opsda{52.7} & \opsda{36.3} & \opsda{$-16.4$} & & \opsda{77.5} & \opsda{64.2} & \opsda{$-13.3$} & & \opsdb{10193} & \opsdb{13649} & \opsdc{$+33.9\%$} \\
\bottomrule
\end{tabular*}
\end{table}

The model-level results support the conclusion in \autoref{subsec:mode_results}: reasoning mode mismatch transfers mode-specific behavior rather than privileged semantics. Qwen3.5-4B is a collapse case under both matched modes and is analyzed in Appendix~\ref{app:qwen35_collapse}.

\subsection{Training Steps, Rollout Length, and Loss Token Position}
\label{app:main_length_details}

Figures~\ref{fig:opsd_extend300} and~\ref{fig:length_sweeps} and Table~\ref{tab:position_selection_summary} report the model-level trends. First-256 reuses OPSD-256, while Uni-256 and Final-256 use 1024-token rollouts and apply the loss to at most 256 uniformly sampled or final valid response positions. The results match the findings in \autoref{subsec:training_steps} and \autoref{subsec:length_results}: longer training and rollouts do not reliably improve accuracy, and early positions perform best under the fixed budget of 256 positions. Appendix~\ref{app:per_dataset_results} gives the benchmark-level values, and Appendix~\ref{app:seed_sensitivity} gives the seed repetitions.

\subsection{Privileged Information}
\label{app:privileged_information}

\subsubsection{Answer-Only Controls}
\label{app:answer_only_controls}

To extend the privileged information comparison in \autoref{subsec:teacher_context}, we vary the form and correctness of the answer supplied to the frozen teacher. The integer-answer controls use the same 3200 OpenThoughts problems and example order, materialized with sampler seed~1. Qwen3-1.7B and OLMo-3-7B-Think use matched \thmode{}/\thmode{} modes, training seed~42, global batch size~64, learning rate $10^{-6}$, contribution cap $\tau_{\mathrm{clip}}=0.05$, and prompt and student rollout caps of 1024 tokens. Each integer-answer run trains for 50 updates. The student receives the original problem, and all four integer-answer conditions retain the same answer-only teacher template in Appendix~\ref{app:training_details}.

Correct integer supplies the original answer $n$. Equivalent form expresses the same value as a fraction or decimal. Wrong (fixed) replaces it with $10n+1$, while Wrong (random) uses seeded selections from 26 numerical transformation rules, including adding one, multiplying by two, and dropping the last digit. The wrong-answer conditions retain the template's verified-answer framing while replacing its answer field. Random problems is the original answer-only setting on randomly sampled OpenThoughts problems without the integer restriction; it supplies their correct answers, rather than random answer values. We evaluate this reference at checkpoint~50 of the standard 100-step run. Its training data and learning-rate schedule therefore differ from the integer controls; the matched answer transformations should be compared against Correct integer.

\autoref{tab:answer_only_controls} reports equal-weight averages over AIME24, AIME25, AIME26, and HMMT25; \autoref{tab:answer_only_per_dataset} gives the breakdown. Each benchmark has 30 questions and eight generations per question. Length averages output tokens over all generations. Evaluation uses the unchanged benchmark questions, temperature~0.6, top-$p=0.95$, and output caps of 38912 tokens for Qwen3-1.7B and 32768 for OLMo-3-7B-Think.

The fixed wrong-answer mapping exceeds Correct integer in Avg@8 for both models, while the randomized mapping produces mixed changes. All tested answer-only conditions improve Avg@8 over the Qwen3-1.7B base but remain below the OLMo-3-7B-Think base. These observations challenge attributing training gains to the literal correctness of privileged answers alone. They do not establish that answer information is unused: $n\mapsto10n+1$ is invertible, and other transformations may retain information about the correct answer. These are single-training-seed comparisons without a statistical significance test.

\subsubsection{Prefix Length Stratification}
\label{app:cotlen}

At a fixed 256-token rollout length, we compare standard OPSD (Full) with training on gold trace length deciles 0--4 (Short prefix) or 7--9 (Long prefix). Consistent with \autoref{subsec:teacher_context}, longer teacher prefixes provide no consistent advantage. The ordering varies across models (\autoref{tab:teacher_cotlen}).

\begin{table}[H]
\caption{Results of teacher prefix length at a 256-token rollout length. Short prefix and Long prefix use gold trace length deciles 0--4 and 7--9 from OpenThoughts, respectively. Using the Qwen3-1.7B tokenizer, their token lengths have mean/$P_{50}$/$P_{75}$ values of 439/454/543 and 992/942/1077, respectively. $\Delta$ is the OPSD--Base change (percentage points for accuracy, relative for Length). Bold marks the best Avg@8 and Pass@8 among Full, Short prefix, and Long prefix within each model. Per-dataset results are in \autoref{tab:cotlen_per_dataset}.}
\label{tab:teacher_cotlen}
\centering
\setlength{\tabcolsep}{2.8pt}
\scriptsize
\renewcommand{\arraystretch}{0.88}
\providecommand{\cotav}[1]{\makebox[2.85em][c]{#1}}
\providecommand{\cotlen}[1]{\makebox[3.8em][c]{#1}}
\providecommand{\cotdel}[1]{\makebox[4.2em][c]{#1}}
\begin{tabular*}{\textwidth}{@{\extracolsep{\fill}}ll ccc c ccc c ccc@{}}
\toprule
\multirow{2}{*}[-2pt]{Model} & \multirow{2}{*}[-2pt]{Subset}
& \multicolumn{3}{c}{Avg@8} & & \multicolumn{3}{c}{Pass@8} & & \multicolumn{3}{c}{Length} \\
\cmidrule(lr){3-5}\cmidrule(lr){7-9}\cmidrule(lr){11-13}
& & \cotav{Base} & \cotav{OPSD} & \cotav{$\Delta$} &
& \cotav{Base} & \cotav{OPSD} & \cotav{$\Delta$} &
& \cotlen{Base} & \cotlen{OPSD} & \cotdel{$\Delta$} \\
\midrule
\multirow{3}{*}{Qwen3-1.7B} & Full & \cotav{34.0} & \cotav{\textbf{38.1}} & \cotav{$+4.1$} &
& \cotav{59.2} & \cotav{\textbf{62.5}} & \cotav{$+3.3$} &
& \cotlen{18.6k} & \cotlen{21.4k} & \cotdel{$+15.2\%$} \\
& Short prefix & \cotav{34.0} & \cotav{37.1} & \cotav{$+3.1$} &
& \cotav{59.2} & \cotav{\textbf{62.5}} & \cotav{$+3.3$} &
& \cotlen{18.6k} & \cotlen{20.9k} & \cotdel{$+12.3\%$} \\
& Long prefix & \cotav{34.0} & \cotav{36.7} & \cotav{$+2.7$} &
& \cotav{59.2} & \cotav{\textbf{62.5}} & \cotav{$+3.3$} &
& \cotlen{18.6k} & \cotlen{20.8k} & \cotdel{$+11.5\%$} \\
\midrule
\multirow{3}{*}{Qwen3-4B} & Full & \cotav{62.3} & \cotav{60.9} & \cotav{$-1.4$} &
& \cotav{79.2} & \cotav{75.8} & \cotav{$-3.4$} &
& \cotlen{17.8k} & \cotlen{19.1k} & \cotdel{$+7.4\%$} \\
& Short prefix & \cotav{62.3} & \cotav{\textbf{61.0}} & \cotav{$-1.3$} &
& \cotav{79.2} & \cotav{\textbf{79.2}} & \cotav{$+0.0$} &
& \cotlen{17.8k} & \cotlen{19.0k} & \cotdel{$+6.9\%$} \\
& Long prefix & \cotav{62.3} & \cotav{60.6} & \cotav{$-1.7$} &
& \cotav{79.2} & \cotav{74.2} & \cotav{$-5.0$} &
& \cotlen{17.8k} & \cotlen{19.1k} & \cotdel{$+7.4\%$} \\
\midrule
\multirow{3}{*}{Qwen3-4B-Instruct} & Full & \cotav{48.5} & \cotav{\textbf{49.6}} & \cotav{$+1.1$} &
& \cotav{71.7} & \cotav{\textbf{70.8}} & \cotav{$-0.9$} &
& \cotlen{9.1k} & \cotlen{10.8k} & \cotdel{$+18.0\%$} \\
& Short prefix & \cotav{48.5} & \cotav{48.9} & \cotav{$+0.4$} &
& \cotav{71.7} & \cotav{68.3} & \cotav{$-3.4$} &
& \cotlen{9.1k} & \cotlen{11.0k} & \cotdel{$+20.5\%$} \\
& Long prefix & \cotav{48.5} & \cotav{48.9} & \cotav{$+0.4$} &
& \cotav{71.7} & \cotav{67.5} & \cotav{$-4.2$} &
& \cotlen{9.1k} & \cotlen{10.6k} & \cotdel{$+15.8\%$} \\
\midrule
\multirow{3}{*}{\ml{OLMo-3-\\7B-Instruct}} & Full & \cotav{39.5} & \cotav{42.9} & \cotav{$+3.4$} &
& \cotav{60.8} & \cotav{\textbf{68.3}} & \cotav{$+7.5$} &
& \cotlen{7.6k} & \cotlen{8.3k} & \cotdel{$+9.8\%$} \\
& Short prefix & \cotav{39.5} & \cotav{\textbf{44.4}} & \cotav{$+4.9$} &
& \cotav{60.8} & \cotav{65.8} & \cotav{$+5.0$} &
& \cotlen{7.6k} & \cotlen{8.1k} & \cotdel{$+7.4\%$} \\
& Long prefix & \cotav{39.5} & \cotav{41.7} & \cotav{$+2.2$} &
& \cotav{60.8} & \cotav{64.2} & \cotav{$+3.4$} &
& \cotlen{7.6k} & \cotlen{8.2k} & \cotdel{$+8.2\%$} \\
\bottomrule
\end{tabular*}
\end{table}

\subsection{Loss Design Ablations}
\label{app:loss_design_details}

These ablations vary the divergence between the student and teacher distributions, the vocabulary support used to compute it, and the response positions included in the OPSD loss.

\subsubsection{Divergence Objective Ablation}
\label{app:divergence_ablation}

The official OPSD implementation uses the forward KL($\beta=0$). We additionally train with symmetric JSD ($\beta=0.5$) and reverse KL ($\beta=1$), keeping other settings fixed.

\paragraph{Loss Definitions and Implementation.}
At a valid response position $i$, write $\pi_i^S(v)=\pi_\theta(v\mid Q,y_{<i}^S)$ for the active student and $\pi_i^T(v)=\pi_{\theta_0}(v\mid Q,C,y_{<i}^S)$ for the teacher. For any two next-token distributions $a,b$ on the vocabulary $\mathcal{V}$,
\begin{equation}
\label{eq:kl_definition}
D_{\mathrm{KL}}(a\|b)=\sum_{v\in\mathcal{V}}a(v)\log\frac{a(v)}{b(v)}.
\end{equation}
Accordingly, the two KL endpoints evaluated here are
\begin{equation}
\label{eq:kl_endpoints}
\ell_i^{\mathrm{fwd}}=D_{\mathrm{KL}}(\pi_i^T\|\pi_i^S),
\qquad
\ell_i^{\mathrm{rev}}=D_{\mathrm{KL}}(\pi_i^S\|\pi_i^T).
\end{equation}
For $0<\beta<1$, let $m_{\beta,i}=(1-\beta)\pi_i^S+\beta\pi_i^T$ and use the generalized-JSD interior
\begin{equation}
\label{eq:generalized_jsd}
\begin{aligned}
\ell_i^{\mathrm{JSD}_\beta}
&=
\beta D_{\mathrm{KL}}(\pi_i^T\|m_{\beta,i})
+(1-\beta)D_{\mathrm{KL}}(\pi_i^S\|m_{\beta,i}), \\
\ell_i^{\mathrm{JSD}_{0.5}}
&=\tfrac12D_{\mathrm{KL}}\!\left(\pi_i^T\middle\|\tfrac{\pi_i^S+\pi_i^T}{2}\right)
+\tfrac12D_{\mathrm{KL}}\!\left(\pi_i^S\middle\|\tfrac{\pi_i^S+\pi_i^T}{2}\right).
\end{aligned}
\end{equation}
At $\beta=0$ and $\beta=1$, the implementation uses $\ell_i^{\mathrm{fwd}}$ and $\ell_i^{\mathrm{rev}}$, respectively. At $\beta=0.5$, it uses the symmetric JSD above. We set \texttt{jsd\_token\_clip} to $0.05$, capping each pointwise vocabulary contribution before summing over $v$ and averaging over valid response positions. Because negative contributions are not lower-clipped, the resulting surrogate can be negative.

The forward KL experiments use the official OPSD loss: full vocabulary distributions with each vocabulary contribution clipped before summation. Student sampling and teacher scoring both use temperature 1.1, and the loss is averaged over all valid response tokens.

Changing divergence loss does not provide a universal solution to improve performance of OPSD, as in \autoref{subsec:divergence_objectives}. Forward KL leads in some settings, while reverse KL or symmetric JSD leads in others (\autoref{tab:divergence_macro}). Appendix~\ref{app:topk_loss} further restricts the divergence to a vocabulary top-$k$ support.

\begin{table}[t]
\caption{Divergence ablation results averaged over four benchmarks. $\Delta$ is the absolute change from Base for Avg@8/Pass@8 and the relative change for Length. Bold marks the best Avg@8 and Pass@8 within each model. Decoding settings are fixed across objectives. Per-dataset results are in \autoref{tab:divergence_per_dataset}.}
\label{tab:divergence_macro}
\centering
\scriptsize
\setlength{\tabcolsep}{5.0pt}
\begin{tabular}{@{}ll cc@{\hspace{1.2em}}cc@{\hspace{1.2em}}cc@{}}
\toprule
Model & Objective & \multicolumn{2}{c}{Avg@8} & \multicolumn{2}{c}{Pass@8} & \multicolumn{2}{c}{Length} \\
\cmidrule(lr){3-4}\cmidrule(lr){5-6}\cmidrule(lr){7-8}
& & Value & $\Delta$ & Value & $\Delta$ & Value & $\Delta$ \\
\midrule
\multirow{3}{*}{Qwen3-1.7B}
& Fwd. KL ($\beta=0$) & \textbf{35.31} & $+1.3$ & \textbf{60.83} & $+1.6$ & 22.4k & $+20.3\%$ \\
& JSD ($\beta=0.5$) & 31.25 & $-2.7$ & 55.00 & $-4.2$ & 15.2k & $-18.4\%$ \\
& Rev. KL ($\beta=1$) & 26.46 & $-7.5$ & 47.50 & $-11.7$ & 24.4k & $+31.0\%$ \\
\midrule
\multirow{3}{*}{Qwen3-4B}
& Fwd. KL ($\beta=0$) & \textbf{57.92} & $-4.4$ & \textbf{75.83} & $-3.4$ & 19.9k & $+12.0\%$ \\
& JSD ($\beta=0.5$) & 56.56 & $-5.7$ & \textbf{75.83} & $-3.4$ & 14.1k & $-20.7\%$ \\
& Rev. KL ($\beta=1$) & 31.98 & $-30.3$ & 56.67 & $-22.5$ & 27.9k & $+57.0\%$ \\
\midrule
\multirow{3}{*}{Qwen3-4B-Thinking}
& Fwd. KL ($\beta=0$) & 65.83 & $-9.6$ & 83.33 & $-3.4$ & 43.4k & $+90.5\%$ \\
& JSD ($\beta=0.5$) & 68.02 & $-7.4$ & 82.50 & $-4.2$ & 17.7k & $-22.3\%$ \\
& Rev. KL ($\beta=1$) & \textbf{74.17} & $-1.2$ & \textbf{85.00} & $-1.7$ & 34.1k & $+49.7\%$ \\
\midrule
\multirow{3}{*}{\ml{OLMo-3-\\7B-Think}}
& Fwd. KL ($\beta=0$) & 58.75 & $-4.7$ & 77.50 & $-4.2$ & 21.7k & $+15.8\%$ \\
& JSD ($\beta=0.5$) & 62.08 & $-1.4$ & 81.67 & $0.0$ & 17.3k & $-7.7\%$ \\
& Rev. KL ($\beta=1$) & \textbf{62.40} & $-1.1$ & \textbf{82.50} & $+0.8$ & 19.9k & $+6.2\%$ \\
\midrule
\multirow{3}{*}{\ml{OLMo-3-\\7B-Instruct}}
& Fwd. KL ($\beta=0$) & 40.83 & $+1.3$ & \textbf{62.50} & $+1.7$ & 7.4k & $-2.1\%$ \\
& JSD ($\beta=0.5$) & 37.19 & $-2.3$ & 58.33 & $-2.5$ & 7.5k & $-0.8\%$ \\
& Rev. KL ($\beta=1$) & \textbf{42.50} & $+3.0$ & 61.67 & $+0.9$ & 8.5k & $+12.4\%$ \\
\midrule
\multirow{3}{*}{Falcon-H1R-7B}
& Fwd. KL ($\beta=0$) & \textbf{89.06} & $-0.4$ & \textbf{96.67} & $0.0$ & 22.1k & $+6.6\%$ \\
& JSD ($\beta=0.5$) & 88.44 & $-1.1$ & 95.83 & $-0.9$ & 20.4k & $-1.6\%$ \\
& Rev. KL ($\beta=1$) & 88.13 & $-1.4$ & \textbf{96.67} & $0.0$ & 20.4k & $-1.3\%$ \\
\midrule
\multirow{3}{*}{MiMo-7B-RL}
& Fwd. KL ($\beta=0$) & 44.27 & $-11.3$ & 64.17 & $-10.0$ & 16.7k & $-0.3\%$ \\
& JSD ($\beta=0.5$) & \textbf{52.60} & $-3.0$ & \textbf{72.50} & $-1.7$ & 14.1k & $-15.5\%$ \\
& Rev. KL ($\beta=1$) & 50.21 & $-5.4$ & \textbf{72.50} & $-1.7$ & 19.6k & $+17.3\%$ \\
\bottomrule
\end{tabular}
\end{table}

\newcommand{\appendixentropy}{%
\subsubsection{Entropy-Based Token Selection}
\label{subsec:entropy_training}

We test whether OPSD should focus on positions where the student is uncertain. At the 256-token rollout length, HE20/HE80 and LE20/LE80 apply the loss to the highest- and lowest-entropy 20\% or 80\% of response positions while Full uses all positions. All other settings are fixed. \autoref{tab:entropy_train} reports selected macro averages.

\begin{table}[t]
\caption{Selected macro average results for student entropy loss masking at a 256-token rollout length. Parentheses show changes from Base (percentage points for Avg@8/Pass@8, relative percentage for Length). HE/LE means the highest/lowest-entropy 20\%/80\%. Bold marks the best accuracy per model. Complete results are in \autoref{tab:entropy_train_per_dataset}.}
\label{tab:entropy_train}
\centering
\scriptsize
\setlength{\tabcolsep}{2.4pt}
\providecommand{\entacc}[2]{##1\,(##2)}
\providecommand{\entlen}[2]{##1\,(##2)}
\resizebox{\linewidth}{!}{%
\begin{tabular}{@{}l ccc@{\hspace{1.2em}}ccc@{\hspace{1.2em}}ccc@{}}
\toprule
\multirow{2}{*}{Subset}
& \multicolumn{3}{c}{Qwen3-1.7B}
& \multicolumn{3}{c}{Qwen3-4B-Thinking}
& \multicolumn{3}{c}{OLMo-3-7B-Instruct} \\
\cmidrule(lr){2-4}\cmidrule(lr){5-7}\cmidrule(lr){8-10}
& Avg@8 & Pass@8 & Length
& Avg@8 & Pass@8 & Length
& Avg@8 & Pass@8 & Length \\
\midrule
Full
& \entacc{\textbf{38.1}}{$+4.1$} & \entacc{62.5}{$+3.3$} & \entlen{21.4k}{$+15.2\%$}
& \entacc{70.3}{$-5.1$} & \entacc{83.3}{$-3.4$} & \entlen{30.5k}{$+34.0\%$}
& \entacc{42.9}{$+3.4$} & \entacc{68.3}{$+7.5$} & \entlen{8.3k}{$+9.8\%$} \\
HE20
& \entacc{36.9}{$+2.9$} & \entacc{\textbf{65.8}}{$+6.6$} & \entlen{21.7k}{$+16.7\%$}
& \entacc{71.8}{$-3.6$} & \entacc{\textbf{85.0}}{$-1.7$} & \entlen{30.1k}{$+32.2\%$}
& \entacc{42.2}{$+2.7$} & \entacc{\textbf{69.2}}{$+8.4$} & \entlen{8.1k}{$+7.0\%$} \\
HE80
& \entacc{37.8}{$+3.8$} & \entacc{61.7}{$+2.5$} & \entlen{20.8k}{$+11.6\%$}
& \entacc{72.1}{$-3.3$} & \entacc{83.3}{$-3.4$} & \entlen{30.2k}{$+32.6\%$}
& \entacc{42.3}{$+2.8$} & \entacc{68.3}{$+7.5$} & \entlen{8.1k}{$+6.4\%$} \\
LE20
& \entacc{37.6}{$+3.6$} & \entacc{64.2}{$+5.0$} & \entlen{18.0k}{$-3.5\%$}
& \entacc{73.9}{$-1.5$} & \entacc{\textbf{85.0}}{$-1.7$} & \entlen{22.5k}{$-1.3\%$}
& \entacc{40.2}{$+0.7$} & \entacc{66.7}{$+5.9$} & \entlen{7.9k}{$+3.7\%$} \\
LE80
& \entacc{33.2}{$-0.8$} & \entacc{59.2}{$+0.0$} & \entlen{17.7k}{$-5.2\%$}
& \entacc{\textbf{74.5}}{$-0.9$} & \entacc{83.3}{$-3.4$} & \entlen{26.0k}{$+14.1\%$}
& \entacc{\textbf{44.1}}{$+4.6$} & \entacc{\textbf{69.2}}{$+8.4$} & \entlen{8.5k}{$+11.3\%$} \\
\bottomrule
\end{tabular}
}
\end{table}

Entropy-based position selection provides no consistent improvement over Full, matching \autoref{subsec:divergence_objectives}. Low-entropy variants more often shorten or stabilize outputs, whereas high-entropy variants more often exhibit the length growth observed under standard OPSD (\autoref{tab:entropy_train_per_dataset}). Appendix~\ref{subsec:token_entropy_analysis} reports the corresponding initial token-level signals.
}

\appendixentropy

\subsubsection{Selective-Supervision Ablations}
\label{app:selective_supervision}

We test two ways to concentrate OPSD supervision: restricting the divergence to a small vocabulary support and applying the loss only at response positions favored by the teacher.

\begin{table}[H]
\caption{Selective OPSD supervision: Avg@8 change from Base (percentage points). Panels (a) and (b) use 1024- and 256-token rollouts. Full metrics are in \autoref{tab:topk_loss_macro} and \autoref{tab:advt4_macro}.}
\label{tab:selective_supervision_app}
\centering
\scriptsize
\begin{minipage}[t]{0.32\textwidth}
\vspace{0pt}
\centering
\textbf{(a) Vocabulary support (1024-token rollout)}\\[1pt]
\setlength{\tabcolsep}{2.5pt}
\begin{tabular}{@{}lccc@{}}
\toprule
Model & Full & Top-16 & Top-4 \\
\midrule
Qwen3-1.7B         & $+1.3$ & $-1.0$ & $-8.2$ \\
Qwen3-4B-Instruct  & $+1.5$ & $+3.5$ & $-2.8$ \\
OLMo-3-7B-Think    & $-1.1$ & $-3.6$ & $-60.0$ \\
\bottomrule
\end{tabular}
\end{minipage}\hspace{0.015\textwidth}%
\begin{minipage}[t]{0.32\textwidth}
\vspace{0pt}
\centering
\textbf{(b) Teacher-favored positions (256-token rollout)}\\[1pt]
\setlength{\tabcolsep}{3.0pt}
\begin{tabular}{@{}lccc@{}}
\toprule
Model & Full & AdvT4 & AdvT16 \\
\midrule
Qwen3-1.7B        & $+4.1$ & $+0.7$ & $+3.2$ \\
Qwen3-4B          & $-1.4$ & $-1.8$ & $-0.1$ \\
Qwen3-4B-Thinking & $-5.1$ & $-1.0$ & $-0.2$ \\
OLMo-3-7B-Think   & $-2.7$ & $-1.2$ & $-1.7$ \\
\bottomrule
\end{tabular}
\end{minipage}
\end{table}

As in \autoref{subsec:divergence_objectives}, neither vocabulary truncation nor teacher-favored position selection improves consistently across models.

\paragraph{Vocabulary Top-$k$ Distillation Loss.}
\label{app:topk_loss}

The default OPSD loss in \autoref{eq:opsd_clipped} is a full vocabulary divergence. We vary only its vocabulary support: the top-$k$ variant evaluates per-token KL on $k$ tokens and renormalizes both student and teacher distributions on that support, while all other settings remain fixed.
Forward KL ($\beta=0$) takes the teacher's top-$k$ tokens and uses $D_{\mathrm{KL}}(\widetilde{\pi}_i^T\|\widetilde{\pi}_i^S)$. Reverse KL ($\beta=1$) takes the student's top-$k$ tokens and uses $D_{\mathrm{KL}}(\widetilde{\pi}_i^S\|\widetilde{\pi}_i^T)$.

We report three models: Qwen3-1.7B in \thmode{}/\thmode{} with forward KL, Qwen3-4B-Instruct in \nth{}/\nth{} with forward KL, and OLMo-3-7B-Think in \thmode{}/\thmode{} with reverse KL, the stronger OLMo endpoint in \autoref{tab:divergence_macro}. \autoref{tab:topk_loss_macro} compares $k\in\{4,16\}$ against the matched full vocabulary run.

\begin{table}[H]
\caption{Vocabulary top-$k$ loss results averaged over four benchmarks. The Qwen3 rows use forward KL with teacher Top-$k$. OLMo-3-7B-Think uses reverse KL with student Top-$k$. Full vocab rows are the matched controls from \autoref{tab:divergence_macro}. $\Delta$ is the absolute change from Base for Avg@8/Pass@8 and the relative change for Length. Bold marks the best Avg@8 and Pass@8 within each model. Per-dataset results are in \autoref{tab:topk_loss_per_dataset}.}
\label{tab:topk_loss_macro}
\centering
\scriptsize
\setlength{\tabcolsep}{5.0pt}
\begin{tabular}{@{}ll cc@{\hspace{1.2em}}cc@{\hspace{1.2em}}cc@{}}
\toprule
Model & Loss support & \multicolumn{2}{c}{Avg@8} & \multicolumn{2}{c}{Pass@8} & \multicolumn{2}{c}{Length} \\
\cmidrule(lr){3-4}\cmidrule(lr){5-6}\cmidrule(lr){7-8}
& & Value & $\Delta$ & Value & $\Delta$ & Value & $\Delta$ \\
\midrule
\multirow{3}{*}{Qwen3-1.7B}
& Full vocab & \textbf{35.31} & $+1.3$ & 60.83 & $+1.6$ & 22.4k & $+20.3\%$ \\
& $k=16$ & 33.02 & $-1.0$ & \textbf{62.50} & $+3.3$ & 22.6k & $+21.4\%$ \\
& $k=4$ & 25.83 & $-8.2$ & 47.50 & $-11.7$ & 28.8k & $+54.6\%$ \\
\midrule
\multirow{3}{*}{\ml{Qwen3-4B-\\Instruct}}
& Full vocab & 50.00 & $+1.5$ & 70.00 & $-1.7$ & 10.6k & $+15.2\%$ \\
& $k=16$ & \textbf{51.98} & $+3.5$ & \textbf{70.83} & $-0.9$ & 10.6k & $+15.4\%$ \\
& $k=4$ & 45.73 & $-2.8$ & 69.17 & $-2.5$ & 8.2k & $-10.9\%$ \\
\midrule
\multirow{3}{*}{\ml{OLMo-3-\\7B-Think}}
& Full vocab & \textbf{62.40} & $-1.1$ & \textbf{82.50} & $+0.8$ & 19.9k & $+6.2\%$ \\
& $k=16$ & 59.90 & $-3.6$ & 81.67 & $0.0$ & 20.2k & $+7.9\%$ \\
& $k=4$ & 3.54 & $-60.0$ & 23.33 & $-58.4$ & 32.2k & $+71.9\%$ \\
\bottomrule
\end{tabular}
\end{table}

Vocabulary truncation provides no reliable improvement: Top-16 remains close to the full vocabulary objective, whereas Top-4 degrades more sharply across the tested models (\autoref{tab:topk_loss_macro}).

\paragraph{Teacher-Favored Position Selection (AdvT4 / AdvT16).}
\label{app:advt4}

The default OPSD loss averages over all valid response tokens. We vary only the position selection rule: AdvT$k$ restricts the per-completion average to positions where the frozen teacher ranks the sampled student token among its top-$k$ vocabulary outcomes, with $k\in\{4,16\}$ and all other settings fixed. \autoref{tab:advt4_macro} compares Base, the matched 256-token full loss control, and both AdvT checkpoints.

\begin{table}[H]
\caption{Teacher-favored position selection (AdvT4 / AdvT16) versus the matched 256-token full loss control. Bold marks the best Avg@8 / Pass@8 within each model.}
\label{tab:advt4_macro}
\centering
\scriptsize
\setlength{\tabcolsep}{1.2pt}
\renewcommand{\arraystretch}{0.95}
\begin{tabular*}{\linewidth}{@{\extracolsep{\fill}}l ccc ccc ccc ccc@{}}
\toprule
\multirow{2}{*}{Stage}
& \multicolumn{3}{c}{Qwen3-1.7B}
& \multicolumn{3}{c}{Qwen3-4B}
& \multicolumn{3}{c}{Qwen3-4B-Thinking}
& \multicolumn{3}{c}{OLMo-3-7B-Think} \\
\cmidrule(lr){2-4}\cmidrule(lr){5-7}\cmidrule(lr){8-10}\cmidrule(lr){11-13}
& Avg@8 & Pass@8 & Length
& Avg@8 & Pass@8 & Length
& Avg@8 & Pass@8 & Length
& Avg@8 & Pass@8 & Length \\
\midrule
Base
& 34.0 & 59.2 & 18.6k
& \textbf{62.3} & \textbf{79.2} & 17.8k
& \textbf{75.4} & \textbf{86.7} & 22.8k
& \textbf{63.5} & 81.7 & 18.7k \\
c256
& \textbf{38.1} & \textbf{62.5} & 21.4k
& 60.9 & 75.8 & 19.1k
& 70.3 & 83.3 & 30.5k
& 60.8 & 80.0 & 21.4k \\
AdvT4
& 34.7 & 59.2 & 19.8k
& 60.5 & 76.7 & 18.4k
& 74.4 & 85.8 & 26.9k
& 62.3 & \textbf{82.5} & 21.2k \\
AdvT16
& 37.2 & 61.7 & 19.5k
& 62.2 & 78.3 & 18.0k
& 75.2 & 85.0 & 26.8k
& 61.9 & 80.8 & 21.4k \\
\bottomrule
\end{tabular*}
\end{table}

Neither AdvT4 nor AdvT16 consistently improves over the matched 256-token full loss control. AdvT16 remains closer to the control on the Qwen3 models, while AdvT4 produces the milder OLMo degradation (\autoref{tab:advt4_macro}).

\subsubsection{Entropy-Stratified Signal}
\label{app:token_entropy}
\label{subsec:token_entropy_analysis}

Signal concentration does not imply training utility: entropy-based position selection provides no consistent benefit, as in \autoref{subsec:divergence_objectives}, although initial divergence concentrates at high-entropy positions (\autoref{tab:token_entropy_signal_app}).
We use the HE20 KL and HE20 $|$adv.$|$ shares defined in Eqs.~(\ref{eq:he20_kl_share}) and~(\ref{eq:he20_abs_adv_share}) in \autoref{subsec:signal_performance}.

\begin{table}[H]
\caption{Token-level entropy tail signals for \autoref{tab:entropy_train}. HE20 KL/Adv are the highest-entropy 20\% shares of total forward KL and absolute advantage. $D>.05$ is the percentage of positions with unclipped forward KL above 0.05 within each entropy group.}
\label{tab:token_entropy_signal_app}
\centering
\setlength{\tabcolsep}{2.8pt}
\scriptsize
\renewcommand{\arraystretch}{0.92}
\begin{tabular}{@{}ll cccc cc@{}}
\toprule
Model & Mode &
\multicolumn{2}{c}{Highest 20\%} &
\multicolumn{2}{c}{Lowest 20\%} &
\multicolumn{2}{c}{HE20 share} \\
\cmidrule(lr){3-4}\cmidrule(lr){5-6}\cmidrule(lr){7-8}
& & KL & $D>.05$ & KL & $D>.05$ & KL & Adv \\
\midrule
Qwen3-1.7B & \thmode/\thmode & 0.371 & 66.4 & 0.004 & 0.1 & 51.5 & 69.8 \\
Qwen3-1.7B & \nth/\nth & 0.277 & 61.5 & 0.003 & 0.1 & 55.3 & 72.9 \\
Qwen3-4B & \thmode/\thmode & 0.349 & 68.4 & 0.002 & 0.1 & 51.2 & 68.4 \\
Qwen3-4B & \nth/\nth & 0.221 & 57.2 & 0.002 & 0.1 & 60.1 & 74.4 \\
Qwen3-4B-Instruct & \nth/\nth & 0.522 & 66.0 & 0.003 & 0.1 & 59.7 & 76.3 \\
\ml{Qwen3-4B-\\Thinking-2507} & \thmode/\thmode & 0.405 & 71.1 & 0.004 & 0.2 & 50.0 & 69.1 \\
OLMo-3-7B-Think & \thmode/\thmode & 0.266 & 62.9 & 0.006 & 0.3 & 45.9 & 60.1 \\
OLMo-3-7B-Instruct & \nth/\nth & 0.347 & 70.5 & 0.004 & 0.3 & 54.8 & 64.7 \\
\bottomrule
\end{tabular}
\end{table}

\subsection{Initial Model Ability}
\label{app:main_ability_details}

\autoref{tab:capability_per_dataset} reports the complete math benchmark results behind \autoref{fig:initial_capability}a, including output length. The corresponding coding breakdown is in \autoref{tab:capability_coding_per_dataset}. The math breakdown supports the conclusion in \autoref{subsec:initial_capability}: gains are dependent among models, with improvements confined to a few less capable or Instruct checkpoints and degradation on most stronger reasoning checkpoints.

\subsection{Post-Training Stage and SFT Initialization Sweep}
\label{app:sft_initialization}
\label{app:sft_grpo_then_opsd}

We initialize OPSD from Qwen3-1.7B-Base and Qwen3-4B-Base checkpoints after 0, 5k, 10k, and 15k SFT steps. Student and teacher use the same checkpoint. Appendix~\ref{app:sft_configuration} gives the SFT configs.

We evaluate each SFT checkpoint directly on four benchmarks, then train and evaluate OPSD with the same checkpoint. Reported values are benchmark averages.

\begin{figure}[H]
\centering
\includegraphics[width=\linewidth]{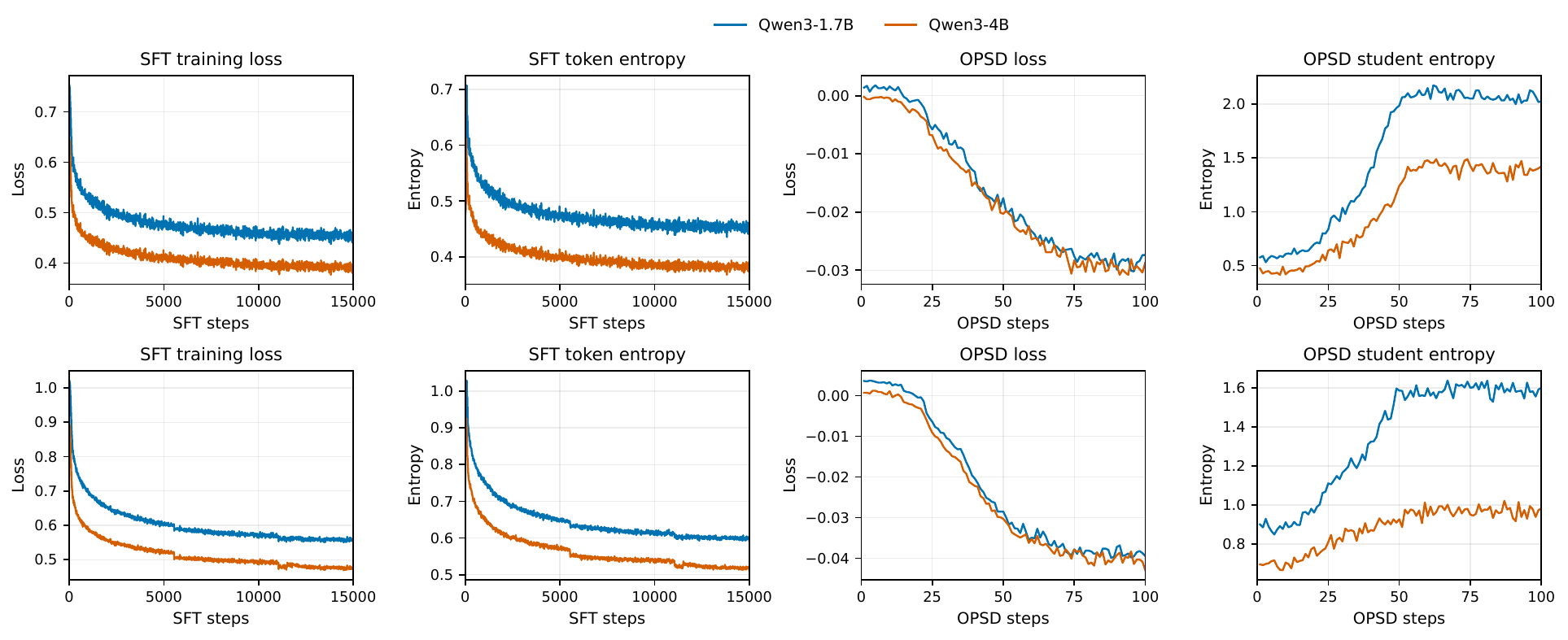}
\caption{Training traces for the SFT initialization sweep on math (top) and coding (bottom). Columns show SFT loss, SFT token entropy, OPSD loss, and OPSD student entropy for Qwen3-1.7B-Base and Qwen3-4B-Base. The OPSD runs are initialized from the corresponding 15k SFT checkpoints.}
\label{fig:sft_training_traces}
\end{figure}

\begin{table}[!ht]
\caption{Macro average direct SFT and downstream OPSD results. Bold marks the best Avg@8 and Pass@8 within each model and stage. Per-checkpoint and per-dataset results are in \autoref{tab:sft_per_dataset_1p7b} and \autoref{tab:sft_per_dataset_4b}.}
\label{tab:sft_initialization}
\centering
\scriptsize
\setlength{\tabcolsep}{3.4pt}
\renewcommand{\arraystretch}{0.92}
\begin{tabular}{@{}llccccccc@{}}
\toprule
Model & SFT steps & \multicolumn{3}{c}{Direct SFT} & \multicolumn{4}{c}{After 100-step OPSD} \\
\cmidrule(lr){3-5}\cmidrule(lr){6-9}
& & Avg@8 & Pass@8 & Format & Avg@8 & Pass@8 & Format & Length \\
\midrule
\multirow{4}{*}{Qwen3-1.7B-Base}
& Base & 2.08 & 8.33 & 80.63 & 1.15 & 7.50 & 63.23 & 8.0k \\
& 5k   & 7.08 & 21.67 & 51.98 & 0.31 & 2.50 & 0.83 & 34.6k \\
& 10k  & 9.59 & 20.83 & 64.58 & \textbf{3.12} & \textbf{11.67} & 6.56 & 35.2k \\
& 15k  & \textbf{11.46} & \textbf{29.17} & 63.54 & 1.56 & 7.50 & 3.85 & 36.2k \\
\midrule
\multirow{4}{*}{Qwen3-4B-Base}
& Base & 7.29 & 20.00 & 87.19 & \textbf{4.79} & \textbf{17.50} & 74.59 & 10.3k \\
& 5k   & 25.63 & 50.83 & 76.04 & 0.83 & 4.17 & 2.08 & 33.1k \\
& 10k  & 32.29 & \textbf{60.84} & 90.52 & 1.67 & 8.33 & 4.17 & 32.8k \\
& 15k  & \textbf{34.69} & 60.00 & 89.27 & 1.98 & 8.33 & 3.44 & 32.9k \\
\bottomrule
\end{tabular}
\end{table}

\FloatBarrier

OPSD remains sensitive to the post-training stage, as in \autoref{sec:pipeline}. Direct SFT ability rises with training, but OPSD collapses from the partially SFT-trained checkpoints (\autoref{tab:sft_initialization}).

We complete the Qwen3-1.7B pipeline by initializing both student and teacher from the GRPO step-100 actor and applying the size-matched OPSD protocol without other changes.

\begin{table}[H]
\caption{Qwen3-1.7B-Base post-training pipeline. SFT denotes the 15k-step checkpoint, and GRPO and OPSD denote 100-step continuations. Per-dataset results are in \autoref{tab:sft_grpo_opsd_per_dataset}.}
\label{tab:sft_grpo_then_opsd}
\centering
\scriptsize
\setlength{\tabcolsep}{5.0pt}
\renewcommand{\arraystretch}{0.92}
\begin{tabular}{@{}lcccc@{}}
\toprule
Model / stage & Avg@8 & Pass@8 & Format & Length \\
\midrule
Qwen3-1.7B-Base & 2.08 & 8.33 & 80.63 & 5.6k \\
$+$ OPSD & 1.15 & 7.50 & 63.23 & 8.0k \\
$+$ SFT & 11.46 & 29.17 & 63.54 & 21.4k \\
$+$ SFT $+$ OPSD & 1.56 & 7.50 & 3.85 & 36.2k \\
$+$ SFT $+$ GRPO & 11.88 & 30.00 & 87.71 & 16.6k \\
$+$ SFT $+$ GRPO $+$ OPSD & 4.69 & 14.17 & 14.06 & 35.5k \\
\bottomrule
\end{tabular}
\end{table}

The 1.7B results are consistent with the conclusion in \autoref{sec:pipeline}: GRPO largely preserves the SFT capability, whereas applying OPSD afterward sharply degrades accuracy and format. The corresponding 4B aggregate pipeline is reported in \autoref{tab:sft_grpo_control}. Its logs separate early, as entropy and gradient norm rise for the SFT-then-GRPO initialization but remain bounded or decrease for Qwen3-4B (\autoref{subsec:training_dynamics}).

\subsubsection{Output Degeneration Examples}
\label{app:sft_output_analysis}

For the SFT training, we inspect all eight samples for each of the 120 evaluation problems (960 generations per row). Consistent with the failure mode in \autoref{sec:pipeline}, \autoref{tab:sft_output_degen} shows a shift from partially successful direct SFT generations to repetitive, non-terminating outputs with few boxed answers after OPSD. The length increase reflects collapse, not longer successful reasoning.

\begin{table}[H]
\caption{Diagnostics at the generation level averaged over four benchmarks. OPSD denotes evaluation after 100-step OPSD from that SFT checkpoint. Boxed denotes Format.}
\label{tab:sft_output_degen}
\centering
\scriptsize
\setlength{\tabcolsep}{2.8pt}
\begin{tabular}{@{}ll cccccc@{}}
\toprule
Checkpoint & Cond. & $N$ & Tokens & $\geq$32k & $\geq$38k & Boxed & Avg@8 \\
\midrule
1.7B SFT-5k  & direct & 960 & 23.8k & 39.8\% & 19.5\% & 52.0\% & 7.1\% \\
1.7B SFT-5k  & OPSD   & 960 & 34.7k & 48.0\% & 32.4\% & 0.8\% & 0.3\% \\
1.7B SFT-10k & OPSD   & 960 & 35.2k & 71.2\% & 55.4\% & 6.6\% & 3.1\% \\
4B SFT-5k    & OPSD   & 960 & 33.1k & 80.0\% & 0.0\% & 2.1\% & 0.8\% \\
\bottomrule
\end{tabular}
\end{table}

A representative correct trajectory from the direct 1.7B SFT-5k checkpoint is shown below:

\begin{promptbox}
[AIME 2024-0, Qwen3-1.7B SFT-5k, direct, excerpt]

9/s = 4 - t/60
9/(s + 2) = 2.4 - t/60

... final answer: \textbackslash boxed\{204\}
\end{promptbox}

After 100-step OPSD, the model exhibits severe output degeneration: responses become excessively long, fail to produce boxed or correct answers, and repeatedly pursue invalid reasoning paths. A representative trajectory is shown below (ellipses denote omitted tokens):

\begin{promptbox}
[AIME 2024-0, same problem, Qwen3-1.7B SFT-5k after 100-step OPSD, excerpt]

Total Time T1 = (9/s) + (t/60) = ?

Alternatively perhaps there is a third condition involving different speeds and times?
Alternatively perhaps I can get another equation from another scenario where she walks slower than s?

Hmm.
Hmm.
...  [repeated 15134x: Hmm. 78x: Alternatively]

.
.
.
[no \textbackslash boxed\{\} answer, 38912 tokens]
\end{promptbox}

The 4B model exhibits the same failure to enter a final answer mode after OPSD. A representative trajectory ending is shown below:

\begin{promptbox}
[AIME 2025-0, Qwen3-4B SFT-5k after 100-step OPSD, final characters]

!!!!!!!!!!!!!!!!!!!!!!!!!!!!!!!! ...
[no \textbackslash boxed\{\} answer, 33583 tokens]
\end{promptbox}

\subsubsection{GRPO on the Collapsing SFT Initializations}
\label{app:sft_grpo_control}

\begingroup
\setlength{\parskip}{.5pc}%
We continue the Qwen3-1.7B-Base and Qwen3-4B-Base SFT-15k checkpoints with 100 GRPO updates under a DAPO-style verifiable math reward. These checkpoints have the strongest direct SFT Avg@8 within their respective sweeps in \autoref{tab:sft_initialization}, and both collapse under 100-step OPSD. Appendix~\ref{app:grpo_configuration} gives the GRPO configuration.

At both model sizes, GRPO remains approximately flat in Avg@8 relative to direct SFT and preserves a high format rate, whereas applying OPSD after GRPO again causes sharp degradation. The 1.7B replication is therefore consistent with the pipeline-stage conclusion in \autoref{sec:pipeline}.
\par
\endgroup

\subsection{OPSD versus GRPO on Qwen3 Thinking Models}
\label{app:released_grpo}

\subsubsection{Math}
For the OPSD--GRPO comparison in \autoref{tab:opsd_grpo_compute}, we train both methods from the Qwen3-1.7B, Qwen3-4B, and Qwen3-8B models in their native \thmode{} mode and evaluate after 50 updates. Both methods consume exactly the same 3200 examples from the OpenMathReasoning integer answer subset in the same order: we materialize the first 3200 rows produced by the GRPO random sampler (seed~1) as the sequential OPSD dataset. GRPO uses the final answer only for reward verification, whereas OPSD supplies the corresponding post-thinking solution as privileged teacher context.

The matched OPSD runs use a frozen same-model teacher, full-parameter updates, a rollout cap of 1024 tokens, global batch size 64, learning rate $1\times10^{-6}$, forward KL, token contribution clipping at 0.05, and seed~42. Qwen3-1.7B trains on 2 A800 GPUs with per-device batch size 8 and gradient accumulation 4; Qwen3-4B uses 2 GPUs, per-device batch size 4, and gradient accumulation 8; and Qwen3-8B uses 4 GPUs, per-device batch size 2, and gradient accumulation 8. The 8B preprocessed dataset sets the maximum prompt length to 2584 tokens to retain the full privileged solution. Appendix~\ref{app:grpo_configuration} gives the GRPO configuration, and \autoref{tab:omr_opsd_per_dataset} and \autoref{tab:released_grpo_per_dataset} report the benchmark-level results.

The cost column in \autoref{tab:opsd_grpo_compute} uses training loop time through the evaluated checkpoint and excludes evaluation and data preprocessing. The results support the tradeoff in \autoref{sec:opsd_grpo_compute}: OPSD uses substantially fewer GPU-hours under these recipes, but its accuracy advantage does not persist as model size increases.

\subsubsection{Coding}
For coding, we pool the APPS train and test splits across all difficulty levels, retain at most 10 execution tests per problem, and format prompts in the native Qwen3 \thmode{} mode. Within each model size, OPSD and GRPO consume the same prompts in the same order for 50 updates. The per-update prompt batches are 96, 48, and 32 for Qwen3-1.7B, Qwen3-4B, and Qwen3-8B, respectively, corresponding to 4800, 2400, and 1600 training prompts. OPSD supplies the official solution as privileged teacher context, whereas GRPO uses execution-based correctness.

The coding OPSD runs use a frozen same-model teacher, full-parameter updates, a 2048-token prompt cap, a 1024-token rollout cap, learning rate $1\times10^{-6}$, forward KL with token-contribution clipping at 0.05, and seed~42. Student and teacher both use \thmode{} mode. The 1.7B run uses 2 A800 GPUs with per-device batch size 8 and gradient accumulation 6; the 4B run uses 2 GPUs with per-device batch size 4 and gradient accumulation 6; and the 8B run uses 4 GPUs with per-device batch size 4 and gradient accumulation 2.

The GRPO settings are given in Appendix~\ref{app:grpo_configuration}. Per-dataset results are reported in \autoref{tab:apps_opsd_grpo_per_dataset}.

\clearpage
\section{Supplementary Analysis Results}
\label{app:analysis_results}
\label{app:empirical_regimes}
\label{app:token_analysis}

This section reports the protocol and per-model results underlying \autoref{sec:token_analysis}. Unless noted otherwise, the analysis uses 2048 prompts with two on-policy rollouts each. The position window study uses a rollout cap of 6144 tokens.

\subsection{Token-Level Analysis Protocol}
\label{app:token_analysis_protocol}

We compute the token-level statistics in \autoref{sec:token_analysis} offline from the same student and teacher distributions used in training. For each initial checkpoint, we sample 2048 training prompts with two on-policy rollouts each ($n=4096$) at temperature $1.1$ and a rollout cap of 1024 tokens. The frozen self-teacher scores the visited response tokens under its privileged context. Reported forward KL is the vocabulary-summed, unclipped $\beta=0$ divergence. Unless explicitly described as shared, the analyses of matched modes, teacher context, and entropy use independent rollout draws under this common protocol. The model type summary in \autoref{tab:checkpoint_type_signal} computes all of its metrics within the entropy analysis draw for each checkpoint.

Throughout the token-level analysis, $D>.05$ denotes $100\,|\{i\in\mathcal{I}:D_i^{\mathrm{fwd}}>0.05\}|/|\mathcal{I}|$, where $D_i^{\mathrm{fwd}}=D_{\mathrm{KL}}(\pi_i^T\|\pi_i^S)$ is the vocabulary-summed, unclipped forward KL and $\mathcal{I}$ is the set of valid response positions being summarized. For HE20/LE20 results, $\mathcal{I}$ is restricted to the corresponding entropy group. This percentage thresholds the full-distribution divergence, not the absolute sampled token log advantage, and is distinct from the training time pointwise contribution cap. Encourage and Discourage are the fractions of positions with sampled token log advantage $a_i>0$ and $a_i<0$. The remainder are ties at stored precision. Because $y_i^S\sim\pi_i^S$, $\mathbb{E}_{y_i^S\sim\pi_i^S}[a_i]=-D_{\mathrm{KL}}(\pi_i^S\|\pi_i^T)\leq0$, so student policy sampling yields a non-positive expectation. Measurements at initialization exclude student drift and optimizer state effects.

For the teacher-side context conditions of \autoref{tab:token_context_signal}, Answer only, Unrelated solution, short and long Full solution, and CoT+Solution share the same student rollouts (sampled under the long solution pool. See Appendix~\ref{app:token_sol_long}). For the position window analysis of Appendix~\ref{subsec:token_length_windows}, the protocol is identical except that completions are capped at 6144 tokens and metrics are aggregated within fixed response position intervals.

\subsection{Additional Token-Level Signals with Matched Modes}
\label{app:token_mode_signal}

\autoref{tab:token_mode_signal_app} complements \autoref{tab:token_signal_summary} with eight additional checkpoints in their native matched modes.
\begin{table}[H]
\caption{Additional signals. Ordered by Base Pass@8. Definitions follow \autoref{tab:token_signal_summary}.}
\label{tab:token_mode_signal_app}
\centering
\setlength{\tabcolsep}{2.8pt}
\scriptsize
\renewcommand{\arraystretch}{0.92}
\begin{tabular}{@{}lccccccc@{}}
\toprule
Model & S & T & Fwd.\ KL & $D>.05$ (\%) & Agree. (\%) & Enc. (\%) & Disc. (\%) \\
\midrule
Qwen3-0.6B & \thmode & \thmode & 0.074 & 21.7 & 92.6 & 36.8 & 62.3 \\
\ml{DeepSeek-R1-\\Distill-Qwen-1.5B} & \thmode & \thmode & 0.033 & 12.5 & 95.1 & 42.7 & 56.9 \\
\ml{OLMo-3-\\7B-Instruct} & \nth & \nth & 0.126 & 24.1 & 92.3 & 38.9 & 51.4 \\
\ml{Qwen3-4B-\\Instruct} & \nth & \nth & 0.175 & 18.9 & 93.5 & 21.8 & 36.8 \\
MiMo-7B-RL & \thmode & \thmode & 0.104 & 23.3 & 93.4 & 20.0 & 75.8 \\
\ml{OLMo-3-\\7B-Think} & \thmode & \thmode & 0.115 & 23.7 & 92.6 & 37.1 & 60.9 \\
\ml{Qwen3-4B-\\Thinking-2507} & \thmode & \thmode & 0.162 & 24.6 & 92.3 & 23.1 & 61.1 \\
Falcon-H1R-7B & \thmode & \thmode & 0.058 & 17.4 & 94.1 & 30.4 & 68.6 \\
\bottomrule
\end{tabular}
\end{table}

Under matched modes, comparable initial forward KL values can accompany both gains and degradation, while top-1 agreement remains high across these outcomes. The initial teacher--student difference therefore does not track downstream performance. This supports the conclusion in \autoref{subsec:signal_performance}: a stronger signal is not necessarily more useful.

\subsection{Where the Divergence Mass Falls in the Vocabulary}
\label{app:token_vocab}

The vocabulary analysis supports the mode transfer conclusion in \autoref{subsec:mode_results}: large cumulative divergence contributions often come from punctuation, delimiters, the frequent word \texttt{the}, LaTeX markers, and reasoning transition tokens such as ``Let,'' ``So,'' ``We,'' and ``I.''

\subsection{Teacher-Side Context Signal}
\label{app:token_teacher_context}

\begin{figure}[H]
\centering
\includegraphics[width=\textwidth]{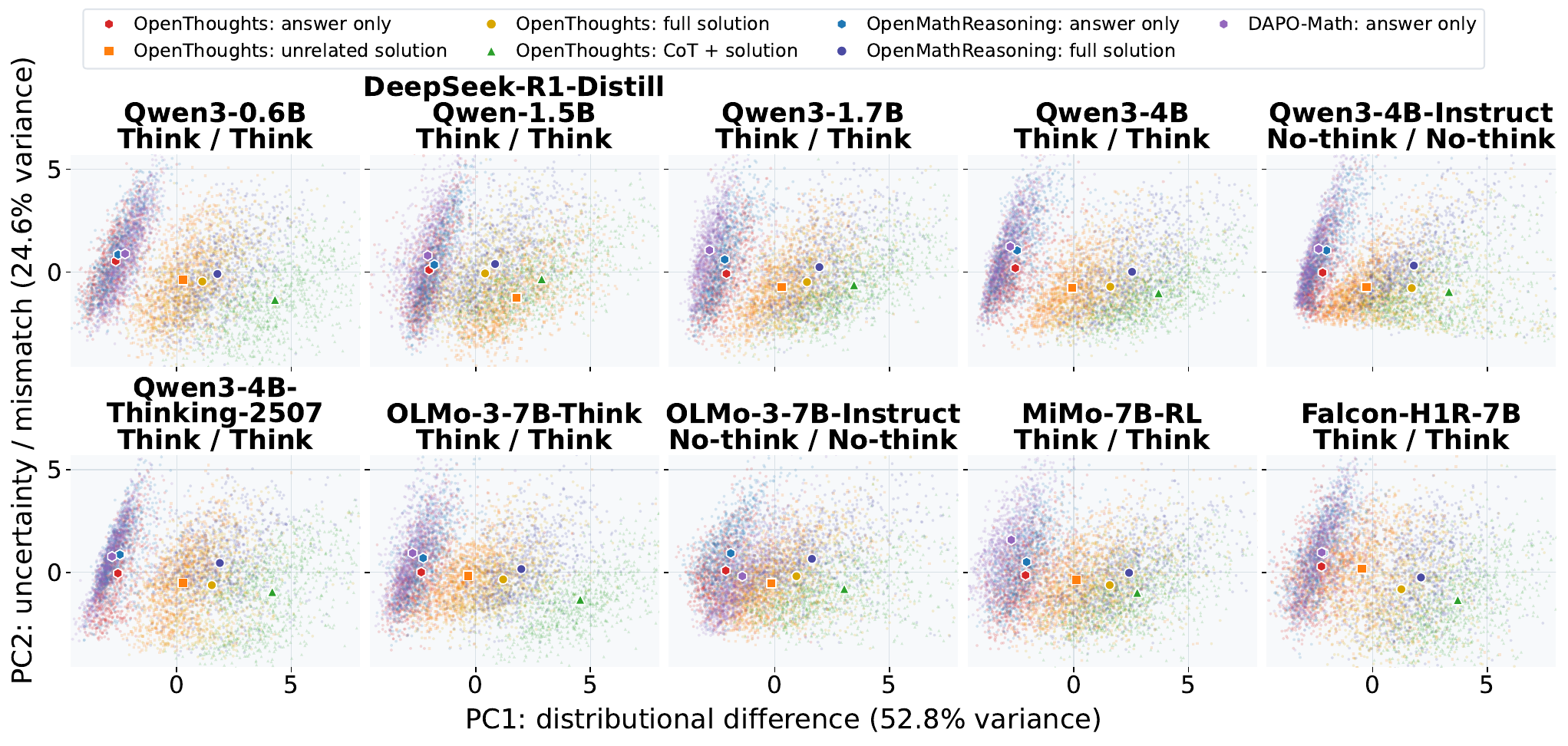}
\caption{Per-model teacher context PCA for the ten checkpoints underlying the aggregate view in \autoref{fig:teacher_context_signals}. Points are rollout-level signal summaries and large markers are centroids for the seven dataset--context conditions. The panels use a common PCA basis and shared axes. Features are standardized within each model--mode setting before the PCA.}
\label{fig:teacher_context_pca_by_model}
\end{figure}

\autoref{tab:token_context_signal_app} reports seven context conditions for ten checkpoints. \autoref{fig:teacher_context_pca_by_model} shows their per-model PCA views on a shared basis. The checkpoint-level results support the context finding in \autoref{subsec:teacher_context}: context type shapes the signal. Within OpenThoughts, CoT+Solution produces the largest signal for every checkpoint, while Full solution and Unrelated solution are closest for DeepSeek-R1-Distill-Qwen-1.5B.

\begin{table}[H]
\caption{Teacher-side context signal for all ten models and the macro average over ten models. Unqualified conditions use OpenThoughts. OMR and DAPO denote OpenMathReasoning and DAPO-Math-17k. CoT+Sol. denotes CoT+Solution.}
\label{tab:token_context_signal_app}
\centering
{\fontsize{5.8pt}{6.5pt}\selectfont
\setlength{\tabcolsep}{1.2pt}
\renewcommand{\arraystretch}{0.96}
\begin{tabular*}{\textwidth}{@{\extracolsep{\fill}}llcccclcccc@{}}
\toprule
Context & Model & Fwd. KL & $D>.05$ (\%) & Abs. adv. & Agree. (\%) & Model & Fwd. KL & $D>.05$ (\%) & Abs. adv. & Agree. (\%) \\
\midrule
Answer & \multirow{7}{*}{Qwen3-0.6B} & 0.010 & 1.8 & 0.036 & 97.9 & \multirow{7}{*}{Qwen3-4B} & 0.025 & 3.7 & 0.047 & 97.8 \\
Unrelated & & 0.046 & 17.5 & 0.112 & 93.8 & & 0.052 & 15.1 & 0.097 & 95.2 \\
Full & & 0.074 & 21.5 & 0.139 & 92.7 & & 0.135 & 22.0 & 0.176 & 92.8 \\
CoT+Sol. & & 0.134 & 36.9 & 0.214 & 89.1 & & 0.203 & 30.3 & 0.242 & 90.3 \\
OMR Answer & & 0.008 & 1.3 & 0.034 & 98.0 & & 0.023 & 3.0 & 0.045 & 97.9 \\
OMR Full & & 0.089 & 24.2 & 0.157 & 91.7 & & 0.167 & 25.3 & 0.208 & 91.5 \\
DAPO Answer & & 0.004 & 0.8 & 0.032 & 98.1 & & 0.010 & 1.8 & 0.035 & 98.2 \\
\specialrule{0.3pt}{0.6pt}{0.6pt}
Answer & \multirow{7}{*}{\shortstack[l]{DeepSeek-R1-\\Distill-Qwen-1.5B}} & 0.004 & 0.9 & 0.024 & 98.6 & \multirow{7}{*}{\shortstack[l]{Qwen3-4B-\\Instruct}} & 0.018 & 3.0 & 0.035 & 98.4 \\
Unrelated & & 0.037 & 11.5 & 0.083 & 95.4 & & 0.050 & 11.1 & 0.080 & 96.4 \\
Full & & 0.032 & 12.3 & 0.086 & 95.3 & & 0.155 & 18.1 & 0.175 & 93.8 \\
CoT+Sol. & & 0.063 & 25.3 & 0.142 & 92.5 & & 0.248 & 23.1 & 0.257 & 91.7 \\
OMR Answer & & 0.004 & 0.7 & 0.023 & 98.6 & & 0.015 & 2.9 & 0.036 & 98.2 \\
OMR Full & & 0.038 & 14.5 & 0.098 & 94.5 & & 0.140 & 21.2 & 0.183 & 92.9 \\
DAPO Answer & & 0.002 & 0.2 & 0.020 & 98.8 & & 0.006 & 1.4 & 0.025 & 98.7 \\
\specialrule{0.3pt}{0.6pt}{0.6pt}
Answer & \multirow{7}{*}{Qwen3-1.7B} & 0.019 & 3.4 & 0.042 & 98.0 & \multirow{7}{*}{\shortstack[l]{Qwen3-4B-\\Thinking-2507}} & 0.018 & 3.2 & 0.039 & 98.1 \\
Unrelated & & 0.059 & 15.4 & 0.107 & 95.0 & & 0.105 & 19.2 & 0.128 & 94.1 \\
Full & & 0.144 & 20.2 & 0.184 & 92.9 & & 0.159 & 24.1 & 0.181 & 92.4 \\
CoT+Sol. & & 0.225 & 27.7 & 0.280 & 90.1 & & 0.275 & 34.8 & 0.270 & 89.0 \\
OMR Answer & & 0.016 & 2.6 & 0.037 & 98.1 & & 0.018 & 2.8 & 0.038 & 98.1 \\
OMR Full & & 0.177 & 22.7 & 0.220 & 91.9 & & 0.176 & 25.3 & 0.208 & 91.5 \\
DAPO Answer & & 0.007 & 1.9 & 0.032 & 98.4 & & 0.007 & 1.3 & 0.026 & 98.6 \\
\specialrule{0.3pt}{0.6pt}{0.6pt}
Answer & \multirow{7}{*}{\shortstack[l]{OLMo-3-\\7B-Think}} & 0.023 & 2.5 & 0.038 & 98.1 & \multirow{7}{*}{MiMo-7B-RL} & 0.019 & 3.6 & 0.046 & 97.9 \\
Unrelated & & 0.055 & 17.5 & 0.107 & 94.6 & & 0.056 & 16.9 & 0.105 & 95.0 \\
Full & & 0.111 & 23.1 & 0.153 & 92.8 & & 0.103 & 23.0 & 0.147 & 93.5 \\
CoT+Sol. & & 0.206 & 38.8 & 0.234 & 89.6 & & 0.154 & 32.5 & 0.190 & 92.0 \\
OMR Answer & & 0.023 & 1.8 & 0.035 & 98.2 & & 0.018 & 2.9 & 0.044 & 97.9 \\
OMR Full & & 0.132 & 25.4 & 0.171 & 91.9 & & 0.122 & 26.0 & 0.171 & 92.4 \\
DAPO Answer & & 0.004 & 0.7 & 0.023 & 98.8 & & 0.009 & 1.6 & 0.036 & 98.2 \\
\specialrule{0.3pt}{0.6pt}{0.6pt}
Answer & \multirow{7}{*}{\shortstack[l]{OLMo-3-\\7B-Instruct}} & 0.015 & 3.8 & 0.043 & 97.7 & \multirow{7}{*}{\shortstack[l]{Falcon-\\H1R-7B}} & 0.012 & 2.7 & 0.038 & 97.9 \\
Unrelated & & 0.048 & 15.5 & 0.099 & 95.0 & & 0.034 & 9.8 & 0.082 & 95.6 \\
Full & & 0.112 & 22.7 & 0.162 & 92.8 & & 0.051 & 15.5 & 0.104 & 94.5 \\
CoT+Sol. & & 0.195 & 30.9 & 0.254 & 89.8 & & 0.086 & 27.4 & 0.158 & 92.0 \\
OMR Answer & & 0.012 & 3.3 & 0.044 & 97.6 & & 0.010 & 1.9 & 0.034 & 98.1 \\
OMR Full & & 0.126 & 27.0 & 0.191 & 91.3 & & 0.059 & 18.0 & 0.120 & 93.6 \\
DAPO Answer & & 0.011 & 3.3 & 0.042 & 97.7 & & 0.007 & 1.3 & 0.029 & 98.4 \\
\specialrule{0.3pt}{0.6pt}{0.6pt}
Answer & \multirow{7}{*}{\shortstack[l]{Macro\\average}} & 0.016 & 2.9 & 0.039 & 98.0 & & & & & \\
Unrelated & & 0.054 & 14.9 & 0.100 & 95.0 & & & & & \\
Full & & 0.108 & 20.3 & 0.151 & 93.3 & & & & & \\
CoT+Sol. & & 0.179 & 30.8 & 0.224 & 90.6 & & & & & \\
OMR Answer & & 0.015 & 2.3 & 0.037 & 98.1 & & & & & \\
OMR Full & & 0.122 & 23.0 & 0.173 & 92.3 & & & & & \\
DAPO Answer & & 0.007 & 1.4 & 0.030 & 98.4 & & & & & \\
\bottomrule
\end{tabular*}
}
\end{table}

\subsection{Long Full Solution Teacher Prefixes}
\label{app:token_sol_long}

The main teacher context analysis uses the prompt cap of 1024 tokens from training. The length control rescores the same rollouts with a 12288-token full solution teacher prefix. Answer only and Unrelated solution use caps of 1024 tokens, while token-level CoT+Solution uses a cap of 10240 tokens. All five prefixes share identical student completions sampled from the long solution pool, with the short condition truncating $C$ in $Q\oplus C$. This offline control differs from the training intervention in \autoref{subsec:teacher_context}, which adds a separate CoT field to matched OpenThoughts rows and uses a teacher cap of 8192 tokens.
Consistent with \autoref{subsec:teacher_context}, prefix length alone changes the initial signal only slightly. The Answer only $<$ Unrelated solution $<$ Full solution ordering remains unchanged (\autoref{tab:token_sol_long_app}).

\begin{table}[H]
\caption{Length control extension of the full solution signal in \autoref{tab:token_context_signal}: short vs.\ long full solution teacher prefixes on shared student rollouts. \textsc{Sol} uses a teacher prompt cap of 1024 tokens. \textsc{Sol-long} raises it to 12288.}
\label{tab:token_sol_long_app}
\centering
\setlength{\tabcolsep}{2.2pt}
\scriptsize
\renewcommand{\arraystretch}{0.96}
\begin{tabular*}{\textwidth}{@{\extracolsep{\fill}}ll *{8}{c}@{}}
\toprule
& & \multicolumn{4}{c}{\textsc{Sol}} & \multicolumn{4}{c}{\textsc{Sol-long}} \\
\cmidrule(lr){3-6}\cmidrule(lr){7-10}
Model & Mode & Fwd KL & $D>.05$ (\%) & Abs. adv. & Agree. (\%) & Fwd KL & $D>.05$ (\%) & Abs. adv. & Agree. (\%) \\
\midrule
Qwen3-1.7B & \thmode/\thmode & 0.144 & 20.2 & 0.184 & 92.9 & 0.147 & 20.4 & 0.188 & 92.9 \\
Qwen3-1.7B & \nth/\nth & 0.090 & 16.7 & 0.125 & 94.7 & 0.094 & 17.2 & 0.129 & 94.6 \\
Qwen3-4B & \thmode/\thmode & 0.135 & 22.0 & 0.176 & 92.8 & 0.137 & 22.3 & 0.178 & 92.7 \\
Qwen3-4B & \nth/\nth & 0.066 & 15.2 & 0.109 & 95.2 & 0.068 & 15.6 & 0.111 & 95.1 \\
Qwen3-4B-Instruct & \nth/\nth & 0.155 & 18.1 & 0.175 & 93.8 & 0.158 & 18.3 & 0.178 & 93.7 \\
OLMo-3-7B-Think & \thmode/\thmode & 0.110 & 23.0 & 0.153 & 92.8 & 0.114 & 23.6 & 0.156 & 92.7 \\
OLMo-3-7B-Instruct & \nth/\nth & 0.112 & 22.6 & 0.162 & 92.8 & 0.115 & 23.1 & 0.165 & 92.7 \\
\midrule
Macro average & -- & 0.116 & 19.7 & 0.155 & 93.6 & 0.119 & 20.1 & 0.158 & 93.5 \\
\bottomrule
\end{tabular*}
\end{table}

\subsection{Position Window Signal}
\label{app:token_length_windows}
\label{subsec:token_length_windows}

The position window results support supervision on early tokens in \autoref{subsec:response_position_signal}: forward KL decreases with response position in every checkpoint (\autoref{tab:token_length_windows_app}).
\begin{table}[H]
\caption{Forward KL by response position window (rollout cap 6144). Each cell reports mean/SNR, where $\mathrm{SNR}=\mu/\sigma$ and $\sigma$ is the token-level standard deviation. The macro row assigns equal weight to each checkpoint. These statistics include all responses reaching each window. \autoref{fig:position_intervention} uses a fixed cohort and response weights.}
\label{tab:token_length_windows_app}
\centering
\setlength{\tabcolsep}{3.0pt}
\scriptsize
\renewcommand{\arraystretch}{0.92}
\setlength{\tabcolsep}{1.5pt}
\begin{tabular*}{0.96\textwidth}{@{\extracolsep{\fill}}ll *{7}{c}@{}}
\toprule
Model & Mode &
$[0,128)$ & $[128,256)$ & $[256,512)$ & $[512,1\mathrm{k})$ & $[1\mathrm{k},2\mathrm{k})$ & $[2\mathrm{k},4\mathrm{k})$ & $[4\mathrm{k},6\mathrm{k})$ \\
\midrule
Qwen3-1.7B & \thmode/\thmode & 0.291/0.238 & 0.224/0.214 & 0.138/0.186 & 0.091/0.153 & 0.066/0.124 & 0.048/0.104 & 0.041/0.090 \\
Qwen3-4B & \thmode/\thmode & 0.249/0.233 & 0.197/0.230 & 0.132/0.203 & 0.094/0.173 & 0.073/0.143 & 0.059/0.124 & 0.048/0.113 \\
Qwen3-4B-Instruct & \nth/\nth & 0.339/0.224 & 0.230/0.202 & 0.157/0.166 & 0.100/0.143 & 0.064/0.125 & 0.046/0.114 & 0.046/0.104 \\
OLMo-3-7B-Think & \thmode/\thmode & 0.258/0.278 & 0.158/0.255 & 0.106/0.221 & 0.077/0.173 & 0.062/0.132 & 0.050/0.106 & 0.039/0.088 \\
OLMo-3-7B-Instruct & \nth/\nth & 0.299/0.312 & 0.139/0.254 & 0.095/0.218 & 0.071/0.196 & 0.071/0.182 & 0.077/0.159 & 0.068/0.120 \\
\midrule
Macro average & -- & 0.287/0.248 & 0.190/0.217 & 0.126/0.185 & 0.087/0.160 & 0.067/0.138 & 0.056/0.121 & 0.048/0.103 \\
\bottomrule
\end{tabular*}
\end{table}

We also compare early and late positions within the same long responses (\autoref{fig:position_intervention}). For each window, we average the measurements within each response and weight the resulting means by the number of valid tokens that response contributes to the final window. With these weights held fixed, the top-1 disagreement rate, mean absolute sampled-token log ratio, and mean forward KL remain lower at later positions across all five models (\autoref{tab:position_intervention_control}).

\begin{table}[H]
\centering
\caption{Early and late metrics. Each metric cell reports early $\to$ late for position windows $[0,128)$ and $[4096,6144)$, respectively. $N$ is the number of responses with more than 4096 valid tokens.}
\label{tab:position_intervention_control}
\scriptsize
\setlength{\tabcolsep}{4pt}
\begin{tabular}{@{}lrccc@{}}
\toprule
Model & $N$ & Top-1 disagreement (\%) & Mean $|a_i|$ & Mean forward KL \\
\midrule
Qwen3-1.7B & 2265 & $9.72 \to 3.71$ & $0.301 \to 0.074$ & $0.299 \to 0.041$ \\
Qwen3-4B & 2225 & $9.08 \to 4.33$ & $0.268 \to 0.081$ & $0.248 \to 0.048$ \\
Qwen3-4B-Instruct & 282 & $8.73 \to 3.64$ & $0.282 \to 0.076$ & $0.240 \to 0.046$ \\
OLMo-3-7B-Think & 2752 & $10.37 \to 3.66$ & $0.260 \to 0.070$ & $0.260 \to 0.039$ \\
OLMo-3-7B-Instruct & 300 & $12.92 \to 4.74$ & $0.346 \to 0.100$ & $0.279 \to 0.068$ \\
\bottomrule
\end{tabular}
\end{table}

\subsection{Initialization-Dependent Training Dynamics}
\enlargethispage{\baselineskip}

\begin{table}[H]
\caption{Mean results over early (steps 1--10) and late (steps 91--100) windows in the initialization block of \autoref{fig:grpo_opsd_dynamics}. Both runs use matched \thmode/\thmode{} and $\tau_{\mathrm{clip}}=0.05$. $H_S$ is student entropy and $\|\nabla\|$ is the pre-clipping gradient norm.}
\label{tab:sft_grpo_opsd_train}
\centering
\scriptsize
\setlength{\tabcolsep}{4.0pt}
\renewcommand{\arraystretch}{0.95}
\begin{tabular}{@{}llcccc@{}}
\toprule
Initialization & Steps & $H_S$ & $D_{\mathrm{KL}}(\pi^T\|\pi^S)$ & $\|\nabla\|$ & Loss \\
\midrule
\multirow{2}{*}{Qwen3-4B (\thmode/\thmode{})}
& 1--10 & 0.24 & 0.137 & 0.39 & $-0.009$ \\
& 91--100 & 0.22 & 0.329 & 0.18 & $-0.020$ \\
\midrule
\multirow{2}{*}{SFT+GRPO then OPSD}
& 1--10 & 0.41 & 0.047 & 0.35 & $-0.001$ \\
& 91--100 & 1.29 & 0.215 & 0.86 & $-0.029$ \\
\bottomrule
\end{tabular}
\end{table}
Forward KL increases for both initializations, whereas student entropy and gradient norm rise together for the collapsing SFT+GRPO actor and decrease for Qwen3-4B. This contrast supports \autoref{subsec:training_dynamics}: the joint entropy--gradient rise is a useful online warning of collapse, but no single trace reliably predicts final performance.
\clearpage
\section{Additional Experiments and Analyses}
\label{app:additional_experiments}
\label{app:alternative_signal_designs}


\subsection{Sensitivity to Training Data}
\label{app:training_dataset_sensitivity}

We repeat 100-step OPSD with a 1024-token rollout cap on GSM8K~\citep{gsm8k} and DeepMath~\citep{deepmath}, using Qwen3-1.7B, Qwen3-4B, and OLMo-3-7B-Think in matched \thmode{}/\thmode{} mode. Within each model, optimization and evaluation settings follow the default OpenThoughts configuration. The teacher receives the reference solution. For DeepMath, we use the post-thinking solution and retain examples with difficulty greater than 5. After applying the 1024-token student and teacher prompt caps, the training sets contain 7473 GSM8K examples and 70477 DeepMath examples, shared across the three models.

\begin{table}[H]
\caption{Training dataset comparison after 100 OPSD steps with a 1024-token rollout cap. Trend ($\uparrow$/$\downarrow$) indicates an increase/decrease in overall metrics relative to Base. Per-dataset results are in \autoref{tab:training_dataset_per_dataset}.}
\label{tab:training_dataset_macro}
\centering
\fontsize{7.0pt}{8.4pt}\selectfont
\setlength{\tabcolsep}{1.5pt}
\renewcommand{\arraystretch}{1.08}
\begin{tabular*}{\textwidth}{@{\extracolsep{\fill}}l*{15}{c}@{}}
\toprule
\multirow{2}{*}{Model} & \multicolumn{3}{c}{Base} & \multicolumn{4}{c}{OpenThoughts} & \multicolumn{4}{c}{GSM8K} & \multicolumn{4}{c}{DeepMath} \\
\cmidrule(lr){2-4}\cmidrule(lr){5-8}\cmidrule(lr){9-12}\cmidrule(lr){13-16}
& Avg@8 & Pass@8 & Length & Avg@8 & Pass@8 & Length & Trend & Avg@8 & Pass@8 & Length & Trend & Avg@8 & Pass@8 & Length & Trend \\
\midrule
Qwen3-1.7B & 34.0 & 59.2 & 18.6k & 35.3 & 60.8 & 22.4k & $\uparrow$ & 35.3 & 60.8 & 19.5k & $\uparrow$ & 34.4 & 60.0 & 22.9k & $\uparrow$ \\
Qwen3-4B & 62.3 & 79.2 & 17.8k & 57.9 & 75.8 & 19.9k & $\downarrow$ & 61.9 & 80.0 & 17.2k & $\downarrow$ & 58.9 & 80.0 & 18.9k & $\downarrow$ \\
OLMo-3-7B-Think & 63.5 & 81.7 & 18.7k & 58.8 & 77.5 & 21.7k & $\downarrow$ & 61.0 & 81.7 & 19.3k & $\downarrow$ & 57.8 & 79.2 & 22.4k & $\downarrow$ \\
\bottomrule
\end{tabular*}
\end{table}

Consistent with \autoref{subsec:initial_capability}, both alternative training datasets slightly increase Avg@8 for Qwen3-1.7B but reduce it for Qwen3-4B and OLMo-3-7B-Think (\autoref{tab:training_dataset_macro}). These results reproduce the model-dependent trend observed with OpenThoughts in the main text.

\subsection{Cross-Dataset Teacher Context Replication}
\label{app:context_omr_replication}

We repeat the solution-only versus CoT-plus-solution comparison on two disjoint OpenMathReasoning subsets using Qwen3-1.7B, with identical examples and row order within each pair. More detailed teacher context does not necessarily improve training, as in \autoref{subsec:teacher_context}: adding CoT lowers Avg@8 in both subsets despite the larger teacher prompt budget (\autoref{tab:context_omr_per_dataset}).

\subsection{Instruction-Only Teacher Prefixes}
\label{app:instruction_prefix}

We remove privileged teacher content and vary only the English student and teacher instructions. Concise-to-detailed (C$\to$D) asks the student for a concise solution and the teacher for a detailed one. D$\to$C reverses the instructions.

\begin{table}[H]
\caption{Instruction-only prefix controls after 100 OPSD steps. Per-dataset results are in \autoref{tab:instruction_prefix_per_dataset}.}
\label{tab:instruction_prefix_macro}
\centering
\scriptsize
\setlength{\tabcolsep}{2.6pt}
\renewcommand{\arraystretch}{0.96}
\begin{tabular*}{\textwidth}{@{\extracolsep{\fill}}l*{12}{c}@{}}
\toprule
\multirow{2}{*}{Model} & \multicolumn{3}{c}{Base} & \multicolumn{3}{c}{Full solution} & \multicolumn{3}{c}{No priv.: C$\to$D} & \multicolumn{3}{c}{No priv.: D$\to$C} \\
\cmidrule(lr){2-4}\cmidrule(lr){5-7}\cmidrule(lr){8-10}\cmidrule(lr){11-13}
& Avg@8 & Pass@8 & Length & Avg@8 & Pass@8 & Length & Avg@8 & Pass@8 & Length & Avg@8 & Pass@8 & Length \\
\midrule
Qwen3-1.7B & 34.0 & 59.2 & 18.6k & 35.3 & 60.8 & 22.4k & 23.1 & 46.7 & 12.6k & 36.3 & 62.5 & 25.0k \\
OLMo-3-7B-Think & 63.5 & 81.7 & 18.7k & 58.8 & 77.5 & 21.7k & 60.7 & 80.8 & 16.5k & 56.4 & 75.0 & 24.1k \\
\bottomrule
\end{tabular*}
\end{table}

Changing only the student and teacher instructions substantially alters accuracy and output length: C$\to$D shortens outputs, whereas D$\to$C lengthens them. These behavioral shifts occur without privileged content, supporting the conclusion in \autoref{subsec:signal_performance} that the teacher's signal reflects the full teacher prefix rather than privileged semantics alone.

\subsection{Additional Training Controls}
\label{app:additional_controls}

\subsubsection{Pointwise Contribution Clip Threshold Sweep}
\label{app:clip_tau}

We vary only the pointwise vocabulary contribution cap, retraining Qwen3-1.7B and OLMo-3-7B-Think with $\tau_{\mathrm{clip}}\in\{0.01,0.1,0.2\}$ while keeping all other settings fixed, and compare against Base and the default $\tau_{\mathrm{clip}}=0.05$ run. We additionally evaluate $\tau_{\mathrm{clip}}\in\{0.5,1.0\}$ on OLMo-3-7B-Think as extended-threshold controls (\autoref{tab:clip_tau_qwen} and \autoref{tab:clip_tau_olmo}).

\begin{table}[H]
\caption{Qwen3-1.7B vocabulary contribution clip threshold sweep after 100 OPSD steps. Bold marks the best Avg@8 / Pass@8. Per-dataset results are in \autoref{tab:clip_tau_per_dataset}.}
\label{tab:clip_tau_qwen}
\centering
\scriptsize
\setlength{\tabcolsep}{6.0pt}
\renewcommand{\arraystretch}{0.95}
\begin{tabular}{@{}lccc@{}}
\toprule
$\tau_{\mathrm{clip}}$ & Avg@8 & Pass@8 & Length \\
\midrule
Base   & 34.0          & 59.2          & 18.6k \\
$0.01$ & 31.4          & 56.7          & 26.2k \\
$0.05$ & \textbf{35.3} & \textbf{60.8} & 22.4k \\
$0.1$  & 32.9          & 60.0          & 16.6k \\
$0.2$  & 30.6          & 53.3          & 15.5k \\
\bottomrule
\end{tabular}
\end{table}

\begin{table}[H]
\caption{OLMo-3-7B-Think vocabulary contribution clip threshold sweep after 100 OPSD steps, including OLMo-only extended-threshold controls at $0.5$ and $1.0$. Bold marks the best Avg@8 / Pass@8. Per-dataset results are in \autoref{tab:clip_tau_per_dataset}.}
\label{tab:clip_tau_olmo}
\centering
\scriptsize
\setlength{\tabcolsep}{6.0pt}
\renewcommand{\arraystretch}{0.95}
\begin{tabular}{@{}lccc@{}}
\toprule
$\tau_{\mathrm{clip}}$ & Avg@8 & Pass@8 & Length \\
\midrule
Base   & \textbf{63.5} & 81.7          & 18.7k \\
$0.01$ & 58.2          & 79.2          & 20.6k \\
$0.05$ & 58.8          & 77.5          & 21.7k \\
$0.1$  & 63.1          & 80.8          & 19.4k \\
$0.2$  & 63.1          & 81.7          & 17.0k \\
$0.5$  & 62.5          & \textbf{82.5} & 17.0k \\
$1.0$  & 62.6          & 80.8          & 16.8k \\
\bottomrule
\end{tabular}
\end{table}

Consistent with \autoref{subsec:divergence_objectives}, changing the clipping threshold provides no reliable remedy. The default $0.05$ leads on Qwen3-1.7B, whereas Base has the highest Avg@8 and $0.5$ the highest Pass@8 on OLMo-3-7B-Think. The OLMo-only extended thresholds do not change the overall conclusion (\autoref{tab:clip_tau_qwen} and \autoref{tab:clip_tau_olmo}).

\subsubsection{Short SFT then OPSD}
\label{app:sft10_same}

To further test OPSD's sensitivity to model initialization, we apply 10 SFT steps before OPSD training. We then run 100-step same-prefix OPSD without privileged teacher context. The primary runs use OpenThoughts for both stages. An additional Qwen3-1.7B control uses OpenMathReasoning for SFT and OpenThoughts for OPSD.

\begin{table}[H]
\caption{Ten-step SFT followed by 100-step no-privilege OPSD. OT/OMR denote the SFT dataset. OPSD uses OpenThoughts throughout. Bold marks the best Avg@8 / Pass@8 within each model--dataset block. Per-dataset results are in \autoref{tab:sft10_same_per_dataset}.}
\label{tab:sft10_same_macro}
\centering
\scriptsize
\setlength{\tabcolsep}{3.8pt}
\renewcommand{\arraystretch}{0.95}
\begin{tabular}{@{}lccc@{}p{3.0em}@{}lccc@{}}
\toprule
\multicolumn{4}{c}{\textbf{Qwen3-1.7B (OT)}} & &
\multicolumn{4}{c}{\textbf{Qwen3-4B-Thinking (OT)}} \\
\cmidrule(lr){1-4}\cmidrule(lr){6-9}
Stage & Avg@8 & Pass@8 & Length & & Stage & Avg@8 & Pass@8 & Length \\
\midrule
Base       & \textbf{34.0} & \textbf{59.2} & 18.6k & & Base       & \textbf{75.4} & \textbf{86.7} & 22.8k \\
SFT-10     & 5.1           & 17.5          & 6.9k  & & SFT-10     & 4.9           & 20.8          & 2.0k \\
+ OPSD-100 & 5.2           & 16.7          & 6.6k  & & + OPSD-100 & 7.8           & 28.3          & 3.1k \\
\midrule
\multicolumn{4}{c}{\textbf{Qwen3-1.7B (OMR)}} & &
\multicolumn{4}{c}{\textbf{OLMo-3-7B-Think (OT)}} \\
\cmidrule(lr){1-4}\cmidrule(lr){6-9}
Stage & Avg@8 & Pass@8 & Length & & Stage & Avg@8 & Pass@8 & Length \\
\midrule
Base       & \textbf{34.0} & \textbf{59.2} & 18.6k & & Base       & \textbf{63.5} & \textbf{81.7} & 18.7k \\
SFT-10     & 28.0          & 54.2          & 16.9k & & SFT-10     & 60.1          & 80.0          & 18.4k \\
+ OPSD-100 & 27.1          & 49.2          & 16.9k & & + OPSD-100 & 60.3          & \textbf{81.7} & 18.3k \\
\bottomrule
\end{tabular}
\end{table}

Models that collapse after SFT remain collapsed after OPSD, while models with degradation do not recover their pre-SFT performance. These results show that OPSD is highly sensitive to the initial teacher and student checkpoints, limiting the regimes in which it can improve reasoning (\autoref{sec:pipeline}).

\subsection{LoRA Diagnostics and Qwen3.5 Collapse}
\label{app:qwen35_collapse}

We compare full-parameter and LoRA OPSD under matched modes across six model settings, then analyze Qwen3.5-4B collapse under matched modes (\autoref{tab:lora_across_models}, \autoref{tab:qwen35_lora}).

\begin{table}[H]
\caption{Full-parameter and LoRA OPSD under matched modes after 100 updates. The arrow columns summarize the overall Avg@8 and Pass@8 change relative to Base. Agree indicates whether Full OPSD and LoRA OPSD have the same direction. Bold marks the best Avg@8 and Pass@8 within each model.}
\label{tab:lora_across_models}
\centering
\scriptsize
\setlength{\tabcolsep}{3.0pt}
\renewcommand{\arraystretch}{0.96}
\begin{tabular*}{\textwidth}{@{\extracolsep{\fill}}l*{12}{c}@{}}
\toprule
\multirow{2}{*}{Model} & \multicolumn{3}{c}{Base} & \multicolumn{4}{c}{Full OPSD} & \multicolumn{4}{c}{LoRA OPSD} & \multirow{2}{*}{Agree} \\
\cmidrule(lr){2-4}\cmidrule(lr){5-8}\cmidrule(lr){9-12}
& Avg@8 & Pass@8 & Length & Avg@8 & Pass@8 & Length & Dir. & Avg@8 & Pass@8 & Length & Dir. & \\
\midrule
Qwen3-1.7B & 34.0 & 59.2 & 18.6k & 35.3 & 60.8 & 22.4k & $\uparrow$ & \textbf{35.8} & \textbf{61.7} & 23.3k & $\uparrow$ & Yes \\
Qwen3-4B & \textbf{62.3} & 79.2 & 17.8k & 57.9 & 75.8 & 19.9k & $\downarrow$ & 59.9 & \textbf{80.0} & 20.2k & $\downarrow$ & Yes \\
Qwen3-4B-Instruct & 48.5 & \textbf{71.7} & 9.1k & \textbf{50.0} & 70.0 & 10.6k & $\downarrow$ & 49.2 & 70.8 & 11.2k & $\downarrow$ & Yes \\
Qwen3-4B-Thinking & \textbf{75.4} & \textbf{86.7} & 22.8k & 65.8 & 83.3 & 43.4k & $\downarrow$ & 68.0 & 82.5 & 40.5k & $\downarrow$ & Yes \\
OLMo-3-7B-Think & \textbf{63.5} & \textbf{81.7} & 18.7k & 58.8 & 77.5 & 21.7k & $\downarrow$ & 62.5 & 77.5 & 21.3k & $\downarrow$ & Yes \\
OLMo-3-7B-Instruct & 39.5 & 60.8 & 7.6k & 40.8 & \textbf{62.5} & 7.4k & $\uparrow$ & \textbf{41.3} & \textbf{62.5} & 7.9k & $\uparrow$ & Yes \\
\bottomrule
\end{tabular*}
\end{table}

OPSD gains remain dependent among models, as in \autoref{subsec:initial_capability}. LoRA reduces drift and keeps most checkpoints closer to Base, but does not yield consistent gains.

\begin{table}[H]
\caption{Qwen3.5-4B full-parameter and LoRA results after 100 OPSD updates under matched modes.}
\label{tab:qwen35_lora}
\centering
\scriptsize
\setlength{\tabcolsep}{5.2pt}
\begin{tabular}{@{}llcccc@{}}
\toprule
Eval mode & Evaluated state & Avg@8 & Pass@8 & Format & Length \\
\midrule
\multirow{3}{*}{\nth}
& Base & 52.71 & 77.50 & 97.71 & 10.2k \\
& Full OPSD (\nth/\nth) & 8.02 & 35.00 & 14.90 & 70.7k \\
& LoRA OPSD (\nth/\nth) & 45.73 & 68.33 & 99.48 & 8.4k \\
\midrule
\multirow{3}{*}{\thmode}
& Base & 80.62 & 91.67 & 90.52 & 38.1k \\
& Full OPSD (\thmode/\thmode) & 0.00 & 0.00 & 5.42 & 81.9k \\
& LoRA OPSD (\thmode/\thmode) & 8.54 & 20.83 & 10.62 & 77.4k \\
\bottomrule
\end{tabular}
\end{table}

\subsubsection{The \nth{} Mode Is Not Strictly Isolated}
Reasoning mode remains a confound rather than a clean semantic control, as in \autoref{subsec:mode_results}. Specifically, Qwen3.5-4B sometimes emits an additional \verb|</think>| delimiter after the chat template closes the empty thinking block and produces longer outputs than Qwen3-4B. The delimiter persists after full-parameter \nth{}/\nth{} training, so the two modes are not operationally isolated.

\subsubsection{The Distinguishing Signal Is Dynamic, Not Initial}
Consistent with \autoref{subsec:signal_performance}, initial KL does not predict training stability. Matched modes start with lower forward KL than their corresponding mismatches, yet both later collapse. The subsequent entropy and forward KL rise separates Qwen3.5-4B from the bounded Qwen3 controls (\autoref{tab:qwen35_token_grid}, \autoref{fig:qwen35_dynamics}).

\subsubsection{LoRA Attenuates the Drift}
\begin{figure}[H]
\centering
\includegraphics[width=\linewidth]{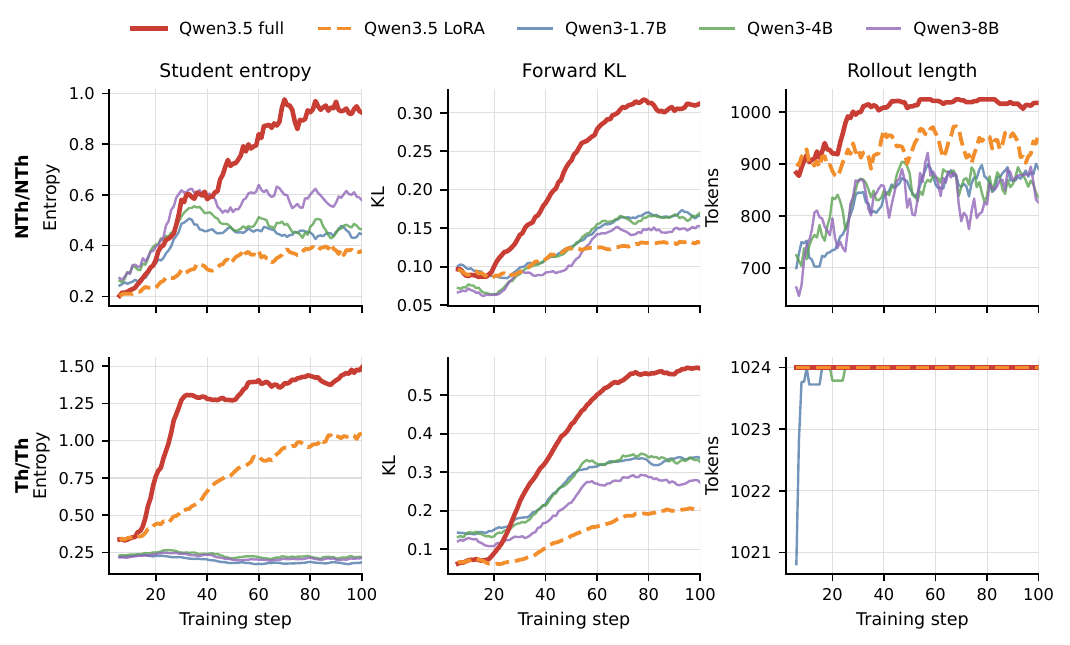}
\caption{Training dynamics under matched modes for full-parameter and LoRA Qwen3.5-4B versus three Qwen3 full-parameter controls. Curves are moving averages over five steps.}
\label{fig:qwen35_dynamics}
\end{figure}

Parameter restriction attenuates but does not remove the collapse behavior. The LoRA controls use rank 64 with scale 128 on \texttt{q\_proj}, \texttt{k\_proj}, \texttt{v\_proj}, \texttt{o\_proj}, \texttt{gate\_proj}, \texttt{up\_proj}, and \texttt{down\_proj}. All other parameters are frozen. LoRA removes length growth but leaves an accuracy drop for \nth{}/\nth{}, while \thmode{}/\thmode{} still collapses (\autoref{fig:qwen35_dynamics}).

\begin{table}[H]
\caption{Initial Qwen3.5-4B token-level signals. Definitions follow \autoref{tab:token_signal_summary}. Per-setting and per-dataset results for the downstream mode grid are in \autoref{tab:mode_per_dataset}.}
\label{tab:qwen35_token_grid}
\centering
\scriptsize
\setlength{\tabcolsep}{4.2pt}
\begin{tabular}{@{}ccccccc@{}}
\toprule
S & T & Fwd. KL & $D>.05$ & Agree. & Enc. & Disc. \\
\midrule
\nth & \nth & 0.087 & 17.7 & 94.7 & 34.1 & 63.4 \\
\nth & \thmode & 0.208 & 31.2 & 92.7 & 14.9 & 84.6 \\
\thmode & \nth & 0.140 & 27.2 & 92.0 & 56.0 & 43.4 \\
\thmode & \thmode & 0.066 & 16.5 & 95.0 & 40.2 & 58.7 \\
\bottomrule
\end{tabular}
\end{table}

\subsection{Replications of Recent OPSD Variants}
\label{app:recent_opsd_replications}

We reproduce two recent OPSD variants using their official implementations. We test whether vanilla OPSD's model and configuration sensitivity persists after changing its supervision target.

\subsubsection{Purified OPSD}
At response position $i$, let $\pi_i^0$ denote the frozen initial policy without privileged context. Let $\pi_i^T$ denote the privileged teacher. Let $\pi_i^R$ denote a reference conditioned only on the privileged field. All three distributions are evaluated on the same student prefix. Purified OPSD first computes
\begin{equation}
\begin{aligned}
\delta_i(v)
&=
\log \pi_i^T(v)-\log \pi_i^R(v), \\
\bar{\delta}_i
&=
\frac{1}{|\mathcal{V}|}\sum_{u\in\mathcal{V}}\delta_i(u), \\
\widetilde{\delta}_i(v)
&=
c\tanh\!\left(
\frac{\delta_i(v)-\bar{\delta}_i}{c}
\right).
\end{aligned}
\label{eq:purified_opsd_correction}
\end{equation}
The corrected target is
\begin{equation}
\pi_i^P(v)
\propto
\pi_i^0(v)
\exp\!\left(
\frac{\widetilde{\delta}_i(v)}{\beta_{\mathrm{PMI}}}
\right).
\label{eq:purified_opsd_target}
\end{equation}
The likelihood ratio removes effects induced by the privileged field alone. The initial policy provides the base distribution. The student is distilled toward $\pi_i^P$ rather than the raw privileged teacher \citep{purified_opsd}.

\subsubsection{\texorpdfstring{$\beta$-OPSD}{beta-OPSD}}
$\beta$-OPSD expresses self-distillation as a sampled-token policy update \citep{beta_opsd}. Let $\pi_i^R=\operatorname{sg}(\pi_i^S)$ be the stop-gradient reference. The method mixes this reference with the privileged teacher,
\begin{equation}
\pi_i^{M}
=
(1-w)\pi_i^R+w\pi_i^T,
\label{eq:beta_opsd_target}
\end{equation}
where $w$ is the teacher weight. For the sampled token $y_i^S$, the token-level signal and its discounted return-to-go are
\begin{equation}
a_i
=
\log \pi_i^{M}(y_i^S)
-
\log \pi_i^S(y_i^S),
\qquad
A_i
=
\sum_{j=i}^{N}\gamma^{j-i}a_j.
\label{eq:beta_opsd_rtg}
\end{equation}
The student objective is
\begin{equation}
\mathcal{L}_{\beta\text{-OPSD}}
=
-\frac{1}{N}\sum_{i=1}^{N}
\operatorname{sg}(A_i)
\log \pi_i^S(y_i^S).
\label{eq:beta_opsd_loss}
\end{equation}
The return-to-go allows later token signals to affect earlier updates.

\subsubsection{Replication Results}

All rows use the protocol with four benchmarks and eight samples.
The replications reproduce the model dependence reported in \autoref{subsec:initial_capability}. Each variant improves upon standard OPSD-1024 in some settings, but neither consistently improves over Base. These methods therefore only partially mitigate OPSD's failure modes. Behavioral collapse and ineffective length growth persist, while some gains over Base remain modest (\autoref{tab:recent_opsd_replications}).

\begin{table}[H]
\caption{Replication results under the original paper configurations, averaged over four benchmarks. Each $\Delta_{\mathrm{B/O}}$ entry gives the change relative to Base / OPSD-1024 in the same native matched reasoning mode. Avg@8 and Pass@8 changes are absolute percentage points, Length changes are relative percentages. Per-dataset results are in \autoref{tab:purified_opsd_per_dataset} and \autoref{tab:beta_opsd_per_dataset}.}
\label{tab:recent_opsd_replications}
\centering
\scriptsize
\setlength{\tabcolsep}{2.2pt}
\renewcommand{\arraystretch}{1.05}
\providecommand{\ropa}[1]{\makebox[2.6em][c]{#1}}
\providecommand{\ropd}[1]{\makebox[5.8em][c]{#1}}
\providecommand{\ropl}[1]{\makebox[3.0em][c]{#1}}
\providecommand{\ropp}[1]{\makebox[6.4em][c]{#1}}
\begin{tabular}{@{}lll cc cc cc@{}}
\toprule
Variant & Model & Privileged field & Avg@8 & $\Delta_{\mathrm{B/O}}$ & Pass@8 & $\Delta_{\mathrm{B/O}}$ & Length & $\Delta_{\mathrm{B/O}}$ \\
\midrule
\multirow{8}{*}[-2pt]{Purified OPSD}
& Qwen3-1.7B & Answer   & \ropa{39.9} & \ropd{$+5.9/+4.6$} & \ropa{63.3} & \ropd{$+4.1/+2.5$} & \ropl{24.2k} & \ropp{$+29.7/+7.9$} \\
& Qwen3-1.7B & Solution & \ropa{37.5} & \ropd{$+3.5/+2.2$} & \ropa{65.0} & \ropd{$+5.8/+4.2$} & \ropl{20.8k} & \ropp{$+11.5/-7.2$} \\
& Qwen3-4B & Answer   & \ropa{63.3} & \ropd{$+1.0/+5.4$} & \ropa{79.2} & \ropd{$+0.0/+3.3$} & \ropl{19.6k} & \ropp{$+10.1/-2.0$} \\
& Qwen3-4B & Solution & \ropa{61.3} & \ropd{$-1.0/+3.4$} & \ropa{77.5} & \ropd{$-1.7/+1.7$} & \ropl{18.5k} & \ropp{$+4.3/-7.1$} \\
& \ml{Qwen3-4B-\\Thinking} & Answer   & \ropa{75.1} & \ropd{$-0.3/+9.3$} & \ropa{84.2} & \ropd{$-2.5/+0.8$} & \ropl{24.8k} & \ropp{$+8.8/-42.9$} \\
& \ml{Qwen3-4B-\\Thinking} & Solution & \ropa{75.3} & \ropd{$-0.1/+9.5$} & \ropa{86.7} & \ropd{$+0.0/+3.3$} & \ropl{24.2k} & \ropp{$+6.2/-44.2$} \\
& \ml{OLMo-3-\\7B-Think} & Answer   & \ropa{64.5} & \ropd{$+1.0/+5.7$} & \ropa{80.8} & \ropd{$-0.9/+3.3$} & \ropl{19.2k} & \ropp{$+2.4/-11.7$} \\
& \ml{OLMo-3-\\7B-Think} & Solution & \ropa{63.8} & \ropd{$+0.3/+5.0$} & \ropa{80.0} & \ropd{$-1.7/+2.5$} & \ropl{18.9k} & \ropp{$+0.6/-13.2$} \\
\midrule
\multirow{5}{*}[-2pt]{$\beta$-OPSD}
& Qwen3-1.7B & Solution & \ropa{25.1} & \ropd{$-8.9/-10.2$} & \ropa{49.2} & \ropd{$-10.0/-11.7$} & \ropl{13.4k} & \ropp{$-28.0/-40.0$} \\
& \ml{Qwen3-4B-\\Thinking} & Solution & \ropa{68.6} & \ropd{$-6.8/+2.8$} & \ropa{82.5} & \ropd{$-4.2/-0.8$} & \ropl{18.2k} & \ropp{$-20.3/-58.1$} \\
& \ml{Qwen3-4B-\\Instruct} & Solution & \ropa{47.2} & \ropd{$-1.3/-2.8$} & \ropa{65.8} & \ropd{$-5.9/-4.2$} & \ropl{11.0k} & \ropp{$+20.3/+4.5$} \\
& \ml{OLMo-3-\\7B-Think} & Solution & \ropa{64.1} & \ropd{$+0.6/+5.3$} & \ropa{81.7} & \ropd{$+0.0/+4.2$} & \ropl{18.8k} & \ropp{$+0.2/-13.5$} \\
& \ml{OLMo-3-\\7B-Instruct} & Solution & \ropa{42.4} & \ropd{$+2.9/+1.6$} & \ropa{65.0} & \ropd{$+4.2/+2.5$} & \ropl{8.3k} & \ropp{$+9.1/+11.9$} \\
\bottomrule
\end{tabular}
\end{table}

\clearpage
\section{Robustness Checks}
\label{app:robustness_checks}

\subsection{Seed Sensitivity}
\label{app:seed_sensitivity}

For matched Qwen3-1.7B and OLMo-3-7B-Think, \autoref{tab:seed_scope} lists the configurations repeated over training or evaluation seeds~$42$, $1024$, and~$65536$. Training seed tables contain independent training runs evaluated with seed~42. Evaluation seed tables reuse the seed-42-trained checkpoints and vary the decoding seed.

\begin{table}[H]
\caption{Seed repetitions for Qwen3-1.7B and OLMo-3-7B-Think. ``Training seeds'' denotes independent training runs. ``Evalualtion seeds'' denotes decoding seeds applied to the seed-42-trained checkpoint.}
\label{tab:seed_scope}
\centering
\small
\setlength{\tabcolsep}{5pt}
\renewcommand{\arraystretch}{1.05}
\begin{tabular}{@{}llll@{}}
\toprule
Experiment family & Conditions & Training seeds & Evaluation seeds \\
\midrule
Base checkpoint & Untrained Base & -- & 42 / 1024 / 65536 \\
Rollout length & 256 / 1024 & 42 / 1024 / 65536 & 42 / 1024 / 65536 \\
Loss token position & First / Uniform / Final-256 & 42 / 1024 / 65536 & 42 / 1024 / 65536 \\
Teacher context & Solution / Answer / Unrelated & 42 / 1024 / 65536 & 42 / 1024 / 65536 \\
\bottomrule
\end{tabular}
\end{table}

\begin{table}[H]
\caption{Base checkpoints under evaluation seeds $42$/$1024$/$65536$. Per-dataset results are in \autoref{tab:seed_base_eval_per_dataset}.}
\label{tab:seed_base_eval}
\centering
{\fontsize{8pt}{9.6pt}\selectfont
\renewcommand{\arraystretch}{1.08}
\setlength{\tabcolsep}{2pt}
\begin{tabular}{@{}l ccc ccc ccc@{}}
\toprule
Model & \multicolumn{9}{c}{Evaluation seed} \\
\cmidrule(lr){2-10}
& \multicolumn{3}{c}{$42$} & \multicolumn{3}{c}{$1024$} & \multicolumn{3}{c}{$65536$} \\
\cmidrule(lr){2-4}\cmidrule(lr){5-7}\cmidrule(lr){8-10}
& Avg@8 & Pass@8 & Length & Avg@8 & Pass@8 & Length & Avg@8 & Pass@8 & Length \\
\midrule
\ml{Qwen3-\\1.7B} & 33.96 & 59.17 & 18.6k & 35.10 & 55.00 & 18.6k & 35.31 & 61.67 & 18.6k \\
\midrule
\ml{OLMo-3-\\7B-Think} & 63.54 & 81.67 & 18.7k & 62.19 & 80.83 & 19.0k & 63.12 & 79.17 & 18.9k \\
\bottomrule
\end{tabular}
}
\end{table}
\subsubsection{Maximum Student Rollout Length}
\label{app:seed_length}

The seed repetitions support the rollout length finding in \autoref{subsec:length_results}: OPSD-256 has higher Avg@8 than OPSD-1024 across all training and evaluation seeds (\autoref{tab:seed_length_train}, \autoref{tab:seed_length_eval}).

\subsubsection{Loss Token Position Selection}
\label{app:seed_position}

The seed repetitions also support the position finding in \autoref{subsec:response_position_signal}: with a fixed budget of 256 positions, First-256 has the highest mean Avg@8 on Qwen3-1.7B, while variants selecting later positions remain below Base on OLMo-3-7B-Think (\autoref{tab:seed_position_train}, \autoref{tab:seed_position_eval}).

\subsubsection{Teacher-Side Context (Prefix) Interventions}
\label{app:seed_context}

The seed repetitions support the context finding in \autoref{subsec:teacher_context}: Answer only does not consistently lead, and Unrelated solution remains comparable to Full solution (\autoref{tab:seed_context_train}, \autoref{tab:seed_context_eval}).

\begin{table}[H]
\caption{Rollout length results across training seeds $42$/$1024$/$65536$, evaluated with seed~$42$. OPSD-256/1024 use full solution context and matched \thmode/\thmode{} modes (\autoref{subsec:length_results}). Throughout the seed sensitivity tables, including the per-dataset breakdowns, $^{\uparrow}$/$^{\downarrow}$ indicate higher/lower Avg@8 or Pass@8 than Base. Training seed comparisons use Base evaluated with seed~42. Evaluation seed comparisons use Base at the matched evaluation seed. Ties and Length receive no arrows. Per-dataset results are in \autoref{tab:seed_length_train_per_dataset}.}
\label{tab:seed_length_train}
\centering
{\fontsize{8pt}{9.6pt}\selectfont
\renewcommand{\arraystretch}{1.08}
\setlength{\tabcolsep}{2pt}
\begin{tabular}{@{}ll ccc ccc ccc@{}}
\toprule
Model & Condition & \multicolumn{9}{c}{Training seed} \\
\cmidrule(lr){3-11}
& & \multicolumn{3}{c}{$42$} & \multicolumn{3}{c}{$1024$} & \multicolumn{3}{c}{$65536$} \\
\cmidrule(lr){3-5}\cmidrule(lr){6-8}\cmidrule(lr){9-11}
& & Avg@8 & Pass@8 & Length & Avg@8 & Pass@8 & Length & Avg@8 & Pass@8 & Length \\
\midrule
\ml{Qwen3-\\1.7B} & OPSD-256 & 38.12$^{\uparrow}$ & 62.50$^{\uparrow}$ & 21.4k & 39.17$^{\uparrow}$ & 65.00$^{\uparrow}$ & 21.1k & 37.08$^{\uparrow}$ & 65.00$^{\uparrow}$ & 21.3k \\
 & OPSD-1024 & 35.31$^{\uparrow}$ & 60.83$^{\uparrow}$ & 22.4k & 35.94$^{\uparrow}$ & 56.67$^{\downarrow}$ & 22.0k & 34.58$^{\uparrow}$ & 62.50$^{\uparrow}$ & 22.5k \\
\midrule
\ml{OLMo-3-\\7B-Think} & OPSD-256 & 60.83$^{\downarrow}$ & 80.00$^{\downarrow}$ & 21.4k & 61.46$^{\downarrow}$ & 79.17$^{\downarrow}$ & 21.5k & 61.98$^{\downarrow}$ & 81.67 & 21.5k \\
 & OPSD-1024 & 58.75$^{\downarrow}$ & 77.50$^{\downarrow}$ & 21.7k & 59.17$^{\downarrow}$ & 78.33$^{\downarrow}$ & 21.8k & 59.69$^{\downarrow}$ & 81.67 & 21.4k \\
\bottomrule
\end{tabular}
}
\end{table}

\begin{table}[H]
\caption{Rollout length results across evaluation seeds $42$/$1024$/$65536$ using the seed-42-trained checkpoints. Per-dataset results are in \autoref{tab:seed_length_eval_per_dataset}.}
\label{tab:seed_length_eval}
\centering
{\fontsize{8pt}{9.6pt}\selectfont
\renewcommand{\arraystretch}{1.08}
\setlength{\tabcolsep}{2pt}
\begin{tabular}{@{}ll ccc ccc ccc@{}}
\toprule
Model & Condition & \multicolumn{9}{c}{Evaluation seed} \\
\cmidrule(lr){3-11}
& & \multicolumn{3}{c}{$42$} & \multicolumn{3}{c}{$1024$} & \multicolumn{3}{c}{$65536$} \\
\cmidrule(lr){3-5}\cmidrule(lr){6-8}\cmidrule(lr){9-11}
& & Avg@8 & Pass@8 & Length & Avg@8 & Pass@8 & Length & Avg@8 & Pass@8 & Length \\
\midrule
\ml{Qwen3-\\1.7B} & OPSD-256 & 38.12$^{\uparrow}$ & 62.50$^{\uparrow}$ & 21.4k & 36.98$^{\uparrow}$ & 59.17$^{\uparrow}$ & 21.0k & 36.67$^{\uparrow}$ & 62.50$^{\uparrow}$ & 21.3k \\
 & OPSD-1024 & 35.31$^{\uparrow}$ & 60.83$^{\uparrow}$ & 22.4k & 36.77$^{\uparrow}$ & 65.00$^{\uparrow}$ & 21.9k & 36.56$^{\uparrow}$ & 61.67 & 22.2k \\
\midrule
\ml{OLMo-3-\\7B-Think} & OPSD-256 & 60.83$^{\downarrow}$ & 80.00$^{\downarrow}$ & 21.4k & 61.98$^{\downarrow}$ & 80.83 & 21.7k & 61.35$^{\downarrow}$ & 77.50$^{\downarrow}$ & 21.6k \\
 & OPSD-1024 & 58.75$^{\downarrow}$ & 77.50$^{\downarrow}$ & 21.7k & 58.85$^{\downarrow}$ & 78.33$^{\downarrow}$ & 21.5k & 60.31$^{\downarrow}$ & 79.17 & 21.6k \\
\bottomrule
\end{tabular}
}
\end{table}

\begin{table}[H]
\caption{Loss token position results across training seeds. First-256 reuses the corresponding OPSD-256 run, while Uni-256 and Final-256 use 1024-token rollouts and apply the loss to at most 256 uniformly sampled or final valid response positions. Per-dataset results are in \autoref{tab:seed_position_train_per_dataset}.}
\label{tab:seed_position_train}
\centering
{\fontsize{8pt}{9.6pt}\selectfont
\renewcommand{\arraystretch}{1.08}
\setlength{\tabcolsep}{2pt}
\begin{tabular}{@{}ll ccc ccc ccc@{}}
\toprule
Model & Condition & \multicolumn{9}{c}{Training seed} \\
\cmidrule(lr){3-11}
& & \multicolumn{3}{c}{$42$} & \multicolumn{3}{c}{$1024$} & \multicolumn{3}{c}{$65536$} \\
\cmidrule(lr){3-5}\cmidrule(lr){6-8}\cmidrule(lr){9-11}
& & Avg@8 & Pass@8 & Length & Avg@8 & Pass@8 & Length & Avg@8 & Pass@8 & Length \\
\midrule
\ml{Qwen3-\\1.7B} & First-256 & 38.12$^{\uparrow}$ & 62.50$^{\uparrow}$ & 21.4k & 39.17$^{\uparrow}$ & 65.00$^{\uparrow}$ & 21.1k & 37.08$^{\uparrow}$ & 65.00$^{\uparrow}$ & 21.3k \\
 & Uni-256 & 35.21$^{\uparrow}$ & 61.67$^{\uparrow}$ & 22.6k & 34.90$^{\uparrow}$ & 59.17 & 22.4k & 36.56$^{\uparrow}$ & 62.50$^{\uparrow}$ & 22.8k \\
 & Final-256 & 34.27$^{\uparrow}$ & 59.17 & 23.7k & 33.23$^{\downarrow}$ & 55.00$^{\downarrow}$ & 23.8k & 34.17$^{\uparrow}$ & 58.33$^{\downarrow}$ & 23.7k \\
\midrule
\ml{OLMo-3-\\7B-Think} & First-256 & 60.83$^{\downarrow}$ & 80.00$^{\downarrow}$ & 21.4k & 61.46$^{\downarrow}$ & 79.17$^{\downarrow}$ & 21.5k & 61.98$^{\downarrow}$ & 81.67 & 21.5k \\
 & Uni-256 & 59.06$^{\downarrow}$ & 82.50$^{\uparrow}$ & 21.5k & 58.54$^{\downarrow}$ & 75.00$^{\downarrow}$ & 21.6k & 60.00$^{\downarrow}$ & 80.83$^{\downarrow}$ & 21.1k \\
 & Final-256 & 61.25$^{\downarrow}$ & 79.17$^{\downarrow}$ & 21.1k & 57.39$^{\downarrow}$ & 78.33$^{\downarrow}$ & 21.7k & 57.29$^{\downarrow}$ & 77.50$^{\downarrow}$ & 21.8k \\
\bottomrule
\end{tabular}
}
\end{table}

\begin{table}[H]
\caption{Loss token position results across evaluation seeds using the seed-42-trained checkpoints. First-256 reuses OPSD-256. Per-dataset results are in \autoref{tab:seed_position_eval_per_dataset}.}
\label{tab:seed_position_eval}
\centering
{\fontsize{8pt}{9.6pt}\selectfont
\renewcommand{\arraystretch}{1.08}
\setlength{\tabcolsep}{2pt}
\begin{tabular}{@{}ll ccc ccc ccc@{}}
\toprule
Model & Condition & \multicolumn{9}{c}{Evaluation seed} \\
\cmidrule(lr){3-11}
& & \multicolumn{3}{c}{$42$} & \multicolumn{3}{c}{$1024$} & \multicolumn{3}{c}{$65536$} \\
\cmidrule(lr){3-5}\cmidrule(lr){6-8}\cmidrule(lr){9-11}
& & Avg@8 & Pass@8 & Length & Avg@8 & Pass@8 & Length & Avg@8 & Pass@8 & Length \\
\midrule
\ml{Qwen3-\\1.7B} & First-256 & 38.12$^{\uparrow}$ & 62.50$^{\uparrow}$ & 21.4k & 36.98$^{\uparrow}$ & 59.17$^{\uparrow}$ & 21.0k & 36.67$^{\uparrow}$ & 62.50$^{\uparrow}$ & 21.3k \\
 & Uni-256 & 35.21$^{\uparrow}$ & 61.67$^{\uparrow}$ & 22.6k & 35.52$^{\uparrow}$ & 59.17$^{\uparrow}$ & 22.3k & 34.27$^{\downarrow}$ & 58.33$^{\downarrow}$ & 22.8k \\
 & Final-256 & 34.27$^{\uparrow}$ & 59.17 & 23.7k & 35.00$^{\downarrow}$ & 61.67$^{\uparrow}$ & 23.5k & 33.85$^{\downarrow}$ & 58.33$^{\downarrow}$ & 23.3k \\
\midrule
\ml{OLMo-3-\\7B-Think} & First-256 & 60.83$^{\downarrow}$ & 80.00$^{\downarrow}$ & 21.4k & 61.98$^{\downarrow}$ & 80.83 & 21.7k & 61.35$^{\downarrow}$ & 77.50$^{\downarrow}$ & 21.6k \\
 & Uni-256 & 59.06$^{\downarrow}$ & 82.50$^{\uparrow}$ & 21.5k & 59.69$^{\downarrow}$ & 78.33$^{\downarrow}$ & 21.2k & 58.54$^{\downarrow}$ & 79.17 & 21.5k \\
 & Final-256 & 61.25$^{\downarrow}$ & 79.17$^{\downarrow}$ & 21.1k & 59.27$^{\downarrow}$ & 79.17$^{\downarrow}$ & 21.1k & 58.23$^{\downarrow}$ & 80.00$^{\uparrow}$ & 21.3k \\
\bottomrule
\end{tabular}
}
\end{table}

\begin{table}[H]
\caption{Teacher context results across training seeds at a rollout cap of 1024 tokens (Full solution / Answer only / Unrelated solution, \autoref{subsec:teacher_context}). Per-dataset results are in \autoref{tab:seed_context_train_per_dataset}.}
\label{tab:seed_context_train}
\centering
{\fontsize{8pt}{9.6pt}\selectfont
\renewcommand{\arraystretch}{1.08}
\setlength{\tabcolsep}{2pt}
\begin{tabular}{@{}ll ccc ccc ccc@{}}
\toprule
Model & Condition & \multicolumn{9}{c}{Training seed} \\
\cmidrule(lr){3-11}
& & \multicolumn{3}{c}{$42$} & \multicolumn{3}{c}{$1024$} & \multicolumn{3}{c}{$65536$} \\
\cmidrule(lr){3-5}\cmidrule(lr){6-8}\cmidrule(lr){9-11}
& & Avg@8 & Pass@8 & Length & Avg@8 & Pass@8 & Length & Avg@8 & Pass@8 & Length \\
\midrule
\ml{Qwen3-\\1.7B} & Solution & 35.31$^{\uparrow}$ & 60.83$^{\uparrow}$ & 22.4k & 35.94$^{\uparrow}$ & 56.67$^{\downarrow}$ & 22.0k & 34.58$^{\uparrow}$ & 62.50$^{\uparrow}$ & 22.5k \\
 & Answer & 35.62$^{\uparrow}$ & 59.17 & 25.8k & 36.67$^{\uparrow}$ & 64.17$^{\uparrow}$ & 25.6k & 34.69$^{\uparrow}$ & 59.17 & 25.5k \\
 & Unrelated & 37.08$^{\uparrow}$ & 60.00$^{\uparrow}$ & 22.0k & 38.96$^{\uparrow}$ & 62.50$^{\uparrow}$ & 21.9k & 37.29$^{\uparrow}$ & 62.50$^{\uparrow}$ & 21.6k \\
\midrule
\ml{OLMo-3-\\7B-Think} & Solution & 58.75$^{\downarrow}$ & 77.50$^{\downarrow}$ & 21.7k & 59.17$^{\downarrow}$ & 78.33$^{\downarrow}$ & 21.8k & 59.69$^{\downarrow}$ & 81.67 & 21.4k \\
 & Answer & 56.15$^{\downarrow}$ & 75.83$^{\downarrow}$ & 24.1k & 56.88$^{\downarrow}$ & 74.17$^{\downarrow}$ & 24.4k & 56.98$^{\downarrow}$ & 74.17$^{\downarrow}$ & 23.9k \\
 & Unrelated & 57.50$^{\downarrow}$ & 76.67$^{\downarrow}$ & 23.4k & 56.46$^{\downarrow}$ & 75.00$^{\downarrow}$ & 23.4k & 56.46$^{\downarrow}$ & 76.67$^{\downarrow}$ & 23.8k \\
\bottomrule
\end{tabular}
}
\end{table}

\begin{table}[H]
\caption{Teacher context results across evaluation seeds using the seed-42-trained checkpoints and a rollout cap of 1024 tokens. Per-dataset results are in \autoref{tab:seed_context_eval_per_dataset}.}
\label{tab:seed_context_eval}
\centering
{\fontsize{8pt}{9.6pt}\selectfont
\renewcommand{\arraystretch}{1.08}
\setlength{\tabcolsep}{2pt}
\begin{tabular}{@{}ll ccc ccc ccc@{}}
\toprule
Model & Condition & \multicolumn{9}{c}{Evaluation seed} \\
\cmidrule(lr){3-11}
& & \multicolumn{3}{c}{$42$} & \multicolumn{3}{c}{$1024$} & \multicolumn{3}{c}{$65536$} \\
\cmidrule(lr){3-5}\cmidrule(lr){6-8}\cmidrule(lr){9-11}
& & Avg@8 & Pass@8 & Length & Avg@8 & Pass@8 & Length & Avg@8 & Pass@8 & Length \\
\midrule
\ml{Qwen3-\\1.7B} & Solution & 35.31$^{\uparrow}$ & 60.83$^{\uparrow}$ & 22.4k & 36.77$^{\uparrow}$ & 65.00$^{\uparrow}$ & 21.9k & 36.56$^{\uparrow}$ & 61.67 & 22.2k \\
 & Answer & 35.62$^{\uparrow}$ & 59.17 & 25.8k & 35.83$^{\uparrow}$ & 55.83$^{\uparrow}$ & 25.8k & 35.94$^{\uparrow}$ & 60.00$^{\downarrow}$ & 25.9k \\
 & Unrelated & 37.08$^{\uparrow}$ & 60.00$^{\uparrow}$ & 22.0k & 35.31$^{\uparrow}$ & 61.67$^{\uparrow}$ & 22.2k & 36.67$^{\uparrow}$ & 59.17$^{\downarrow}$ & 22.0k \\
\midrule
\ml{OLMo-3-\\7B-Think} & Solution & 58.75$^{\downarrow}$ & 77.50$^{\downarrow}$ & 21.7k & 58.85$^{\downarrow}$ & 78.33$^{\downarrow}$ & 21.5k & 60.31$^{\downarrow}$ & 79.17 & 21.6k \\
 & Answer & 56.15$^{\downarrow}$ & 75.83$^{\downarrow}$ & 24.1k & 58.85$^{\downarrow}$ & 79.17$^{\downarrow}$ & 23.9k & 58.12$^{\downarrow}$ & 80.00$^{\uparrow}$ & 24.1k \\
 & Unrelated & 57.50$^{\downarrow}$ & 76.67$^{\downarrow}$ & 23.4k & 58.02$^{\downarrow}$ & 80.00$^{\downarrow}$ & 23.4k & 58.65$^{\downarrow}$ & 75.00$^{\downarrow}$ & 23.3k \\
\bottomrule
\end{tabular}
}
\end{table}
\subsection{Temperature Robustness Checks}
\label{app:temperature_checks}

We vary the evaluation decoding temperature, student rollout sampling temperature, and teacher scoring temperature while holding all other settings fixed.

\subsubsection{Decoding Temperature}
The decoding temperature check leaves the dependency conclusion in \autoref{subsec:initial_capability} unchanged: small Instruct model gains can reverse, while Qwen3-4B-Thinking degrades at both temperatures with substantial length growth (\autoref{tab:temperature_eval}).

\begin{table}[H]
\caption{Decoding temperature robustness check, grouped by identical decode temperature within each model. $\Delta$ is OPSD minus Base. Each OPSD row uses the checkpoint trained under the default configuration. Per-dataset results are in \autoref{tab:temperature_eval_per_dataset}.}
\label{tab:temperature_eval}
\centering
\scriptsize
\setlength{\tabcolsep}{2.4pt}
\renewcommand{\arraystretch}{0.92}
\begin{tabular*}{\textwidth}{@{\extracolsep{\fill}}l c ccc ccc ccc@{}}
\toprule
Model & Decode $\tau$ & \multicolumn{3}{c}{Base} & \multicolumn{3}{c}{100-step OPSD} & \multicolumn{3}{c}{$\Delta$ (OPSD -- Base)} \\
\cmidrule(lr){3-5}\cmidrule(lr){6-8}\cmidrule(lr){9-11}
& & Avg@8 & Pass@8 & Length & Avg@8 & Pass@8 & Length & Avg@8 & Pass@8 & Length \\
\midrule
\multirow{2}{*}{Qwen3-1.7B}
& 0.6 & 33.96 & 59.17 & 18.6k & 35.31 & 60.83 & 22.4k & $+1.35$ & $+1.66$ & $+3.8$k \\
& 1.0 & 38.12 & 62.50 & 18.0k & 36.67 & 65.83 & 21.4k & $-1.45$ & $+3.33$ & $+3.4$k \\
\midrule
\multirow{2}{*}{Qwen3-4B-Instruct}
& 0.7 & 48.54 & 71.67 & 9.2k & 50.00 & 70.00 & 10.6k & $+1.46$ & $-1.67$ & $+1.4$k \\
& 1.0 & 49.17 & 70.83 & 9.2k & 48.75 & 66.67 & 10.7k & $-0.42$ & $-4.16$ & $+1.5$k \\
\midrule
\multirow{2}{*}{Qwen3-4B-Thinking}
& 0.6 & 75.42 & 86.67 & 22.8k & 65.83 & 83.33 & 43.4k & $-9.59$ & $-3.34$ & $+20.6$k \\
& 1.0 & 76.04 & 86.67 & 23.1k & 66.77 & 82.50 & 43.7k & $-9.27$ & $-4.17$ & $+20.6$k \\
\midrule
\multirow{2}{*}{OLMo-3-7B-Instruct}
& 0.6 & 39.48 & 60.83 & 7.6k & 40.83 & 62.50 & 7.4k & $+1.35$ & $+1.67$ & $-0.2$k \\
& 1.0 & 41.88 & 68.33 & 7.6k & 38.02 & 61.67 & 7.2k & $-3.86$ & $-6.66$ & $-0.4$k \\
\bottomrule
\end{tabular*}
\end{table}

\subsubsection{Training Sampling and Scoring Temperatures}
\begin{table}[H]
\caption{Training temperature check for Qwen3-1.7B OPSD. $\tau_S$ is the student rollout sampling temperature and $\tau_T$ is the teacher scoring temperature. All other settings are fixed. Bold marks the best Avg@8 and Pass@8. Per-dataset results are in \autoref{tab:temperature_train_per_dataset}.}
\label{tab:temperature_train}
\centering
\small
\setlength{\tabcolsep}{4.0pt}
\renewcommand{\arraystretch}{0.92}
\begin{tabular}{@{}cccccc@{}}
\toprule
$\tau_S$ & $\tau_T$ & Eval.\ $\tau$ & Avg@8 & Pass@8 & Length \\
\midrule
1.1 & 1.1 & 0.6 & 35.31 & \textbf{60.83} & 22.4k \\
0.6 & 0.6 & 0.6 & \textbf{35.42} & \textbf{60.83} & 20.0k \\
0.6 & 1.1 & 0.6 & 35.21 & 56.67 & 22.3k \\
1.1 & 0.6 & 0.6 & 32.29 & 57.50 & 18.3k \\
\bottomrule
\end{tabular}
\end{table}
Training temperature changes likewise provide no consistent remedy: lowering both temperatures preserves accuracy while shortening outputs, whereas asymmetric settings do not improve accuracy consistently (\autoref{tab:temperature_train}).

\clearpage
\section{Per-Dataset Evaluation Results}
\label{app:per_dataset_results}

The tables are grouped into main text, seed sensitivity, and additional experiment results. Dataset averages use unrounded benchmark-level metrics, so averages of displayed values can differ by 0.1.


\subsection{Main Text Experiment Breakdowns}

\subsubsection{Student--Teacher Reasoning Mode Alignment}
\label{app:mode_per_dataset}

S/T denote student/teacher modes (\thmode{}/\nth{}), and Eval denotes the inference mode. Each model lists its \thmode{} and \nth{} Base rows once because the Base checkpoint does not depend on teacher mode. \textsc{No-priv.} keeps the mismatched \nth{}/\thmode{} modes and problem context fixed but removes the privileged solution.

\begin{table}[H]
\caption{Per-dataset reasoning mode results for \autoref{tab:qwen_mode} and \autoref{tab:qwen_mode_extended}, including the no-privilege control. Bold marks the best Dataset Avg. for Avg@8 and Pass@8 within each model. Dataset Avg. denotes the aggregate result.}
\label{tab:mode_per_dataset}
\centering
{\fontsize{6.3pt}{7.0pt}\selectfont
\setlength{\tabcolsep}{0.5pt}
\renewcommand{\arraystretch}{1.04}
\begin{tabular*}{\textwidth}{@{\extracolsep{\fill}}lllll*{15}{c}@{}}
\toprule
\multirow{2}{*}{Model} & \multirow{2}{*}{S} & \multirow{2}{*}{T} & \multirow{2}{*}{Eval} & \multirow{2}{*}{Checkpoint}
& \multicolumn{3}{c}{AIME 2024} & \multicolumn{3}{c}{AIME 2025} & \multicolumn{3}{c}{AIME 2026} & \multicolumn{3}{c}{HMMT25} & \multicolumn{3}{c}{Dataset Avg.} \\
\cmidrule(lr){6-8}\cmidrule(lr){9-11}\cmidrule(lr){12-14}\cmidrule(lr){15-17}\cmidrule(lr){18-20}
& & & & & Avg@8 & Pass@8 & Length & Avg@8 & Pass@8 & Length & Avg@8 & Pass@8 & Length & Avg@8 & Pass@8 & Length & Avg@8 & Pass@8 & Length \\
\midrule
\multirow{9}{*}{Qwen3-1.7B} & \thmode & -- & \thmode & Base & 43.8 & 76.7 & 18647 & 35.4 & 60.0 & 18266 & 35.0 & 53.3 & 18161 & 21.7 & 46.7 & 19421 & 34.0 & 59.2 & 18624 \\
 & \thmode & \thmode & \thmode & OPSD & 45.8 & 73.3 & 22175 & 36.7 & 66.7 & 22143 & 36.7 & 63.3 & 21288 & 22.1 & 40.0 & 23913 & 35.3 & 60.8 & 22380 \\
 & \thmode & \nth & \thmode & OPSD & 28.8 & 53.3 & 12469 & 19.2 & 40.0 & 9756 & 17.5 & 36.7 & 11343 & 11.3 & 26.7 & 10867 & 19.2 & 39.2 & 11109 \\
 & \nth & -- & \nth & Base & 14.2 & 30.0 & 4958 & 10.4 & 23.3 & 2808 & 8.8 & 20.0 & 4977 & 5.8 & 10.0 & 2948 & 9.8 & 20.8 & 3923 \\
 & \nth & \thmode & \nth & OPSD & 20.0 & 36.7 & 10349 & 15.0 & 33.3 & 8806 & 11.7 & 23.3 & 11657 & 9.6 & 20.0 & 8112 & 14.1 & 28.3 & 9731 \\
 & \nth & \nth & \nth & OPSD & 10.8 & 23.3 & 10015 & 10.8 & 30.0 & 7387 & 9.6 & 20.0 & 10503 & 5.4 & 10.0 & 5946 & 9.2 & 20.8 & 8463 \\
 & \nth & \thmode & \thmode & OPSD & 50.4 & 76.7 & 19576 & 42.1 & 60.0 & 19946 & 42.5 & 66.7 & 19793 & 28.3 & 50.0 & 20668 & 40.8 & \textbf{63.3} & 19996 \\
 & \nth & \thmode & \thmode & \textsc{No-priv.} & 53.3 & 70.0 & 19748 & 42.1 & 63.3 & 20385 & 42.9 & 70.0 & 20423 & 25.8 & 46.7 & 21387 & \textbf{41.0} & 62.5 & 20486 \\
 & \thmode & \nth & \nth & OPSD & 8.8 & 23.3 & 3512 & 11.3 & 26.7 & 2458 & 6.3 & 13.3 & 4045 & 3.8 & 16.7 & 1999 & 7.5 & 20.0 & 3003 \\
\midrule
\multirow{9}{*}{Qwen3-4B} & \thmode & -- & \thmode & Base & 74.6 & 90.0 & 15495 & 65.4 & 80.0 & 18696 & 65.8 & 80.0 & 17249 & 43.3 & 66.7 & 19636 & 62.3 & \textbf{79.2} & 17769 \\
 & \thmode & \thmode & \thmode & OPSD & 66.3 & 80.0 & 18395 & 62.9 & 80.0 & 19655 & 62.9 & 83.3 & 18119 & 39.6 & 60.0 & 23625 & 57.9 & 75.8 & 19949 \\
 & \thmode & \nth & \thmode & OPSD & 57.5 & 83.3 & 14327 & 46.3 & 70.0 & 15505 & 52.5 & 73.3 & 14392 & 35.4 & 50.0 & 16211 & 47.9 & 69.2 & 15108 \\
 & \nth & -- & \nth & Base & 25.0 & 50.0 & 8239 & 21.7 & 36.7 & 5727 & 19.6 & 43.3 & 7309 & 12.1 & 23.3 & 5500 & 19.6 & 38.3 & 6694 \\
 & \nth & \thmode & \nth & OPSD & 26.2 & 53.3 & 17508 & 25.8 & 53.3 & 15433 & 20.4 & 50.0 & 16489 & 13.3 & 26.7 & 19388 & 21.5 & 45.8 & 17204 \\
 & \nth & \nth & \nth & OPSD & 22.9 & 40.0 & 15764 & 23.3 & 36.7 & 15308 & 14.6 & 30.0 & 15289 & 7.9 & 23.3 & 17266 & 17.2 & 32.5 & 15907 \\
 & \nth & \thmode & \thmode & OPSD & 72.5 & 86.7 & 16943 & 68.3 & 80.0 & 20399 & 67.1 & 86.7 & 18510 & 44.2 & 63.3 & 21324 & \textbf{63.0} & \textbf{79.2} & 19294 \\
 & \nth & \thmode & \thmode & \textsc{No-priv.} & 72.9 & 83.3 & 17976 & 65.0 & 76.7 & 19795 & 65.4 & 83.3 & 19041 & 40.4 & 56.7 & 22048 & 60.9 & 75.0 & 19715 \\
 & \thmode & \nth & \nth & OPSD & 21.3 & 40.0 & 7058 & 19.6 & 30.0 & 4629 & 14.6 & 30.0 & 6524 & 8.8 & 20.0 & 3940 & 16.0 & 30.0 & 5538 \\
\midrule
\multirow{9}{*}{Qwen3-8B} & \thmode & -- & \thmode & Base & 76.3 & 86.7 & 14860 & 69.6 & 80.0 & 18059 & 67.1 & 86.7 & 16744 & 43.3 & 60.0 & 20924 & 64.1 & 78.3 & 17647 \\
 & \thmode & \thmode & \thmode & OPSD & 67.5 & 76.7 & 19649 & 62.5 & 83.3 & 21001 & 65.0 & 86.7 & 19844 & 40.4 & 60.0 & 25193 & 58.9 & 76.7 & 21422 \\
 & \thmode & \nth & \thmode & OPSD & 55.8 & 83.3 & 12931 & 45.0 & 70.0 & 14801 & 49.2 & 80.0 & 13775 & 21.3 & 46.7 & 13113 & 42.8 & 70.0 & 13655 \\
 & \nth & -- & \nth & Base & 24.2 & 43.3 & 7391 & 19.2 & 40.0 & 4697 & 16.2 & 40.0 & 6901 & 12.5 & 26.7 & 4667 & 18.0 & 37.5 & 5914 \\
 & \nth & \thmode & \nth & OPSD & 28.8 & 56.7 & 15015 & 27.9 & 43.3 & 14525 & 23.3 & 53.3 & 14755 & 12.9 & 26.7 & 17077 & 23.2 & 45.0 & 15343 \\
 & \nth & \nth & \nth & OPSD & 25.0 & 43.3 & 17311 & 23.8 & 36.7 & 14711 & 18.3 & 33.3 & 17127 & 9.2 & 16.7 & 16985 & 19.1 & 32.5 & 16533 \\
 & \nth & \thmode & \thmode & OPSD & 76.3 & 80.0 & 16687 & 67.1 & 80.0 & 19952 & 70.0 & 83.3 & 18159 & 45.8 & 63.3 & 22370 & \textbf{64.8} & 76.7 & 19292 \\
 & \nth & \thmode & \thmode & \textsc{No-priv.} & 71.3 & 86.7 & 17480 & 72.1 & 83.3 & 20107 & 70.8 & 90.0 & 18597 & 42.9 & 63.3 & 22612 & 64.3 & \textbf{80.8} & 19699 \\
 & \thmode & \nth & \nth & OPSD & 25.0 & 43.3 & 8132 & 23.8 & 36.7 & 5237 & 15.8 & 33.3 & 8612 & 9.6 & 20.0 & 5103 & 18.5 & 33.3 & 6771 \\
\midrule
\multirow{8}{*}{Qwen3.5-4B} & \thmode & -- & \thmode & Base & 87.5 & 96.7 & 33160 & 80.8 & 90.0 & 38872 & 84.6 & 93.3 & 35106 & 69.6 & 86.7 & 45246 & \textbf{80.6} & \textbf{91.7} & 38096 \\
 & \thmode & \thmode & \thmode & OPSD & 0.0 & 0.0 & 81920 & 0.0 & 0.0 & 81920 & 0.0 & 0.0 & 81920 & 0.0 & 0.0 & 81920 & 0.0 & 0.0 & 81920 \\
 & \thmode & \nth & \thmode & OPSD & 10.8 & 33.3 & 71442 & 8.8 & 30.0 & 73966 & 6.7 & 20.0 & 75485 & 8.3 & 26.7 & 71964 & 8.6 & 27.5 & 73214 \\
 & \nth & -- & \nth & Base & 67.1 & 93.3 & 8634 & 47.5 & 66.7 & 9523 & 57.1 & 80.0 & 9623 & 39.2 & 70.0 & 12991 & 52.7 & 77.5 & 10193 \\
 & \nth & \thmode & \nth & OPSD & 0.0 & 0.0 & 81920 & 0.0 & 0.0 & 81920 & 0.0 & 0.0 & 81920 & 0.0 & 0.0 & 81920 & 0.0 & 0.0 & 81920 \\
 & \nth & \nth & \nth & OPSD & 9.2 & 43.3 & 71749 & 7.5 & 30.0 & 70294 & 6.7 & 40.0 & 69655 & 8.8 & 26.7 & 71144 & 8.0 & 35.0 & 70710 \\
 & \nth & \thmode & \thmode & OPSD & 0.0 & 0.0 & 81920 & 0.0 & 0.0 & 81920 & 0.0 & 0.0 & 81920 & 0.0 & 0.0 & 81920 & 0.0 & 0.0 & 81920 \\
 & \thmode & \nth & \nth & OPSD & 45.8 & 73.3 & 12692 & 38.8 & 70.0 & 12564 & 42.1 & 70.0 & 13516 & 18.3 & 43.3 & 15825 & 36.3 & 64.2 & 13649 \\
\bottomrule
\end{tabular*}
}
\end{table}

\subsubsection{Maximum Student Rollout Length}
\label{app:length_per_dataset}

Dataset Avg. gives the solid curve points and dashed Base levels in \autoref{fig:length_sweeps}. \thmode/\thmode{} and \nth/\nth{} denote matched student--teacher modes. The \nth/\nth{} Qwen3 length sweeps are a diagnostic exception to the native \thmode/\thmode{} setting used for subsequent hybrid model conclusions: they test whether the length trend also appears when the same hybrid checkpoints are operated in \nth{} mode. The 1024~/~Uni-256 and 1024~/~Final-256 rows keep the rollout cap at 1024 tokens and apply the loss to at most 256 positions. Uniform samples without replacement, whereas Final selects the last valid positions. Both include every valid position in shorter completions. The sweep curves exclude these position controls. \autoref{tab:position_selection_summary} averages the Qwen3-1.7B and OLMo-3-7B-Think results.

\begin{table}[H]
\caption{Per-dataset rollout length results for the matched \nth/\nth Qwen3 sweeps in \autoref{fig:length_sweeps}.}
\label{tab:length_per_dataset_nothink}
\centering
{\fontsize{6.3pt}{7.0pt}\selectfont
\setlength{\tabcolsep}{0.5pt}
\renewcommand{\arraystretch}{1.04}
\begin{tabular*}{\textwidth}{@{\extracolsep{\fill}}ll*{15}{c}@{}}
\toprule
\multirow{2}{*}{Model} & \multirow{2}{*}{Training rollout}
& \multicolumn{3}{c}{AIME 2024} & \multicolumn{3}{c}{AIME 2025} & \multicolumn{3}{c}{AIME 2026} & \multicolumn{3}{c}{HMMT25} & \multicolumn{3}{c}{Dataset Avg.} \\
\cmidrule(lr){3-5}\cmidrule(lr){6-8}\cmidrule(lr){9-11}\cmidrule(lr){12-14}\cmidrule(lr){15-17}
& & Avg@8 & Pass@8 & Length & Avg@8 & Pass@8 & Length & Avg@8 & Pass@8 & Length & Avg@8 & Pass@8 & Length & Avg@8 & Pass@8 & Length \\
\midrule
\multirow{8}{*}{\shortstack{Qwen3-1.7B\\\nth/\nth}} & Base & 14.2 & 30.0 & 4958 & 10.4 & 23.3 & 2808 & 8.8 & 20.0 & 4977 & 5.8 & 10.0 & 2948 & 9.8 & 20.8 & 3923 \\
 & 128 & 12.1 & 26.7 & 5258 & 10.0 & 23.3 & 3116 & 6.7 & 16.7 & 5283 & 4.2 & 13.3 & 3335 & 8.2 & 20.0 & 4248 \\
 & 256 & 14.2 & 33.3 & 6162 & 12.9 & 30.0 & 5492 & 9.6 & 20.0 & 7519 & 5.0 & 16.7 & 3751 & 10.4 & 25.0 & 5731 \\
 & 512 & 13.3 & 36.7 & 9160 & 10.4 & 30.0 & 8869 & 13.3 & 26.7 & 9046 & 4.2 & 13.3 & 5936 & 10.3 & 26.7 & 8253 \\
 & 1024 & 10.8 & 23.3 & 10015 & 10.8 & 30.0 & 7387 & 9.6 & 20.0 & 10503 & 5.4 & 10.0 & 5946 & 9.2 & 20.8 & 8463 \\
 & 2048 & 15.4 & 33.3 & 7255 & 11.7 & 23.3 & 6224 & 11.2 & 23.3 & 8313 & 5.4 & 20.0 & 4573 & 10.9 & 25.0 & 6591 \\
 & 4096 & 13.8 & 30.0 & 9799 & 12.9 & 36.7 & 6252 & 6.2 & 20.0 & 10046 & 7.1 & 16.7 & 5354 & 10.0 & 25.8 & 7863 \\
 & 6144 & 12.1 & 26.7 & 11950 & 14.2 & 33.3 & 7612 & 11.7 & 16.7 & 9960 & 4.6 & 13.3 & 5514 & 10.6 & 22.5 & 8759 \\
\midrule
\multirow{8}{*}{\shortstack{Qwen3-4B\\\nth/\nth}} & Base & 25.0 & 50.0 & 8239 & 21.7 & 36.7 & 5727 & 19.6 & 43.3 & 7309 & 12.1 & 23.3 & 5500 & 19.6 & 38.3 & 6694 \\
 & 128 & 20.0 & 40.0 & 11529 & 22.1 & 40.0 & 8779 & 15.0 & 23.3 & 10465 & 7.9 & 16.7 & 9715 & 16.2 & 30.0 & 10122 \\
 & 256 & 19.6 & 36.7 & 16505 & 17.5 & 36.7 & 15102 & 14.6 & 33.3 & 16479 & 8.8 & 20.0 & 17498 & 15.1 & 31.7 & 16396 \\
 & 512 & 20.8 & 36.7 & 16896 & 19.2 & 33.3 & 16487 & 13.3 & 30.0 & 17239 & 7.1 & 16.7 & 21014 & 15.1 & 29.2 & 17909 \\
 & 1024 & 22.9 & 40.0 & 15764 & 23.3 & 36.7 & 15308 & 14.6 & 30.0 & 15289 & 7.9 & 23.3 & 17266 & 17.2 & 32.5 & 15907 \\
 & 2048 & 18.8 & 40.0 & 17382 & 20.0 & 33.3 & 15974 & 17.1 & 36.7 & 16931 & 6.7 & 16.7 & 20779 & 15.6 & 31.7 & 17767 \\
 & 4096 & 21.2 & 36.7 & 17786 & 20.4 & 36.7 & 16162 & 14.2 & 36.7 & 18063 & 6.7 & 16.7 & 20520 & 15.6 & 31.7 & 18133 \\
 & 6144 & 18.3 & 33.3 & 15444 & 23.8 & 40.0 & 15505 & 15.4 & 36.7 & 15878 & 9.6 & 20.0 & 13964 & 16.8 & 32.5 & 15198 \\
\midrule
\multirow{8}{*}{\shortstack{Qwen3-8B\\\nth/\nth}} & Base & 24.2 & 43.3 & 7391 & 19.2 & 40.0 & 4697 & 16.2 & 40.0 & 6901 & 12.5 & 26.7 & 4667 & 18.0 & 37.5 & 5914 \\
 & 128 & 21.2 & 36.7 & 13958 & 17.9 & 33.3 & 11013 & 12.1 & 26.7 & 14225 & 7.5 & 20.0 & 10226 & 14.7 & 29.2 & 12356 \\
 & 256 & 22.1 & 33.3 & 16930 & 18.3 & 30.0 & 13522 & 14.2 & 30.0 & 17638 & 7.9 & 20.0 & 15279 & 15.6 & 28.3 & 15842 \\
 & 512 & 25.4 & 36.7 & 15883 & 25.8 & 40.0 & 14532 & 18.8 & 46.7 & 16116 & 10.0 & 20.0 & 17036 & 20.0 & 35.8 & 15891 \\
 & 1024 & 25.0 & 43.3 & 17311 & 23.8 & 36.7 & 14710 & 18.3 & 33.3 & 17127 & 9.2 & 16.7 & 16985 & 19.1 & 32.5 & 16533 \\
 & 2048 & 25.4 & 53.3 & 17870 & 21.7 & 33.3 & 14804 & 18.3 & 36.7 & 16582 & 8.3 & 20.0 & 18155 & 18.4 & 35.8 & 16853 \\
 & 4096 & 24.6 & 46.7 & 17374 & 21.7 & 36.7 & 14721 & 15.8 & 36.7 & 16801 & 9.6 & 23.3 & 20046 & 17.9 & 35.8 & 17235 \\
 & 6144 & 25.4 & 50.0 & 16549 & 20.4 & 43.3 & 14972 & 18.3 & 33.3 & 17073 & 8.3 & 16.7 & 18578 & 18.1 & 35.8 & 16793 \\
\bottomrule
\end{tabular*}
}
\end{table}

\begin{table}[H]
\caption{Per-dataset rollout length results for the remaining settings in \autoref{fig:length_sweeps} and 256-position controls for Qwen3-1.7B and OLMo-3-7B-Think. First-256 reuses the 256-token rollout row, Uni-256 and Final-256 use a 1024-token rollout.}
\label{tab:length_per_dataset}
\label{tab:position_selection_per_dataset}
\centering
{\fontsize{6.3pt}{7.0pt}\selectfont
\setlength{\tabcolsep}{0.5pt}
\renewcommand{\arraystretch}{1.04}
\begin{tabular*}{\textwidth}{@{\extracolsep{\fill}}ll*{15}{c}@{}}
\toprule
\multirow{2}{*}{Model} & \multirow{2}{*}{Training rollout}
& \multicolumn{3}{c}{AIME 2024} & \multicolumn{3}{c}{AIME 2025} & \multicolumn{3}{c}{AIME 2026} & \multicolumn{3}{c}{HMMT25} & \multicolumn{3}{c}{Dataset Avg.} \\
\cmidrule(lr){3-5}\cmidrule(lr){6-8}\cmidrule(lr){9-11}\cmidrule(lr){12-14}\cmidrule(lr){15-17}
& & Avg@8 & Pass@8 & Length & Avg@8 & Pass@8 & Length & Avg@8 & Pass@8 & Length & Avg@8 & Pass@8 & Length & Avg@8 & Pass@8 & Length \\
\midrule
\multirow{10}{*}{\shortstack{Qwen3-1.7B\\\thmode/\thmode}} & Base & 43.8 & 76.7 & 18647 & 35.4 & 60.0 & 18266 & 35.0 & 53.3 & 18161 & 21.7 & 46.7 & 19421 & 34.0 & 59.2 & 18624 \\
 & 128 & 47.5 & 70.0 & 19086 & 37.1 & 56.7 & 20269 & 31.7 & 60.0 & 20086 & 21.7 & 46.7 & 21820 & 34.5 & 58.3 & 20315 \\
 & 256 & 46.3 & 70.0 & 20783 & 36.7 & 66.7 & 21697 & 45.0 & 73.3 & 20995 & 24.6 & 40.0 & 22306 & 38.1 & 62.5 & 21445 \\
 & 512 & 50.0 & 76.7 & 20560 & 31.7 & 56.7 & 21037 & 38.3 & 63.3 & 20711 & 22.9 & 50.0 & 22648 & 35.7 & 61.7 & 21239 \\
 & 1024 & 45.8 & 73.3 & 22175 & 36.7 & 66.7 & 22143 & 36.7 & 63.3 & 21288 & 22.1 & 40.0 & 23913 & 35.3 & 60.8 & 22380 \\
 & 1024 / Uni-256 & 48.8 & 76.7 & 21875 & 31.7 & 60.0 & 21873 & 36.7 & 63.3 & 22032 & 23.8 & 46.7 & 24462 & 35.2 & 61.7 & 22560 \\
 & 1024 / Final-256 & 42.5 & 73.3 & 23510 & 34.6 & 53.3 & 22599 & 37.1 & 63.3 & 22987 & 22.9 & 46.7 & 25800 & 34.3 & 59.2 & 23724 \\
 & 2048 & 42.5 & 73.3 & 22598 & 32.5 & 63.3 & 22135 & 36.3 & 63.3 & 21822 & 19.6 & 40.0 & 25466 & 32.7 & 60.0 & 23005 \\
 & 4096 & 41.3 & 70.0 & 23134 & 32.5 & 53.3 & 22858 & 32.9 & 56.7 & 23500 & 21.3 & 36.7 & 26796 & 32.0 & 54.2 & 24072 \\
 & 6144 & 36.7 & 63.3 & 23970 & 32.1 & 56.7 & 22655 & 33.8 & 53.3 & 22585 & 20.8 & 36.7 & 27047 & 30.8 & 52.5 & 24064 \\
\midrule
\multirow{8}{*}{\shortstack{Qwen3-4B\\\thmode/\thmode}} & Base & 74.6 & 90.0 & 15495 & 65.4 & 80.0 & 18696 & 65.8 & 80.0 & 17249 & 43.3 & 66.7 & 19636 & 62.3 & 79.2 & 17769 \\
 & 128 & 70.8 & 86.7 & 16054 & 67.5 & 76.7 & 18469 & 67.9 & 83.3 & 17432 & 47.1 & 66.7 & 19740 & 63.3 & 78.3 & 17924 \\
 & 256 & 72.9 & 83.3 & 16485 & 63.8 & 83.3 & 19818 & 63.3 & 83.3 & 18075 & 43.8 & 53.3 & 21951 & 60.9 & 75.8 & 19082 \\
 & 512 & 71.7 & 86.7 & 16861 & 60.0 & 76.7 & 20223 & 62.9 & 83.3 & 18215 & 41.7 & 66.7 & 22330 & 59.1 & 78.3 & 19407 \\
 & 1024 & 66.3 & 80.0 & 18395 & 62.9 & 80.0 & 19655 & 62.9 & 83.3 & 18119 & 39.6 & 60.0 & 23625 & 57.9 & 75.8 & 19949 \\
 & 2048 & 63.3 & 83.3 & 17684 & 56.7 & 80.0 & 18741 & 57.9 & 76.7 & 17374 & 35.0 & 53.3 & 21690 & 53.2 & 73.3 & 18872 \\
 & 4096 & 60.0 & 83.3 & 18233 & 60.8 & 80.0 & 17913 & 55.0 & 83.3 & 17629 & 28.3 & 56.7 & 22830 & 51.0 & 75.8 & 19151 \\
 & 6144 & 59.2 & 83.3 & 18954 & 57.5 & 83.3 & 18694 & 57.5 & 73.3 & 17645 & 32.1 & 60.0 & 22426 & 51.6 & 75.0 & 19430 \\
\bottomrule
\end{tabular*}
}
\end{table}

\begin{table}[H]
\centering
{\fontsize{6.3pt}{7.0pt}\selectfont
\setlength{\tabcolsep}{0.5pt}
\renewcommand{\arraystretch}{1.04}
\textit{Table~\ref{tab:length_per_dataset} (continued).}\par\vspace{2pt}
\begin{tabular*}{\textwidth}{@{\extracolsep{\fill}}ll*{15}{c}@{}}
\toprule
\multirow{2}{*}{Model} & \multirow{2}{*}{Training rollout}
& \multicolumn{3}{c}{AIME 2024} & \multicolumn{3}{c}{AIME 2025} & \multicolumn{3}{c}{AIME 2026} & \multicolumn{3}{c}{HMMT25} & \multicolumn{3}{c}{Dataset Avg.} \\
\cmidrule(lr){3-5}\cmidrule(lr){6-8}\cmidrule(lr){9-11}\cmidrule(lr){12-14}\cmidrule(lr){15-17}
& & Avg@8 & Pass@8 & Length & Avg@8 & Pass@8 & Length & Avg@8 & Pass@8 & Length & Avg@8 & Pass@8 & Length & Avg@8 & Pass@8 & Length \\
\midrule
\multirow{8}{*}{\shortstack{Qwen3-4B-\\Instruct\\\nth/\nth}} & Base & 62.5 & 86.7 & 9412 & 45.8 & 70.0 & 8558 & 55.0 & 80.0 & 8434 & 30.8 & 50.0 & 10254 & 48.5 & 71.7 & 9165 \\
 & 128 & 60.0 & 83.3 & 9790 & 47.1 & 70.0 & 8749 & 58.8 & 80.0 & 9095 & 32.1 & 50.0 & 11587 & 49.5 & 70.8 & 9805 \\
 & 256 & 63.8 & 86.7 & 10378 & 47.5 & 73.3 & 10414 & 55.8 & 73.3 & 10143 & 31.2 & 50.0 & 12315 & 49.6 & 70.8 & 10813 \\
 & 512 & 62.5 & 86.7 & 10748 & 47.1 & 73.3 & 10873 & 55.0 & 73.3 & 11005 & 30.4 & 53.3 & 11977 & 48.8 & 71.7 & 11151 \\
 & 1024 & 65.0 & 83.3 & 10141 & 45.8 & 73.3 & 10539 & 58.3 & 76.7 & 10130 & 30.8 & 46.7 & 11409 & 50.0 & 70.0 & 10555 \\
 & 2048 & 62.5 & 83.3 & 10870 & 45.8 & 63.3 & 10014 & 53.8 & 73.3 & 9558 & 32.9 & 56.7 & 11763 & 48.8 & 69.2 & 10551 \\
 & 4096 & 60.4 & 76.7 & 10090 & 42.9 & 66.7 & 11031 & 60.0 & 80.0 & 9245 & 30.8 & 43.3 & 11911 & 48.5 & 66.7 & 10569 \\
 & 6144 & 58.8 & 80.0 & 10459 & 44.2 & 63.3 & 11440 & 58.8 & 80.0 & 9921 & 31.7 & 46.7 & 11483 & 48.3 & 67.5 & 10826 \\
\midrule
\multirow{10}{*}{\shortstack{OLMo-3-7B-\\Think\\\thmode/\thmode}} & Base & 71.7 & 86.7 & 17296 & 66.7 & 80.0 & 19137 & 71.2 & 90.0 & 17560 & 44.6 & 70.0 & 20950 & 63.5 & 81.7 & 18736 \\
 & 128 & 71.2 & 83.3 & 19310 & 68.8 & 83.3 & 20503 & 66.2 & 86.7 & 19815 & 45.4 & 70.0 & 23308 & 62.9 & 80.8 & 20734 \\
 & 256 & 67.9 & 86.7 & 20282 & 62.5 & 80.0 & 21406 & 67.5 & 83.3 & 20212 & 45.4 & 70.0 & 23691 & 60.8 & 80.0 & 21398 \\
 & 512 & 67.1 & 80.0 & 20896 & 64.2 & 83.3 & 22492 & 67.9 & 86.7 & 20801 & 43.8 & 63.3 & 24783 & 60.7 & 78.3 & 22243 \\
 & 1024 & 68.8 & 76.7 & 20502 & 55.4 & 80.0 & 22027 & 67.1 & 86.7 & 20285 & 43.8 & 66.7 & 24083 & 58.8 & 77.5 & 21724 \\
 & 1024 / Uni-256 & 67.1 & 90.0 & 19995 & 60.4 & 83.3 & 21781 & 66.3 & 86.7 & 20468 & 42.5 & 70.0 & 23707 & 59.1 & 82.5 & 21488 \\
 & 1024 / Final-256 & 70.0 & 83.3 & 19567 & 64.6 & 83.3 & 21531 & 67.1 & 86.7 & 20365 & 43.3 & 63.3 & 23137 & 61.3 & 79.2 & 21150 \\
 & 2048 & 64.6 & 83.3 & 20833 & 54.6 & 76.7 & 23336 & 66.2 & 86.7 & 20502 & 38.8 & 60.0 & 25313 & 56.0 & 76.7 & 22496 \\
 & 4096 & 57.5 & 76.7 & 21608 & 47.5 & 76.7 & 24474 & 56.7 & 83.3 & 22804 & 34.6 & 56.7 & 25445 & 49.1 & 73.3 & 23583 \\
 & 6144 & 52.5 & 73.3 & 23037 & 44.2 & 70.0 & 25369 & 49.2 & 86.7 & 24015 & 32.1 & 56.7 & 26872 & 44.5 & 71.7 & 24823 \\
\midrule
\multirow{8}{*}{\shortstack{OLMo-3-7B-\\Instruct\\\nth/\nth}} & Base & 48.8 & 76.7 & 7459 & 39.6 & 60.0 & 7182 & 42.5 & 66.7 & 7277 & 27.1 & 40.0 & 8528 & 39.5 & 60.8 & 7612 \\
 & 128 & 55.0 & 80.0 & 8329 & 41.7 & 63.3 & 7560 & 48.8 & 73.3 & 7592 & 25.0 & 43.3 & 9661 & 42.6 & 65.0 & 8286 \\
 & 256 & 52.1 & 83.3 & 8329 & 41.2 & 73.3 & 7928 & 50.4 & 70.0 & 7865 & 27.9 & 46.7 & 9323 & 42.9 & 68.3 & 8361 \\
 & 512 & 54.6 & 80.0 & 7619 & 39.6 & 66.7 & 7554 & 49.2 & 66.7 & 7329 & 24.6 & 46.7 & 9160 & 42.0 & 65.0 & 7915 \\
 & 1024 & 53.8 & 76.7 & 7201 & 39.6 & 63.3 & 6818 & 46.2 & 66.7 & 6983 & 23.8 & 43.3 & 8698 & 40.8 & 62.5 & 7425 \\
 & 2048 & 52.1 & 76.7 & 7034 & 39.2 & 60.0 & 6763 & 45.0 & 76.7 & 6945 & 24.6 & 43.3 & 8600 & 40.2 & 64.2 & 7335 \\
 & 4096 & 50.0 & 80.0 & 7374 & 37.9 & 63.3 & 6591 & 44.2 & 63.3 & 6828 & 25.0 & 40.0 & 8047 & 39.3 & 61.7 & 7210 \\
 & 6144 & 48.8 & 73.3 & 7589 & 39.6 & 70.0 & 7030 & 44.2 & 70.0 & 7063 & 20.4 & 36.7 & 8200 & 38.2 & 62.5 & 7471 \\
\bottomrule
\end{tabular*}
}
\end{table}

\subsubsection{Extended-Training OPSD Checkpoints}
\label{app:opsd_extend300_per_dataset}

Step~0 is Base, step~100 is the main text checkpoint, and later rows continue the same matched \thmode/\thmode{} run to 300 updates. The math sweep trains on OpenThoughts, and the coding sweep trains on OpenCodeReasoning.

\begin{table}[H]
\caption{Per-dataset extended-training OPSD results for \autoref{fig:opsd_extend300}. Len denotes Length.}
\label{tab:opsd_extend300_per_dataset}
\centering
{\fontsize{6.2pt}{7.0pt}\selectfont
\setlength{\tabcolsep}{1.6pt}
\renewcommand{\arraystretch}{1.02}
\begin{tabular*}{\textwidth}{@{\extracolsep{\fill}}l*{15}{c}@{}}
\toprule
\multirow{2}{*}{Steps}
& \multicolumn{3}{c}{AIME 2024} & \multicolumn{3}{c}{AIME 2025} & \multicolumn{3}{c}{AIME 2026} & \multicolumn{3}{c}{HMMT25} & \multicolumn{3}{c}{Dataset Avg.} \\
\cmidrule(lr){2-4}\cmidrule(lr){5-7}\cmidrule(lr){8-10}\cmidrule(lr){11-13}\cmidrule(lr){14-16}
& Avg@8 & Pass@8 & Len & Avg@8 & Pass@8 & Len & Avg@8 & Pass@8 & Len & Avg@8 & Pass@8 & Len & Avg@8 & Pass@8 & Len \\
\midrule
\multicolumn{16}{@{}l}{\textit{Qwen3-1.7B}} \\
Base & 43.8 & 76.7 & 18647 & 35.4 & 60.0 & 18266 & 35.0 & 53.3 & 18161 & 21.7 & 46.7 & 19421 & 34.0 & 59.2 & 18624 \\
50 & 47.9 & 80.0 & 23206 & 35.4 & 60.0 & 23071 & 38.3 & 63.3 & 23010 & 22.1 & 43.3 & 25737 & 35.9 & 61.7 & 23756 \\
100 & 45.8 & 73.3 & 22175 & 36.7 & 66.7 & 22143 & 36.7 & 63.3 & 21288 & 22.1 & 40.0 & 23913 & 35.3 & 60.8 & 22380 \\
150 & 43.3 & 73.3 & 19383 & 31.7 & 56.7 & 19359 & 36.3 & 60.0 & 18536 & 20.4 & 43.3 & 21521 & 32.9 & 58.3 & 19700 \\
200 & 41.7 & 73.3 & 19724 & 32.5 & 46.7 & 18206 & 30.4 & 60.0 & 18620 & 21.3 & 40.0 & 20508 & 31.5 & 55.0 & 19264 \\
250 & 40.8 & 70.0 & 19622 & 27.5 & 43.3 & 19806 & 34.2 & 63.3 & 17825 & 21.3 & 33.3 & 20405 & 30.9 & 52.5 & 19414 \\
300 & 42.1 & 73.3 & 19243 & 29.6 & 60.0 & 18902 & 33.8 & 56.7 & 18687 & 23.8 & 40.0 & 20578 & 32.3 & 57.5 & 19353 \\
\midrule
\multicolumn{16}{@{}l}{\textit{OLMo-3-7B-Think}} \\
Base & 71.7 & 86.7 & 17296 & 66.7 & 80.0 & 19137 & 71.2 & 90.0 & 17560 & 44.6 & 70.0 & 20950 & 63.5 & 81.7 & 18736 \\
50 & 66.7 & 80.0 & 20029 & 66.3 & 83.3 & 20831 & 65.8 & 86.7 & 19779 & 43.3 & 70.0 & 24332 & 60.5 & 80.0 & 21243 \\
100 & 68.8 & 76.7 & 20502 & 55.4 & 80.0 & 22027 & 67.1 & 86.7 & 20285 & 43.8 & 66.7 & 24083 & 58.8 & 77.5 & 21724 \\
150 & 66.7 & 80.0 & 19975 & 61.7 & 86.7 & 22177 & 62.1 & 86.7 & 21138 & 39.2 & 60.0 & 24007 & 57.4 & 78.3 & 21824 \\
200 & 59.6 & 83.3 & 20854 & 49.2 & 76.7 & 23271 & 58.8 & 86.7 & 21628 & 38.8 & 56.7 & 24472 & 51.6 & 75.8 & 22556 \\
250 & 62.9 & 83.3 & 20572 & 49.6 & 76.7 & 23718 & 52.1 & 83.3 & 22067 & 34.6 & 53.3 & 25399 & 49.8 & 74.2 & 22939 \\
300 & 61.3 & 80.0 & 20792 & 50.0 & 73.3 & 22752 & 55.0 & 83.3 & 21970 & 38.3 & 56.7 & 25127 & 51.1 & 73.3 & 22660 \\
\bottomrule
\end{tabular*}
}
\end{table}

\begin{table}[H]
\caption{Per-dataset coding results for \autoref{fig:opsd_extend300}. LCB v6 denotes LiveCodeBench v6.}
\label{tab:opsd_extend300_coding_per_dataset}
\centering
{\fontsize{6.2pt}{7.0pt}\selectfont
\setlength{\tabcolsep}{1.6pt}
\renewcommand{\arraystretch}{1.02}
\begin{tabular*}{\textwidth}{@{\extracolsep{\fill}}l*{12}{c}@{}}
\toprule
\multirow{2}{*}{Steps}
& \multicolumn{3}{c}{HumanEvalPlus} & \multicolumn{3}{c}{MBPPPlus} & \multicolumn{3}{c}{LCB v6} & \multicolumn{3}{c}{Dataset Avg.} \\
\cmidrule(lr){2-4}\cmidrule(lr){5-7}\cmidrule(lr){8-10}\cmidrule(lr){11-13}
& Avg@4 & Pass@4 & Len & Avg@4 & Pass@4 & Len & Avg@4 & Pass@4 & Len & Avg@4 & Pass@4 & Len \\
\midrule
\multicolumn{13}{@{}l}{\textit{Qwen3-1.7B}} \\
Base & 77.9 & 86.6 & 3886 & 67.6 & 77.0 & 3191 & 31.0 & 38.3 & 14924 & 58.8 & 67.3 & 7334 \\
50 & 80.5 & 87.8 & 9057 & 66.9 & 76.2 & 7186 & 30.9 & 35.4 & 26686 & 59.4 & 66.5 & 14310 \\
100 & 78.4 & 87.8 & 8363 & 68.2 & 77.5 & 6769 & 31.4 & 38.3 & 25701 & 59.3 & 67.9 & 13611 \\
150 & 77.1 & 85.4 & 8645 & 65.5 & 74.9 & 6542 & 28.9 & 37.7 & 23755 & 57.2 & 66.0 & 12981 \\
200 & 75.9 & 89.0 & 8594 & 65.7 & 74.3 & 6301 & 28.9 & 37.7 & 21973 & 56.8 & 67.0 & 12289 \\
250 & 75.8 & 85.4 & 8248 & 65.5 & 74.9 & 6054 & 27.9 & 36.6 & 22022 & 56.4 & 65.6 & 12108 \\
300 & 75.6 & 88.4 & 8785 & 66.1 & 76.2 & 6310 & 29.3 & 38.3 & 22105 & 57.0 & 67.6 & 12400 \\
\midrule
\multicolumn{13}{@{}l}{\textit{OLMo-3-7B-Think}} \\
Base & 84.6 & 94.5 & 5810 & 73.2 & 84.1 & 5523 & 47.4 & 60.6 & 17552 & 68.4 & 79.7 & 9628 \\
50 & 85.1 & 96.3 & 7877 & 72.8 & 85.2 & 8102 & 43.4 & 52.6 & 22339 & 67.1 & 78.0 & 12773 \\
100 & 84.9 & 93.9 & 8199 & 73.5 & 84.9 & 8164 & 45.3 & 55.4 & 19450 & 67.9 & 78.1 & 11938 \\
150 & 85.2 & 96.3 & 8635 & 72.2 & 84.4 & 8819 & 41.3 & 53.7 & 22926 & 66.2 & 78.1 & 13460 \\
200 & 85.7 & 93.9 & 9060 & 72.6 & 84.1 & 8954 & 41.9 & 53.1 & 22830 & 66.7 & 77.1 & 13615 \\
250 & 83.1 & 95.1 & 9305 & 71.9 & 84.1 & 8972 & 41.9 & 51.4 & 22739 & 65.6 & 76.9 & 13672 \\
300 & 86.1 & 95.7 & 8849 & 71.8 & 84.1 & 9212 & 42.0 & 53.7 & 22623 & 66.7 & 77.9 & 13561 \\
\bottomrule
\end{tabular*}
}
\end{table}

\subsubsection{Teacher-Side Context}
\label{app:context_per_dataset}

All rows use OpenThoughts Math 30k. None denotes Base without privileged context. For Qwen3-1.7B and OLMo-3-7B-Think, CoT~+ solution uses the same examples and order as Full solution. The teacher context and prompt cap (8192 instead of 1024 tokens) change.

\begin{table}[H]
\caption{Per-dataset teacher context results for \autoref{tab:teacher_context} and \autoref{tab:teacher_context_1p7b_cot}.}
\label{tab:context_per_dataset}
\centering
{\fontsize{6.3pt}{7.0pt}\selectfont
\setlength{\tabcolsep}{0.5pt}
\renewcommand{\arraystretch}{1.04}
\begin{tabular*}{\textwidth}{@{\extracolsep{\fill}}lll*{15}{c}@{}}
\toprule
\multirow{2}{*}{Model} & \multirow{2}{*}{Rollout} & \multirow{2}{*}{\begin{tabular}[c]{@{}c@{}}Privileged\\context\end{tabular}}
& \multicolumn{3}{c}{AIME 2024} & \multicolumn{3}{c}{AIME 2025} & \multicolumn{3}{c}{AIME 2026} & \multicolumn{3}{c}{HMMT25} & \multicolumn{3}{c}{Dataset Avg.} \\
\cmidrule(lr){4-6}\cmidrule(lr){7-9}\cmidrule(lr){10-12}\cmidrule(lr){13-15}\cmidrule(lr){16-18}
& & & Avg@8 & Pass@8 & Length & Avg@8 & Pass@8 & Length & Avg@8 & Pass@8 & Length & Avg@8 & Pass@8 & Length & Avg@8 & Pass@8 & Length \\
\midrule
\multirow{8}{*}{Qwen3-1.7B} & -- & None & 43.8 & 76.7 & 18647 & 35.4 & 60.0 & 18266 & 35.0 & 53.3 & 18161 & 21.7 & 46.7 & 19421 & 34.0 & 59.2 & 18624 \\
 & 256 & Full solution & 46.3 & 70.0 & 20783 & 36.7 & 66.7 & 21697 & 45.0 & 73.3 & 20995 & 24.6 & 40.0 & 22306 & 38.1 & 62.5 & 21445 \\
 & 256 & Answer only & 49.6 & 70.0 & 23223 & 38.8 & 60.0 & 24084 & 38.3 & 66.7 & 22631 & 25.0 & 40.0 & 25702 & 37.9 & 59.2 & 23910 \\
 & 256 & Unrelated & 47.9 & 73.3 & 19074 & 40.0 & 63.3 & 19943 & 36.3 & 66.7 & 20098 & 23.3 & 46.7 & 20165 & 36.9 & 62.5 & 19820 \\
 & 1024 & Full solution & 45.8 & 73.3 & 22175 & 36.7 & 66.7 & 22143 & 36.7 & 63.3 & 21288 & 22.1 & 40.0 & 23913 & 35.3 & 60.8 & 22380 \\
 & 1024 & Answer only & 43.8 & 63.3 & 26399 & 34.2 & 63.3 & 26574 & 38.8 & 63.3 & 24559 & 23.8 & 43.3 & 28577 & 35.1 & 58.3 & 26527 \\
 & 1024 & Unrelated & 47.5 & 70.0 & 22011 & 32.1 & 53.3 & 23665 & 43.8 & 63.3 & 21083 & 23.8 & 36.7 & 23133 & 36.8 & 55.8 & 22473 \\
 & 1024 & CoT + solution & 35.4 & 70.0 & 18947 & 30.0 & 53.3 & 17435 & 32.9 & 56.7 & 18752 & 25.0 & 43.3 & 19299 & 30.8 & 55.8 & 18608 \\
\midrule
\multirow{7}{*}{Qwen3-4B} & -- & None & 74.6 & 90.0 & 15495 & 65.4 & 80.0 & 18696 & 65.8 & 80.0 & 17249 & 43.3 & 66.7 & 19636 & 62.3 & 79.2 & 17769 \\
 & 256 & Full solution & 72.9 & 83.3 & 16485 & 63.8 & 83.3 & 19818 & 63.3 & 83.3 & 18075 & 43.8 & 53.3 & 21951 & 60.9 & 75.8 & 19082 \\
 & 256 & Answer only & 68.3 & 86.7 & 19266 & 62.9 & 73.3 & 21919 & 65.8 & 83.3 & 20063 & 40.8 & 56.7 & 23971 & 59.5 & 75.0 & 21305 \\
 & 256 & Unrelated & 70.8 & 83.3 & 16952 & 66.3 & 80.0 & 20057 & 67.9 & 83.3 & 18752 & 42.1 & 60.0 & 21451 & 61.8 & 76.7 & 19303 \\
 & 1024 & Full solution & 66.3 & 80.0 & 18395 & 62.9 & 80.0 & 19655 & 62.9 & 83.3 & 18119 & 39.6 & 60.0 & 23625 & 57.9 & 75.8 & 19949 \\
 & 1024 & Answer only & 66.7 & 76.7 & 20339 & 64.2 & 83.3 & 22500 & 62.5 & 83.3 & 20465 & 36.3 & 53.3 & 25267 & 57.4 & 74.2 & 22143 \\
 & 1024 & Unrelated & 68.3 & 83.3 & 22068 & 62.9 & 80.0 & 24667 & 64.2 & 86.7 & 23108 & 42.9 & 56.7 & 25404 & 59.6 & 76.7 & 23812 \\
\midrule
\multirow{7}{*}{\ml{Qwen3-4B-\\Instruct}} & -- & None & 62.5 & 86.7 & 9412 & 45.8 & 70.0 & 8558 & 55.0 & 80.0 & 8434 & 30.8 & 50.0 & 10254 & 48.5 & 71.7 & 9165 \\
 & 256 & Full solution & 63.8 & 86.7 & 10378 & 47.5 & 73.3 & 10414 & 55.8 & 73.3 & 10143 & 31.2 & 50.0 & 12315 & 49.6 & 70.8 & 10813 \\
 & 256 & Answer only & 59.6 & 83.3 & 9867 & 44.2 & 70.0 & 9542 & 55.4 & 80.0 & 9266 & 31.7 & 46.7 & 11185 & 47.7 & 70.0 & 9965 \\
 & 256 & Unrelated & 63.3 & 83.3 & 9095 & 47.9 & 73.3 & 8854 & 53.8 & 76.7 & 8264 & 27.9 & 43.3 & 9605 & 48.2 & 69.2 & 8954 \\
 & 1024 & Full solution & 65.0 & 83.3 & 10141 & 45.8 & 73.3 & 10539 & 58.3 & 76.7 & 10130 & 30.8 & 46.7 & 11409 & 50.0 & 70.0 & 10555 \\
 & 1024 & Answer only & 64.2 & 83.3 & 10101 & 42.5 & 70.0 & 10618 & 48.8 & 73.3 & 9566 & 29.6 & 43.3 & 12786 & 46.3 & 67.5 & 10768 \\
 & 1024 & Unrelated & 59.2 & 80.0 & 10370 & 40.0 & 60.0 & 10839 & 52.9 & 66.7 & 10422 & 28.8 & 46.7 & 12188 & 45.2 & 63.3 & 10955 \\
\midrule
\multirow{8}{*}{\ml{OLMo-3-\\7B-Think}} & -- & None & 71.7 & 86.7 & 17296 & 66.7 & 80.0 & 19137 & 71.2 & 90.0 & 17560 & 44.6 & 70.0 & 20950 & 63.5 & 81.7 & 18736 \\
 & 256 & Full solution & 67.9 & 86.7 & 20282 & 62.5 & 80.0 & 21406 & 67.5 & 83.3 & 20212 & 45.4 & 70.0 & 23691 & 60.8 & 80.0 & 21398 \\
 & 256 & Answer only & 69.2 & 80.0 & 20708 & 63.8 & 80.0 & 22298 & 66.2 & 86.7 & 21301 & 42.1 & 60.0 & 23954 & 60.3 & 76.7 & 22065 \\
 & 256 & Unrelated & 66.7 & 83.3 & 19971 & 63.3 & 83.3 & 21395 & 63.3 & 83.3 & 20782 & 44.6 & 63.3 & 23084 & 59.5 & 78.3 & 21308 \\
 & 1024 & Full solution & 68.8 & 76.7 & 20502 & 55.4 & 80.0 & 22027 & 67.1 & 86.7 & 20285 & 43.8 & 66.7 & 24083 & 58.8 & 77.5 & 21724 \\
 & 1024 & Answer only & 64.2 & 90.0 & 23176 & 65.0 & 80.0 & 24229 & 61.3 & 86.7 & 23307 & 40.0 & 66.7 & 26068 & 57.6 & 80.8 & 24195 \\
 & 1024 & Unrelated & 67.9 & 80.0 & 21942 & 63.8 & 80.0 & 23527 & 66.7 & 86.7 & 22130 & 39.6 & 63.3 & 25263 & 59.5 & 77.5 & 23216 \\
 & 1024 & CoT + solution & 64.6 & 80.0 & 19337 & 64.6 & 80.0 & 21006 & 62.1 & 90.0 & 20053 & 43.3 & 60.0 & 22843 & 58.6 & 77.5 & 20810 \\
\midrule
\multirow{7}{*}{\ml{OLMo-3-\\7B-Instruct}} & -- & None & 48.8 & 76.7 & 7459 & 39.6 & 60.0 & 7182 & 42.5 & 66.7 & 7277 & 27.1 & 40.0 & 8528 & 39.5 & 60.8 & 7612 \\
 & 256 & Full solution & 52.1 & 83.3 & 8329 & 41.2 & 73.3 & 7928 & 50.4 & 70.0 & 7865 & 27.9 & 46.7 & 9323 & 42.9 & 68.3 & 8361 \\
 & 256 & Answer only & 32.9 & 60.0 & 4603 & 22.1 & 40.0 & 3889 & 26.7 & 56.7 & 4482 & 14.2 & 33.3 & 4500 & 24.0 & 47.5 & 4368 \\
 & 256 & Unrelated & 50.4 & 80.0 & 8152 & 37.1 & 60.0 & 7741 & 44.6 & 70.0 & 7590 & 20.8 & 36.7 & 8035 & 38.2 & 61.7 & 7880 \\
 & 1024 & Full solution & 53.8 & 76.7 & 7201 & 39.6 & 63.3 & 6818 & 46.2 & 66.7 & 6983 & 23.8 & 43.3 & 8698 & 40.8 & 62.5 & 7425 \\
 & 1024 & Answer only & 30.0 & 56.7 & 4861 & 28.3 & 46.7 & 3766 & 31.7 & 63.3 & 4875 & 16.3 & 23.3 & 5271 & 26.6 & 47.5 & 4693 \\
 & 1024 & Unrelated & 53.8 & 80.0 & 9794 & 38.8 & 63.3 & 8761 & 47.1 & 66.7 & 8847 & 26.7 & 50.0 & 10405 & 41.6 & 65.0 & 9452 \\
\bottomrule
\end{tabular*}
}
\end{table}

\paragraph{OpenMathReasoning Replication.}
We repeat the context comparison on Qwen3-1.7B using two disjoint OpenMathReasoning subsets defined by post-thinking length. Solution and CoT~+ solution use identical examples and ordering within each subset. Adding CoT lowers Avg@8 in both subsets (\autoref{tab:context_omr_per_dataset}).
The results proves that providing more privileged information does not help improving the performance of OPSD.

\begin{table}[H]
\caption{Qwen3-1.7B teacher context replication on the OpenMathReasoning \texttt{cot} split with a student rollout cap of 1024 tokens. Each pair uses identical examples. Subset labels give the post-thinking solution length before CoT is added. CoT~+ solution uses a teacher cap of 12288 tokens.}
\label{tab:context_omr_per_dataset}
\centering
{\fontsize{6.3pt}{7.0pt}\selectfont
\setlength{\tabcolsep}{0.5pt}
\renewcommand{\arraystretch}{1.04}
\begin{tabular*}{\textwidth}{@{\extracolsep{\fill}}lll*{15}{c}@{}}
\toprule
\multirow{2}{*}{\begin{tabular}[c]{@{}c@{}}Post-thinking\\subset\end{tabular}} & \multirow{2}{*}{\begin{tabular}[c]{@{}c@{}}Privileged\\context\end{tabular}} & \multirow{2}{*}{\begin{tabular}[c]{@{}c@{}}Teacher\\cap\end{tabular}}
& \multicolumn{3}{c}{AIME 2024} & \multicolumn{3}{c}{AIME 2025} & \multicolumn{3}{c}{AIME 2026} & \multicolumn{3}{c}{HMMT25} & \multicolumn{3}{c}{Dataset Avg.} \\
\cmidrule(lr){4-6}\cmidrule(lr){7-9}\cmidrule(lr){10-12}\cmidrule(lr){13-15}\cmidrule(lr){16-18}
& & & Avg@8 & Pass@8 & Length & Avg@8 & Pass@8 & Length & Avg@8 & Pass@8 & Length & Avg@8 & Pass@8 & Length & Avg@8 & Pass@8 & Length \\
\midrule
\multirow{2}{*}{$\leq$1k} & Solution & 1024 & 45.8 & 73.3 & 22865 & 35.0 & 60.0 & 22109 & 34.6 & 63.3 & 21651 & 22.1 & 40.0 & 24262 & 34.4 & 59.2 & 22722 \\
 & CoT + solution & 12288 & 37.9 & 76.7 & 16997 & 27.9 & 53.3 & 16101 & 28.3 & 56.7 & 16444 & 17.1 & 36.7 & 18137 & 27.8 & 55.8 & 16920 \\
\midrule
\multirow{2}{*}{2--4k} & Solution & 4096 & 45.4 & 66.7 & 21721 & 37.1 & 66.7 & 22442 & 32.5 & 56.7 & 22517 & 23.8 & 46.7 & 24077 & 34.7 & 59.2 & 22689 \\
 & CoT + solution & 12288 & 38.3 & 80.0 & 17884 & 27.9 & 43.3 & 16633 & 27.1 & 53.3 & 16678 & 22.5 & 40.0 & 17620 & 29.0 & 54.2 & 17204 \\
\bottomrule
\end{tabular*}
}
\end{table}

\needspace{23\baselineskip}
\paragraph{Answer-Only Controls.}
The conditions below correspond to the step-50 comparison in \autoref{tab:answer_only_controls}. Random problems uses correct answers from the unrestricted sampling pool; the remaining answer-only conditions share the integer-answer subset. Details are in Appendix~\ref{app:answer_only_controls}.

\begin{table}[H]
\caption{Per-dataset answer-only controls. Avg@8 and Pass@8 are percentages; Length is the mean number of output tokens. Base denotes the model before OPSD, and all trained checkpoints are evaluated at step 50.}
\label{tab:answer_only_per_dataset}
\centering
{\fontsize{6.5pt}{7.3pt}\selectfont
\setlength{\tabcolsep}{1pt}
\renewcommand{\arraystretch}{1.04}
\begin{tabular*}{\textwidth}{@{\extracolsep{\fill}}l*{12}{c}@{}}
\toprule
\multirow{2}{*}{Answer condition} & \multicolumn{3}{c}{AIME 2024} & \multicolumn{3}{c}{AIME 2025} & \multicolumn{3}{c}{AIME 2026} & \multicolumn{3}{c}{HMMT25} \\
\cmidrule(lr){2-4}\cmidrule(lr){5-7}\cmidrule(lr){8-10}\cmidrule(lr){11-13}
& Avg@8 & Pass@8 & Length & Avg@8 & Pass@8 & Length & Avg@8 & Pass@8 & Length & Avg@8 & Pass@8 & Length \\
\midrule
\multicolumn{13}{l}{\textbf{Qwen3-1.7B}} \\
None (Base) & 43.75 & 76.67 & 18647 & 35.42 & 60.00 & 18266 & 35.00 & 53.33 & 18161 & 21.67 & 46.67 & 19421 \\
Random problems & 40.83 & 60.00 & 27798 & 37.08 & 53.33 & 27585 & 41.25 & 63.33 & 26394 & 23.75 & 36.67 & 29944 \\
Correct integer & 50.83 & 73.33 & 22797 & 40.00 & 66.67 & 24131 & 43.33 & 70.00 & 22510 & 25.42 & 43.33 & 25077 \\
Equivalent form & 50.83 & 76.67 & 23475 & 44.17 & 70.00 & 24112 & 45.42 & 63.33 & 22650 & 25.42 & 46.67 & 25574 \\
Wrong (fixed) & 48.75 & 73.33 & 22901 & 42.50 & 73.33 & 23789 & 48.33 & 73.33 & 22606 & 29.17 & 53.33 & 25148 \\
Wrong (random) & 54.58 & 76.67 & 22603 & 38.33 & 60.00 & 23762 & 43.75 & 66.67 & 22604 & 26.67 & 43.33 & 25426 \\
\midrule
\multicolumn{13}{l}{\textbf{OLMo-3-7B-Think}} \\
None (Base) & 71.67 & 86.67 & 17296 & 66.67 & 80.00 & 19137 & 71.25 & 90.00 & 17560 & 44.58 & 70.00 & 20950 \\
Random problems & 64.17 & 83.33 & 23773 & 62.50 & 80.00 & 24803 & 67.08 & 86.67 & 23773 & 40.00 & 66.67 & 26845 \\
Correct integer & 65.00 & 80.00 & 22647 & 61.67 & 83.33 & 24115 & 63.33 & 86.67 & 23236 & 39.58 & 56.67 & 26307 \\
Equivalent form & 67.08 & 86.67 & 22286 & 61.25 & 83.33 & 23903 & 64.58 & 86.67 & 22667 & 40.83 & 66.67 & 26391 \\
Wrong (fixed) & 66.25 & 76.67 & 22051 & 61.25 & 80.00 & 23998 & 64.58 & 90.00 & 22388 & 43.33 & 63.33 & 25899 \\
Wrong (random) & 62.92 & 83.33 & 22668 & 60.00 & 83.33 & 23802 & 64.17 & 86.67 & 22803 & 40.42 & 63.33 & 25891 \\
\bottomrule
\end{tabular*}
}
\end{table}

\subsubsection{Divergence Objective Ablation}
\label{app:divergence_per_dataset}

Forward KL uses $\beta=0$, symmetric JSD uses $\beta=0.5$, and reverse KL uses $\beta=1$.

\begin{table}[H]
\caption{Per-dataset divergence ablation results for \autoref{tab:divergence_main} and \autoref{tab:divergence_macro}.}
\label{tab:divergence_per_dataset}
\centering
{\fontsize{6.3pt}{7.0pt}\selectfont
\setlength{\tabcolsep}{0.5pt}
\renewcommand{\arraystretch}{1.04}
\begin{tabular*}{\textwidth}{@{\extracolsep{\fill}}ll*{15}{c}@{}}
\toprule
\multirow{2}{*}{Model} & \multirow{2}{*}{Objective} & \multicolumn{3}{c}{AIME 2024} & \multicolumn{3}{c}{AIME 2025} & \multicolumn{3}{c}{AIME 2026} & \multicolumn{3}{c}{HMMT25} & \multicolumn{3}{c}{Dataset Avg.} \\
\cmidrule(lr){3-5}\cmidrule(lr){6-8}\cmidrule(lr){9-11}\cmidrule(lr){12-14}\cmidrule(lr){15-17}
& & Avg@8 & Pass@8 & Length & Avg@8 & Pass@8 & Length & Avg@8 & Pass@8 & Length & Avg@8 & Pass@8 & Length & Avg@8 & Pass@8 & Length \\
\midrule
\multirow{3}{*}{Qwen3-1.7B}
& Fwd. KL ($\beta=0$) & 45.8 & 73.3 & 22175 & 36.7 & 66.7 & 22143 & 36.7 & 63.3 & 21288 & 22.1 & 40.0 & 23913 & 35.3 & 60.8 & 22380 \\
& JSD ($\beta=0.5$) & 40.0 & 70.0 & 14950 & 32.9 & 53.3 & 15222 & 33.3 & 60.0 & 14764 & 18.8 & 36.7 & 15931 & 31.3 & 55.0 & 15217 \\
& Rev. KL ($\beta=1$) & 37.5 & 63.3 & 23381 & 31.7 & 46.7 & 22391 & 23.8 & 46.7 & 23906 & 12.9 & 33.3 & 27968 & 26.5 & 47.5 & 24411 \\
\midrule
\multirow{3}{*}{Qwen3-4B}
& Fwd. KL ($\beta=0$) & 66.3 & 80.0 & 18395 & 62.9 & 80.0 & 19655 & 62.9 & 83.3 & 18119 & 39.6 & 60.0 & 23625 & 57.9 & 75.8 & 19949 \\
& JSD ($\beta=0.5$) & 67.1 & 83.3 & 12485 & 59.6 & 76.7 & 14881 & 60.8 & 83.3 & 13550 & 38.8 & 60.0 & 15634 & 56.6 & 75.8 & 14137 \\
& Rev. KL ($\beta=1$) & 37.9 & 73.3 & 27816 & 36.3 & 56.7 & 27897 & 33.8 & 60.0 & 26749 & 20.0 & 36.7 & 29080 & 32.0 & 56.7 & 27886 \\
\bottomrule
\end{tabular*}
}
\end{table}

\subsubsection{Student Entropy Loss Masking}
\label{app:entropy_train_per_dataset}

All runs use a 256-token rollout.

\begin{table}[H]
\caption{Per-dataset student entropy loss masking results for \autoref{tab:entropy_train}. Full uses all tokens. HE20/HE80 and LE20/LE80 select the highest- or lowest-entropy fraction of each completion.}
\label{tab:entropy_train_per_dataset}
\centering
{\fontsize{6.0pt}{6.7pt}\selectfont
\setlength{\tabcolsep}{0.4pt}
\renewcommand{\arraystretch}{1.02}
\begin{tabular*}{\textwidth}{@{\extracolsep{\fill}}ll*{15}{c}@{}}
\toprule
\multirow{2}{*}{Model} & \multirow{2}{*}{Subset}
& \multicolumn{3}{c}{AIME 2024} & \multicolumn{3}{c}{AIME 2025} & \multicolumn{3}{c}{AIME 2026} & \multicolumn{3}{c}{HMMT25} & \multicolumn{3}{c}{Dataset Avg.} \\
\cmidrule(lr){3-5}\cmidrule(lr){6-8}\cmidrule(lr){9-11}\cmidrule(lr){12-14}\cmidrule(lr){15-17}
& & Avg@8 & Pass@8 & Length & Avg@8 & Pass@8 & Length & Avg@8 & Pass@8 & Length & Avg@8 & Pass@8 & Length & Avg@8 & Pass@8 & Length \\
\midrule
\multirow{6}{*}{Qwen3-1.7B} & Base & 43.8 & 76.7 & 18647 & 35.4 & 60.0 & 18266 & 35.0 & 53.3 & 18161 & 21.7 & 46.7 & 19421 & 34.0 & 59.2 & 18624 \\
 & Full & 46.3 & 70.0 & 20783 & 36.7 & 66.7 & 21697 & 45.0 & 73.3 & 20995 & 24.6 & 40.0 & 22306 & 38.1 & 62.5 & 21445 \\
 & HE20 & 44.6 & 76.7 & 21009 & 36.7 & 63.3 & 22166 & 39.6 & 73.3 & 21696 & 26.7 & 50.0 & 22074 & 36.9 & 65.8 & 21736 \\
 & HE80 & 52.5 & 76.7 & 20083 & 34.6 & 56.7 & 21365 & 40.0 & 70.0 & 20013 & 24.2 & 43.3 & 21676 & 37.8 & 61.7 & 20784 \\
 & LE20 & 49.6 & 73.3 & 17001 & 37.9 & 63.3 & 17766 & 39.2 & 73.3 & 18538 & 23.8 & 46.7 & 18619 & 37.6 & 64.2 & 17981 \\
 & LE80 & 45.8 & 73.3 & 17275 & 37.5 & 60.0 & 16975 & 30.0 & 70.0 & 17773 & 19.6 & 33.3 & 18633 & 33.2 & 59.2 & 17664 \\
\bottomrule
\end{tabular*}
}
\end{table}

\subsubsection{Vocabulary Top-$k$ Distillation Loss}
\label{app:topk_loss_per_dataset}

Full vocab rows are matched 100-step controls: forward KL for Qwen3-1.7B and Qwen3-4B-Instruct, and reverse KL for OLMo-3-7B-Think.

\begin{table}[H]
\caption{Per-dataset vocabulary top-$k$ results for \autoref{tab:topk_loss_macro}.}
\label{tab:topk_loss_per_dataset}
\centering
{\fontsize{6.3pt}{7.0pt}\selectfont
\setlength{\tabcolsep}{0.5pt}
\renewcommand{\arraystretch}{1.04}
\begin{tabular*}{\textwidth}{@{\extracolsep{\fill}}ll*{15}{c}@{}}
\toprule
\multirow{2}{*}{Model} & \multirow{2}{*}{Support} & \multicolumn{3}{c}{AIME 2024} & \multicolumn{3}{c}{AIME 2025} & \multicolumn{3}{c}{AIME 2026} & \multicolumn{3}{c}{HMMT25} & \multicolumn{3}{c}{Dataset Avg.} \\
\cmidrule(lr){3-5}\cmidrule(lr){6-8}\cmidrule(lr){9-11}\cmidrule(lr){12-14}\cmidrule(lr){15-17}
& & Avg@8 & Pass@8 & Length & Avg@8 & Pass@8 & Length & Avg@8 & Pass@8 & Length & Avg@8 & Pass@8 & Length & Avg@8 & Pass@8 & Length \\
\midrule
\multirow{3}{*}{Qwen3-1.7B}
& Full vocab & 45.8 & 73.3 & 22175 & 36.7 & 66.7 & 22143 & 36.7 & 63.3 & 21288 & 22.1 & 40.0 & 23913 & 35.3 & 60.8 & 22380 \\
& $k=16$ & 42.5 & 80.0 & 22299 & 31.7 & 56.7 & 22440 & 34.6 & 63.3 & 21396 & 23.3 & 50.0 & 24293 & 33.0 & 62.5 & 22607 \\
& $k=4$ & 28.3 & 60.0 & 28891 & 28.8 & 46.7 & 28108 & 31.3 & 53.3 & 26761 & 15.0 & 30.0 & 31383 & 25.8 & 47.5 & 28786 \\
\midrule
\multirow{3}{*}{\ml{Qwen3-4B-\\Instruct}}
& Full vocab & 65.0 & 83.3 & 10141 & 45.8 & 73.3 & 10539 & 58.3 & 76.7 & 10130 & 30.8 & 46.7 & 11409 & 50.0 & 70.0 & 10555 \\
& $k=16$ & 63.8 & 80.0 & 10156 & 50.8 & 70.0 & 10317 & 59.6 & 80.0 & 9726 & 33.8 & 53.3 & 12112 & 52.0 & 70.8 & 10578 \\
& $k=4$ & 57.5 & 80.0 & 8275 & 45.4 & 66.7 & 7677 & 49.6 & 76.7 & 6874 & 30.4 & 53.3 & 9850 & 45.7 & 69.2 & 8169 \\
\midrule
\multirow{3}{*}{\ml{OLMo-3-\\7B-Think}}
& Full vocab & 72.9 & 93.3 & 18338 & 65.0 & 83.3 & 20245 & 66.3 & 86.7 & 19107 & 45.4 & 66.7 & 22097 & 62.4 & 82.5 & 19947 \\
& $k=16$ & 70.0 & 86.7 & 18976 & 62.9 & 86.7 & 19850 & 64.6 & 86.7 & 18940 & 42.1 & 66.7 & 23117 & 59.9 & 81.7 & 20221 \\
& $k=4$ & 4.2 & 23.3 & 32285 & 2.9 & 23.3 & 32015 & 4.6 & 26.7 & 32082 & 2.5 & 20.0 & 32482 & 3.5 & 23.3 & 32216 \\
\bottomrule
\end{tabular*}
}
\end{table}

\subsubsection{Initial Model Ability}
\label{app:capability_per_dataset}

Checkpoint denotes Base or the rollout cap used for full solution OPSD. Rows follow the Base Pass@8 ordering in \autoref{fig:initial_capability}.

\begin{table}[H]
\caption{Per-dataset math results for \autoref{fig:initial_capability}a. 256 and 1024 are full solution OPSD conditions.}
\label{tab:capability_per_dataset}
\centering
{\fontsize{6.3pt}{7.0pt}\selectfont
\setlength{\tabcolsep}{0.5pt}
\renewcommand{\arraystretch}{1.04}
\begin{tabular*}{\textwidth}{@{\extracolsep{\fill}}ll*{15}{c}@{}}
\toprule
\multirow{2}{*}{Model} & \multirow{2}{*}{Checkpoint}
& \multicolumn{3}{c}{AIME 2024} & \multicolumn{3}{c}{AIME 2025} & \multicolumn{3}{c}{AIME 2026} & \multicolumn{3}{c}{HMMT25} & \multicolumn{3}{c}{Dataset Avg.} \\
\cmidrule(lr){3-5}\cmidrule(lr){6-8}\cmidrule(lr){9-11}\cmidrule(lr){12-14}\cmidrule(lr){15-17}
& & Avg@8 & Pass@8 & Length & Avg@8 & Pass@8 & Length & Avg@8 & Pass@8 & Length & Avg@8 & Pass@8 & Length & Avg@8 & Pass@8 & Length \\
\midrule
\multirow{3}{*}{Qwen3-0.6B} & Base & 8.8 & 30.0 & 16815 & 17.5 & 40.0 & 15188 & 12.9 & 26.7 & 16334 & 8.8 & 23.3 & 15532 & 12.0 & 30.0 & 15967 \\
 & 256 & 10.8 & 30.0 & 21812 & 18.3 & 36.7 & 19027 & 10.4 & 30.0 & 21177 & 9.2 & 26.7 & 19708 & 12.2 & 30.8 & 20431 \\
 & 1024 & 9.2 & 30.0 & 20924 & 12.5 & 30.0 & 18855 & 8.3 & 16.7 & 19339 & 7.5 & 16.7 & 20224 & 9.4 & 23.3 & 19835 \\
\midrule
\multirow{3}{*}{\shortstack[l]{DeepSeek-R1-\\Distill-Qwen-1.5B}} & Base & 28.3 & 53.3 & 17211 & 24.6 & 46.7 & 15944 & 19.6 & 43.3 & 17906 & 14.2 & 40.0 & 18742 & 21.7 & 45.8 & 17451 \\
 & 256 & 28.8 & 66.7 & 17814 & 23.3 & 43.3 & 17795 & 22.5 & 50.0 & 18387 & 15.0 & 26.7 & 18842 & 22.4 & 46.7 & 18209 \\
 & 1024 & 30.4 & 63.3 & 17405 & 21.7 & 40.0 & 16524 & 20.4 & 46.7 & 18662 & 16.7 & 30.0 & 18357 & 22.3 & 45.0 & 17737 \\
\bottomrule
\end{tabular*}
}
\end{table}

\autoref{tab:capability_coding_per_dataset} reports the coding results behind \autoref{fig:initial_capability}b. Rows follow the Base Pass@4 ordering.

\input{initial_capability_coding/per_dataset}
\FloatBarrier

\subsubsection{SFT Initialization Sweep}
\label{app:sft_per_dataset}

Direct evaluates the SFT checkpoint. After OPSD evaluates the matched 100-step frozen-self-teacher checkpoint. Base denotes zero SFT updates. The 7.5k and 12.5k rows report direct SFT.

\begin{table}[H]
\caption{Per-dataset Qwen3-1.7B-Base SFT initialization results for \autoref{tab:sft_initialization}. Thinking mode evaluation uses a 38912-token maximum output length.}
\label{tab:sft_per_dataset_1p7b}
\centering
{\fontsize{6.3pt}{7.0pt}\selectfont
\setlength{\tabcolsep}{0.5pt}
\renewcommand{\arraystretch}{1.04}
\begin{tabular*}{\textwidth}{@{\extracolsep{\fill}}ll*{15}{c}@{}}
\toprule
\multirow{2}{*}{SFT steps} & \multirow{2}{*}{Evaluation}
& \multicolumn{3}{c}{AIME 2024} & \multicolumn{3}{c}{AIME 2025} & \multicolumn{3}{c}{AIME 2026} & \multicolumn{3}{c}{HMMT25} & \multicolumn{3}{c}{Dataset Avg.} \\
\cmidrule(lr){3-5}\cmidrule(lr){6-8}\cmidrule(lr){9-11}\cmidrule(lr){12-14}\cmidrule(lr){15-17}
& & Avg@8 & Pass@8 & Length & Avg@8 & Pass@8 & Length & Avg@8 & Pass@8 & Length & Avg@8 & Pass@8 & Length & Avg@8 & Pass@8 & Length \\
\midrule
Base & Direct & 3.8 & 13.3 & 6629 & 1.2 & 10.0 & 5214 & 3.3 & 10.0 & 5654 & 0.0 & 0.0 & 4805 & 2.1 & 8.3 & 5576 \\
5k & Direct & 8.8 & 26.7 & 24373 & 9.6 & 23.3 & 23444 & 7.5 & 23.3 & 23689 & 2.5 & 13.3 & 23591 & 7.1 & 21.7 & 23774 \\
7.5k & Direct & 9.2 & 26.7 & 24317 & 10.0 & 26.7 & 23141 & 9.6 & 26.7 & 25677 & 4.2 & 20.0 & 23953 & 8.2 & 25.0 & 24272 \\
10k & Direct & 10.4 & 20.0 & 22854 & 12.9 & 23.3 & 18654 & 10.4 & 23.3 & 20890 & 4.6 & 16.7 & 20950 & 9.6 & 20.8 & 20837 \\
12.5k & Direct & 12.9 & 33.3 & 20910 & 16.2 & 36.7 & 19194 & 10.8 & 30.0 & 21869 & 4.2 & 13.3 & 20361 & 11.0 & 28.3 & 20584 \\
15k & Direct & 12.9 & 36.7 & 21830 & 15.8 & 36.7 & 20261 & 12.1 & 30.0 & 22976 & 5.0 & 13.3 & 20620 & 11.5 & 29.2 & 21422 \\
\midrule
Base & After OPSD & 2.9 & 16.7 & 8339 & 0.4 & 3.3 & 7601 & 1.2 & 10.0 & 8881 & 0.0 & 0.0 & 6987 & 1.1 & 7.5 & 7952 \\
5k & After OPSD & 0.8 & 6.7 & 35018 & 0.4 & 3.3 & 34134 & 0.0 & 0.0 & 34849 & 0.0 & 0.0 & 34599 & 0.3 & 2.5 & 34650 \\
10k & After OPSD & 5.4 & 16.7 & 34934 & 3.8 & 13.3 & 34525 & 2.5 & 10.0 & 35201 & 0.8 & 6.7 & 36296 & 3.1 & 11.7 & 35239 \\
15k & After OPSD & 2.5 & 13.3 & 35896 & 1.7 & 10.0 & 36323 & 2.1 & 6.7 & 35962 & 0.0 & 0.0 & 36550 & 1.6 & 7.5 & 36183 \\
\bottomrule
\end{tabular*}
}
\end{table}

\begin{table}[H]
\caption{Per-dataset Qwen3-4B-Base SFT initialization results for \autoref{tab:sft_initialization}.}
\label{tab:sft_per_dataset_4b}
\centering
{\fontsize{6.3pt}{7.0pt}\selectfont
\setlength{\tabcolsep}{0.5pt}
\renewcommand{\arraystretch}{1.04}
\begin{tabular*}{\textwidth}{@{\extracolsep{\fill}}ll*{15}{c}@{}}
\toprule
\multirow{2}{*}{SFT steps} & \multirow{2}{*}{Evaluation}
& \multicolumn{3}{c}{AIME 2024} & \multicolumn{3}{c}{AIME 2025} & \multicolumn{3}{c}{AIME 2026} & \multicolumn{3}{c}{HMMT25} & \multicolumn{3}{c}{Dataset Avg.} \\
\cmidrule(lr){3-5}\cmidrule(lr){6-8}\cmidrule(lr){9-11}\cmidrule(lr){12-14}\cmidrule(lr){15-17}
& & Avg@8 & Pass@8 & Length & Avg@8 & Pass@8 & Length & Avg@8 & Pass@8 & Length & Avg@8 & Pass@8 & Length & Avg@8 & Pass@8 & Length \\
\midrule
Base & Direct & 12.1 & 26.7 & 3692 & 7.5 & 26.7 & 2707 & 8.3 & 20.0 & 2889 & 1.2 & 6.7 & 2863 & 7.3 & 20.0 & 3038 \\
5k & Direct & 30.4 & 66.7 & 16251 & 30.0 & 53.3 & 16427 & 25.4 & 50.0 & 16250 & 16.7 & 33.3 & 18485 & 25.6 & 50.8 & 16853 \\
7.5k & Direct & 35.0 & 63.3 & 14817 & 33.8 & 56.7 & 14477 & 31.2 & 56.7 & 15043 & 20.0 & 33.3 & 15893 & 30.0 & 52.5 & 15057 \\
10k & Direct & 37.1 & 70.0 & 13999 & 34.6 & 56.7 & 13907 & 36.7 & 66.7 & 14128 & 20.8 & 50.0 & 14470 & 32.3 & 60.8 & 14126 \\
12.5k & Direct & 42.5 & 73.3 & 13632 & 32.5 & 56.7 & 14781 & 33.3 & 66.7 & 14865 & 20.0 & 33.3 & 15521 & 32.1 & 57.5 & 14700 \\
15k & Direct & 45.8 & 73.3 & 13843 & 34.2 & 63.3 & 13968 & 36.2 & 60.0 & 13625 & 22.5 & 43.3 & 15708 & 34.7 & 60.0 & 14286 \\
\midrule
Base & After OPSD & 7.9 & 26.7 & 9927 & 4.6 & 23.3 & 10265 & 4.2 & 10.0 & 10319 & 2.5 & 10.0 & 10718 & 4.8 & 17.5 & 10308 \\
5k & After OPSD & 0.4 & 3.3 & 33234 & 1.7 & 6.7 & 32905 & 1.2 & 6.7 & 33055 & 0.0 & 0.0 & 33277 & 0.8 & 4.2 & 33118 \\
10k & After OPSD & 1.2 & 6.7 & 33153 & 2.9 & 13.3 & 32525 & 1.7 & 10.0 & 32934 & 0.8 & 3.3 & 32766 & 1.7 & 8.3 & 32844 \\
15k & After OPSD & 2.1 & 13.3 & 33095 & 3.3 & 10.0 & 32466 & 2.5 & 10.0 & 32770 & 0.0 & 0.0 & 33220 & 2.0 & 8.3 & 32888 \\
\bottomrule
\end{tabular*}
}
\end{table}

The table expands the first six rows of \autoref{tab:sft_grpo_control}, from Qwen3-4B-Base through SFT+GRPO+OPSD.

\begin{table}[H]
\caption{Per-dataset post-training pipeline results for the first six rows of \autoref{tab:sft_grpo_control}. \autoref{tab:sft_grpo_opsd_per_dataset} reports the corresponding Qwen3-1.7B pipeline and the matched Qwen3-4B control. Fmt denotes Format.}
\label{tab:sft_grpo_per_dataset}
\centering
{\fontsize{6.0pt}{6.7pt}\selectfont
\setlength{\tabcolsep}{0.45pt}
\renewcommand{\arraystretch}{1.02}
\begin{tabular*}{\textwidth}{@{\extracolsep{\fill}}l*{16}{c}@{}}
\toprule
\multirow{2}{*}{Init.\ / method} & \multicolumn{4}{c}{AIME 2024} & \multicolumn{4}{c}{AIME 2025} & \multicolumn{4}{c}{AIME 2026} & \multicolumn{4}{c}{HMMT25} \\
\cmidrule(lr){2-5}\cmidrule(lr){6-9}\cmidrule(lr){10-13}\cmidrule(lr){14-17}
& Avg@8 & Pass@8 & Fmt & Len & Avg@8 & Pass@8 & Fmt & Len & Avg@8 & Pass@8 & Fmt & Len & Avg@8 & Pass@8 & Fmt & Len \\
\midrule
Qwen3-4B-Base (direct) & 12.1 & 26.7 & 86.2 & 3692 & 7.5 & 26.7 & 89.2 & 2707 & 8.3 & 20.0 & 86.7 & 2889 & 1.2 & 6.7 & 86.7 & 2863 \\
+ 100-step OPSD & 7.9 & 26.7 & 74.2 & 9927 & 4.6 & 23.3 & 76.2 & 10265 & 4.2 & 10.0 & 73.8 & 10319 & 2.5 & 10.0 & 74.2 & 10718 \\
\midrule
SFT-15k (direct) & 45.8 & 73.3 & 89.6 & 13843 & 34.2 & 63.3 & 90.4 & 13968 & 36.2 & 60.0 & 88.8 & 13625 & 22.5 & 43.3 & 88.3 & 15708 \\
+ 100-step OPSD & 2.1 & 13.3 & 3.3 & 33095 & 3.3 & 10.0 & 5.0 & 32466 & 2.5 & 10.0 & 3.3 & 32770 & 0.0 & 0.0 & 2.1 & 33220 \\
+ GRPO & 43.8 & 73.3 & 93.3 & 14430 & 34.6 & 56.7 & 96.2 & 13451 & 35.8 & 70.0 & 93.3 & 13860 & 20.4 & 43.3 & 96.2 & 14990 \\
+ GRPO + OPSD & 3.3 & 10.0 & 7.1 & 37973 & 5.0 & 13.3 & 5.4 & 37043 & 3.3 & 20.0 & 5.4 & 37485 & 2.5 & 13.3 & 4.6 & 37812 \\
\bottomrule
\end{tabular*}
}
\end{table}

The coding breakdown below expands all Qwen3-1.7B and Qwen3-4B rows in \autoref{tab:sft_grpo_control}.

\begin{table}[H]
\caption{Per-dataset coding post-training pipeline results for \autoref{tab:sft_grpo_control}.}
\label{tab:sft_grpo_coding_per_dataset}
\centering
{\fontsize{6.0pt}{6.7pt}\selectfont
\setlength{\tabcolsep}{1.0pt}
\renewcommand{\arraystretch}{1.02}
\begin{tabular*}{\textwidth}{@{\extracolsep{\fill}}l*{12}{c}@{}}
\toprule
\multirow{2}{*}{Model / stage}
& \multicolumn{3}{c}{HumanEvalPlus} & \multicolumn{3}{c}{MBPPPlus} & \multicolumn{3}{c}{LCB v6} & \multicolumn{3}{c}{Dataset Avg.} \\
\cmidrule(lr){2-4}\cmidrule(lr){5-7}\cmidrule(lr){8-10}\cmidrule(lr){11-13}
& Avg@4 & Pass@4 & Len & Avg@4 & Pass@4 & Len & Avg@4 & Pass@4 & Len & Avg@4 & Pass@4 & Len \\
\midrule
\multicolumn{13}{@{}l}{\textit{Qwen3-1.7B}} \\
Qwen3-1.7B-Base & 42.7 & 74.4 & 3069 & 40.6 & 68.8 & 5794 & 5.6 & 16.0 & 3705 & 29.6 & 53.1 & 4189 \\
$+$ OPSD & 37.5 & 68.3 & 4003 & 34.6 & 64.0 & 6549 & 4.4 & 13.7 & 6248 & 25.5 & 48.7 & 5600 \\
$+$ SFT & 54.9 & 78.0 & 5786 & 53.5 & 70.4 & 4374 & 17.0 & 22.9 & 19000 & 41.8 & 57.1 & 9720 \\
$+$ SFT $+$ OPSD & 30.8 & 57.3 & 23020 & 32.8 & 57.1 & 22570 & 9.3 & 16.0 & 32341 & 24.3 & 43.5 & 25977 \\
$+$ SFT $+$ GRPO & 58.2 & 80.5 & 4340 & 37.1 & 54.5 & 3438 & 19.3 & 27.4 & 14822 & 38.2 & 54.1 & 7533 \\
$+$ SFT $+$ GRPO $+$ OPSD & 30.2 & 61.0 & 23005 & 23.9 & 46.0 & 19647 & 12.3 & 22.3 & 26504 & 22.1 & 43.1 & 23052 \\
\midrule
Qwen3-1.7B & 77.9 & 86.6 & 3886 & 67.6 & 77.0 & 3191 & 31.0 & 38.3 & 14924 & 58.8 & 67.3 & 7334 \\
$+$ OPSD & 77.4 & 86.6 & 8594 & 68.0 & 77.8 & 6656 & 30.3 & 37.7 & 25075 & 58.6 & 67.4 & 13442 \\
\midrule
\multicolumn{13}{@{}l}{\textit{Qwen3-4B}} \\
Qwen3-4B-Base & 60.8 & 84.8 & 3350 & 52.4 & 74.6 & 5110 & 17.1 & 22.3 & 1670 & 43.5 & 60.5 & 3377 \\
$+$ OPSD & 47.6 & 81.1 & 7449 & 42.9 & 73.8 & 8643 & 11.3 & 21.7 & 8145 & 33.9 & 58.9 & 8079 \\
$+$ SFT & 81.3 & 91.5 & 2870 & 75.7 & 83.3 & 2281 & 35.6 & 45.7 & 13870 & 64.2 & 73.5 & 6340 \\
$+$ SFT $+$ OPSD & 79.4 & 89.6 & 6242 & 71.6 & 80.7 & 5714 & 32.7 & 40.6 & 19579 & 61.3 & 70.3 & 10512 \\
$+$ SFT $+$ GRPO & 85.2 & 92.7 & 2833 & 64.5 & 78.6 & 2187 & 36.9 & 45.7 & 13691 & 62.2 & 72.3 & 6237 \\
$+$ SFT $+$ GRPO $+$ OPSD & 80.5 & 93.3 & 6017 & 61.7 & 75.4 & 5449 & 33.4 & 43.4 & 20150 & 58.5 & 70.7 & 10538 \\
\midrule
Qwen3-4B & 87.0 & 92.1 & 3723 & 76.5 & 80.7 & 2698 & 48.3 & 57.7 & 14614 & 70.6 & 76.8 & 7012 \\
$+$ OPSD & 86.0 & 92.1 & 4831 & 76.7 & 83.6 & 3794 & 44.9 & 57.1 & 21610 & 69.2 & 77.6 & 10078 \\
\bottomrule
\end{tabular*}
}
\end{table}

The 50-step OPSD and GRPO runs reported below start from Qwen3-1.7B, Qwen3-4B, and Qwen3-8B \thmode{} checkpoints and consume exactly the same 3200 OpenMathReasoning problems in the same order. \autoref{tab:mode_per_dataset} gives the Base rows.

\begin{table}[H]
\caption{Per-dataset 50-step GRPO results on Qwen3 thinking checkpoints for \autoref{tab:opsd_grpo_compute}.}
\label{tab:released_grpo_per_dataset}
\centering
{\fontsize{6.0pt}{6.7pt}\selectfont
\setlength{\tabcolsep}{0.45pt}
\renewcommand{\arraystretch}{1.02}
\begin{tabular*}{\textwidth}{@{\extracolsep{\fill}}l*{16}{c}@{}}
\toprule
\multirow{2}{*}{Checkpoint} & \multicolumn{4}{c}{AIME 2024} & \multicolumn{4}{c}{AIME 2025} & \multicolumn{4}{c}{AIME 2026} & \multicolumn{4}{c}{HMMT25} \\
\cmidrule(lr){2-5}\cmidrule(lr){6-9}\cmidrule(lr){10-13}\cmidrule(lr){14-17}
& Avg@8 & Pass@8 & Fmt & Len & Avg@8 & Pass@8 & Fmt & Len & Avg@8 & Pass@8 & Fmt & Len & Avg@8 & Pass@8 & Fmt & Len \\
\midrule
Qwen3-1.7B + GRPO-50 & 47.1 & 76.7 & 95.8 & 19761 & 37.1 & 63.3 & 97.5 & 19985 & 38.8 & 66.7 & 97.1 & 20334 & 25.4 & 46.7 & 95.8 & 21751 \\
Qwen3-4B + GRPO-50   & 72.5 & 83.3 & 97.5 & 16342 & 67.9 & 86.7 & 92.9 & 19731 & 67.1 & 83.3 & 92.5 & 18507 & 43.8 & 60.0 & 91.3 & 22044 \\
Qwen3-8B + GRPO-50   & 74.2 & 83.3 & 98.8 & 14857 & 68.3 & 86.7 & 93.8 & 17667 & 67.5 & 90.0 & 93.8 & 16346 & 43.8 & 70.0 & 95.8 & 20193 \\
\bottomrule
\end{tabular*}
}
\end{table}

\begin{table}[H]
\caption{Per-dataset 50-step OPSD results on the same OpenMathReasoning examples and in the same order as GRPO, corresponding to \autoref{tab:opsd_grpo_compute}. OPSD uses the post-thinking solution as privileged teacher context.}
\label{tab:omr_opsd_per_dataset}
\centering
{\fontsize{6.0pt}{6.7pt}\selectfont
\setlength{\tabcolsep}{0.45pt}
\renewcommand{\arraystretch}{1.02}
\begin{tabular*}{\textwidth}{@{\extracolsep{\fill}}l*{16}{c}@{}}
\toprule
\multirow{2}{*}{Checkpoint} & \multicolumn{4}{c}{AIME 2024} & \multicolumn{4}{c}{AIME 2025} & \multicolumn{4}{c}{AIME 2026} & \multicolumn{4}{c}{HMMT25} \\
\cmidrule(lr){2-5}\cmidrule(lr){6-9}\cmidrule(lr){10-13}\cmidrule(lr){14-17}
& Avg@8 & Pass@8 & Fmt & Len & Avg@8 & Pass@8 & Fmt & Len & Avg@8 & Pass@8 & Fmt & Len & Avg@8 & Pass@8 & Fmt & Len \\
\midrule
Qwen3-1.7B + OPSD-50 & 48.3 & 76.7 & 87.9 & 21217 & 40.4 & 66.7 & 93.3 & 21169 & 40.0 & 63.3 & 88.3 & 21876 & 23.3 & 46.7 & 91.3 & 23860 \\
Qwen3-4B + OPSD-50   & 70.4 & 83.3 & 91.7 & 17829 & 63.8 & 86.7 & 88.3 & 20209 & 61.3 & 83.3 & 88.8 & 18498 & 45.0 & 66.7 & 89.2 & 22175 \\
Qwen3-8B + OPSD-50   & 72.9 & 86.7 & 93.8 & 17205 & 66.2 & 80.0 & 88.8 & 20340 & 62.9 & 90.0 & 90.4 & 18798 & 41.2 & 60.0 & 86.7 & 22985 \\
\bottomrule
\end{tabular*}
}
\end{table}

\begin{table}[H]
\caption{Per-dataset 50-step coding results for \autoref{tab:opsd_grpo_compute}.}
\label{tab:apps_opsd_grpo_per_dataset}
\centering
{\fontsize{6.0pt}{6.7pt}\selectfont
\setlength{\tabcolsep}{1.0pt}
\renewcommand{\arraystretch}{1.02}
\begin{tabular*}{\textwidth}{@{\extracolsep{\fill}}l*{12}{c}@{}}
\toprule
\multirow{2}{*}{Checkpoint}
& \multicolumn{3}{c}{HumanEvalPlus} & \multicolumn{3}{c}{MBPPPlus} & \multicolumn{3}{c}{LCB v6} & \multicolumn{3}{c}{Dataset Avg.} \\
\cmidrule(lr){2-4}\cmidrule(lr){5-7}\cmidrule(lr){8-10}\cmidrule(lr){11-13}
& Avg@4 & Pass@4 & Len & Avg@4 & Pass@4 & Len & Avg@4 & Pass@4 & Len & Avg@4 & Pass@4 & Len \\
\midrule
Qwen3-1.7B Base      & 77.9 & 86.6 & 3886 & 67.6 & 77.0 & 3191 & 31.0 & 38.3 & 14924 & 58.8 & 67.3 & 7334 \\
Qwen3-1.7B + OPSD-50 & 76.2 & 83.5 & 9301 & 67.3 & 76.5 & 6986 & 29.0 & 36.0 & 26505 & 57.5 & 65.3 & 14264 \\
Qwen3-1.7B + GRPO-50 & 78.4 & 89.0 & 3954 & 67.4 & 77.2 & 2927 & 30.6 & 37.1 & 14848 & 58.8 & 67.8 & 7243 \\
\midrule
Qwen3-4B Base      & 87.0 & 92.1 & 3723 & 76.5 & 80.7 & 2698 & 48.3 & 57.7 & 14614 & 70.6 & 76.8 & 7012 \\
Qwen3-4B + OPSD-50 & 86.7 & 92.7 & 5689 & 77.3 & 82.5 & 4486 & 45.3 & 54.9 & 22025 & 69.8 & 76.7 & 10734 \\
Qwen3-4B + GRPO-50 & 89.3 & 93.9 & 3484 & 77.0 & 81.7 & 2731 & 48.0 & 56.6 & 14723 & 71.4 & 77.4 & 6979 \\
\midrule
Qwen3-8B Base      & 86.1 & 92.1 & 3936 & 78.0 & 84.4 & 2932 & 49.9 & 59.4 & 14373 & 71.3 & 78.6 & 7080 \\
Qwen3-8B + OPSD-50 & 86.7 & 92.7 & 4193 & 78.9 & 84.9 & 3237 & 50.1 & 58.9 & 17272 & 71.9 & 78.8 & 8234 \\
Qwen3-8B + GRPO-50 & 85.7 & 90.2 & 3999 & 77.4 & 83.6 & 2958 & 48.4 & 56.0 & 14366 & 70.5 & 76.6 & 7108 \\
\bottomrule
\end{tabular*}
}
\end{table}

\subsubsection{Post-Training Pipeline Details}
\label{app:sft_grpo_opsd_per_dataset}

The first block expands the Qwen3-1.7B pipeline in \autoref{tab:sft_grpo_then_opsd}. SFT is the 15k-step checkpoint. GRPO and OPSD are 100-step continuations. The final two rows retain the matched Qwen3-4B \thmode/\thmode{} control from the comparison.

\begin{table}[H]
\caption{Per-dataset Qwen3-1.7B post-training pipeline results for \autoref{tab:sft_grpo_then_opsd}, followed by the matched Qwen3-4B control.}
\label{tab:sft_grpo_opsd_per_dataset}
\centering
{\fontsize{6.0pt}{6.7pt}\selectfont
\setlength{\tabcolsep}{0.45pt}
\renewcommand{\arraystretch}{1.02}
\begin{tabular*}{\textwidth}{@{\extracolsep{\fill}}l*{16}{c}@{}}
\toprule
\multirow{2}{*}{Model / stage}
& \multicolumn{4}{c}{AIME 2024} & \multicolumn{4}{c}{AIME 2025} & \multicolumn{4}{c}{AIME 2026} & \multicolumn{4}{c}{HMMT25} \\
\cmidrule(lr){2-5}\cmidrule(lr){6-9}\cmidrule(lr){10-13}\cmidrule(lr){14-17}
& Avg@8 & Pass@8 & Fmt & Len & Avg@8 & Pass@8 & Fmt & Len & Avg@8 & Pass@8 & Fmt & Len & Avg@8 & Pass@8 & Fmt & Len \\
\midrule
Qwen3-1.7B-Base & 3.8 & 13.3 & 77.5 & 6629 & 1.2 & 10.0 & 81.3 & 5214 & 3.3 & 10.0 & 80.0 & 5654 & 0.0 & 0.0 & 83.8 & 4805 \\
$+$ OPSD & 2.9 & 16.7 & 62.9 & 8339 & 0.4 & 3.3 & 67.1 & 7601 & 1.2 & 10.0 & 59.6 & 8881 & 0.0 & 0.0 & 63.3 & 6987 \\
$+$ SFT & 12.9 & 36.7 & 61.3 & 21830 & 15.8 & 36.7 & 65.8 & 20261 & 12.1 & 30.0 & 58.3 & 22976 & 5.0 & 13.3 & 68.8 & 20620 \\
$+$ SFT $+$ OPSD & 2.5 & 13.3 & 5.8 & 35896 & 1.7 & 10.0 & 3.3 & 36323 & 2.1 & 6.7 & 4.2 & 35962 & 0.0 & 0.0 & 2.1 & 36550 \\
$+$ SFT $+$ GRPO & 14.2 & 36.7 & 85.4 & 16850 & 16.7 & 36.7 & 90.8 & 15476 & 11.3 & 26.7 & 84.2 & 17573 & 5.4 & 20.0 & 90.4 & 16523 \\
$+$ SFT $+$ GRPO $+$ OPSD & 8.3 & 16.7 & 21.7 & 33812 & 4.6 & 23.3 & 11.3 & 35917 & 5.4 & 13.3 & 12.9 & 35610 & 0.4 & 3.3 & 10.4 & 36757 \\
\midrule
Qwen3-4B (\thmode/\thmode{}) & 74.6 & 90.0 & 98.8 & 15495 & 65.4 & 80.0 & 96.2 & 18696 & 65.8 & 80.0 & 95.0 & 17249 & 43.3 & 66.7 & 98.3 & 19636 \\
$+$ OPSD & 66.3 & 80.0 & 86.7 & 18395 & 62.9 & 80.0 & 86.2 & 19655 & 62.9 & 83.3 & 89.6 & 18119 & 39.6 & 60.0 & 82.9 & 23625 \\
\bottomrule
\end{tabular*}
}
\end{table}

\subsection{Seed Sensitivity Breakdowns}

\subsubsection{Seed Sensitivity: Base Checkpoints (Evaluation Seed)}
\label{app:seed_base_eval_per_dataset}

\begin{table}[H]
\caption{Per-dataset Base results for \autoref{tab:seed_base_eval}.}
\label{tab:seed_base_eval_per_dataset}
\centering
{\fontsize{6.2pt}{7.0pt}\selectfont
\renewcommand{\arraystretch}{1.08}
\setlength{\tabcolsep}{0.8pt}
\begin{tabular*}{\textwidth}{@{\extracolsep{\fill}}@{}ll*{15}{c}@{}}
\toprule
\multirow{2}{*}{Model} & \multirow{2}{*}{Eval seed} & \multicolumn{3}{c}{AIME 2024} & \multicolumn{3}{c}{AIME 2025} & \multicolumn{3}{c}{AIME 2026} & \multicolumn{3}{c}{HMMT25} & \multicolumn{3}{c}{Dataset Avg.} \\
\cmidrule(lr){3-5}\cmidrule(lr){6-8}\cmidrule(lr){9-11}\cmidrule(lr){12-14}\cmidrule(lr){15-17}
& & Avg@8 & Pass@8 & Length & Avg@8 & Pass@8 & Length & Avg@8 & Pass@8 & Length & Avg@8 & Pass@8 & Length & Avg@8 & Pass@8 & Length \\
\midrule
\ml{Qwen3-\\1.7B} & 42 & 43.8 & 76.7 & 18647 & 35.4 & 60.0 & 18266 & 35.0 & 53.3 & 18161 & 21.7 & 46.7 & 19421 & 34.0 & 59.2 & 18624 \\
 & 1024 & 47.5 & 76.7 & 17907 & 36.7 & 53.3 & 18615 & 35.8 & 56.7 & 17919 & 20.4 & 33.3 & 19812 & 35.1 & 55.0 & 18563 \\
 & 65536 & 42.9 & 73.3 & 18102 & 37.9 & 56.7 & 18542 & 33.3 & 66.7 & 18614 & 27.1 & 50.0 & 19196 & 35.3 & 61.7 & 18613 \\
\midrule
\ml{OLMo-3-\\7B-Think} & 42 & 71.7 & 86.7 & 17296 & 66.7 & 80.0 & 19137 & 71.2 & 90.0 & 17560 & 44.6 & 70.0 & 20950 & 63.5 & 81.7 & 18736 \\
 & 1024 & 67.9 & 83.3 & 17617 & 64.2 & 86.7 & 19148 & 72.1 & 86.7 & 17955 & 44.6 & 66.7 & 21317 & 62.2 & 80.8 & 19009 \\
 & 65536 & 70.8 & 86.7 & 17101 & 65.8 & 76.7 & 19099 & 70.0 & 86.7 & 17775 & 45.8 & 66.7 & 21497 & 63.1 & 79.2 & 18868 \\
\bottomrule
\end{tabular*}
}
\end{table}

\subsubsection{Seed Sensitivity: Rollout Length (Training Seed)}
\label{app:seed_length_train_per_dataset}

\begin{table}[H]
\caption{Per-dataset training seed results for \autoref{tab:seed_length_train}.}
\label{tab:seed_length_train_per_dataset}
\centering
{\fontsize{6.2pt}{7.0pt}\selectfont
\renewcommand{\arraystretch}{1.08}
\setlength{\tabcolsep}{0.8pt}
\begin{tabular*}{\textwidth}{@{\extracolsep{\fill}}@{}lll*{15}{c}@{}}
\toprule
\multirow{2}{*}{Model} & \multirow{2}{*}{Condition} & \multirow{2}{*}{Train seed} & \multicolumn{3}{c}{AIME 2024} & \multicolumn{3}{c}{AIME 2025} & \multicolumn{3}{c}{AIME 2026} & \multicolumn{3}{c}{HMMT25} & \multicolumn{3}{c}{Dataset Avg.} \\
\cmidrule(lr){4-6}\cmidrule(lr){7-9}\cmidrule(lr){10-12}\cmidrule(lr){13-15}\cmidrule(lr){16-18}
& & & Avg@8 & Pass@8 & Length & Avg@8 & Pass@8 & Length & Avg@8 & Pass@8 & Length & Avg@8 & Pass@8 & Length & Avg@8 & Pass@8 & Length \\
\midrule
\ml{Qwen3-\\1.7B} & OPSD-256 & 42 & 46.2$^{\uparrow}$ & 70.0$^{\downarrow}$ & 20783 & 36.7$^{\uparrow}$ & 66.7$^{\uparrow}$ & 21697 & 45.0$^{\uparrow}$ & 73.3$^{\uparrow}$ & 20995 & 24.6$^{\uparrow}$ & 40.0$^{\downarrow}$ & 22306 & 38.1$^{\uparrow}$ & 62.5$^{\uparrow}$ & 21445 \\
 &  & 1024 & 51.2$^{\uparrow}$ & 83.3$^{\uparrow}$ & 20253 & 36.2$^{\uparrow}$ & 63.3$^{\uparrow}$ & 21118 & 42.1$^{\uparrow}$ & 66.7$^{\uparrow}$ & 21295 & 27.1$^{\uparrow}$ & 46.7 & 21828 & 39.2$^{\uparrow}$ & 65.0$^{\uparrow}$ & 21124 \\
 &  & 65536 & 50.4$^{\uparrow}$ & 76.7 & 20253 & 37.1$^{\uparrow}$ & 63.3$^{\uparrow}$ & 20962 & 38.3$^{\uparrow}$ & 70.0$^{\uparrow}$ & 21160 & 22.5$^{\uparrow}$ & 50.0$^{\uparrow}$ & 22975 & 37.1$^{\uparrow}$ & 65.0$^{\uparrow}$ & 21338 \\
\midrule
\ml{Qwen3-\\1.7B} & OPSD-1024 & 42 & 45.8$^{\uparrow}$ & 73.3$^{\downarrow}$ & 22175 & 36.7$^{\uparrow}$ & 66.7$^{\uparrow}$ & 22143 & 36.7$^{\uparrow}$ & 63.3$^{\uparrow}$ & 21288 & 22.1$^{\uparrow}$ & 40.0$^{\downarrow}$ & 23913 & 35.3$^{\uparrow}$ & 60.8$^{\uparrow}$ & 22380 \\
 &  & 1024 & 49.2$^{\uparrow}$ & 73.3$^{\downarrow}$ & 21091 & 36.7$^{\uparrow}$ & 63.3$^{\uparrow}$ & 21896 & 37.5$^{\uparrow}$ & 60.0$^{\uparrow}$ & 21149 & 20.4$^{\downarrow}$ & 30.0$^{\downarrow}$ & 23892 & 35.9$^{\uparrow}$ & 56.7$^{\downarrow}$ & 22007 \\
\bottomrule
\end{tabular*}
}
\end{table}

\subsubsection{Seed Sensitivity: Rollout Length (Evaluation Seed)}
\label{app:seed_length_eval_per_dataset}

\begin{table}[H]
\caption{Per-dataset evaluation seed results for \autoref{tab:seed_length_eval}.}
\label{tab:seed_length_eval_per_dataset}
\centering
{\fontsize{6.2pt}{7.0pt}\selectfont
\renewcommand{\arraystretch}{1.08}
\setlength{\tabcolsep}{0.8pt}
\begin{tabular*}{\textwidth}{@{\extracolsep{\fill}}@{}lll*{15}{c}@{}}
\toprule
\multirow{2}{*}{Model} & \multirow{2}{*}{Condition} & \multirow{2}{*}{Eval seed} & \multicolumn{3}{c}{AIME 2024} & \multicolumn{3}{c}{AIME 2025} & \multicolumn{3}{c}{AIME 2026} & \multicolumn{3}{c}{HMMT25} & \multicolumn{3}{c}{Dataset Avg.} \\
\cmidrule(lr){4-6}\cmidrule(lr){7-9}\cmidrule(lr){10-12}\cmidrule(lr){13-15}\cmidrule(lr){16-18}
& & & Avg@8 & Pass@8 & Length & Avg@8 & Pass@8 & Length & Avg@8 & Pass@8 & Length & Avg@8 & Pass@8 & Length & Avg@8 & Pass@8 & Length \\
\midrule
\ml{Qwen3-\\1.7B} & OPSD-256 & 42 & 46.2$^{\uparrow}$ & 70.0$^{\downarrow}$ & 20783 & 36.7$^{\uparrow}$ & 66.7$^{\uparrow}$ & 21697 & 45.0$^{\uparrow}$ & 73.3$^{\uparrow}$ & 20995 & 24.6$^{\uparrow}$ & 40.0$^{\downarrow}$ & 22306 & 38.1$^{\uparrow}$ & 62.5$^{\uparrow}$ & 21445 \\
 &  & 1024 & 47.9$^{\uparrow}$ & 70.0$^{\downarrow}$ & 20152 & 36.7 & 60.0$^{\uparrow}$ & 21390 & 37.9$^{\uparrow}$ & 56.7 & 20678 & 25.4$^{\uparrow}$ & 50.0$^{\uparrow}$ & 21863 & 37.0$^{\uparrow}$ & 59.2$^{\uparrow}$ & 21021 \\
 &  & 65536 & 47.5$^{\uparrow}$ & 73.3 & 20297 & 34.2$^{\downarrow}$ & 63.3$^{\uparrow}$ & 21394 & 41.7$^{\uparrow}$ & 66.7 & 21027 & 23.3$^{\downarrow}$ & 46.7$^{\downarrow}$ & 22365 & 36.7$^{\uparrow}$ & 62.5$^{\uparrow}$ & 21271 \\
\midrule
\ml{Qwen3-\\1.7B} & OPSD-1024 & 42 & 45.8$^{\uparrow}$ & 73.3$^{\downarrow}$ & 22175 & 36.7$^{\uparrow}$ & 66.7$^{\uparrow}$ & 22143 & 36.7$^{\uparrow}$ & 63.3$^{\uparrow}$ & 21288 & 22.1$^{\uparrow}$ & 40.0$^{\downarrow}$ & 23913 & 35.3$^{\uparrow}$ & 60.8$^{\uparrow}$ & 22380 \\
 &  & 1024 & 49.2$^{\uparrow}$ & 76.7 & 21279 & 34.2$^{\downarrow}$ & 63.3$^{\uparrow}$ & 21384 & 38.8$^{\uparrow}$ & 66.7$^{\uparrow}$ & 21373 & 25.0$^{\uparrow}$ & 53.3$^{\uparrow}$ & 23458 & 36.8$^{\uparrow}$ & 65.0$^{\uparrow}$ & 21874 \\
\bottomrule
\end{tabular*}
}
\end{table}

\subsubsection{Seed Sensitivity: Loss Token Position (Training Seed)}
\label{app:seed_position_train_per_dataset}

\begin{table}[H]
\caption{Per-dataset training seed results for \autoref{tab:seed_position_train}. First-256 reuses the corresponding OPSD-256 run.}
\label{tab:seed_position_train_per_dataset}
\centering
{\fontsize{6.2pt}{7.0pt}\selectfont
\renewcommand{\arraystretch}{1.08}
\setlength{\tabcolsep}{0.8pt}
\begin{tabular*}{\textwidth}{@{\extracolsep{\fill}}@{}lll*{15}{c}@{}}
\toprule
\multirow{2}{*}{Model} & \multirow{2}{*}{Condition} & \multirow{2}{*}{Train seed} & \multicolumn{3}{c}{AIME 2024} & \multicolumn{3}{c}{AIME 2025} & \multicolumn{3}{c}{AIME 2026} & \multicolumn{3}{c}{HMMT25} & \multicolumn{3}{c}{Dataset Avg.} \\
\cmidrule(lr){4-6}\cmidrule(lr){7-9}\cmidrule(lr){10-12}\cmidrule(lr){13-15}\cmidrule(lr){16-18}
& & & Avg@8 & Pass@8 & Length & Avg@8 & Pass@8 & Length & Avg@8 & Pass@8 & Length & Avg@8 & Pass@8 & Length & Avg@8 & Pass@8 & Length \\
\midrule
\ml{Qwen3-\\1.7B} & \shortstack{First-256 /\\OPSD-256} & 42 & 46.2$^{\uparrow}$ & 70.0$^{\downarrow}$ & 20783 & 36.7$^{\uparrow}$ & 66.7$^{\uparrow}$ & 21697 & 45.0$^{\uparrow}$ & 73.3$^{\uparrow}$ & 20995 & 24.6$^{\uparrow}$ & 40.0$^{\downarrow}$ & 22306 & 38.1$^{\uparrow}$ & 62.5$^{\uparrow}$ & 21445 \\
 &  & 1024 & 51.2$^{\uparrow}$ & 83.3$^{\uparrow}$ & 20253 & 36.2$^{\uparrow}$ & 63.3$^{\uparrow}$ & 21118 & 42.1$^{\uparrow}$ & 66.7$^{\uparrow}$ & 21295 & 27.1$^{\uparrow}$ & 46.7 & 21828 & 39.2$^{\uparrow}$ & 65.0$^{\uparrow}$ & 21124 \\
 &  & 65536 & 50.4$^{\uparrow}$ & 76.7 & 20253 & 37.1$^{\uparrow}$ & 63.3$^{\uparrow}$ & 20962 & 38.3$^{\uparrow}$ & 70.0$^{\uparrow}$ & 21160 & 22.5$^{\uparrow}$ & 50.0$^{\uparrow}$ & 22975 & 37.1$^{\uparrow}$ & 65.0$^{\uparrow}$ & 21338 \\
\midrule
\ml{Qwen3-\\1.7B} & Uni-256 & 42 & 48.8$^{\uparrow}$ & 76.7 & 21874 & 31.7$^{\downarrow}$ & 60.0 & 21873 & 36.7$^{\uparrow}$ & 63.3$^{\uparrow}$ & 22032 & 23.8$^{\uparrow}$ & 46.7 & 24462 & 35.2$^{\uparrow}$ & 61.7$^{\uparrow}$ & 22560 \\
 &  & 1024 & 47.5$^{\uparrow}$ & 73.3$^{\downarrow}$ & 21809 & 34.2$^{\downarrow}$ & 63.3$^{\uparrow}$ & 22245 & 34.6$^{\downarrow}$ & 60.0$^{\uparrow}$ & 22053 & 23.3$^{\uparrow}$ & 40.0$^{\downarrow}$ & 23502 & 34.9$^{\uparrow}$ & 59.2 & 22402 \\
\bottomrule
\end{tabular*}
}
\end{table}

\subsubsection{Seed Sensitivity: Loss Token Position (Evaluation Seed)}
\label{app:seed_position_eval_per_dataset}

\begin{table}[H]
\caption{Per-dataset evaluation seed results for \autoref{tab:seed_position_eval}. First-256 reuses OPSD-256.}
\label{tab:seed_position_eval_per_dataset}
\centering
{\fontsize{6.2pt}{7.0pt}\selectfont
\renewcommand{\arraystretch}{1.08}
\setlength{\tabcolsep}{0.8pt}
\begin{tabular*}{\textwidth}{@{\extracolsep{\fill}}@{}lll*{15}{c}@{}}
\toprule
\multirow{2}{*}{Model} & \multirow{2}{*}{Condition} & \multirow{2}{*}{Eval seed} & \multicolumn{3}{c}{AIME 2024} & \multicolumn{3}{c}{AIME 2025} & \multicolumn{3}{c}{AIME 2026} & \multicolumn{3}{c}{HMMT25} & \multicolumn{3}{c}{Dataset Avg.} \\
\cmidrule(lr){4-6}\cmidrule(lr){7-9}\cmidrule(lr){10-12}\cmidrule(lr){13-15}\cmidrule(lr){16-18}
& & & Avg@8 & Pass@8 & Length & Avg@8 & Pass@8 & Length & Avg@8 & Pass@8 & Length & Avg@8 & Pass@8 & Length & Avg@8 & Pass@8 & Length \\
\midrule
\ml{Qwen3-\\1.7B} & \shortstack{First-256 /\\OPSD-256} & 42 & 46.2$^{\uparrow}$ & 70.0$^{\downarrow}$ & 20783 & 36.7$^{\uparrow}$ & 66.7$^{\uparrow}$ & 21697 & 45.0$^{\uparrow}$ & 73.3$^{\uparrow}$ & 20995 & 24.6$^{\uparrow}$ & 40.0$^{\downarrow}$ & 22306 & 38.1$^{\uparrow}$ & 62.5$^{\uparrow}$ & 21445 \\
 &  & 1024 & 47.9$^{\uparrow}$ & 70.0$^{\downarrow}$ & 20152 & 36.7 & 60.0$^{\uparrow}$ & 21390 & 37.9$^{\uparrow}$ & 56.7 & 20678 & 25.4$^{\uparrow}$ & 50.0$^{\uparrow}$ & 21863 & 37.0$^{\uparrow}$ & 59.2$^{\uparrow}$ & 21021 \\
 &  & 65536 & 47.5$^{\uparrow}$ & 73.3 & 20297 & 34.2$^{\downarrow}$ & 63.3$^{\uparrow}$ & 21394 & 41.7$^{\uparrow}$ & 66.7 & 21027 & 23.3$^{\downarrow}$ & 46.7$^{\downarrow}$ & 22365 & 36.7$^{\uparrow}$ & 62.5$^{\uparrow}$ & 21271 \\
\midrule
\ml{Qwen3-\\1.7B} & Uni-256 & 42 & 48.8$^{\uparrow}$ & 76.7 & 21874 & 31.7$^{\downarrow}$ & 60.0 & 21873 & 36.7$^{\uparrow}$ & 63.3$^{\uparrow}$ & 22032 & 23.8$^{\uparrow}$ & 46.7 & 24462 & 35.2$^{\uparrow}$ & 61.7$^{\uparrow}$ & 22560 \\
 &  & 1024 & 47.1$^{\downarrow}$ & 73.3$^{\downarrow}$ & 22073 & 34.6$^{\downarrow}$ & 63.3$^{\uparrow}$ & 21959 & 36.7$^{\uparrow}$ & 60.0$^{\uparrow}$ & 21197 & 23.8$^{\uparrow}$ & 40.0$^{\uparrow}$ & 24054 & 35.5$^{\uparrow}$ & 59.2$^{\uparrow}$ & 22321 \\
\bottomrule
\end{tabular*}
}
\end{table}

\subsubsection{Seed Sensitivity: Teacher-Side Context (Training Seed)}
\label{app:seed_context_train_per_dataset}

\begin{table}[H]
\caption{Per-dataset training seed results for \autoref{tab:seed_context_train}.}
\label{tab:seed_context_train_per_dataset}
\centering
{\fontsize{6.2pt}{7.0pt}\selectfont
\renewcommand{\arraystretch}{1.08}
\setlength{\tabcolsep}{0.8pt}
\begin{tabular*}{\textwidth}{@{\extracolsep{\fill}}@{}lll*{15}{c}@{}}
\toprule
\multirow{2}{*}{Model} & \multirow{2}{*}{Condition} & \multirow{2}{*}{Train seed} & \multicolumn{3}{c}{AIME 2024} & \multicolumn{3}{c}{AIME 2025} & \multicolumn{3}{c}{AIME 2026} & \multicolumn{3}{c}{HMMT25} & \multicolumn{3}{c}{Dataset Avg.} \\
\cmidrule(lr){4-6}\cmidrule(lr){7-9}\cmidrule(lr){10-12}\cmidrule(lr){13-15}\cmidrule(lr){16-18}
& & & Avg@8 & Pass@8 & Length & Avg@8 & Pass@8 & Length & Avg@8 & Pass@8 & Length & Avg@8 & Pass@8 & Length & Avg@8 & Pass@8 & Length \\
\midrule
\ml{Qwen3-\\1.7B} & Solution & 42 & 45.8$^{\uparrow}$ & 73.3$^{\downarrow}$ & 22175 & 36.7$^{\uparrow}$ & 66.7$^{\uparrow}$ & 22143 & 36.7$^{\uparrow}$ & 63.3$^{\uparrow}$ & 21288 & 22.1$^{\uparrow}$ & 40.0$^{\downarrow}$ & 23913 & 35.3$^{\uparrow}$ & 60.8$^{\uparrow}$ & 22380 \\
 &  & 1024 & 49.2$^{\uparrow}$ & 73.3$^{\downarrow}$ & 21091 & 36.7$^{\uparrow}$ & 63.3$^{\uparrow}$ & 21896 & 37.5$^{\uparrow}$ & 60.0$^{\uparrow}$ & 21149 & 20.4$^{\downarrow}$ & 30.0$^{\downarrow}$ & 23892 & 35.9$^{\uparrow}$ & 56.7$^{\downarrow}$ & 22007 \\
 &  & 65536 & 41.7$^{\downarrow}$ & 76.7 & 22332 & 35.8$^{\uparrow}$ & 53.3$^{\downarrow}$ & 21324 & 37.1$^{\uparrow}$ & 70.0$^{\uparrow}$ & 21537 & 23.8$^{\uparrow}$ & 50.0$^{\uparrow}$ & 24818 & 34.6$^{\uparrow}$ & 62.5$^{\uparrow}$ & 22503 \\
\midrule
\ml{Qwen3-\\1.7B} & Answer & 42 & 47.1$^{\uparrow}$ & 76.7 & 25291 & 34.2$^{\downarrow}$ & 56.7$^{\downarrow}$ & 25710 & 39.6$^{\uparrow}$ & 70.0$^{\uparrow}$ & 24298 & 21.7 & 33.3$^{\downarrow}$ & 27985 & 35.6$^{\uparrow}$ & 59.2 & 25821 \\
 &  & 1024 & 44.2$^{\uparrow}$ & 73.3$^{\downarrow}$ & 25539 & 37.1$^{\uparrow}$ & 63.3$^{\uparrow}$ & 25758 & 40.8$^{\uparrow}$ & 70.0$^{\uparrow}$ & 23793 & 24.6$^{\uparrow}$ & 50.0$^{\uparrow}$ & 27365 & 36.7$^{\uparrow}$ & 64.2$^{\uparrow}$ & 25614 \\
\bottomrule
\end{tabular*}
}
\end{table}

\subsubsection{Seed Sensitivity: Teacher-Side Context (Evaluation Seed)}
\label{app:seed_context_eval_per_dataset}

\begin{table}[H]
\caption{Per-dataset evaluation seed results for \autoref{tab:seed_context_eval}.}
\label{tab:seed_context_eval_per_dataset}
\centering
{\fontsize{6.2pt}{7.0pt}\selectfont
\renewcommand{\arraystretch}{1.08}
\setlength{\tabcolsep}{0.8pt}
\begin{tabular*}{\textwidth}{@{\extracolsep{\fill}}@{}lll*{15}{c}@{}}
\toprule
\multirow{2}{*}{Model} & \multirow{2}{*}{Condition} & \multirow{2}{*}{Eval seed} & \multicolumn{3}{c}{AIME 2024} & \multicolumn{3}{c}{AIME 2025} & \multicolumn{3}{c}{AIME 2026} & \multicolumn{3}{c}{HMMT25} & \multicolumn{3}{c}{Dataset Avg.} \\
\cmidrule(lr){4-6}\cmidrule(lr){7-9}\cmidrule(lr){10-12}\cmidrule(lr){13-15}\cmidrule(lr){16-18}
& & & Avg@8 & Pass@8 & Length & Avg@8 & Pass@8 & Length & Avg@8 & Pass@8 & Length & Avg@8 & Pass@8 & Length & Avg@8 & Pass@8 & Length \\
\midrule
\ml{Qwen3-\\1.7B} & Solution & 42 & 45.8$^{\uparrow}$ & 73.3$^{\downarrow}$ & 22175 & 36.7$^{\uparrow}$ & 66.7$^{\uparrow}$ & 22143 & 36.7$^{\uparrow}$ & 63.3$^{\uparrow}$ & 21288 & 22.1$^{\uparrow}$ & 40.0$^{\downarrow}$ & 23913 & 35.3$^{\uparrow}$ & 60.8$^{\uparrow}$ & 22380 \\
 &  & 1024 & 49.2$^{\uparrow}$ & 76.7 & 21279 & 34.2$^{\downarrow}$ & 63.3$^{\uparrow}$ & 21384 & 38.8$^{\uparrow}$ & 66.7$^{\uparrow}$ & 21373 & 25.0$^{\uparrow}$ & 53.3$^{\uparrow}$ & 23458 & 36.8$^{\uparrow}$ & 65.0$^{\uparrow}$ & 21874 \\
 &  & 65536 & 49.6$^{\uparrow}$ & 73.3 & 21563 & 35.0$^{\downarrow}$ & 63.3$^{\uparrow}$ & 21611 & 38.8$^{\uparrow}$ & 66.7 & 21850 & 22.9$^{\downarrow}$ & 43.3$^{\downarrow}$ & 23902 & 36.6$^{\uparrow}$ & 61.7 & 22231 \\
\midrule
\ml{Qwen3-\\1.7B} & Answer & 42 & 47.1$^{\uparrow}$ & 76.7 & 25291 & 34.2$^{\downarrow}$ & 56.7$^{\downarrow}$ & 25710 & 39.6$^{\uparrow}$ & 70.0$^{\uparrow}$ & 24298 & 21.7 & 33.3$^{\downarrow}$ & 27985 & 35.6$^{\uparrow}$ & 59.2 & 25821 \\
 &  & 1024 & 46.7$^{\downarrow}$ & 70.0$^{\downarrow}$ & 25361 & 39.2$^{\uparrow}$ & 66.7$^{\uparrow}$ & 25576 & 37.1$^{\uparrow}$ & 53.3$^{\downarrow}$ & 24341 & 20.4 & 33.3 & 27802 & 35.8$^{\uparrow}$ & 55.8$^{\uparrow}$ & 25770 \\
\bottomrule
\end{tabular*}
}
\end{table}

\subsection{Additional Experiment Breakdowns}

\subsubsection{Training Dataset Comparison}
\label{app:training_dataset_per_dataset}

\begin{table}[H]
\caption{Per-dataset results for the training dataset comparison in \autoref{tab:training_dataset_macro}. All OPSD rows use 100 training steps and a 1024-token rollout cap.}
\label{tab:training_dataset_per_dataset}
\centering
{\fontsize{6.3pt}{7.0pt}\selectfont
\setlength{\tabcolsep}{0.5pt}
\renewcommand{\arraystretch}{1.04}
\begin{tabular*}{\textwidth}{@{\extracolsep{\fill}}ll*{15}{c}@{}}
\toprule
\multirow{2}{*}{Model} & \multirow{2}{*}{Training data}
& \multicolumn{3}{c}{AIME 2024} & \multicolumn{3}{c}{AIME 2025} & \multicolumn{3}{c}{AIME 2026} & \multicolumn{3}{c}{HMMT25} & \multicolumn{3}{c}{Dataset Avg.} \\
\cmidrule(lr){3-5}\cmidrule(lr){6-8}\cmidrule(lr){9-11}\cmidrule(lr){12-14}\cmidrule(lr){15-17}
& & Avg@8 & Pass@8 & Length & Avg@8 & Pass@8 & Length & Avg@8 & Pass@8 & Length & Avg@8 & Pass@8 & Length & Avg@8 & Pass@8 & Length \\
\midrule
\multirow{4}{*}{Qwen3-1.7B} & Base & 43.8 & 76.7 & 18647 & 35.4 & 60.0 & 18266 & 35.0 & 53.3 & 18161 & 21.7 & 46.7 & 19421 & 34.0 & 59.2 & 18624 \\
 & OpenThoughts & 45.8 & 73.3 & 22175 & 36.7 & 66.7 & 22143 & 36.7 & 63.3 & 21288 & 22.1 & 40.0 & 23913 & 35.3 & 60.8 & 22380 \\
 & GSM8K & 51.2 & 83.3 & 18562 & 36.2 & 63.3 & 18628 & 32.9 & 56.7 & 19823 & 20.8 & 40.0 & 21169 & 35.3 & 60.8 & 19545 \\
 & DeepMath & 47.1 & 76.7 & 22711 & 33.8 & 56.7 & 22129 & 35.8 & 60.0 & 22314 & 20.8 & 46.7 & 24558 & 34.4 & 60.0 & 22928 \\
\midrule
\multirow{4}{*}{Qwen3-4B} & Base & 74.6 & 90.0 & 15495 & 65.4 & 80.0 & 18696 & 65.8 & 80.0 & 17249 & 43.3 & 66.7 & 19636 & 62.3 & 79.2 & 17769 \\
 & OpenThoughts & 66.2 & 80.0 & 18395 & 62.9 & 80.0 & 19655 & 62.9 & 83.3 & 18119 & 39.6 & 60.0 & 23625 & 57.9 & 75.8 & 19949 \\
\bottomrule
\end{tabular*}
}
\end{table}

\subsubsection{Teacher Prefix Length Stratification}
\label{app:cotlen_per_dataset}

\begin{table}[H]
\caption{Per-dataset results by teacher prefix length for \autoref{tab:teacher_cotlen}. Full uses the complete training set and a 256-token rollout. Short prefix and Long prefix are defined by gold trace length and contain different problems.}
\label{tab:cotlen_per_dataset}
\centering
{\fontsize{6.3pt}{7.0pt}\selectfont
\setlength{\tabcolsep}{0.5pt}
\renewcommand{\arraystretch}{1.04}
\begin{tabular*}{\textwidth}{@{\extracolsep{\fill}}ll*{15}{c}@{}}
\toprule
\multirow{2}{*}{Model} & \multirow{2}{*}{Subset}
& \multicolumn{3}{c}{AIME 2024} & \multicolumn{3}{c}{AIME 2025} & \multicolumn{3}{c}{AIME 2026} & \multicolumn{3}{c}{HMMT25} & \multicolumn{3}{c}{Dataset Avg.} \\
\cmidrule(lr){3-5}\cmidrule(lr){6-8}\cmidrule(lr){9-11}\cmidrule(lr){12-14}\cmidrule(lr){15-17}
& & Avg@8 & Pass@8 & Length & Avg@8 & Pass@8 & Length & Avg@8 & Pass@8 & Length & Avg@8 & Pass@8 & Length & Avg@8 & Pass@8 & Length \\
\midrule
\multirow{4}{*}{Qwen3-1.7B} & Base & 43.8 & 76.7 & 18647 & 35.4 & 60.0 & 18266 & 35.0 & 53.3 & 18161 & 21.7 & 46.7 & 19421 & 34.0 & 59.2 & 18624 \\
 & Full & 46.3 & 70.0 & 20783 & 36.7 & 66.7 & 21697 & 45.0 & 73.3 & 20995 & 24.6 & 40.0 & 22306 & 38.1 & 62.5 & 21445 \\
 & Short prefix & 49.2 & 80.0 & 20157 & 38.8 & 66.7 & 21312 & 36.3 & 60.0 & 20889 & 24.2 & 43.3 & 21264 & 37.1 & 62.5 & 20905 \\
 & Long prefix & 48.8 & 73.3 & 19618 & 35.4 & 63.3 & 21613 & 37.5 & 66.7 & 20203 & 25.0 & 46.7 & 21638 & 36.7 & 62.5 & 20768 \\
\midrule
\multirow{4}{*}{Qwen3-4B} & Base & 74.6 & 90.0 & 15495 & 65.4 & 80.0 & 18696 & 65.8 & 80.0 & 17249 & 43.3 & 66.7 & 19636 & 62.3 & 79.2 & 17769 \\
 & Full & 72.9 & 83.3 & 16485 & 63.8 & 83.3 & 19818 & 63.3 & 83.3 & 18075 & 43.8 & 53.3 & 21951 & 60.9 & 75.8 & 19082 \\
\bottomrule
\end{tabular*}
}
\end{table}

\subsubsection{Temperature Robustness Checks}
\label{app:temperature_per_dataset}

\begin{table}[H]
\caption{Per-dataset decoding temperature results for \autoref{tab:temperature_eval}. Base and OPSD use the listed temperature.}
\label{tab:temperature_eval_per_dataset}
\centering
{\fontsize{6.3pt}{7.0pt}\selectfont
\setlength{\tabcolsep}{0.5pt}
\renewcommand{\arraystretch}{1.04}
\begin{tabular*}{\textwidth}{@{\extracolsep{\fill}}ll*{15}{c}@{}}
\toprule
\multirow{2}{*}{Model} & \multirow{2}{*}{Decode $\tau$ / ckpt} & \multicolumn{3}{c}{AIME 2024} & \multicolumn{3}{c}{AIME 2025} & \multicolumn{3}{c}{AIME 2026} & \multicolumn{3}{c}{HMMT25} & \multicolumn{3}{c}{Dataset Avg.} \\
\cmidrule(lr){3-5}\cmidrule(lr){6-8}\cmidrule(lr){9-11}\cmidrule(lr){12-14}\cmidrule(lr){15-17}
& & Avg@8 & Pass@8 & Length & Avg@8 & Pass@8 & Length & Avg@8 & Pass@8 & Length & Avg@8 & Pass@8 & Length & Avg@8 & Pass@8 & Length \\
\midrule
\multirow{4}{*}{Qwen3-1.7B}
 & 0.6 / Base & 43.8 & 76.7 & 18647 & 35.4 & 60.0 & 18266 & 35.0 & 53.3 & 18161 & 21.7 & 46.7 & 19421 & 34.0 & 59.2 & 18624 \\
 & 0.6 / OPSD & 45.8 & 73.3 & 22175 & 36.7 & 66.7 & 22143 & 36.7 & 63.3 & 21288 & 22.1 & 40.0 & 23913 & 35.3 & 60.8 & 22380 \\
 & 1.0 / Base & 49.2 & 73.3 & 17458 & 39.2 & 63.3 & 17941 & 40.4 & 66.7 & 17523 & 23.8 & 46.7 & 19009 & 38.1 & 62.5 & 17983 \\
 & 1.0 / OPSD & 51.2 & 80.0 & 20477 & 35.4 & 73.3 & 21752 & 34.6 & 60.0 & 20409 & 25.4 & 50.0 & 22901 & 36.7 & 65.8 & 21385 \\
\midrule
\multirow{4}{*}{\ml{Qwen3-4B-\\Instruct}}
 & 0.7 / Base & 62.5 & 86.7 & 9412 & 45.8 & 70.0 & 8558 & 55.0 & 80.0 & 8434 & 30.8 & 50.0 & 10254 & 48.5 & 71.7 & 9165 \\
 & 0.7 / OPSD & 65.0 & 83.3 & 10141 & 45.8 & 73.3 & 10539 & 58.3 & 76.7 & 10130 & 30.8 & 46.7 & 11409 & 50.0 & 70.0 & 10555 \\
\bottomrule
\end{tabular*}
}
\end{table}

\begin{table}[H]
\caption{Per-dataset Qwen3-1.7B training temperature results for \autoref{tab:temperature_train}. All rows use decoding temperature $0.6$.}
\label{tab:temperature_train_per_dataset}
\centering
{\fontsize{6.3pt}{7.0pt}\selectfont
\setlength{\tabcolsep}{0.5pt}
\renewcommand{\arraystretch}{1.04}
\begin{tabular*}{\textwidth}{@{\extracolsep{\fill}}ll*{15}{c}@{}}
\toprule
\multirow{2}{*}{$\tau_S$} & \multirow{2}{*}{$\tau_T$} & \multicolumn{3}{c}{AIME 2024} & \multicolumn{3}{c}{AIME 2025} & \multicolumn{3}{c}{AIME 2026} & \multicolumn{3}{c}{HMMT25} & \multicolumn{3}{c}{Dataset Avg.} \\
\cmidrule(lr){3-5}\cmidrule(lr){6-8}\cmidrule(lr){9-11}\cmidrule(lr){12-14}\cmidrule(lr){15-17}
& & Avg@8 & Pass@8 & Length & Avg@8 & Pass@8 & Length & Avg@8 & Pass@8 & Length & Avg@8 & Pass@8 & Length & Avg@8 & Pass@8 & Length \\
\midrule
1.1 & 1.1 & 45.8 & 73.3 & 22175 & 36.7 & 66.7 & 22143 & 36.7 & 63.3 & 21288 & 22.1 & 40.0 & 23913 & 35.3 & 60.8 & 22380 \\
0.6 & 0.6 & 46.7 & 76.7 & 19889 & 37.1 & 63.3 & 19212 & 35.4 & 63.3 & 19985 & 22.5 & 40.0 & 21010 & 35.4 & 60.8 & 20024 \\
0.6 & 1.1 & 46.2 & 76.7 & 21424 & 34.2 & 53.3 & 22346 & 38.3 & 56.7 & 21255 & 22.1 & 40.0 & 24204 & 35.2 & 56.7 & 22307 \\
1.1 & 0.6 & 43.8 & 76.7 & 17488 & 34.2 & 60.0 & 18115 & 30.4 & 53.3 & 18741 & 20.8 & 40.0 & 19019 & 32.3 & 57.5 & 18341 \\
\bottomrule
\end{tabular*}
}
\end{table}

\subsubsection{Replications of Recent OPSD Variants}
\label{app:recent_opsd_per_dataset}

These replications use the recent papers' LoRA based 200-step configurations.

\begin{table}[H]
\caption{Per-dataset Purified OPSD results for \autoref{tab:recent_opsd_replications}. Priv. is the privileged teacher field used to construct the PMI target.}
\label{tab:purified_opsd_per_dataset}
\centering
{\fontsize{5.8pt}{6.6pt}\selectfont
\setlength{\tabcolsep}{0.45pt}
\renewcommand{\arraystretch}{1.04}
\begin{tabular*}{\textwidth}{@{\extracolsep{\fill}}ll*{15}{c}@{}}
\toprule
\multirow{2}{*}{Model} & \multirow{2}{*}{Priv.} & \multicolumn{3}{c}{AIME24} & \multicolumn{3}{c}{AIME25} & \multicolumn{3}{c}{AIME26} & \multicolumn{3}{c}{HMMT25} & \multicolumn{3}{c}{Overall} \\
\cmidrule(lr){3-5}\cmidrule(lr){6-8}\cmidrule(lr){9-11}\cmidrule(lr){12-14}\cmidrule(lr){15-17}
 &  & Avg@8 & Pass@8 & Length & Avg@8 & Pass@8 & Length & Avg@8 & Pass@8 & Length & Avg@8 & Pass@8 & Length & Avg@8 & Pass@8 & Length \\
\midrule
Qwen3-1.7B & Answer & 52.9 & 76.7 & 23.1k & 41.7 & 63.3 & 24.1k & 41.7 & 70.0 & 24.0k & 23.3 & 43.3 & 25.4k & 39.9 & 63.3 & 24.2k \\
Qwen3-1.7B & Solution & 46.3 & 76.7 & 19.6k & 37.9 & 60.0 & 21.0k & 36.7 & 63.3 & 20.6k & 29.2 & 60.0 & 21.8k & 37.5 & 65.0 & 20.8k \\
Qwen3-4B & Answer & 75.0 & 83.3 & 17.0k & 70.8 & 80.0 & 19.7k & 66.7 & 86.7 & 18.6k & 40.8 & 66.7 & 23.0k & 63.3 & 79.2 & 19.6k \\
Qwen3-4B & Solution & 72.5 & 80.0 & 16.6k & 62.1 & 80.0 & 19.6k & 64.6 & 83.3 & 17.7k & 46.3 & 66.7 & 20.1k & 61.3 & 77.5 & 18.5k \\
Qwen3-4B-Thinking-2507 & Answer & 83.8 & 90.0 & 22.1k & 82.5 & 86.7 & 24.6k & 78.3 & 86.7 & 23.7k & 55.8 & 73.3 & 28.8k & 75.1 & 84.2 & 24.8k \\
Qwen3-4B-Thinking-2507 & Solution & 84.2 & 93.3 & 21.7k & 79.6 & 86.7 & 23.9k & 81.3 & 90.0 & 23.1k & 56.3 & 76.7 & 28.1k & 75.3 & 86.7 & 24.2k \\
OLMo-3-7B-Think & Answer & 73.8 & 86.7 & 17.5k & 69.2 & 80.0 & 19.1k & 70.4 & 90.0 & 18.1k & 44.6 & 66.7 & 22.0k & 64.5 & 80.8 & 19.2k \\
OLMo-3-7B-Think & Solution & 73.3 & 86.7 & 17.0k & 66.3 & 86.7 & 18.6k & 71.3 & 86.7 & 18.0k & 44.2 & 60.0 & 21.8k & 63.8 & 80.0 & 18.9k \\
\bottomrule
\end{tabular*}
}
\end{table}

\begin{table}[H]
\caption{Per-dataset $\beta$-OPSD results for \autoref{tab:recent_opsd_replications}.}
\label{tab:beta_opsd_per_dataset}
\centering
{\fontsize{5.8pt}{6.6pt}\selectfont
\setlength{\tabcolsep}{0.45pt}
\renewcommand{\arraystretch}{1.04}
\begin{tabular*}{\textwidth}{@{\extracolsep{\fill}}l*{15}{c}@{}}
\toprule
\multirow{2}{*}{Model} & \multicolumn{3}{c}{AIME24} & \multicolumn{3}{c}{AIME25} & \multicolumn{3}{c}{AIME26} & \multicolumn{3}{c}{HMMT25} & \multicolumn{3}{c}{Overall} \\
\cmidrule(lr){2-4}\cmidrule(lr){5-7}\cmidrule(lr){8-10}\cmidrule(lr){11-13}\cmidrule(lr){14-16}
 & Avg@8 & Pass@8 & Length & Avg@8 & Pass@8 & Length & Avg@8 & Pass@8 & Length & Avg@8 & Pass@8 & Length & Avg@8 & Pass@8 & Length \\
\midrule
Qwen3-1.7B & 32.5 & 63.3 & 13.6k & 27.9 & 56.7 & 12.6k & 25.4 & 50.0 & 14.1k & 14.6 & 26.7 & 13.4k & 25.1 & 49.2 & 13.4k \\
Qwen3-4B-Thinking-2507 & 76.7 & 86.7 & 16.5k & 74.6 & 86.7 & 18.0k & 77.1 & 90.0 & 18.1k & 46.3 & 66.7 & 20.1k & 68.6 & 82.5 & 18.2k \\
Qwen3-4B-Instruct & 59.2 & 76.7 & 11.1k & 43.3 & 60.0 & 10.3k & 57.5 & 83.3 & 9.9k & 28.8 & 43.3 & 12.8k & 47.2 & 65.8 & 11.0k \\
OLMo-3-7B-Think & 69.2 & 80.0 & 17.5k & 69.2 & 86.7 & 18.6k & 72.1 & 90.0 & 18.2k & 45.8 & 70.0 & 20.8k & 64.1 & 81.7 & 18.8k \\
OLMo-3-7B-Instruct & 55.0 & 83.3 & 8.3k & 45.0 & 60.0 & 7.9k & 45.0 & 66.7 & 8.0k & 24.6 & 50.0 & 8.9k & 42.4 & 65.0 & 8.3k \\
\bottomrule
\end{tabular*}
}
\end{table}

\subsubsection{Additional Training Controls}
\label{app:additional_controls_per_dataset}

\begin{table}[H]
\caption{Per-dataset English instruction prefix controls for \autoref{tab:instruction_prefix_macro}. C/D denote concise/detailed student-to-teacher instructions. Both instruction-only rows contain no privileged teacher context.}
\label{tab:instruction_prefix_per_dataset}
\centering
{\fontsize{5.8pt}{6.6pt}\selectfont
\setlength{\tabcolsep}{0.45pt}
\renewcommand{\arraystretch}{1.02}
\begin{tabular*}{\textwidth}{@{\extracolsep{\fill}}ll*{15}{c}@{}}
\toprule
\multirow{2}{*}{Model} & \multirow{2}{*}{Condition}
& \multicolumn{3}{c}{AIME 2024} & \multicolumn{3}{c}{AIME 2025} & \multicolumn{3}{c}{AIME 2026} & \multicolumn{3}{c}{HMMT25} & \multicolumn{3}{c}{Dataset Avg.} \\
\cmidrule(lr){3-5}\cmidrule(lr){6-8}\cmidrule(lr){9-11}\cmidrule(lr){12-14}\cmidrule(lr){15-17}
& & Avg@8 & Pass@8 & Len & Avg@8 & Pass@8 & Len & Avg@8 & Pass@8 & Len & Avg@8 & Pass@8 & Len & Avg@8 & Pass@8 & Len \\
\midrule
\multirow{4}{*}{Qwen3-1.7B}
& Base & 43.8 & 76.7 & 18647 & 35.4 & 60.0 & 18266 & 35.0 & 53.3 & 18161 & 21.7 & 46.7 & 19421 & 34.0 & 59.2 & 18624 \\
& Full solution & 45.8 & 73.3 & 22175 & 36.7 & 66.7 & 22143 & 36.7 & 63.3 & 21288 & 22.1 & 40.0 & 23913 & 35.3 & 60.8 & 22380 \\
& No priv.: C$\to$D & 30.0 & 53.3 & 12632 & 25.0 & 43.3 & 12005 & 22.5 & 53.3 & 12903 & 15.0 & 36.7 & 12791 & 23.1 & 46.7 & 12583 \\
& No priv.: D$\to$C & 43.3 & 73.3 & 25307 & 37.5 & 66.7 & 25255 & 42.1 & 70.0 & 23577 & 22.1 & 40.0 & 25965 & 36.3 & 62.5 & 25026 \\
\midrule
\multirow{4}{*}{\ml{OLMo-3-\\7B-Think}}
& Base & 71.7 & 86.7 & 17296 & 66.7 & 80.0 & 19137 & 71.2 & 90.0 & 17560 & 44.6 & 70.0 & 20950 & 63.5 & 81.7 & 18736 \\
& Full solution & 68.8 & 76.7 & 20502 & 55.4 & 80.0 & 22027 & 67.1 & 86.7 & 20285 & 43.8 & 66.7 & 24083 & 58.8 & 77.5 & 21724 \\
& No priv.: C$\to$D & 72.9 & 86.7 & 14621 & 63.3 & 83.3 & 17110 & 65.0 & 90.0 & 15766 & 41.7 & 63.3 & 18436 & 60.7 & 80.8 & 16483 \\
& No priv.: D$\to$C & 62.1 & 80.0 & 23097 & 62.5 & 80.0 & 23728 & 61.7 & 86.7 & 23608 & 39.2 & 53.3 & 25808 & 56.4 & 75.0 & 24060 \\
\bottomrule
\end{tabular*}
}
\end{table}

\begin{table}[H]
\caption{Per-dataset AdvT4 / AdvT16 results for \autoref{tab:advt4_macro}.}
\label{tab:advt4_per_dataset}
\centering
{\fontsize{6.0pt}{6.8pt}\selectfont
\setlength{\tabcolsep}{1.2pt}
\renewcommand{\arraystretch}{1.02}
\begin{tabular*}{\textwidth}{@{\extracolsep{\fill}}ll*{15}{c}@{}}
\toprule
\multirow{2}{*}{Model} & \multirow{2}{*}{Stage}
& \multicolumn{3}{c}{AIME 2024} & \multicolumn{3}{c}{AIME 2025} & \multicolumn{3}{c}{AIME 2026} & \multicolumn{3}{c}{HMMT25} & \multicolumn{3}{c}{Dataset Avg.} \\
\cmidrule(lr){3-5}\cmidrule(lr){6-8}\cmidrule(lr){9-11}\cmidrule(lr){12-14}\cmidrule(lr){15-17}
& & Avg@8 & Pass@8 & Len & Avg@8 & Pass@8 & Len & Avg@8 & Pass@8 & Len & Avg@8 & Pass@8 & Len & Avg@8 & Pass@8 & Len \\
\midrule
\multirow{4}{*}{Qwen3-1.7B}
& Base & 43.8 & 76.7 & 18647 & 35.4 & 60.0 & 18266 & 35.0 & 53.3 & 18161 & 21.7 & 46.7 & 19421 & 34.0 & 59.2 & 18624 \\
& c256 & 46.2 & 70.0 & 20783 & 36.7 & 66.7 & 21697 & 45.0 & 73.3 & 20995 & 24.6 & 40.0 & 22306 & 38.1 & 62.5 & 21445 \\
& AdvT4 & 47.1 & 80.0 & 19206 & 34.6 & 50.0 & 20145 & 35.8 & 56.7 & 18732 & 21.2 & 50.0 & 20992 & 34.7 & 59.2 & 19769 \\
& AdvT16 & 51.2 & 80.0 & 18716 & 37.1 & 60.0 & 19128 & 36.2 & 60.0 & 19273 & 24.2 & 46.7 & 20824 & 37.2 & 61.7 & 19485 \\
\midrule
\multirow{4}{*}{Qwen3-4B}
& Base & 74.6 & 90.0 & 15495 & 65.4 & 80.0 & 18696 & 65.8 & 80.0 & 17249 & 43.3 & 66.7 & 19636 & 62.3 & 79.2 & 17769 \\
& c256 & 72.9 & 83.3 & 16485 & 63.8 & 83.3 & 19818 & 63.3 & 83.3 & 18075 & 43.8 & 53.3 & 21951 & 60.9 & 75.8 & 19082 \\
\bottomrule
\end{tabular*}
}
\end{table}

\begin{table}[H]
\caption{Per-dataset results for pointwise vocabulary contribution caps in \autoref{tab:clip_tau_qwen} and \autoref{tab:clip_tau_olmo}.}
\label{tab:clip_tau_per_dataset}
\centering
{\fontsize{6.0pt}{6.8pt}\selectfont
\setlength{\tabcolsep}{1.2pt}
\renewcommand{\arraystretch}{1.02}
\begin{tabular*}{\textwidth}{@{\extracolsep{\fill}}ll*{15}{c}@{}}
\toprule
\multirow{2}{*}{Model} & \multirow{2}{*}{$\tau_{\mathrm{clip}}$}
& \multicolumn{3}{c}{AIME 2024} & \multicolumn{3}{c}{AIME 2025} & \multicolumn{3}{c}{AIME 2026} & \multicolumn{3}{c}{HMMT25} & \multicolumn{3}{c}{Dataset Avg.} \\
\cmidrule(lr){3-5}\cmidrule(lr){6-8}\cmidrule(lr){9-11}\cmidrule(lr){12-14}\cmidrule(lr){15-17}
& & Avg@8 & Pass@8 & Len & Avg@8 & Pass@8 & Len & Avg@8 & Pass@8 & Len & Avg@8 & Pass@8 & Len & Avg@8 & Pass@8 & Len \\
\midrule
\multirow{5}{*}{Qwen3-1.7B}
& Base & 43.8 & 76.7 & 18647 & 35.4 & 60.0 & 18266 & 35.0 & 53.3 & 18161 & 21.7 & 46.7 & 19421 & 34.0 & 59.2 & 18624 \\
& $0.01$ & 40.4 & 66.7 & 25792 & 30.8 & 56.7 & 25552 & 32.1 & 63.3 & 24715 & 22.1 & 40.0 & 28918 & 31.4 & 56.7 & 26244 \\
& $0.05$ & 45.8 & 73.3 & 22175 & 36.7 & 66.7 & 22143 & 36.7 & 63.3 & 21288 & 22.1 & 40.0 & 23913 & 35.3 & 60.8 & 22380 \\
& $0.1$ & 44.2 & 76.7 & 16599 & 32.1 & 53.3 & 16320 & 32.9 & 60.0 & 16266 & 22.5 & 50.0 & 17404 & 32.9 & 60.0 & 16647 \\
& $0.2$ & 34.6 & 63.3 & 14852 & 32.5 & 53.3 & 15336 & 31.7 & 53.3 & 15600 & 23.8 & 43.3 & 16392 & 30.6 & 53.3 & 15545 \\
\midrule
\multirow{7}{*}{\ml{OLMo-3-\\7B-Think}}
& Base & 71.7 & 86.7 & 17296 & 66.7 & 80.0 & 19137 & 71.2 & 90.0 & 17560 & 44.6 & 70.0 & 20950 & 63.5 & 81.7 & 18736 \\
\bottomrule
\end{tabular*}
}
\end{table}

\begin{table}[H]
\caption{Per-dataset 10-step SFT followed by 100-step no-privilege OPSD for \autoref{tab:sft10_same_macro}. OT/OMR denote the SFT dataset. OPSD uses OpenThoughts.}
\label{tab:sft10_same_per_dataset}
\centering
{\fontsize{5.5pt}{6.3pt}\selectfont
\setlength{\tabcolsep}{0.35pt}
\renewcommand{\arraystretch}{1.02}
\begin{tabular*}{\textwidth}{@{\extracolsep{\fill}}lll*{15}{c}@{}}
\toprule
\multirow{2}{*}{Model} & \multirow{2}{*}{SFT data} & \multirow{2}{*}{Stage}
& \multicolumn{3}{c}{AIME 2024} & \multicolumn{3}{c}{AIME 2025} & \multicolumn{3}{c}{AIME 2026} & \multicolumn{3}{c}{HMMT25} & \multicolumn{3}{c}{Dataset Avg.} \\
\cmidrule(lr){4-6}\cmidrule(lr){7-9}\cmidrule(lr){10-12}\cmidrule(lr){13-15}\cmidrule(lr){16-18}
& & & Avg@8 & Pass@8 & Len & Avg@8 & Pass@8 & Len & Avg@8 & Pass@8 & Len & Avg@8 & Pass@8 & Len & Avg@8 & Pass@8 & Len \\
\midrule
\multirow{3}{*}{Qwen3-1.7B} & \multirow{3}{*}{OT}
& Base & 43.8 & 76.7 & 18647 & 35.4 & 60.0 & 18266 & 35.0 & 53.3 & 18161 & 21.7 & 46.7 & 19421 & 34.0 & 59.2 & 18624 \\
& & SFT-10 & 7.1 & 23.3 & 9991 & 7.5 & 20.0 & 5382 & 5.8 & 26.7 & 6527 & 0.0 & 0.0 & 5897 & 5.1 & 17.5 & 6949 \\
& & + OPSD-100 & 7.9 & 23.3 & 8030 & 5.8 & 20.0 & 6372 & 5.8 & 16.7 & 5708 & 1.3 & 6.7 & 6183 & 5.2 & 16.7 & 6573 \\
\midrule
\multirow{3}{*}{Qwen3-1.7B} & \multirow{3}{*}{OMR}
& Base & 43.8 & 76.7 & 18647 & 35.4 & 60.0 & 18266 & 35.0 & 53.3 & 18161 & 21.7 & 46.7 & 19421 & 34.0 & 59.2 & 18624 \\
& & SFT-10 & 35.4 & 66.7 & 17957 & 30.0 & 46.7 & 15927 & 29.6 & 63.3 & 17051 & 17.1 & 40.0 & 16562 & 28.0 & 54.2 & 16874 \\
& & + OPSD-100 & 36.7 & 63.3 & 17613 & 29.2 & 53.3 & 16041 & 26.3 & 53.3 & 17416 & 16.3 & 26.7 & 16416 & 27.1 & 49.2 & 16872 \\
\bottomrule
\end{tabular*}
}
\end{table}

%% file: initial_capability_coding/per_dataset.tex
\begin{table}[htbp]
\centering
\caption{Per-dataset coding results for \autoref{fig:initial_capability}b.}
\label{tab:capability_coding_per_dataset}
{\fontsize{6.3pt}{7.0pt}\selectfont
\setlength{\tabcolsep}{1pt}
\begin{tabular*}{\textwidth}{@{\extracolsep{\fill}}ll*{12}{r}@{}}
\toprule
Model & Checkpoint & \multicolumn{3}{c}{HumanEval+} & \multicolumn{3}{c}{MBPP+} & \multicolumn{3}{c}{LCB v6} & \multicolumn{3}{c}{Dataset Avg.} \\
\cmidrule(lr){3-5}\cmidrule(lr){6-8}\cmidrule(lr){9-11}\cmidrule(lr){12-14}
 & & Avg@4 & Pass@4 & Length & Avg@4 & Pass@4 & Length & Avg@4 & Pass@4 & Length & Avg@4 & Pass@4 & Length \\
\midrule
Qwen3-0.6B & Base & 45.7 & 66.5 & 3590 & 32.1 & 52.6 & 2582 & 14.3 & 21.1 & 14257 & 30.7 & 46.8 & 6810 \\
 & OPSD-256 & 47.7 & 68.9 & 10168 & 34.8 & 59.5 & 7615 & 15.0 & 22.9 & 21067 & 32.5 & 50.4 & 12950 \\
 & OPSD-1024 & 49.2 & 73.2 & 11946 & 35.6 & 59.5 & 7691 & 14.0 & 18.9 & 20027 & 33.0 & 50.5 & 13221 \\
\addlinespace[2pt]
DeepSeek-R1-Distill-Qwen-1.5B & Base & 59.6 & 77.4 & 5336 & 44.8 & 65.1 & 4641 & 18.9 & 25.7 & 14275 & 41.1 & 56.1 & 8084 \\
 & OPSD-256 & 58.7 & 76.2 & 4729 & 41.9 & 62.7 & 3857 & 19.7 & 26.3 & 13816 & 40.1 & 55.1 & 7467 \\
 & OPSD-1024 & 57.8 & 79.3 & 4488 & 44.0 & 63.8 & 4023 & 18.0 & 27.4 & 13257 & 39.9 & 56.8 & 7256 \\
\bottomrule
\end{tabular*}}
\end{table}